%% file: main.tex
\documentclass[11pt]{article}

\usepackage[utf8]{inputenc} 
\usepackage[T1]{fontenc}    
\usepackage{hyperref}       
\usepackage{url}            
\usepackage{booktabs}       
\usepackage{amsfonts}       
\usepackage{nicefrac}       
\usepackage{microtype}      
\usepackage{xcolor}         
\usepackage{bbm}
\usepackage{amsfonts}       
\usepackage{nicefrac}       
\usepackage{microtype}  
\usepackage{amsmath}
\usepackage{varwidth}
\usepackage{lscape} 
\usepackage{graphicx}
\usepackage{adjustbox}
\usepackage{wrapfig}
\usepackage{overpic}
\usepackage{fullpage}
\usepackage{natbib}
\usepackage{textcomp}
\usepackage{subcaption} 

\usepackage{tikz}

\usepackage{abstract}

\usepackage{enumitem}
\setlist{nolistsep}

\usepackage{xcolor}
\usepackage{color}
\definecolor{deepblue}{rgb}{0,0,0.5}
\definecolor{deepred}{rgb}{0.6,0,0}
\definecolor{deepgreen}{rgb}{0,0.5,0}
\definecolor{darkgreen}{rgb}{0,0.6,0}
\definecolor{darkred}{rgb}{0.7,0.0,0}
\definecolor{darkblue}{rgb}{0,0.0,0.6}
\definecolor{magenta}{rgb}{0.8,0.1,0.8}
\definecolor{darksomething}{rgb}{0,0.4,0.6}
\definecolor{darkother}{rgb}{0.4,0.6,0}

\makeatletter
\renewcommand{\paragraph}{%
  \@startsection{paragraph}{4}%
  {\z@}{0ex \@plus 1ex \@minus .2ex}{-0.75em}%
  {\normalfont\normalsize\bfseries}%
}

\newcommand\blfootnote[1]{%
  \begingroup
  \renewcommand\thefootnote{}\footnote{#1}%
  \addtocounter{footnote}{-1}%
  \endgroup
}

\definecolor{citecolor}{rgb}{.259,.659,1}
\definecolor{mydarkblue}{rgb}{0,0.08,0.45}
\definecolor{urlcolor}{rgb}{0,.145,.698}
\definecolor{linkcolor}{rgb}{0.01,0.31,.65}
\hypersetup{
    colorlinks=true,
    breaklinks=true,  %
    bookmarksnumbered=true,
    linkcolor=linkcolor,
    citecolor=citecolor,
    filecolor=mydarkblue,
    urlcolor=urlcolor,
    pdftitle={VeLO: Training Versatile Learned Optimizers by Scaling Up},
    pdfpagemode=FullScreen,
    pdfview=FitH
    }

\newcommand{\pp}[1]{\left(#1 \right)}

\usepackage{listings}

\definecolor{gray}{gray}{0.5}
\colorlet{commentcolour}{green!50!black}

\colorlet{stringcolour}{red!60!black}
\colorlet{keywordcolour}{magenta!90!black}
\colorlet{exceptioncolour}{yellow!50!red}
\colorlet{commandcolour}{blue!60!black}
\colorlet{numpycolour}{blue!60!green}
\colorlet{literatecolour}{magenta!90!black}
\colorlet{promptcolour}{green!50!black}
\colorlet{specmethodcolour}{violet}

\newcommand*{\literatecolour}{\textcolor{literatecolour}}

\newcommand*{\pythonprompt}{\textcolor{promptcolour}{{>}{>}{>}}}

\lstdefinestyle{mypython}{
language=python,
showtabs=true,
tab=,
tabsize=2,
basicstyle=\ttfamily\footnotesize,
stringstyle=\color{stringcolour},
showstringspaces=false,
alsoletter={1234567890},
otherkeywords={\%, \}, \{, \&, \|},
keywordstyle=\color{keywordcolour}\bfseries,
emph={and,break,class,continue,def,yield,del,elif ,else,%
except,exec,finally,for,from,global,if,import,in,%
lambda,not,or,pass,print,raise,return,try,while,assert,with},
emphstyle=\color{blue}\bfseries,
emph={[2]True, False, None},
emphstyle=[2]\color{keywordcolour},
emph={[3]object,type,isinstance,copy,deepcopy,zip,enumerate,reversed,list,set,len,dict,tuple,xrange,append,execfile,real,imag,reduce,str,repr},
emphstyle=[3]\color{commandcolour},
emph={Exception,NameError,IndexError,SyntaxError,TypeError,ValueError,OverflowError,ZeroDivisionError},
emphstyle=\color{exceptioncolour}\bfseries,
morecomment=[s]{"""}{"""},
commentstyle=\color{commentcolour}\slshape,
emph={[4]ode, fsolve, sqrt, exp, sin, cos,arctan, arctan2, arccos, pi,  array, norm, solve, dot, arange, isscalar, max, sum, flatten, shape, reshape, find, any, all, abs, plot, linspace, legend, quad, polyval,polyfit, hstack, concatenate,vstack,column_stack,empty,zeros,ones,rand,vander,grid,pcolor,eig,eigs,eigvals,svd,qr,tan,det,logspace,roll,min,mean,cumsum,cumprod,diff,vectorize,lstsq,cla,eye,xlabel,ylabel,squeeze},
emphstyle=[4]\color{numpycolour},
emph={[5]__init__,__add__,__mul__,__div__,__sub__,__call__,__getitem__,__setitem__,__eq__,__ne__,__nonzero__,__rmul__,__radd__,__repr__,__str__,__get__,__truediv__,__pow__,__name__,__future__,__all__},
emphstyle=[5]\color{specmethodcolour},
emph={[6]assert,yield},
emphstyle=[6]\color{keywordcolour}\bfseries,
emph={[7]range},
emphstyle={[7]\color{keywordcolour}\bfseries},
literate=*%
{:}{{\literatecolour:}}{1}%
{=}{{\literatecolour=}}{1}%
{-}{{\literatecolour-}}{1}%
{+}{{\literatecolour+}}{1}%
{*}{{\literatecolour*}}{1}%
{**}{{\literatecolour{**}}}2%
{/}{{\literatecolour/}}{1}%
{//}{{\literatecolour{//}}}2%
{!}{{\literatecolour!}}{1}%
{[}{{\literatecolour[}}{1}%
{]}{{\literatecolour]}}{1}%
{<}{{\literatecolour<}}{1}%
{>}{{\literatecolour>}}{1}%
{>>>}{\pythonprompt}{3}%
,%
frame=trbl,
rulecolor=\color{black!40},
backgroundcolor=\color{white},
breakindent=.5\textwidth,frame=single,breaklines=true%
}

\lstnewenvironment{python}[1][]{\lstset{style=mypython}}{}

\lstdefinestyle{mypythoninline}{
style=mypython,%
basicstyle=\ttfamily,%
keywordstyle=\color{keywordcolour},%
emphstyle={[7]\color{keywordcolour}},%
emphstyle=\color{exceptioncolour},%
literate=*%
{:}{{\literatecolour:}}{2}%
{=}{{\literatecolour=}}{2}%
{-}{{\literatecolour-}}{2}%
{+}{{\literatecolour+}}{2}%
{*}{{\literatecolour*}}2%
{**}{{\literatecolour{**}}}3%
{/}{{\literatecolour/}}{2}%
{//}{{\literatecolour{//}}}{2}%
{!}{{\literatecolour!}}{2}%
{[}{{\literatecolour[}}{2}%
{]}{{\literatecolour]}}{2}%
{<}{{\literatecolour<}}{2}%
{<=}{{\literatecolour{<=}}}3%
{>}{{\literatecolour>}}{2}%
{>=}{{\literatecolour{>=}}}3%
{==}{{\literatecolour{==}}}3%
{!=}{{\literatecolour{!=}}}3%
{+=}{{\literatecolour{+=}}}3%
{-=}{{\literatecolour{-=}}}3%
{*=}{{\literatecolour{*=}}}3%
{/=}{{\literatecolour{/=}}}3%
}

\newcommand{\e}{\texttt{e}}

\title{\textbf{VeLO}: Training Versatile Learned Optimizers by Scaling Up}

\author{%
  Luke Metz\footnotemark[1],
  James Harrison\footnotemark[2],
  C. Daniel Freeman,
  Amil Merchant, \\
  Lucas Beyer,
  James Bradbury,
  Naman Agarwal,
  Ben Poole,\\
  Igor Mordatch,
  Adam Roberts,
  Jascha Sohl-Dickstein\footnotemark[3]
}

\date{Google Research, Brain Team}

\begin{document}

\maketitle

\begin{abstract}
While deep learning models have replaced hand-designed features across many domains, these models are still trained with hand-designed optimizers. In this work, we leverage the same scaling approach behind the success of deep learning to learn versatile optimizers.
We train an optimizer for deep learning which is itself a small neural network that ingests gradients and outputs parameter updates. 
Meta-trained with approximately four thousand TPU-months of compute on a wide variety of optimization tasks,
our optimizer not only exhibits compelling performance, but optimizes in interesting and unexpected ways.
It requires no hyperparameter tuning, instead automatically adapting to the specifics of the problem being optimized.
We open source our learned optimizer, meta-training code, the associated train and test data, and an extensive optimizer benchmark suite with baselines at \href{https://velo-code.github.io}{\texttt{velo-code.github.io}}.
\vspace{2em}
\end{abstract}

\begin{figure}[h!]
\centering
\begin{tikzpicture}
 
    \node[above right, inner sep=0] (image) at (0,0) {
        \includegraphics[width=\textwidth]{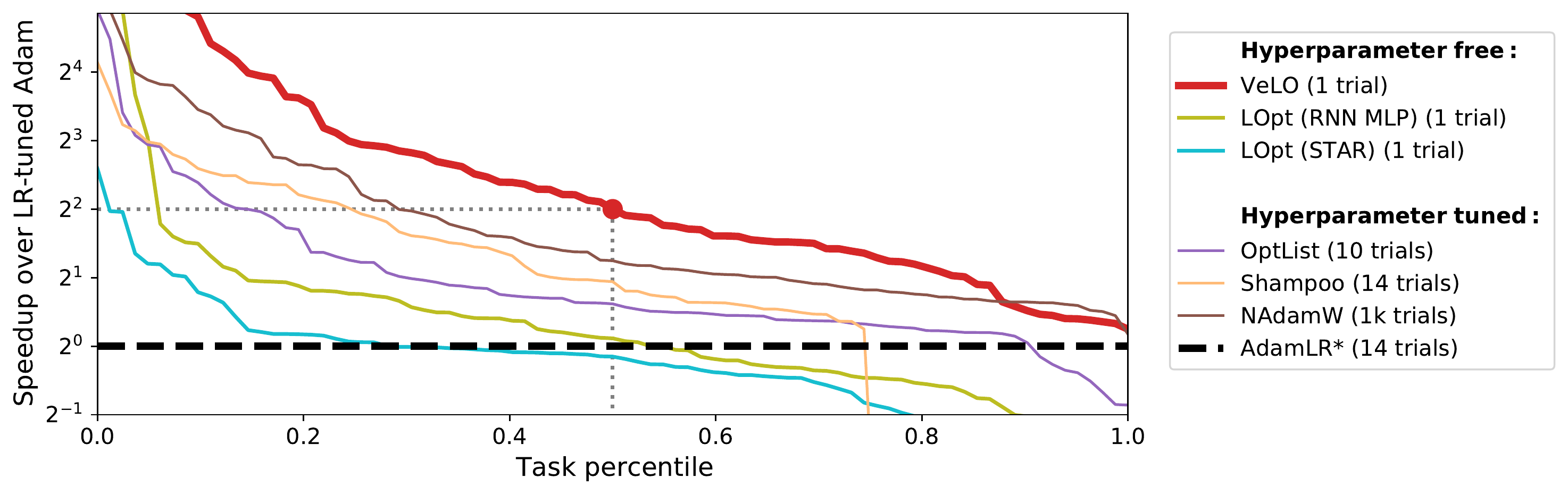}
    };
    \draw [->,line width=2pt] (7.3+1.5,5.1-0.3) to [out=180,in=90] (6.47,3.2);
    \node [below right,align=left,font={\footnotesize},draw=gray,draw opacity=0.]
    (text) at (7.35+1.5,5.4-0.3){
        VeLO is more than\\
        $2^2$ times faster than\\
        LR-tuned Adam on\\
        50\% of tasks 
        };
\end{tikzpicture}
    \vspace{-2.em}
    \caption{\small{ \textbf{Optimizer performance on the 83 canonical tasks in the VeLOdrome benchmark.}
    Our learned optimizer VeLO (red) with no hyperparameters optimizes models dramatically faster than learning rate-tuned baselines (orange, black dashed), and 
    usually surpasses the performance of NAdamW (brown) with one thousand trials of per-problem hyperparameter tuning. We exceed the performance of previous work on learned optimizers: the RNN MLP from \citet{metz2020tasks}, and the STAR learned optimizer from \citet{harrison2022closer}.
    The $y$-axis shows the relative number of steps it takes learning rate-tuned Adam to achieve the same loss each optimizer reaches after 10K training steps (e.g. a $y$-axis value of 2 means that it takes Adam 20K training iterations to reach the same loss).
    The $x$-axis shows the fraction of tasks for which the optimizer achieves at least that large a speedup over learning rate-tuned Adam. On all tasks, we train faster than learning rate-tuned Adam (all values >1). On about half of the tasks, we are more than 4x faster than learning rate-tuned Adam. On more than 14\% of the tasks, we are more than 16x times faster.
    \vspace{-2.5em}
    }
    \label{fig:aggregate_multitask_cond}
    }
\end{figure}

\blfootnote{Contact: $^*$\texttt{luke.s.metz@gmail.com}, $^\dagger$\texttt{jamesharrison@google.com}, $^\ddagger$\texttt{jaschasd@google.com}}

\clearpage
\setcounter{tocdepth}{1}
\tableofcontents
\clearpage

\section{Introduction}
Scaling up has been crucial to the success of deep learning across many domains \citep{krizhevsky2012imagenet, hannun2014deep,radford2018improving, radford2019language, brown2020language, devlin2018bert}.
However, scaling brings with it several challenges: increased compute, larger datasets and considerably more engineering effort~\citep{zhang2022opt, barham2022pathways}. 
The field of meta-learning, or the study of learning machine learning algorithms, has not seen this same explosion of scale.
Scaling meta-learning systems is fundamentally harder for several reasons.
In meta-learning, a large training dataset corresponds to a large set of \emph{tasks}, which are representative of the tasks a practitioner might want to optimize. 
Unlike image and text data that can be gathered from the internet, there is no standardized or automated way to collect these tasks. 
Even worse, meta-training over a diverse set of \emph{realistic} tasks can be extraordinarily computationally costly, as individual problems within the task distribution are often themselves computationally expensive.
As a result of these difficulties, very few large scale meta-learning systems exist.

While in supervised learning model sizes are often increased to improve performance, simply scaling up the model size of a learned optimizer can be problematic.
Larger optimizers may require fewer iterations to achieve good performance, but the overhead per step may increase. A simpler hand-designed optimizer with less overhead could be run for more training steps or more carefully tuned to achieve competitive performance.
Thus, a balance between overhead and performance must be struck~\citep{metz2022practical}.

In this paper, we present VeLO, a versatile learned optimizer that is parameterized by a neural network, and meta-trained at a far greater scale than has previously been investigated. 
We build on our prior work scaling learned optimizers~\citep{metz2020tasks} and scale even further: we meta-train on three orders of magnitude more tasks, use two orders of magnitude more compute, and develop a considerably faster learned optimizer architecture.
The resulting learned optimizer performs better with less computational overhead, enabling training of much larger models.

VeLO requires no hyperparameter tuning, and works well on a wide variety of neural network training tasks. 
We evaluate VeLO's generalization abilities with VeLOdrome, a new optimization benchmark, and show VeLO's ability to generalize to new problems not seen during meta-training. 
We also evaluate on a wide range of real-world models, including language, vision, and decision Transformers; vision models such as ResNets, NERF models, and detection models; and other models such as recurrent and graph networks. 
VeLO represents the first general-purpose learned optimizer for deep learning, and serves as concrete evidence of the viability of learned optimization.

\subsection{Try VeLO}
We designed VeLO to be easy to try on any JAX model that uses Optax~\citep{deepmind2020jax}: 
\begin{python}
from learned_optimization.research.general_lopt import prefab
opt = prefab.optax_lopt(total_training_steps)
opt_state = opt.init(params)

updates, opt_state = opt.update(grads, opt_state, params=params, 
                                    extra_args={"loss": loss})
params = optax.apply_updates(params, updates)
\end{python}

We also provide a simple \href{https://colab.sandbox.google.com/github/google/learned_optimization/blob/main/learned_optimization/research/general_lopt/Demo_for_training_a_model_with_a_learned_optimizer.ipynb}{Colab notebook} that trains one of a variety of test problems we have implemented in the \texttt{learned\_optimization} package \citep{metz2022practical}.

\section{Problem Setting: Learned Optimization}

Consider using an optimizer to train a neural network with parameters $\phi^t$ indexed by training step $t$, with total number of optimization steps $N$. The training loss for the minibatch at time $t$ is written $\ell_t(\phi^t)$.
Training with minibatch stochastic gradient descent (SGD), the update dynamics take the form:
\begin{equation}
    \phi^{t+1} = \phi^t - \alpha \nabla_{\phi^t} \ell_t(\phi^t)
\end{equation}
where the learning rate $\alpha$ is the only meta-parameter (or hyperparameter). 
Defining $U_\text{SGD}(g; \alpha) = \alpha g$, we can write the SGD update as: 
\begin{equation}
    \phi^{t+1} = \phi^t - U_\text{SGD}\left(\nabla_{\phi^t} \ell_t(\phi^t); \alpha\right).
\end{equation}

\paragraph{Learned update rules.} The core idea behind learned optimizers is to replace the fixed-form update rule $U_\text{SGD}$, which is a function of one (meta-)parameter (learning rate $\alpha$) with a more flexible form, parameterized by \textit{many} more meta-parameters. 
In this work, we parameterize the update $U(g, ...; \theta)$ as a neural network with meta-parameters $\theta$, which takes as input gradients $g$.
By allowing for a more expressive update function, it is possible to have faster training and thus a more useful optimizer.
This comes at the cost of making it harder to find the values of the meta-parameters which result in the best performance.

The update rule $U(\cdot; \theta)$ can take additional inputs beyond just the gradient. 
For example, the update can also depend on the current parameter values $\phi$, or the value of the loss at the current timestep.
It can also utilize recurrence across training steps, either using a recurrent neural network, or more simply by accumulating exponential moving averages of gradients (as done by Adam~\citep{kingma2014adam} and momentum), iterates, or any other statistic available during training. 

These learned update rules are often parameterized in a manner that applies the same computation across each parameter of the model being optimized~\citep{andrychowicz2016learning, metz2019understanding}.
This allows them to be applied to networks of different sizes than those used during training.

\paragraph{Meta-training.} Meta-training is the process of finding the (meta-)parameters $\theta$ of the update rule $U(\cdot; \theta)$ such that the resulting optimizer performs well on some specified meta-objective.
Intuitively, this meta-objective defines what it means for an optimizer to be ``good at optimizing''; in this work we write the meta-loss as $L(\theta)$.
In prior work, the meta-loss is commonly defined as the average training loss throughout training $\frac{1}{N}\sum_{t=1}^N \ell_t(\phi^t)$ or the loss $\ell_N(\phi^N)$ at the end of training.
It can also be non-standard measurements such as the final validation loss $\ell_\text{valid}(\phi^N)$, which would encourage the learned optimizer to train models in such a way that they generalize well~\citep{metz2019understanding}.
Methods used to train learned optimizers, or modify the parameters of the update rule ($\theta$) to improve this meta-objective, include  backprop~\citep{andrychowicz2016learning}, reinforcement learning~\citep{li2016learning, li2017learning}, and evolution~\citep{metz2019understanding, metz2020tasks, metz2021training}.

In this work, we leverage gradient-based meta-learning, but with gradients computed with Evolution Strategies (ES)~\citep{rechenberg1973evolutionsstrategie, nesterov2011random, salimans2017evolution} rather than backpropagation.
The primary benefit of ES over analytic gradients---in addition to improved memory efficiency---is that it provides unbiased estimates of the gradient of a \emph{Gaussian-smoothed} meta-loss.
This Gaussian-smoothing averages over the extreme sensitivity of the optimization trajectory $\{\phi^t : t \in [1,\dots,N]\}$ to the exact value of the meta-parameters $\theta$.
Without this smoothing, meta-training is often extremely unstable \citep{metz2019understanding}.

Much like in standard hyperparameter tuning, meta-training can be done on a single task.
While this results in an extremely performant optimizer \emph{for that task}, the amortized cost 
including meta-training is too large to make 
this worthwhile for most applications. 
Instead, we amortize~\citep{amos2022tutorial} the meta-training cost over a large distribution of tasks (Section \ref{sec:data}), with the goal of learning an optimizer that generalizes well to new tasks.

\section{Methods: Large Scale Optimizer Training}
In this section, we highlight the most important aspects of our system: 
\begin{enumerate}
    \item the learned optimizer architecture, i.e. the functional form of $U(\cdot\,; \theta)$;
    \item the distribution of tasks on which the optimizer is meta-trained; and
    \item the details of the meta-training, such as gradient estimation and curricula.
\end{enumerate}
This section provides only a brief discussion of each element, with more detailed discussion reserved for the Appendix. 
See \href{https://github.com/google/learned_optimization/blob/main/learned_optimization/learned_optimizers/adafac_mlp_lopt.py}{our code repository} for the complete implementation of our architecture; links to specific components of the training infrastructure are provided throughout the appendix.

\begin{figure}[t!]
    \centering
    \begin{overpic}[width=\textwidth]{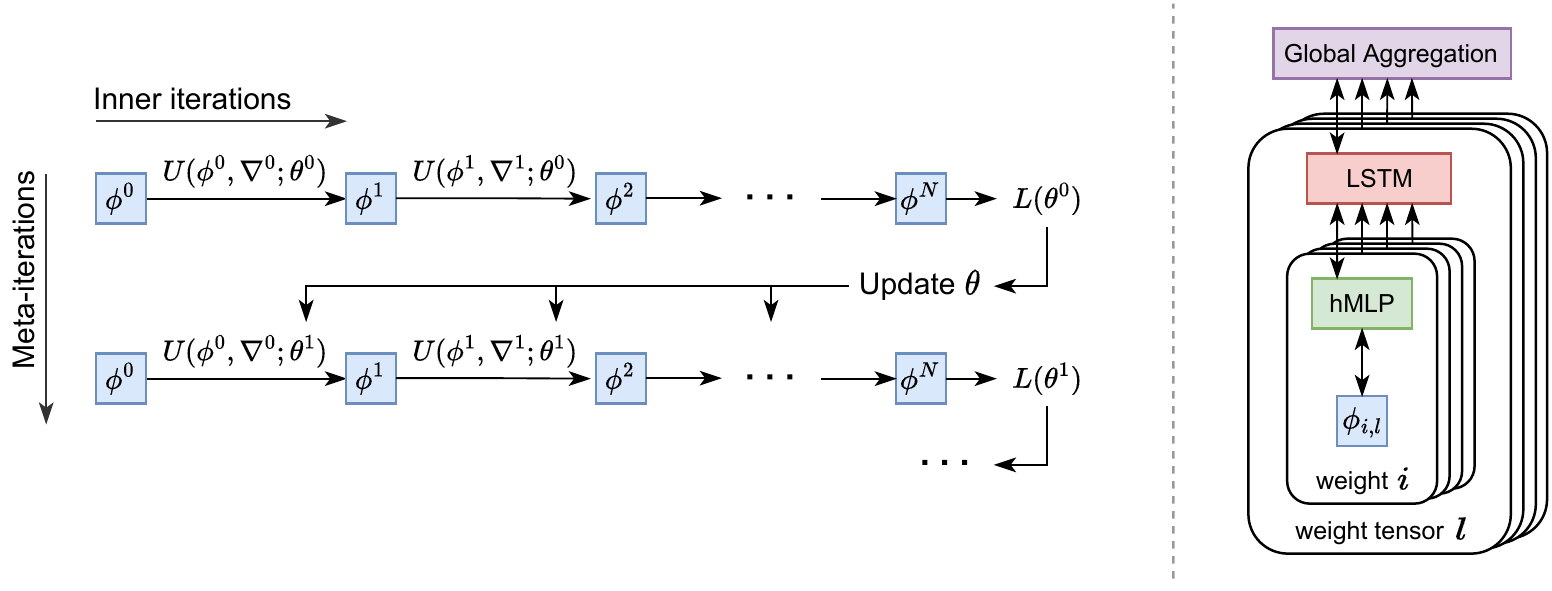}
    \put (0,1) {\textbf{\small(a)}}
    \put (77,1) {\textbf{\small(b)}}
    \end{overpic}
    \caption{
    \textbf{(a) Training and meta-training.}
    The learned optimizer's update rule $U\pp{\cdot\,;\theta}$ has meta-parameters $\theta$, and generates updates to the parameters $\phi$ of a model it is training. In the figure we only show the update rule taking parameter values and gradients as inputs, but the update may take any features of the problem as an input. 
    Inner-training consists of applying the learned optimizer for $N$ optimization steps, producing final parameters $\phi^N$, which is a function of the meta-parameters $\theta$. 
    After inner-training, the meta-loss $L\pp{\phi^N\pp{\theta}}$ is used to evaluate the performance of the trained model. 
    An estimate of the gradient of the meta-loss (in our case estimated using Evolution Strategies) is then used to update $\theta$. 
    This process is repeated in order to meta-train the parameters $\theta$ of the learned optimizer.
    \textbf{(b) The hierarchical architecture of the learned optimizer.} The learned optimizer's architecture is adapted to match the architecture of the problem it is optimizing. 
    For each scalar weight $\phi_{i,l}$, a tiny hypernetwork MLP (hMLP) takes as input information about the weight's gradients and iterates, and outputs a scalar update to the weight's value. The parameters of the tiny MLP are set by an LSTM with parameters $\theta$. One copy of the LSTM is constructed for each weight tensor $\phi_{\,\cdot,l}$ in the inner problem, and each LSTM sets the parameters for many MLPs.
    The LSTMs for all weight tensors coordinate with each other by outputing a global context signal. This signal is max-pooled over all LSTMs, before being provided back as input to each LSTM. 
    \label{fig:velo_combined}
    }
\end{figure}

\subsection{Learned Optimizer Architecture}
\paragraph{Hierarchical hypernetwork.}
To be useful, our learned optimizer must both be computationally efficient and expressive.
We leverage a two-layer hierarchy of computation: a ``per-tensor'' LSTM~\citep{hochreiter1997long} which operates on features derived by aggregating information from each parameter tensor in various ways, and a ``per-parameter'' MLP which operates on each parameter scalar.
To increase the capacity of this network, we can add computation to the per-parameter network, which scales linearly with the number of parameters, or to the per-tensor network, which scales linearly with the number of tensors, and thus is considerably more efficient.
Figure~\ref{fig:velo_combined} visualizes our learned optimizer architecture.

Next, we consider how to route information through the hierarchy.
Past work~\citep{wichrowska2017learned, metz2020tasks} passed the results of the per-tensor network directly to the per-parameter MLP as additional conditioning.
This introduces a number of additional input features, which not only slow down the per-parameter model, but can also be hard to effectively use, given that the per-parameter model is applied to \textit{every} parameter, and thus must be tiny to reduce overhead (in our case, it is an MLP with 4 hidden units).
As a solution to this, we leverage hypernetworks~\citep{ha2016hypernetworks}. 
Instead of generating information used to condition the MLP, the per-tensor network generates the \textit{weight matrices} of the per-parameter MLP.
This allows for more expressive per-tensor computation without increasing the cost of the per-parameter network.

\paragraph{Per-tensor LSTM.} We use a 512 hidden-unit LSTM with a variety of input features inspired by~\citet{metz2020tasks}.
First, we use the mean and variance of parameter values, the exponential moving averages of the gradient and squared gradient (as used in the Adam update).
The per-tensor network also takes as input a series of additional features representing the current fraction of training completed, so that it can learn training-time dependent strategies, 
such as learning rate schedules.
Finally, our per-tensor network has access to the training loss which can enable complex behaviors such as detecting divergence of the loss.

\paragraph{Per-parameter MLP.}
Our per-parameter MLP follows~\citet{metz2022practical}, and leverages an extremely small MLP (2-hidden layer, 4-hidden unit) operating on the collection of features specifically found to be both fast to compute and performant. 
Unlike in \citet{metz2022practical}, the weights of this per-parameter network are not fixed for all tasks, but instead generated by the per-tensor model. 

\subsection{Data: A Diverse Distribution of Tasks} \label{sec:data}
Unlike in supervised learning, there are no standard, large-scale distributions of tasks for learned optimizer training.
Following \citet{metz2020tasks}, we construct a parametric task distribution for meta-training.
Tasks are generated by sampling a model family, training dataset, training loss function, and architectural hyperparameters including hidden layer widths, depth, and activation function.
Model families include MLPs, ConvNets, ResNets~\citep{he2015delving}, Transformers~\citep{vaswani2017attention}, Vision Transformers~\citep{dosovitskiy2020image}, RNNs, auto-encoders, variational auto-encoders~\citep{kingma2013auto}, and even other learned optimizers.

To provide additional variation during meta-training, and analogous to data-augmentation, we perform a series of ``task-augmentations''---programmatic modifications to tasks which change the training dynamics.
Examples of these task augmentations include: re-parameterizing weight tensors, estimating gradients only in subspaces, introducing asynchronicity in gradient calculation, and changing floating-point precision.

Because tasks are sampled from a wide distribution, their run times vary greatly, sometimes by more than 3 orders of magnitude. To lower the cost of meta-training, we use rejection sampling based on the estimated training time in order to meta-train on fast tasks more frequently than slow ones. 

\subsection{Meta-Training}
\paragraph{Meta-objective.}
The measure of optimization performance we focus on is the training loss at the end of training (setting $L(\theta) = \ell_N(\phi^N)$, where $\phi^N$ depends on $\theta$).
Empirically, we found that targeting final loss yields optimizers which train models to lower loss values than would be possible by targeting average loss as is done in most previous work~\citep{metz2020tasks}.
While we believe final loss is often most important for users, the resulting learned optimizers can exhibit counter-intuitive behavior at intermediate training times.
For example, they may not monotonically lower the loss, and may plateau or in extreme cases even increase loss over part of training. 

\paragraph{Meta-gradient estimation.}
We leverage ES to estimate gradients of this meta-objective.
Unlike past work \citep{metz2022practical,metz2019understanding}, we use full length unrolls, and train each model to completion for each meta-gradient evaluation. This is in contrast to truncated methods, which yield a meta-gradient for a subset of an inner training run. 
Full length unrolls are significantly less compute efficient. 
However, using them makes it straightforward to target the final training loss, and has the added benefit of reducing communication overhead when doing distributed training.

\paragraph{Multi-task training.}
To encourage meta-generalization, we meta-train on a wide variety of tasks.
This is challenging as each task has a different loss scale and thus gradients can not be directly averaged.
Instead of manually normalizing each loss to a uniform scale, we normalize the gradients from each task to be unit-length before averaging them across different tasks.

\paragraph{Curriculum.}
We use an increasing curriculum over both the number of training iterations and problem size (as measured by the time required for a forward pass, as is used in our task rejection sampling) to dramatically speed up meta-training. 

\paragraph{Vectorization and compilation.}
We make extensive use of JAX's vectorization (\texttt{vmap}) to parallelize compute across 
batches of different tasks, and thus make better use of accelerators.
We additionally use JAX's compilation (\texttt{jit}) which greatly accelerates these non-standard workloads on both TPU and GPU.

\paragraph{Data-parallel training on a massive cluster.}
Meta-training occurs on anywhere from 1K to 4K TPU-based accelerators, physically distributed around the world.
Gradients are computed and applied in an asynchronous, batched manner with an increasing batch size between 10K-40K tasks.
Training took approximately 4 weeks; meta-training curves are presented in Appendix \ref{sec:infra}.

\section{Evaluating Learned Optimizers}
Evaluation of optimizers in machine learning is notoriously difficult~\citep{choi2019empirical, schmidt2020descending}.
Evaluating learned optimizers is more difficult still, as one has to additionally consider the degree to which evaluation tasks are ``out of distribution'' relative to the training task distribution. 
As such, in this section we present several different evaluations of VeLO, with varying degrees and types of distribution shift in the evaluation tasks. 
In particular, we present three distinct benchmarks. First, we present VeLOdrome, a diverse evaluation set that can be run relatively efficiently. Second, we benchmark VeLO on the MLCommons algorithms test problems. Finally, we present evaluations on a broad range of real-world state of the art models including vision, language, and decision Transformers among several others. We also present an investigation of problems in which VeLO fails or underperforms baselines. 

\subsection{VeLOdrome: A Canonical Evaluation Set of 83 Tasks}
\label{sec:83}
For our first set of evaluations, we use a test set of relatively small deep learning models, which we refer to as VeLOdrome. 
These are similar to tasks used for meta-training, though they are hand-designed to be more canonical than the often unusual task specifications sampled during meta-training.
Our main consideration for size here is training time; each model is designed to be able to be trained on a single accelerator in under an hour. This enables both rapid evaluation of our learned optimizer, but also aggressive tuning of baseline optimizers for fair comparison. 
This set contains 83 tasks, including convolutional networks, variational auto-encoders, residual networks, and language models such as recurrent networks and Transformers.
In addition to evaluating learned optimizers, we hope this distribution of tasks can serve as a starting point for hand-designed optimizer evaluation.
For full details on VeLOdrome, see Appendix~\ref{app:opt_benchmark}.

\subsubsection{Baseline Optimizers}
One area that makes optimizer comparison difficult is hyperparameter tuning.
By design, our learned optimizers require \emph{no per-problem tuning}, whereas hand-designed optimizers are typically ineffective unless hyperparameter-tuned on a new task.
We explore different levels of hyperparameter tuning, ranging from learning rate tuning---evaluating 15 different trials logarithmically spaced with half powers of 10---to searching over a wider search space with 1K trials.
For our learning rate-tuned optimizers, we evaluate 15 hand-designed optimizers: Adam \citep{kingma2014adam}, AdaBelief \citep{zhuang2020adabelief}, SGD with momentum, RMSProp \citep{Tieleman2012}, SM3 \citep{anil2019memory}, SM3 with momentum, Yogi \citep{zaheer2018adaptive}, RAdam \citep{liu2019variance},  LARS \citep{you2017large},  LAMB \citep{you2019large}, Fromage \citep{bernstein2020distance}, AdamW \citep{loshchilov2017decoupled}, AdaFactor \citep{shazeer2018adafactor}, Adagrad \citep{duchi2011adaptive}, and Shampoo \citep{gupta2018shampoo, anil2020second} with 6 different grafting types \citep{agarwal2020disentangling}.

As a more aggressively-tuned baseline optimizer, we use Nesterov accelerated AdamW \citep{dozat2016incorporating} with tunable learning rate, $\beta_1$, $\beta_2$, $\epsilon$, weight decay (both applied separately from momentum as in AdamW, and not), and a cosine learning rate schedule (with optional warm-up). 
We searched over 1000 hyperparameter configurations for this optimizer; see Appendix C.2 of \citet{metz2019using} for a complete description. 
NAdamW is a superset of many popular optimizers (see discussion of NAdam in \citet{choi2019empirical}), and so with sufficient tuning will achieve best-case performance across a variety of hand-designed optimizers.
We also evaluate a meta-learned list of hyperparameter configurations---``OptList'', described by \citet{metz2019using}.
OptList achieves much better performance than the learning rate-tuned optimizers, while only requiring 10 hyperparameter evaluation trials.

To the best of our knowledge, this is the largest optimizer benchmark to date with respect to both the number of tasks, and the number of optimizers evaluated.
Learning curves for >1 million trained models are \href{https://learned-optimization.readthedocs.io/en/latest/optimizer_baselines.html}{open-sourced}.

\subsubsection{Normalized Performance Across Tasks}
These problems all have dramatically different scales for losses, making comparisons across different tasks difficult.
To enable easy comparison of optimizers across diverse tasks, we report the improvement in training time an optimizer achieves on each task compared to a baseline optimizer---in this case, learning rate-tuned Adam.
For example, a value of 2.0 indicates the final loss achieved by a target optimizer could be achieved by running the baseline for double the amount of training time.
See Appendix~\ref{app:fixed_task_normalizer} for a complete description of this metric.

In Figure~\ref{fig:aggregate_multitask_cond}, we take all tasks, normalize performance relative to the tuned Adam baseline, sort the values (independently for each optimizer), and plot.
In this visualization, we can quickly read off the fraction of time an optimizer outperforms a particular baseline.
We find that VeLO outperforms all learning rate-tuned optimizers on all problems.
On >85\% of tasks, it also outperforms the extensive 1000 trial NAdamW hyperparameter search with only a single training run. 
We also compare to two previous learned optimizers, the RNN MLP LOpt from \citet{metz2020tasks}\footnote{For RNN MLP LOpt we use the final pre-trained model from that effort. This model was meta-trained targeting validation rather than training loss, so is being evaluated slightly out of distribution.} and a STAR learned optimizer from \citet{harrison2022closer}\footnote{Note that the STAR optimizer was only meta-trained on a single task.}.

\begin{figure}
    \centering
    \makebox[\textwidth]{%
    \begin{overpic}[width=0.25\textwidth]{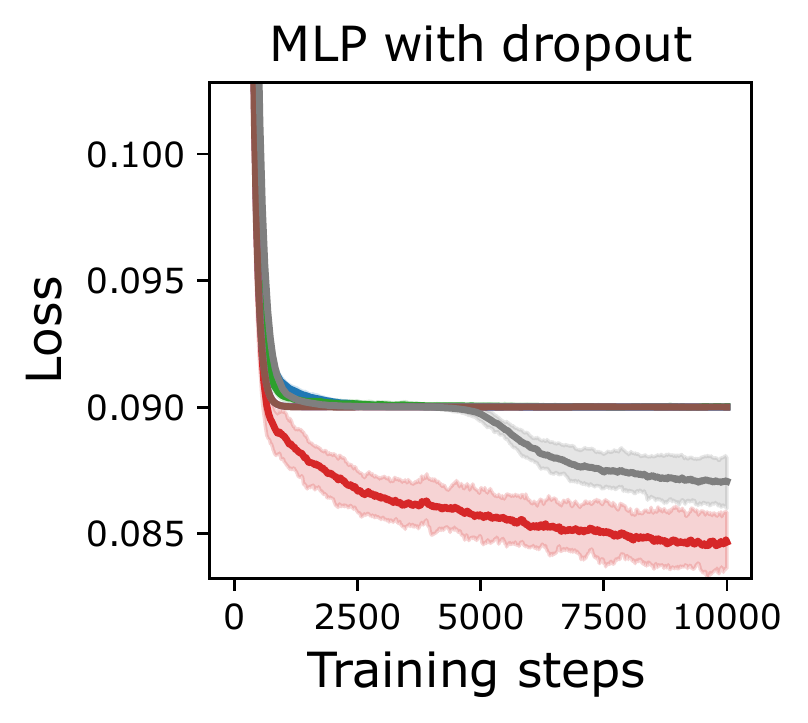}
    \put (29,24.0) {\textbf{\small(a)}}
    \end{overpic}
    \begin{overpic}[width=0.25\textwidth]{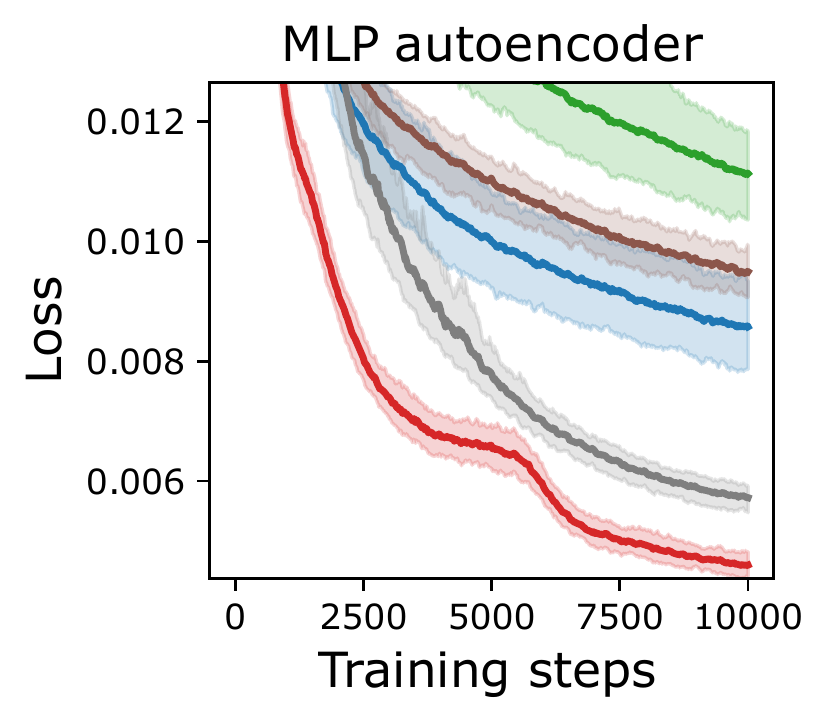}
    \put (29,24.0) {\textbf{\small(b)}}
    \end{overpic}
    \begin{overpic}[width=0.25\textwidth]{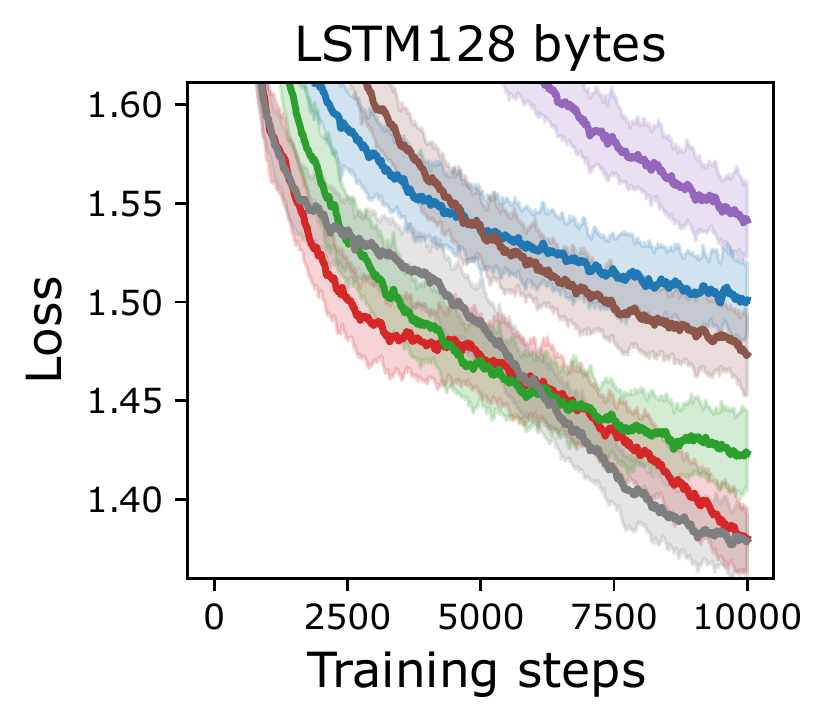}
    \put (27,24.0) {\textbf{\small(c)}}
    \end{overpic}
    \begin{overpic}[width=0.25\textwidth]{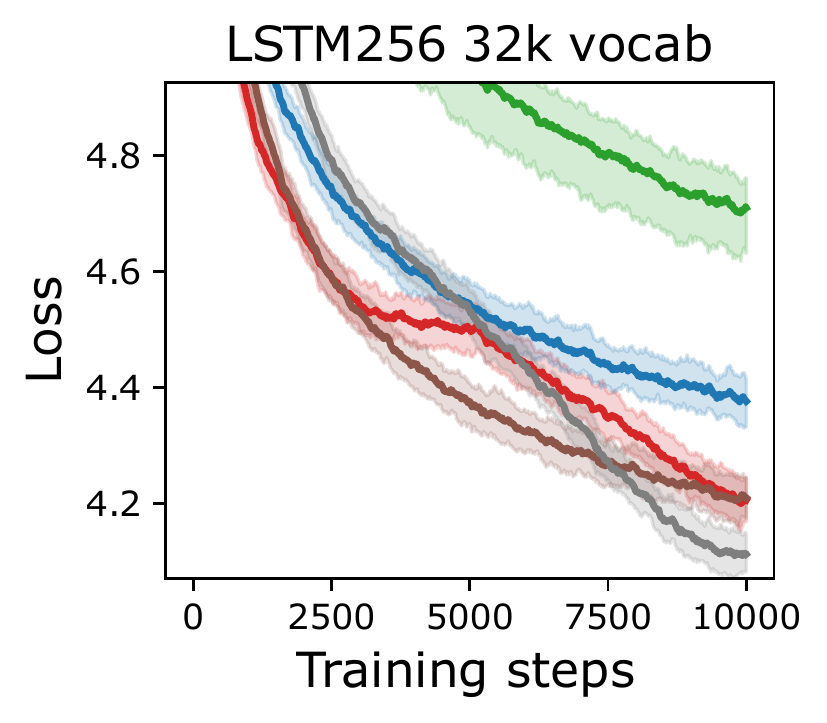}
    \put (24,24.0) {\textbf{\small(d)}}
    \end{overpic}
    }
    \begin{overpic}[width=0.6\textwidth]{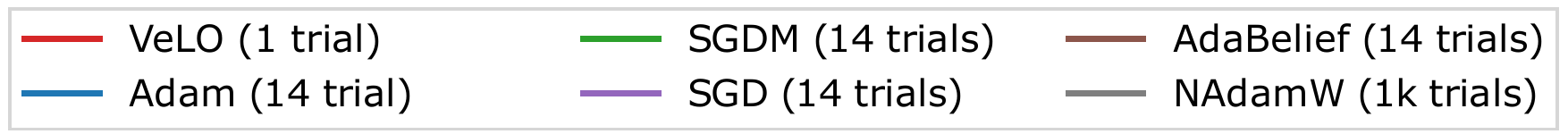}
    \end{overpic}
    \caption{\textbf{Best- and worst-case VeLO performance.} 
    We plot the two optimization tasks where the learned optimizer performs \textbf{(a,b)} best and \textbf{(c,d)} worst compared to baseline optimizers, from the 83 VeLOdrome evaluation tasks described in Section \ref{sec:83}.
    \label{fig:two_good_two_bad}
    }
\end{figure}

\paragraph{Best and worst case tasks.}
Next, we explore the specific tasks on which VeLO performs best and worst.
To do this, we sort the tasks by the difference in performance between VeLO and the best baseline, and show the two tasks at each extreme (Figure~\ref{fig:two_good_two_bad}).
On the task with the best relative performance---an MLP with dropout---all optimizers except VeLO and heavily tuned NAdamW fail. 
On the task with the worst relative performance---an LSTM with a large vocabulary---VeLO still performs second best out of all optimizer classes.

\subsection{MLCommons Tasks}
We investigate a set of six different tasks from the \href{https://mlcommons.org/en/groups/research-algorithms}{MLCommons algorithms track}.
These tasks are out-of-distribution relative to training tasks, primarily due to their scale---they are significantly larger than those seen during meta-training. 
We present only the training loss in the main text.
Other metrics are presented in Appendix~\ref{app:extended_mlcommons}. 

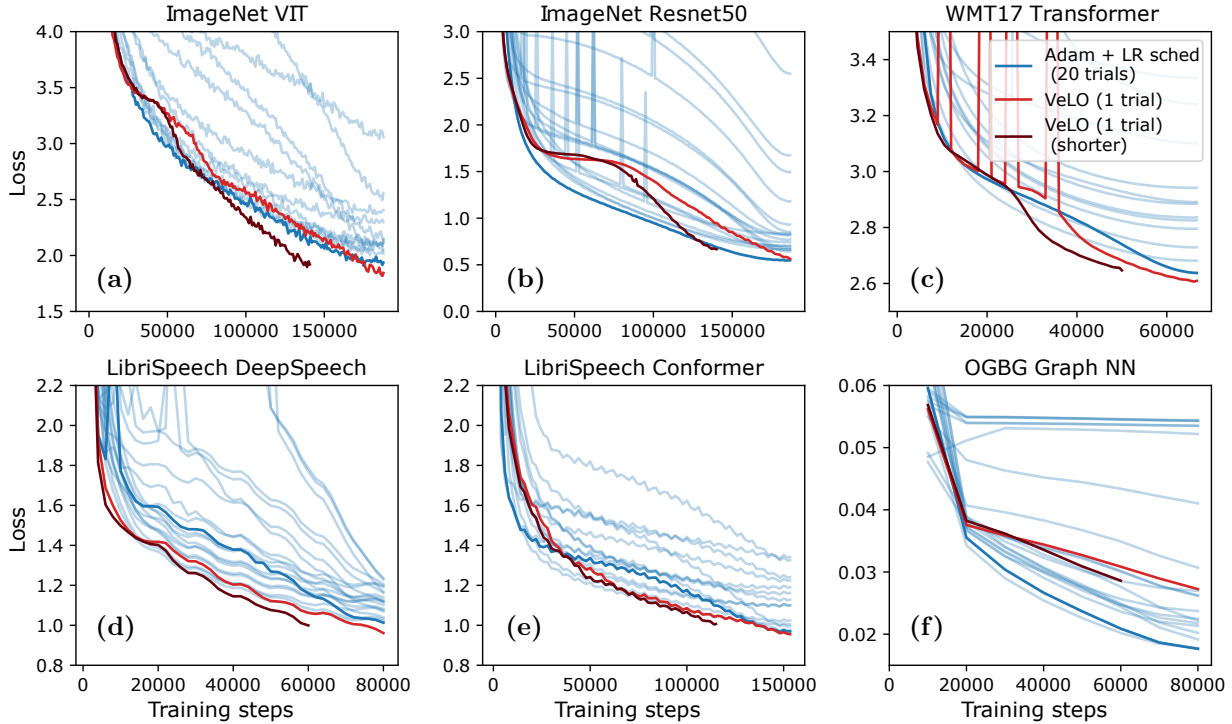
\begin{figure}
    \centering
    \makebox[\textwidth]{%
    \begin{overpic}[width=\textwidth]{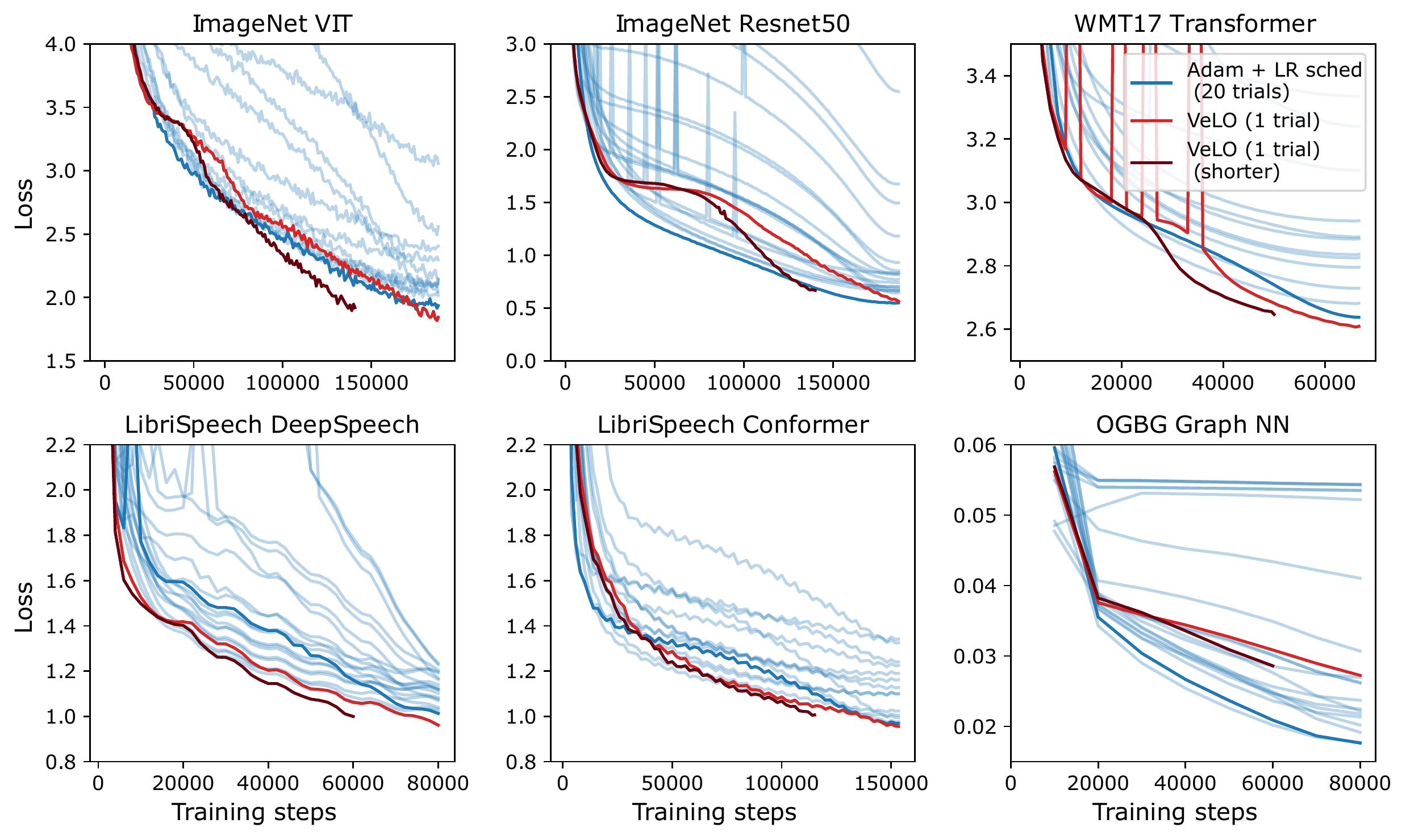}
    \put (8,36.0) {\textbf{\small(a)}}
    \put (41,36.0) {\textbf{\small(b)}}
    \put (73.5,36.0) {\textbf{\small(c)}}
    \put (8,8.0) {\textbf{\small(d)}}
    \put (41,8.0) {\textbf{\small(e)}}
    \put (73.5,8.0) {\textbf{\small(f)}}
    \end{overpic}
    }
    \caption{\textbf{VeLO performance on six MLCommons workloads.} On all but one of the workloads (OGBG Graph NN), VeLO matches or outperforms the best trial of tuned Adam (with learning rate schedule and weight decay).
    \label{fig:init2winit_train}
    }
\end{figure}

For each model family, we compare to an Adam baseline with a learning rate warm up of 5\% of total training iterations, 
and a cosine decay with a learning rate and weight decay searched in log space between $[10^{-2}, 10^{-4}]$ and $[10^{-2}, 1]$ respectively with 20 random trials. 
This search space itself was chosen by the MLCommons organizers, 
based on the top performing hyperparameters across the tasks from a larger search space with 100 trials each. 
In addition to comparing against VeLO applied for the same number of training steps as the baseline, we also compare to VeLO applied for only 75\% as many training iterations.
This shows VeLO's ability to adapt its update steps to the total training time.
Results for all tasks in the evaluation are presented in Figure~\ref{fig:init2winit_train}.

\subsection{Generalization to Tasks Unlike Any Used for Meta-Training} 

\begin{figure}
    \centering
    \makebox[\textwidth]{%
    \begin{overpic}[width=0.25\textwidth]{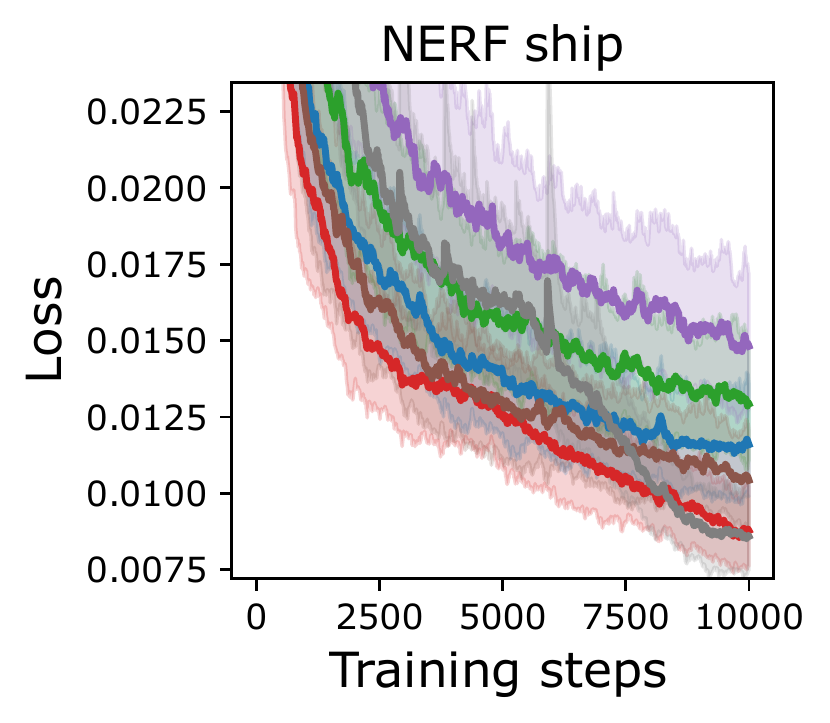}
    \put (31,24.0) {\textbf{\small(a)}}
    \end{overpic}
    \begin{overpic}[width=0.25\textwidth]{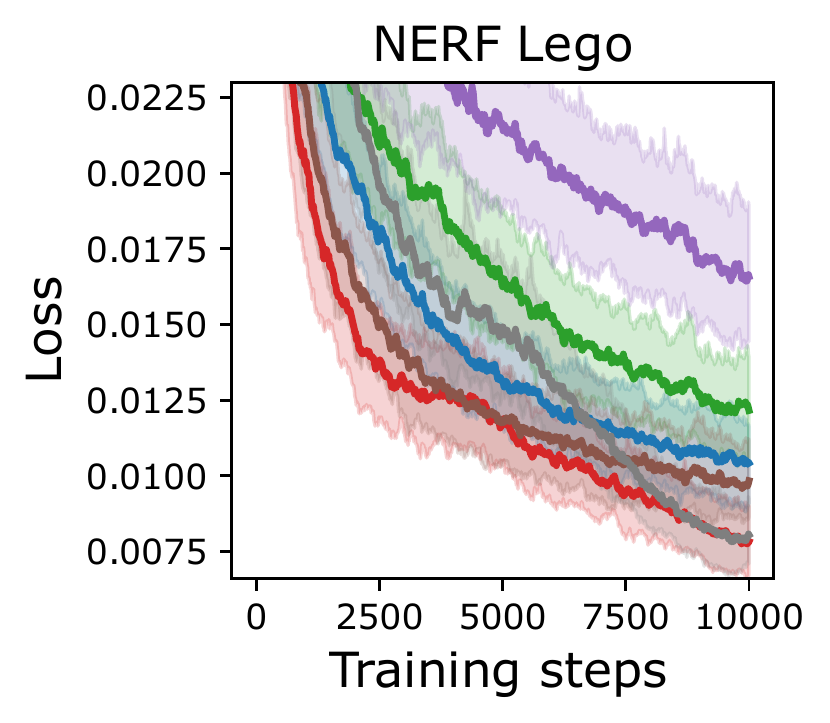}
    \put (31,24.0) {\textbf{\small(b)}}
    \end{overpic}
    \begin{overpic}[width=0.25\textwidth]{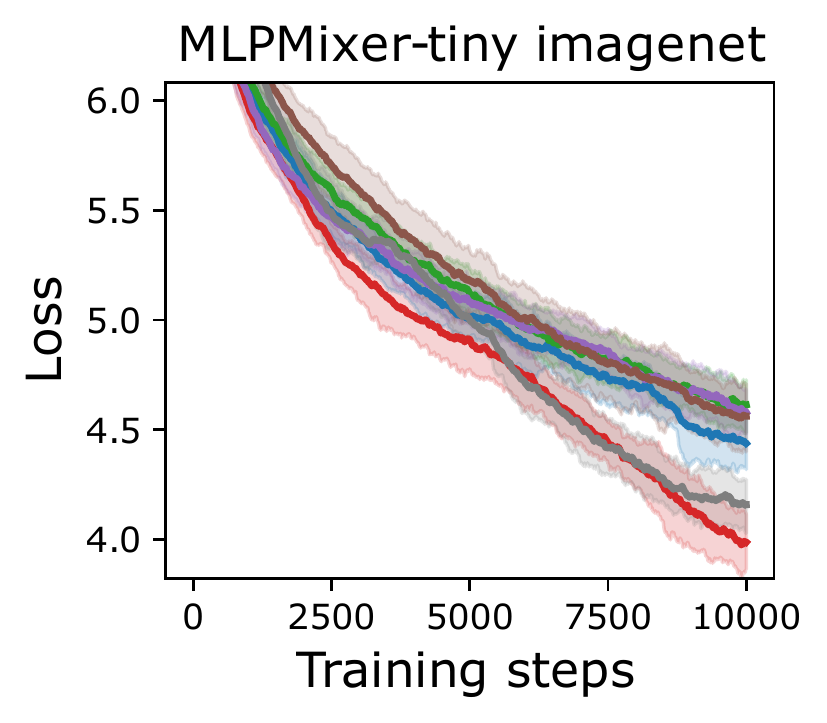}
    \put (24,24.0) {\textbf{\small(c)}}
    \end{overpic}
    \begin{overpic}[width=0.25\textwidth]{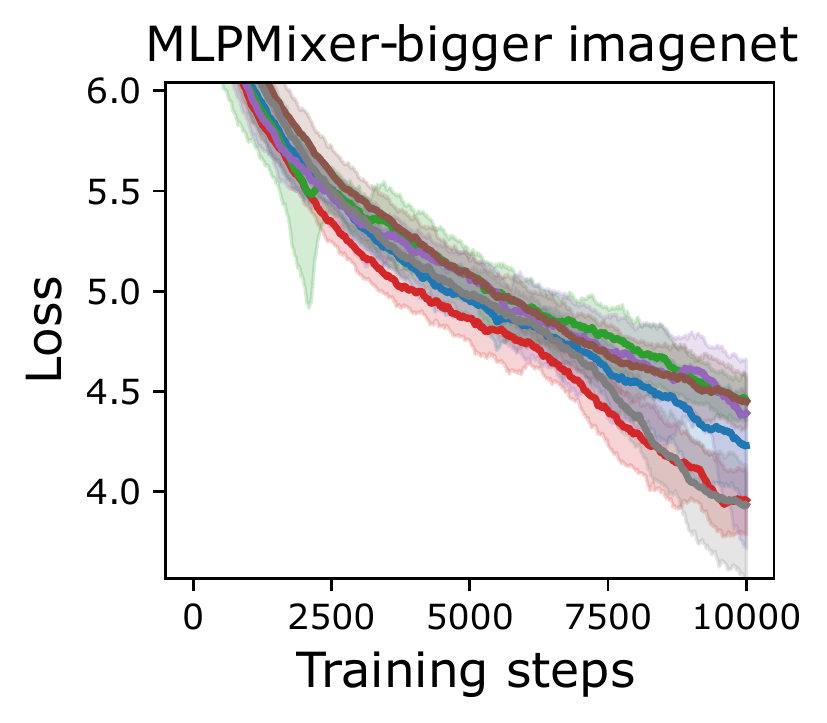}
    \put (25,24.0) {\textbf{\small(d)}}
    \end{overpic}
    }
    \begin{overpic}[width=0.6\textwidth]{figs/fig_pc/legend_bot.pdf}
    \end{overpic}
    \caption{\textbf{VeLO performs well on out of distribution Nerf and MLP-Mixer tasks.} We show performance on \textbf{(a,b)} two NERF tasks, and \textbf{(c,d)} two MLP-Mixer models trained on ImageNet. No NERF or MLP-Mixer tasks were included in the meta-training distribution.
    \label{fig:nerf}
    }
\end{figure}
In this subsection we evaluate VeLO on a diverse set of tasks, many of which are much larger than those seen during meta-training, and all of which are outside of the meta-training distribution. 
As elsewhere in the paper body, we present only training loss.
Validation loss, and other performance metrics, are provided for selected tasks in Appendix \ref{sec:ext_exp_res}.

\subsubsection{NERF}
We test VeLO on NERF tasks~\citep{mildenhall2020nerf}, a family of algorithms never seen during meta-training (Figure \ref{fig:nerf}).
We use the JAXNerf~\citep{jaxnerf2020github} code base, and compare VeLO against a variety of tuned baseline optimizers.
VeLO outperforms learning rate-tuned baseline optimizers without any tuning, and performs comparably to the extensively-tuned NAdamW optimizer.

\subsubsection{MLP-Mixer}
Next we apply VeLO to optimize MLP-Mixer models~\citep{tolstikhin2021mlp}, trained on 64x64 ImageNet (Figure~\ref{fig:nerf}).
This family of models was never seen during meta-training. 
VeLO outperforms learning rate-tuned baseline optimizers, and performs comparably to or better than the extensively tuned NAdamW optimizer.

\begin{figure}[t]
    \makebox[\textwidth]{%
        \centering
        \begin{overpic}[width=0.6\textwidth]{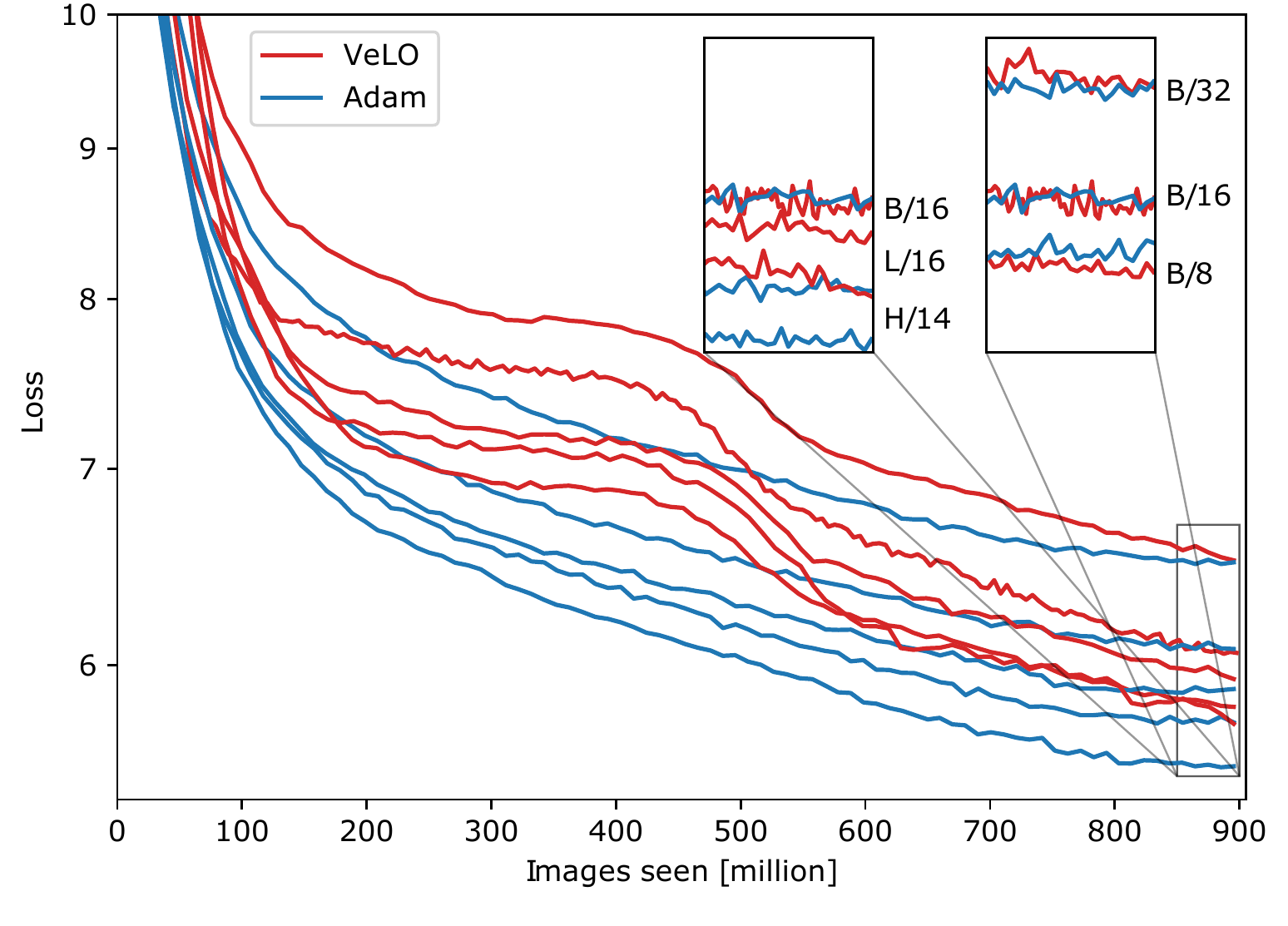} 
        \put (11,13.5) {\textbf{(a)}}
        \end{overpic}
    \quad
        \centering
    \begin{overpic}[width=0.37\textwidth]{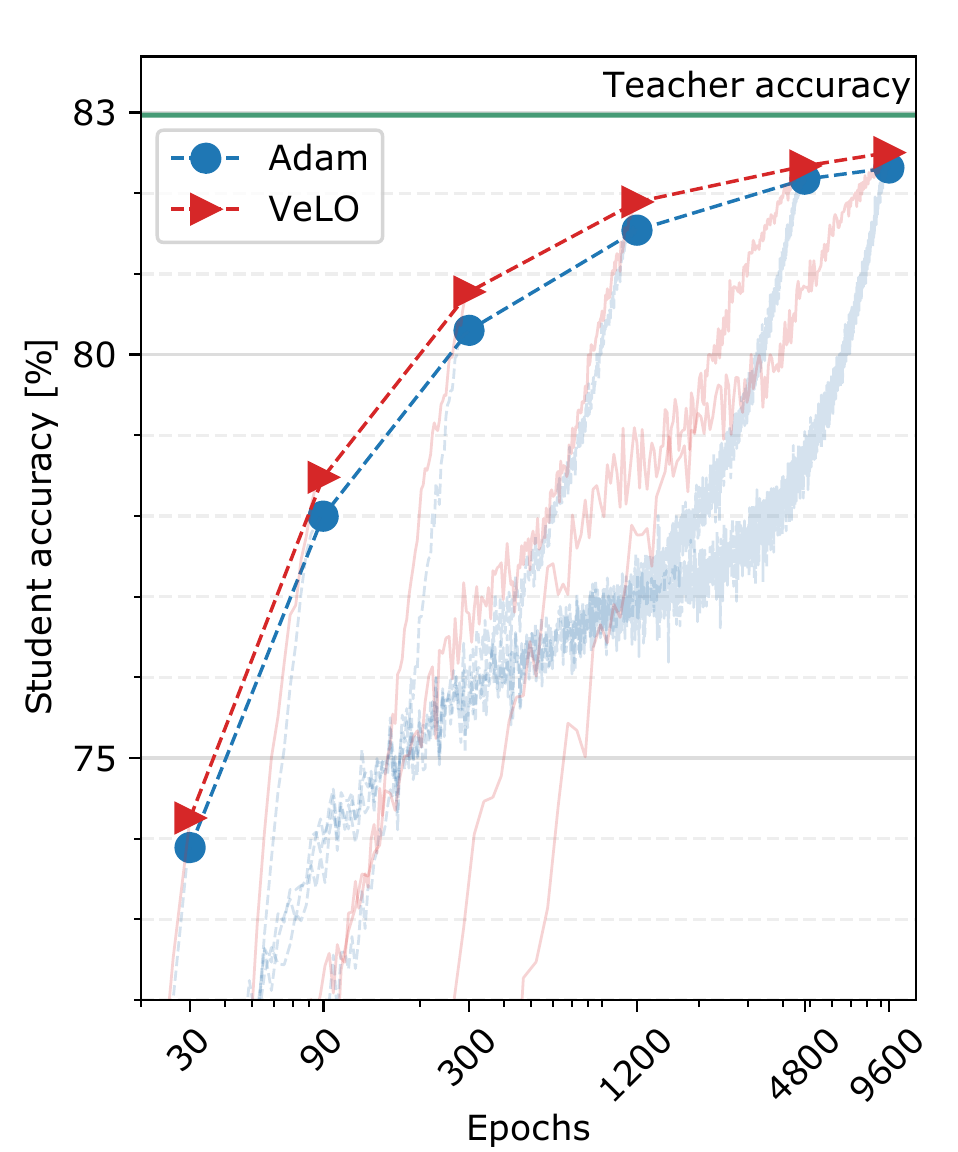}
    \put (69,18.0) {\textbf{(b)}}
    \end{overpic}
    }
    \caption{\textbf{(a) Large-scale ViT training on JFT}. VeLO matches heavily tuned Adam with bells and whistles in most cases, but falls behind for the largest ViT-H/14 (650M parameters). \textbf{(b) Distilling a ResNet-152x2 to a ResNet-50 student} on ImageNet following \citet{distillation_ours}.
    VeLO performs better than the standard baseline.
    \label{fig:vit_jft4b}
    }
\end{figure}

\subsubsection{Large-scale Vision Transformers}
We test VeLO on much larger ViT models (up to B/8 and H/14), which are difficult to optimize with existing techniques and often encounter instabilities \citep{mocov3}.
We train them on the JFT-3B dataset for 900M examples (i.e.\ less than one epoch) following \citet{zhai2022scaling}, to avoid the need to tweak augmentation and regularization and focus solely on applying the learned optimizer at scale.
Figure~\ref{fig:vit_jft4b} (left) shows training curves for the learned optimizer along with the heavily tuned Adam baseline
currently used for ViT research.
The learned optimizer matches or outperforms Adam on all ViT-B models, but starts falling behind on the larger ViT-L and ViT-H variants. 
The larger models ViT-L and ViT-H have approximately 300M and 650M parameters, respectively.
It is encouraging that this first attempt using a learned optimizer for this class of problems, without any tuning, matches or approaches these very strong Adam baselines. The Adam baseline includes weight decay, gradient clipping, learning rate schedules, and warm up/cooldown schedules, all of which have been hand-tuned over the course of more than a year.

\subsubsection{Knowledge Distillation}
Knowledge distillation, especially for model compression, is an especially difficult optimization task \citep{distillation_ours,distillation_stanton} with immediate practical application. 
At the same time, it is a task which was not included in VeLO's meta-training curriculum, and hence a good test of its generalization.
We closely follow (and use the codebase of) \citet{distillation_ours}, and distill the teacher into the (BiT) ResNet-50 student using the ``function matching'' approach.
Figure~\ref{fig:vit_jft4b} (right) shows that VeLO consistently outperforms the published baseline across all durations. Note that for the longer training runs, we increase the batch-size to 65K in order to stay within VeLO's step budget. 
See Section \ref{sec:bs} for discussion of VeLO's ability to make use of larger training batches than baseline methods.

\subsubsection{Object Detection with Faster R-CNN}
\begin{figure}
    \centering
    \begin{overpic}[width=0.48\textwidth]{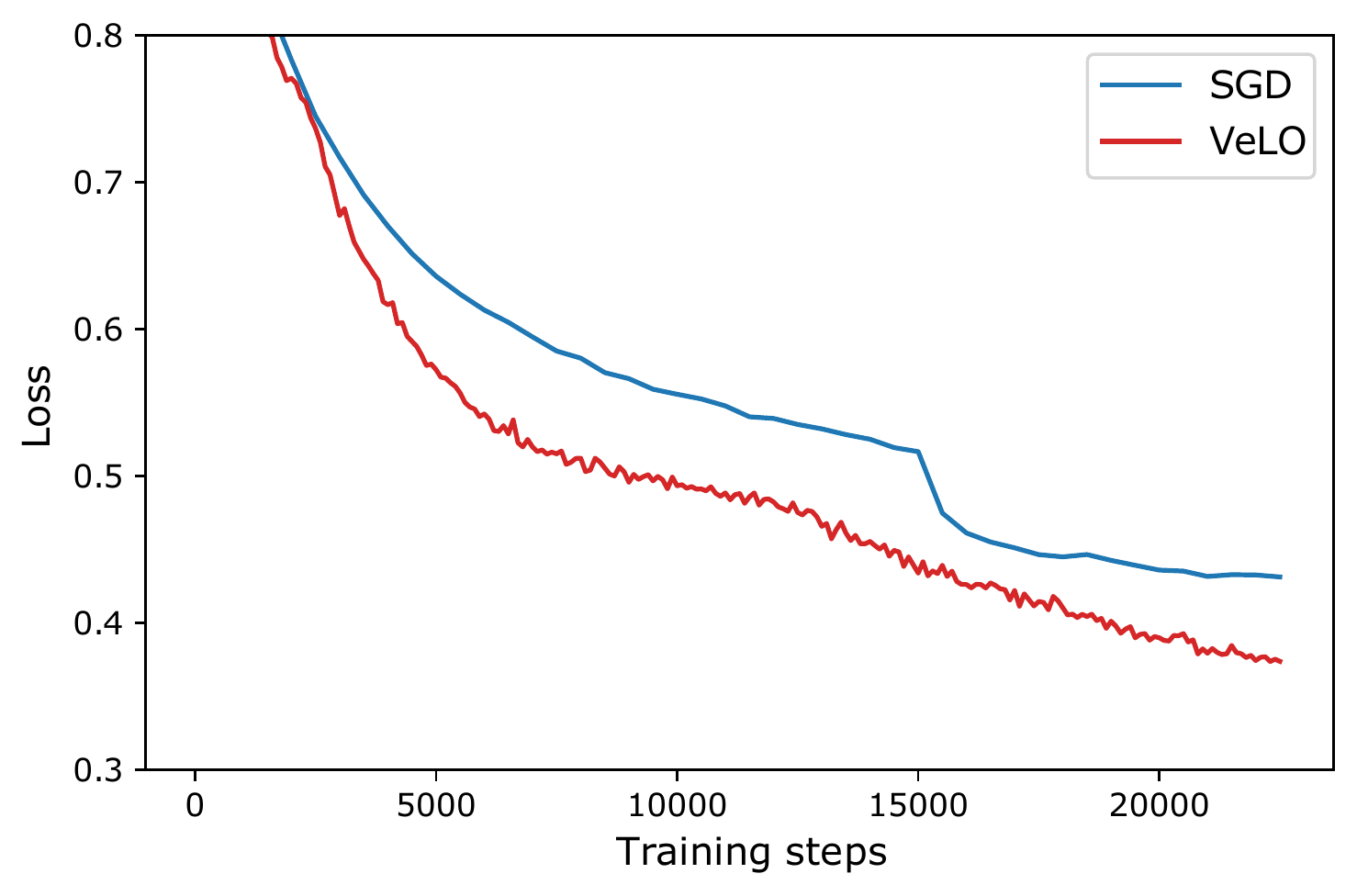}
    \put (13.3,13.0) {\textbf{(a)}}
    \end{overpic}
    \begin{overpic}[width=0.49\textwidth]{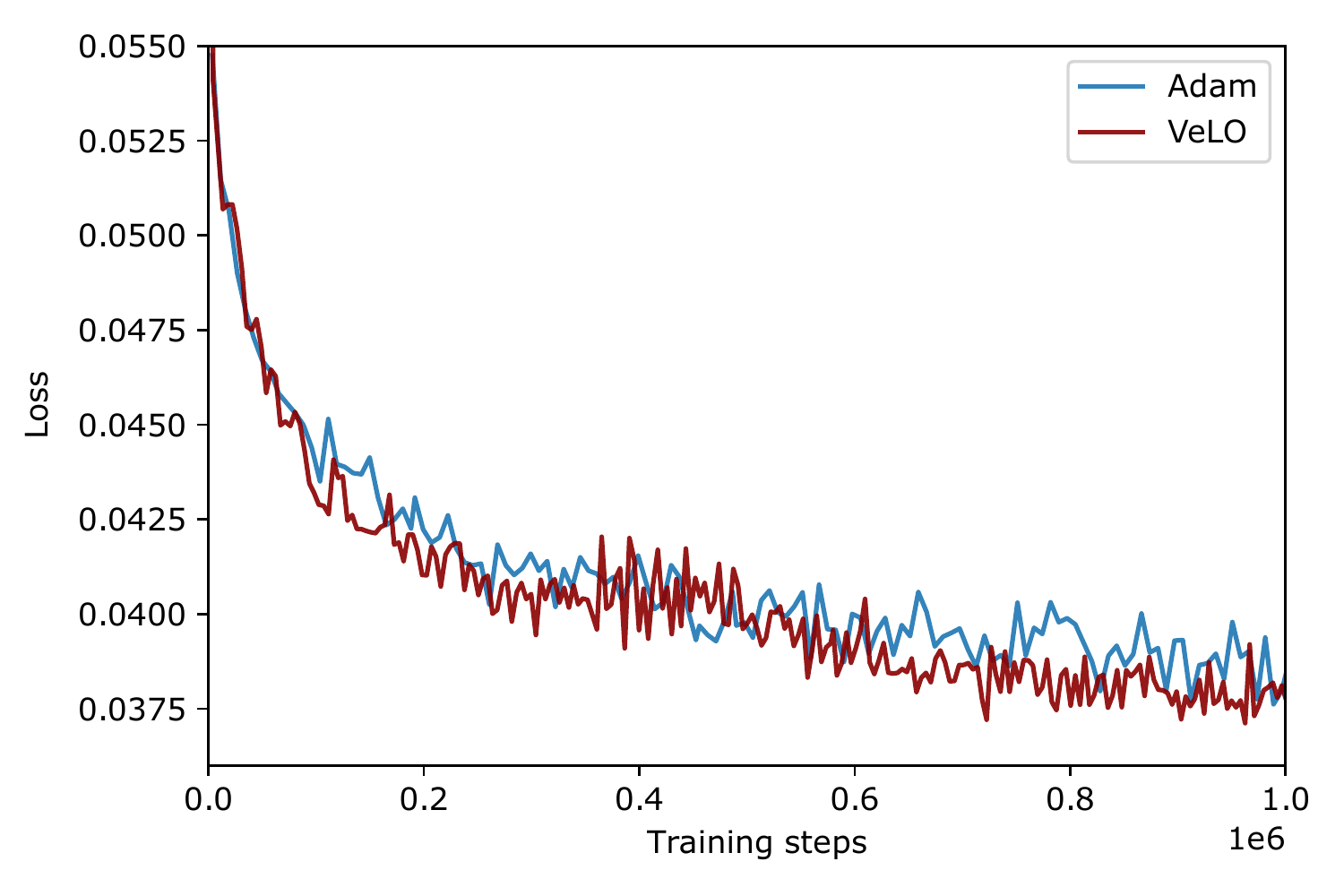}
    \put (18,13.0) {\textbf{(b)}}
    \end{overpic}
    \caption{\textbf{(a) Object detection with Faster-RCNN.} VeLO performs better on training loss than the typical SGD staircase learning rate schedule.
    \label{fig:object_detection_train}
    \textbf{(b) Learned optimizer performance on Decision Transformers.} VeLO matches the current hand-tuned Adam optimizer.
    \label{fig:decision_transformer}
    }
\end{figure}
There are no direct object detection models included in the meta-training set, making training popular models such as Faster R-CNN \citep{ren2015faster} an interesting out-of-distribution benchmark.
In Figure \ref{fig:object_detection_train}a, VeLO outperforms the standard stochastic gradient descent (SGD) optimizer with piecewise constant learning rates which is typically used for this task.

\subsubsection{Multi-Game Decision Transformers}
We also test VeLO on training large-scale Decision Transformers---a class of model which solves reinforcement learning problems via sequence modeling \citep{chen2021decision}. We train a 200M decoder-only Transformer model on sequences of interleaved images, returns, and actions. Following \citet{lee2022multi}, a single model is trained to simultaneously play 46 Atari games \citep{bellemare2013arcade}. This was previously done using a tuned Adam optimizer with custom learning rate schedule. Figure~\ref{fig:decision_transformer}b shows training curves for the learned optimizer compared with the tuned Adam baseline, showing comparable performance between the tuned baseline and VeLO. 

\subsubsection{Large Language Models}
\begin{figure}
    \centering
    \begin{overpic}[width=0.51\textwidth]{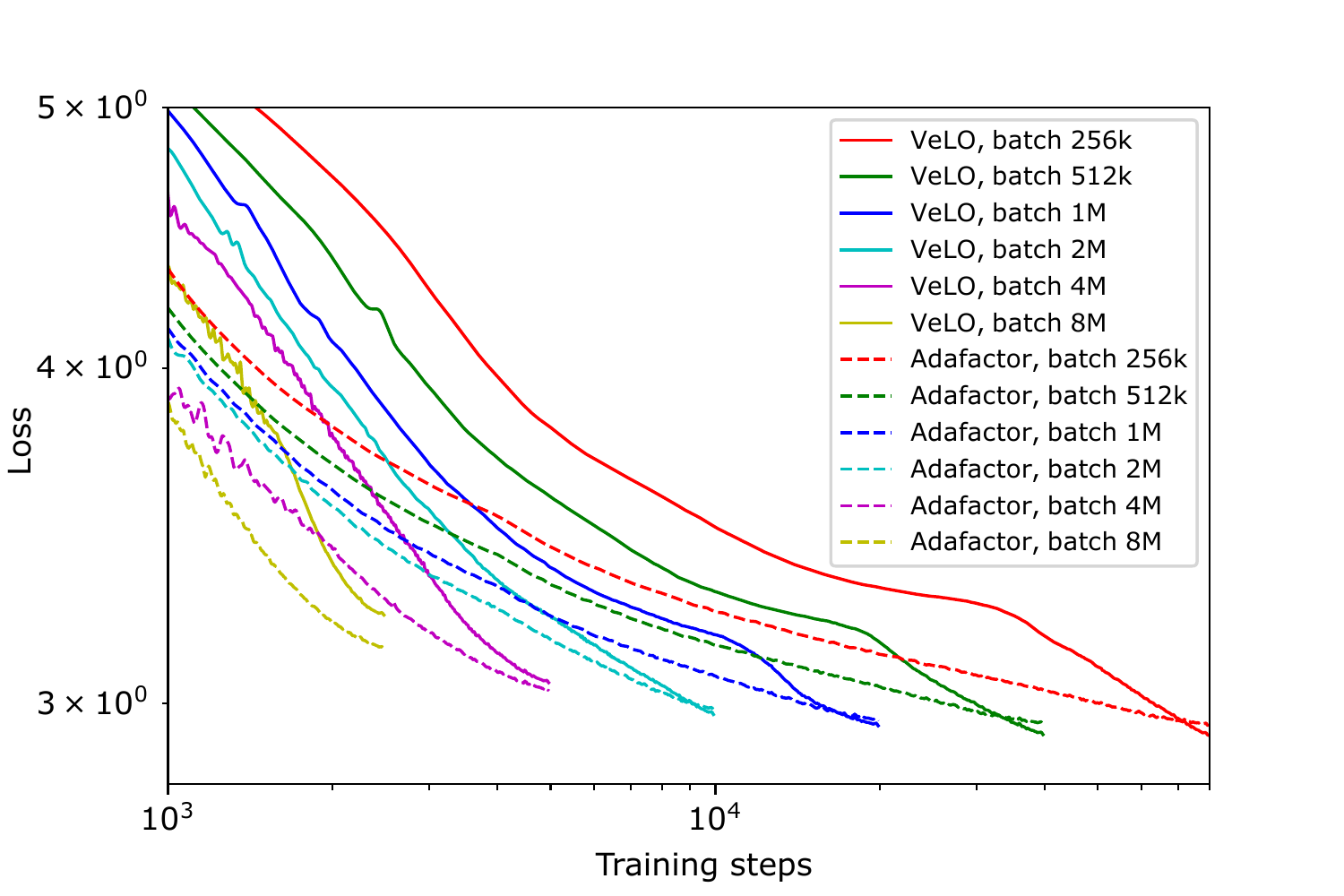}
    \put (16,13.0) {\textbf{(a)}}
    \end{overpic}
    \begin{overpic}[width=0.475\textwidth]{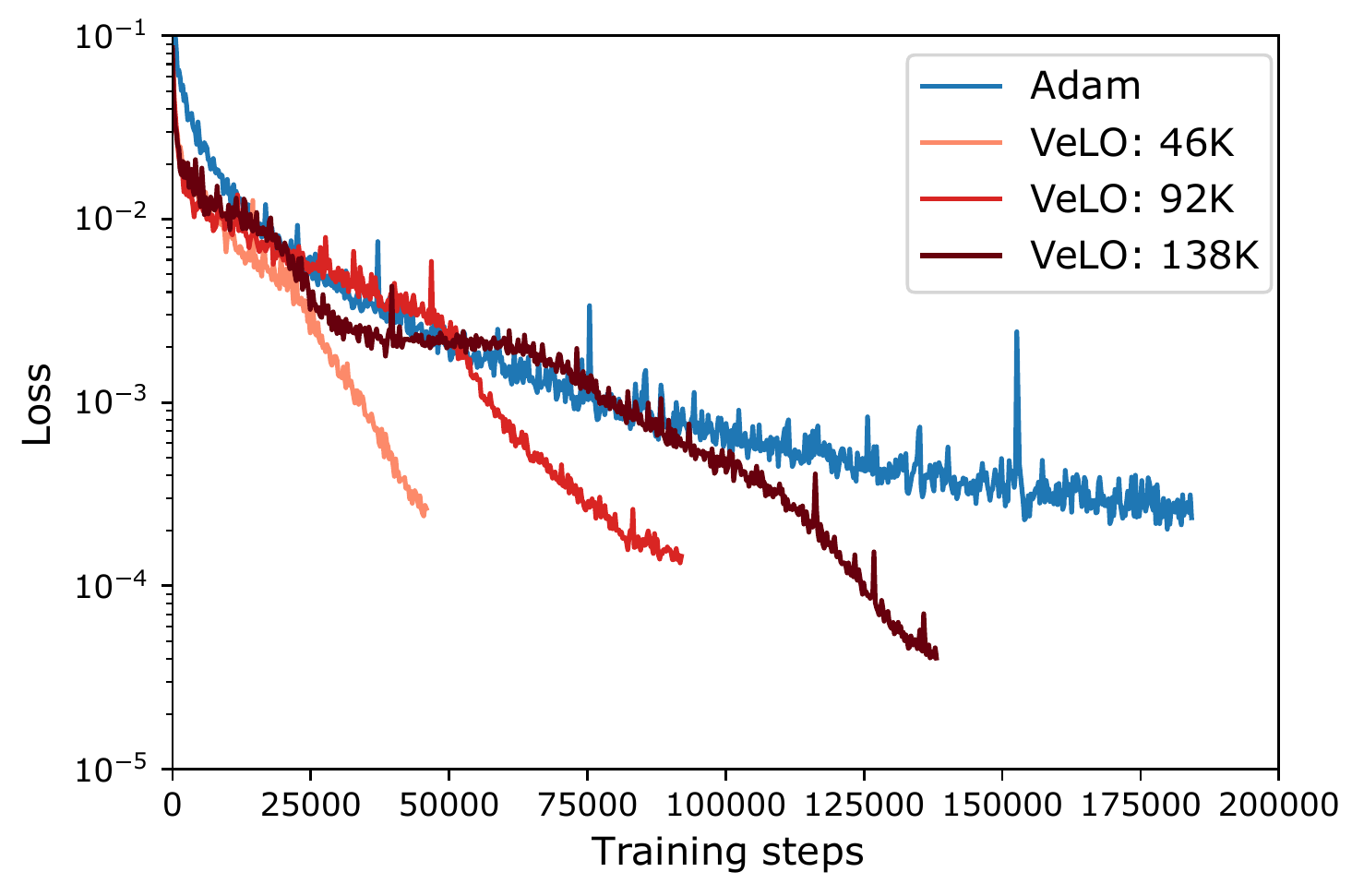}
    \put (16,13.0) {\textbf{(b)}}
    \end{overpic}
    \caption{\textbf{(a) Transformer LM training with VeLO and Adafactor.} Models with 100M parameters were trained to 20B tokens with various batch sizes. VeLO with weight decay performs better for small batch but underperforms an LR-tuned Adafactor baseline with cosine decay at large batch.
    \label{fig:100m_transformer}
    \textbf{(b) Learned optimizer applied to Graph NN predicting chemical properties of materials.} We show different training horizons in color. We find VeLO is more than 3x faster than a hyperparameter-tuned Adam for shorter training horizons, and reaches lower loss for longer horizons.
    \label{fig:physics_graph_nets}
    }
\end{figure}
A common setting for optimizing Transformer language models is a single epoch training run over a dataset with a large fixed number of tokens. In this setting, two key focus areas for optimization are increasing the maximum batch size that can be used without significantly reducing convergence, and (for conventional optimizers) identifying a learning rate schedule that anneals down to the best possible model after the specified number of steps. We show that VeLO does well on the latter (after resolving an issue with activation scales) but underperforms a heavily-tuned baseline (Adafactor with LR tuning and cosine decay) on the former.

Out of the box, VeLO performed relatively poorly on a 100M parameter Transformer language model trained on 20B tokens of C4 \citep{raffel2020exploring} at various batch sizes, with divergence on many runs. We identified that this divergence was due to unbounded growth in activation magnitudes which eventually cause precision issues, and addressed it by adding weight decay (discussed in more detail in Section \ref{sec:llm}). This growth in parameter magnitudes (and thus activation magnitudes) is likely a consequence of the these experiments being much larger than the tasks on which VeLO was trained. With weight decay of 1\e-6 (which unfortunately adds a tunable hyperparameter to VeLO), the optimization is stable and outperforms an LR-tuned Adafactor baseline at small batch, but begins to underperform at very large batch sizes (Figure \ref{fig:100m_transformer}a). 

\subsubsection{Chemical Properties of Materials with Graph Networks}
\label{sec chem gnns}

Both graph neural networks (GNNs), and scientific data are out-of-distribution tasks for VeLO, and it showed its weakest MLCommons performance on the GNN task (Section \ref{sec mlc gnn}).
Despite this, in a separate experiment with a GNN applied to scientific data, VeLO performed better than the hand-tuned baseline currently in use (Figure \ref{fig:physics_graph_nets}b). When applied for only 46K training steps, VeLO outperformed 184K training steps of the baseline optimizer.
However, as discussed in Section \ref{sec long training failure}, VeLO performs less well on longer training runs.
The model for this figure is a message-passing GNN \citep{gilmer2017neural} with 3 layers, trained to predict energies based on a dataset of known inorganic crystals from Materials Project \citep{jain2013commentary}. The input graph representation has nodes representing atoms and edges representing interatomic distances.

This evaluation differs from the MLCommons ogbg \citep{hu2020open} benchmark due to multi-edges arising from periodic boundary conditions instead of isolated molecules. Additionally, the associated task is a regression task, in comparison to the binary label prediction in ogbg. See Appendix~\ref{app:gnn_materials} for more information and generalization performance.

\subsection{Limitations and Failure Cases} \label{subsec:limitations}

In this subsection we discuss the observed limitations and failure cases of VeLO. We define failure broadly: if VeLO is not comparable in performance to a tuned baseline, we consider this to be a failure. 
This is a high bar, especially since VeLO has no hyperparameter tuning.
The failures below all occur when VeLO is asked to optimize tasks which are very unlike tasks in its meta-training distribution. 

\subsubsection{Scaling to Models Larger than 500M Parameters}

Across several domains, we observed performance decreases relative to baselines with larger model size (as measured by number of parameters). In our ViT experiments, VeLO's performance lagged behind tuned baselines for the largest models. In particular, the H/14 model (650M) notably underperformed relative to the baseline model. 

\begin{figure}[!t]
    \begin{minipage}[c]{0.625\textwidth}
    \centering
    \makebox[\textwidth]{%
    \begin{overpic}[width=\textwidth]{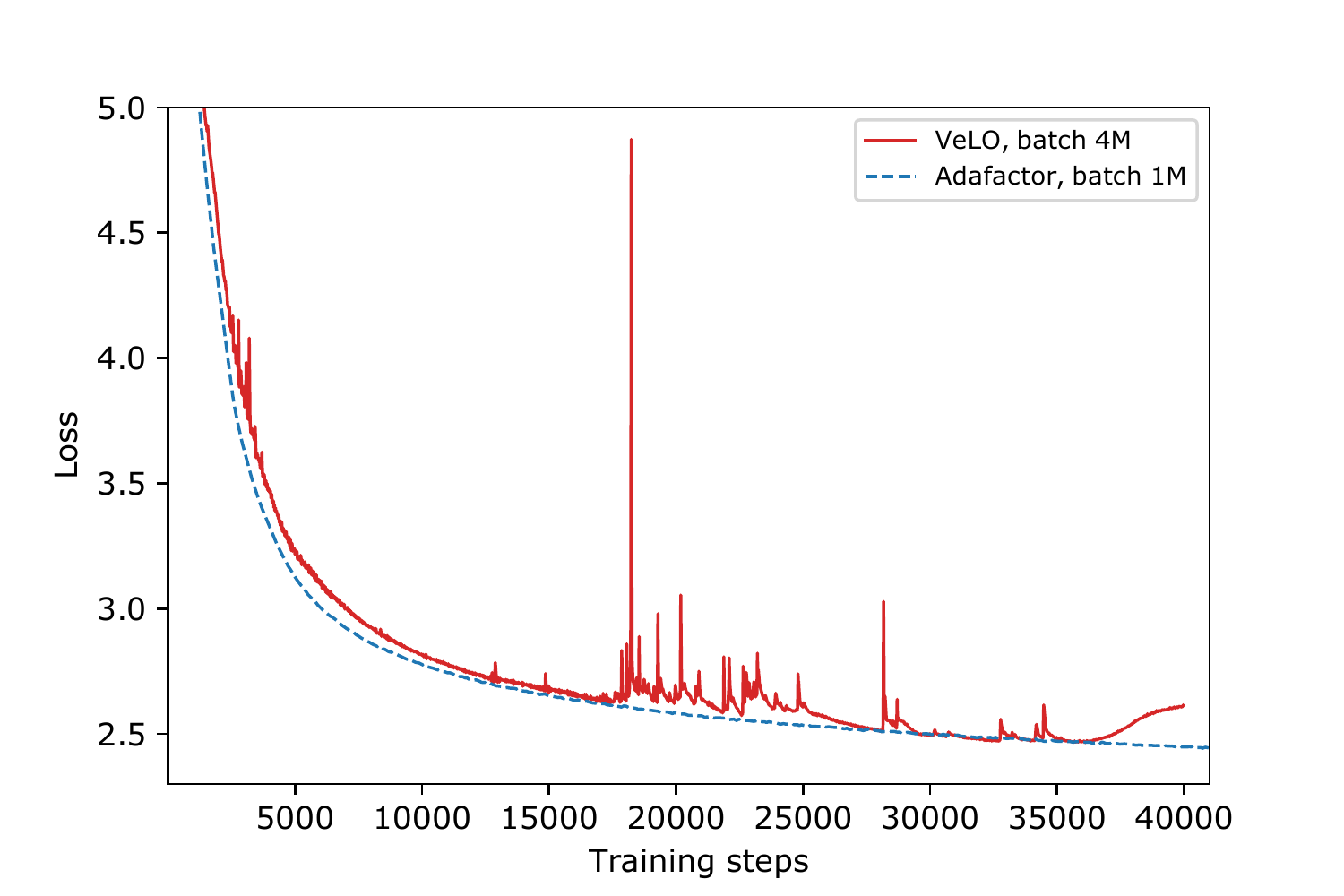}
    \end{overpic}
    }
    \end{minipage}\hfill
    \begin{minipage}[c]{0.375\textwidth}
    \caption{
    \textbf{VeLO struggles to train models which are much larger than those used for meta-training} Plot shows training of an 8B parameter Transformer language model, trained to 160B tokens. VeLO exhibits instability even with weight decay, and underperforms an untuned Adafactor baseline with exponential learning rate decay on a step-for-step basis despite 4x larger batch.
    \label{fig:8b_transformer}
    }
    \end{minipage}
\end{figure}

In our LLM experiments, VeLO performed relatively poorly and was fairly unstable for the largest evaluated models. Figure \ref{fig:8b_transformer} shows a single run of an 8B parameter Transformer trained on 160B tokens of C4. VeLO (with weight decay of 1\e-6) underperforms an untuned Adafactor baseline with 1/4 the batch, even on a step-for-step basis, although we emphasize that this model size is far out of the meta-training distribution.

The performance of VeLO lags behind baselines, or even decreases, as model size is increased beyon approximately 500M parameters. These models are far out of the meta-training distribution, which contains only a small number of tasks which are even 
5\% this size. 
Achieving better meta-generalization to training large scale models---and ideally, consistent performance for any size of model---is an important direction for future work.

\subsubsection{Scaling to Longer than 200K Training Steps}
\label{sec long training failure}

VeLO was originally meta-trained on optimization tasks which involved 20K or fewer steps of inner training. VeLO was then finetuned on problems with up to 200K inner training steps (see Appendix \ref{app:interactive_hparams} for more details). 
We find that VeLO's performance relative to baseline optimizers worsens as the number of training steps approaches, and then exceeds, 200K.

A particularly dramatic example of this is the GNN task from Section \ref{sec chem gnns}. There, VeLO is far more effective than the tuned baseline when used for up to $\sim$150K training steps. When applied for 200K steps or longer however, it begins to perform worse not only relative to the baseline, but also in an absolute sense. 

\subsubsection{Extended Training}

It is common to extend the training of a model after an initial run. For example, in transfer learning, a pretrained model is finetuned on a different dataset and/or objective to perform a specific task or set of tasks. Underfit models may also have their training continued on additional data or epochs of the same dataset.

Since VeLO is conditioned on the total number of iterations during inner-training and the inner parameters were always initialized to a random state during meta-training, we find that it struggles to extend training beyond its initially specified number of iterations. The resulting behavior differs depending on how the optimizer state is set for the extended training, but in many cases it is possible to partially remedy the poor performance. We explored the following options for continuation from a completed VeLO training run:
\begin{itemize}
    \item \emph{Na\"ive Continue}: Continue from the final optimizer state of the previous run, allowing the iteration number to be greater than the total number of steps the optimizer is conditioned on: a state that was never observed during meta-training.
    \item \emph{Full Reset}: Initialize the optimizer state from scratch, as is done at the beginning of a standard training run, conditioning on the length of the continuation run.
    \item \emph{Reset Steps}: Continue from the final optimizer state of the previous run but reset the iteration to 0 and the number of steps to the length of the continuation run. 
    \item \emph{Increase Steps}: Continue from the final optimizer state of the previous run but increase the number of steps the optimizer is conditioned on to be the sum of the lengths of the initial run and continuation.
\end{itemize}

\begin{figure}
    \centering
    \begin{overpic}[width=0.49\textwidth]{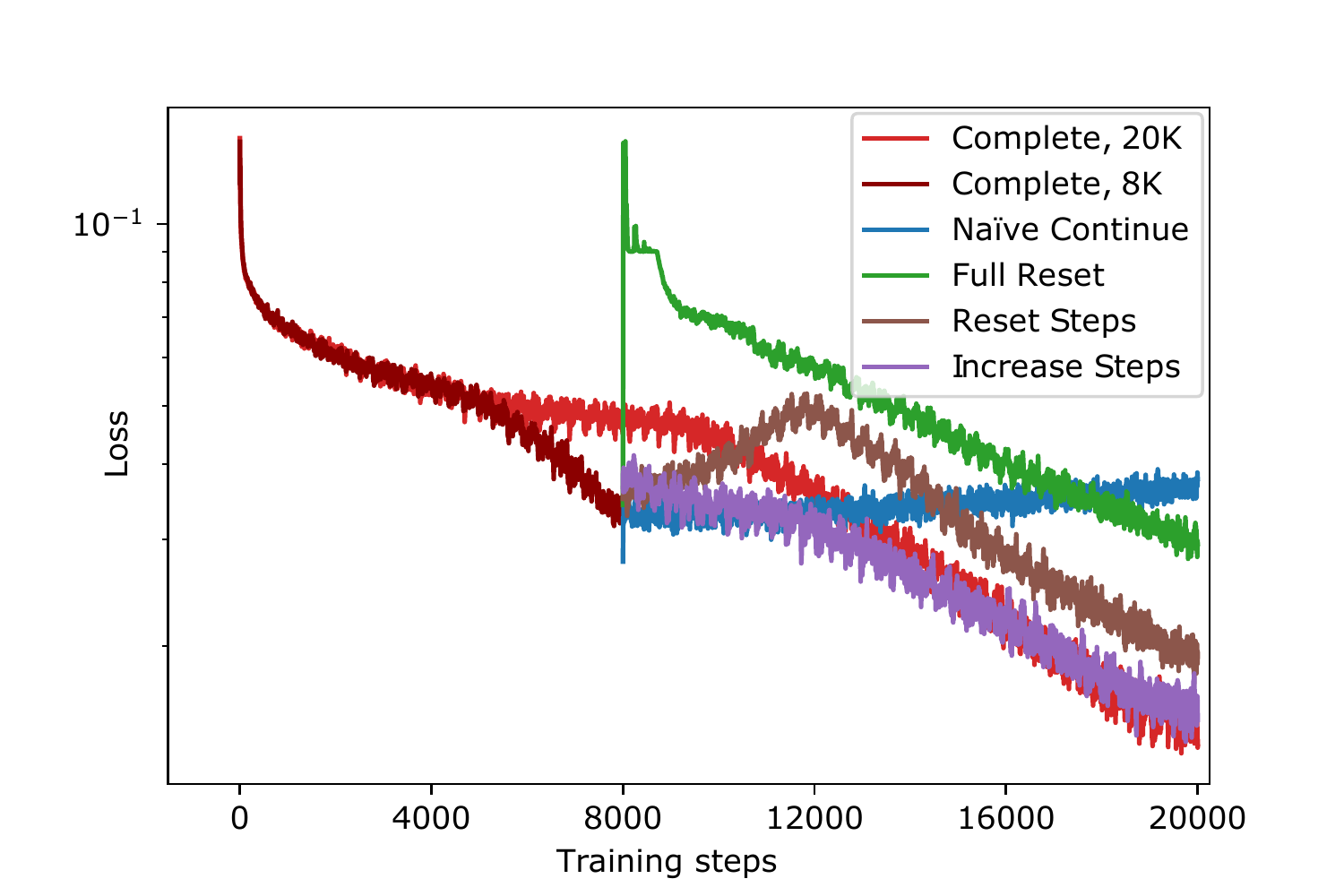}
    \put (15,12.0) {\textbf{(a)}}
    \end{overpic}
    \begin{overpic}[width=0.49\textwidth]{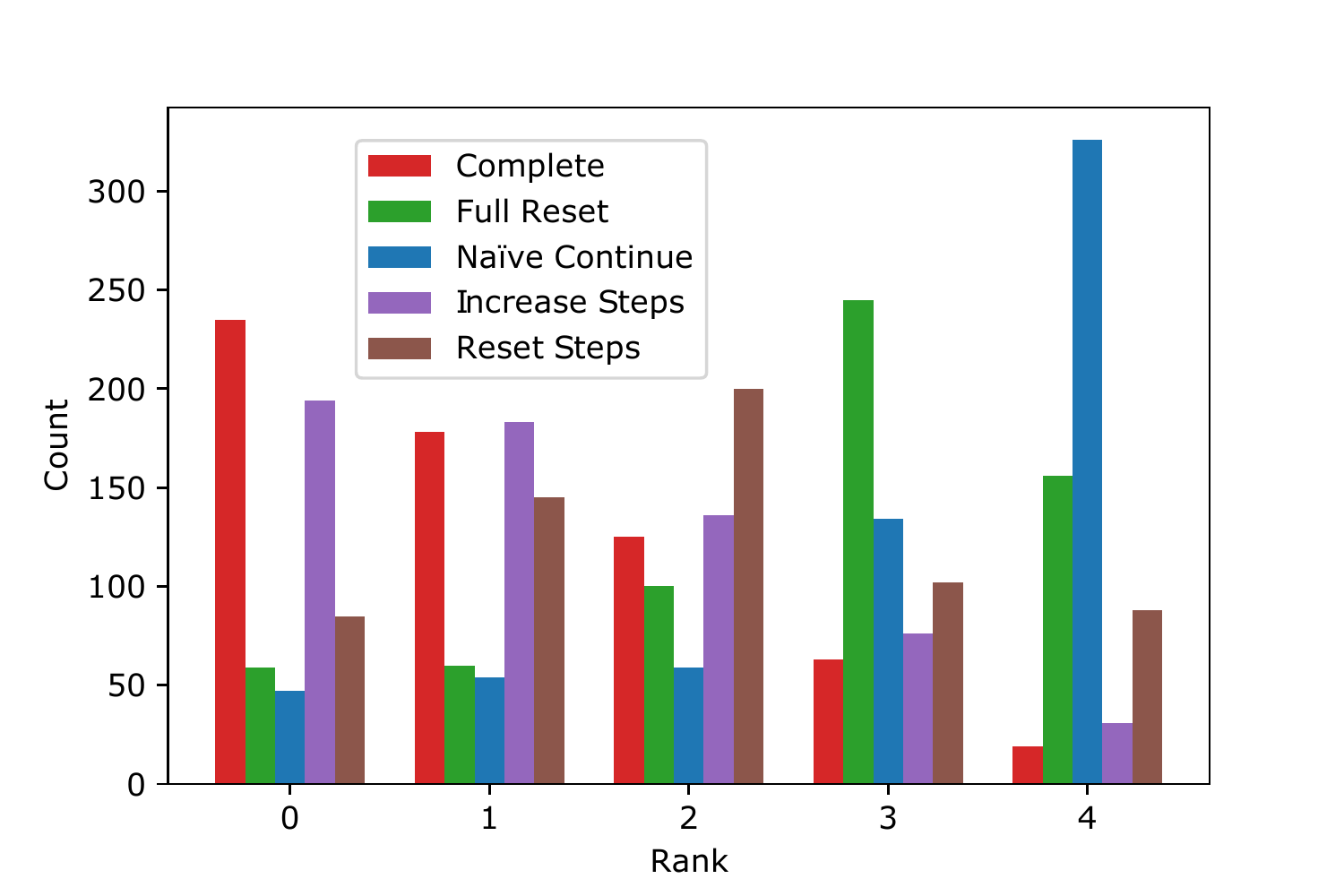}
    \put (15,52.0) {\textbf{(b)}}
    \end{overpic}
    \caption{\textbf{(a) Comparison of methods for extending a VeLO training run on example problem.} We train an image MLP model on a CIFAR10 for 8,000 steps with VeLO (``Complete, 8K") and then extend training for an additional 12,000 steps with VeLO, adjusting the optimizer state in several ways, as described in the text.
    \textbf{(b) Histogram of ranks of final training loss for each continuation method in large-scale evaluation of 620 tasks.} We train each task for 20,000 total steps, splitting the steps across 2 VeLO runs (except for "Complete") at a point sampled from $N(10000, 5000^2)$, and applying each of the continuation methods.}
    \label{fig:continuation_comparisons}
\end{figure}

We compare these approaches anecdotally (Figure~\ref{fig:continuation_comparisons}, left) and by evaluating them on 620 different tasks (Figure~\ref{fig:continuation_comparisons}, right) with continuation points sampled from a Normal distribution centered at half of the total training steps. When comparing the final training losses, we find that \emph{na\"ive continue} performs the worst in 52\% of experiments and observe anecdotally that it tends to diverge slowly throughout the run. \emph{Full reset} is second worst, often resulting in a large initial spike in the loss, followed by a slow partial recovery. \emph{Increase steps} performs better than all other continuation methods in 49\% of the experiments, the next best being \emph{reset steps} for 25\%. However, doing the complete training in a single run performs best overall in 38\% of the experiments, beating \emph{increase steps} in 54\% of them.

We also explore the behavior of VeLO when extending training from a model partially trained with a different optimizer, which is sometimes done during finetuning. In Figure~\ref{fig:adam_continuation} we see that, similar to the \emph{full reset} VeLO continuation, the loss immediately spikes. This demonstrates that VeLO has limited ability to generalize from a non-random initial state.

\subsubsection{Reinforcement Learning}

\begin{figure}
\centering    
\makebox[\textwidth]{%
    \begin{overpic}[width=0.5\textwidth]{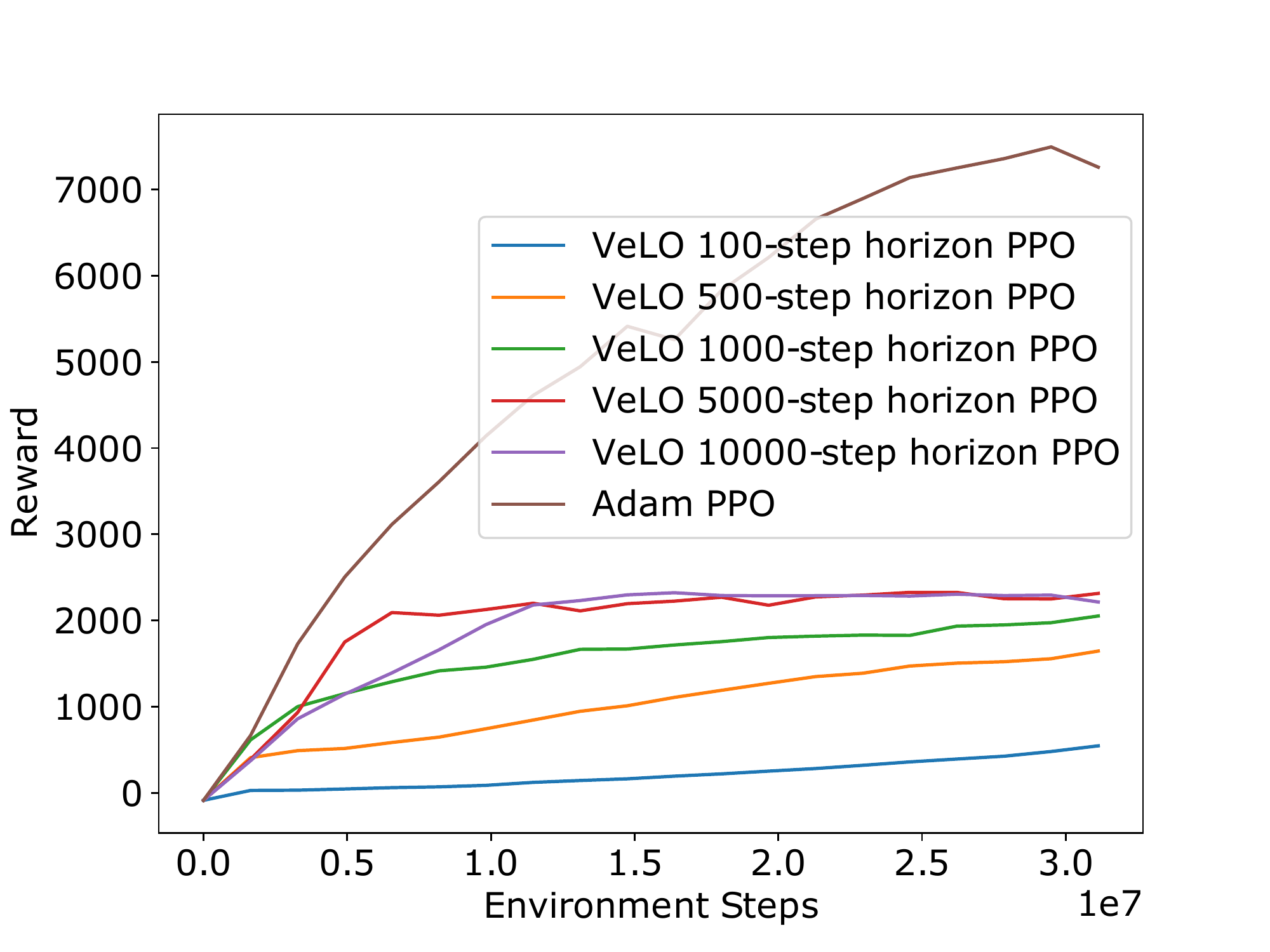}
    \put (14.5,60.0) {\textbf{(a)}}
    \end{overpic}
    \begin{overpic}[width=0.5\textwidth]{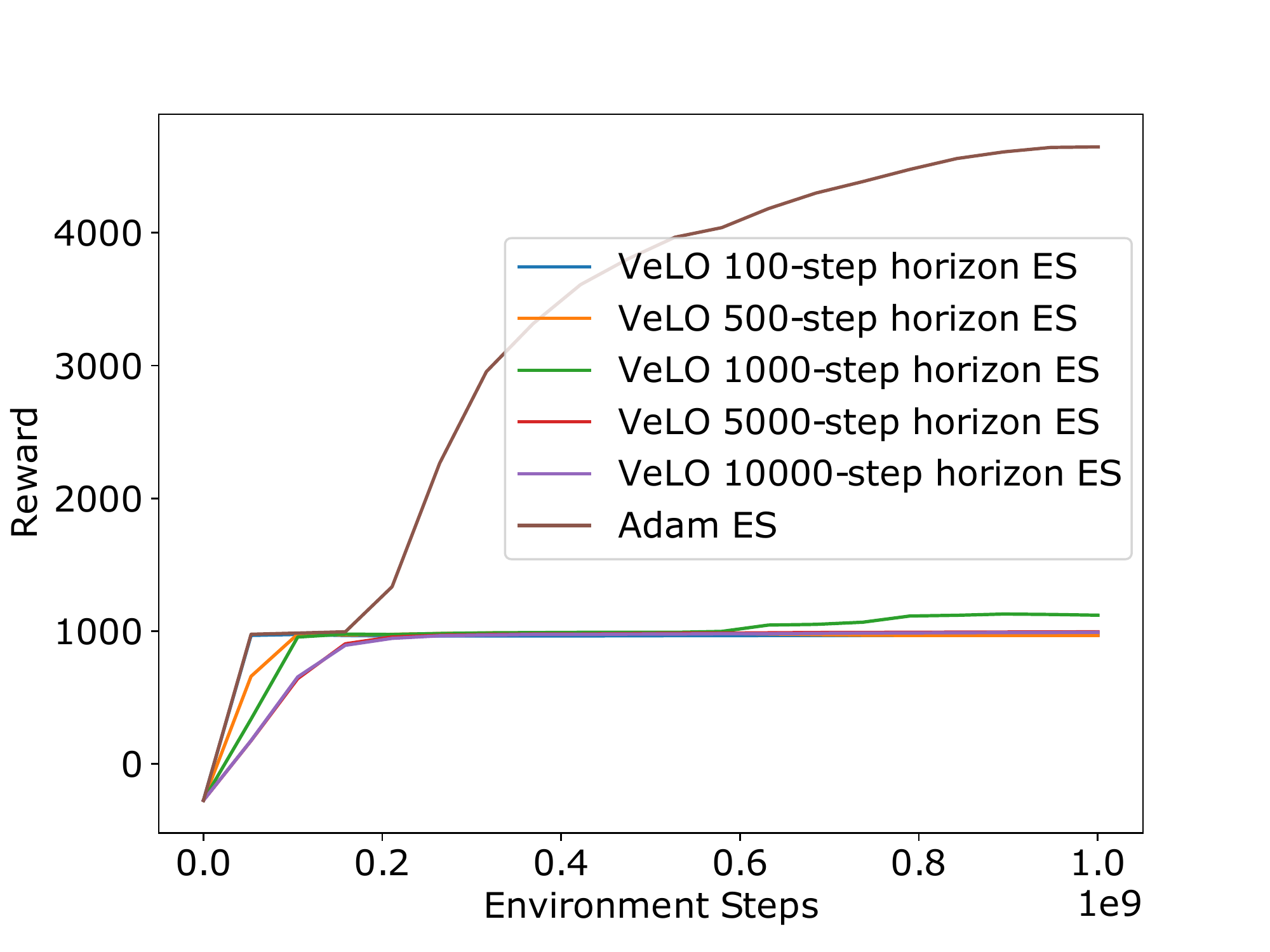}
    \put (14.5,60.0) {\textbf{(b)}}
    \end{overpic}
    }
    \caption{\textbf{VeLO struggles to train the standard Ant task in reinforcement learning. VeLO was not meta-trained on any reinforcement learning tasks, and so this corresponds to out-of-domain generalization.} 
    We compare VeLO against Adam-aggregated gradients in \textbf{(a)} PPO and \textbf{(b)} ES.  While the learned optimizer is able to learn a locomotive gait when aggregating PPO gradients, it does not perform as well as Adam, and in the case of ES, it fails to learn a locomotive gait at all, learning only to stand in place.  Different lopt curves indicate different target number of training steps fed into the learned optimizer as a feature.}
    \label{fig:brax_rl}
\end{figure}

To test how far out of the meta-training domain we could push VeLO, we also considered continuous control reinforcement learning tasks using the Brax physics engine \citep{freeman2021brax}. We consider the Ant task, which requires optimizing a locomotion policy for an 8-degree-of-freedom quadruped (Figure \ref{fig:brax_rl}). We consider two optimization problems: (a) PPO \citep{schulman2017proximal}, which optimizes a 4-layer, 32-neuron fully connected policy network as well as a 5-layer, 256-neuron value network, and (b) ES \citep{rechenberg1973evolutionsstrategie}, which only optimizes the policy network. Both are targeting the same reward function, which roughly measures how far the ant locomotes to the right. For both PPO and ES, we compare VeLO against Adam, a standard baseline. While the learned optimizer is able to find a locomotive gait when aggregating PPO gradients, the reward it reaches is considerably lower than the score achieved by default Adam. Additionally, in the case of ES, VeLO fails to escape the local minimum of ``standing still'', where the Ant does not move at all at initialization.

\section{Understanding Learned Optimizer Behavior}

Learned optimizers can behave in more diverse ways than hand-designed optimizers. 
At the same time, they are often even more inscrutable than hand-designed optimizers, since their complex functional form means their behavior must be characterized with techniques designed to study black box systems
\citep{maheswaranathan2020reverse}.
In this section we experimentally characterize aspects of VeLO's behavior.

\subsection{VeLO Adapts to Training Horizon}
\begin{figure}
    \centering
    \makebox[0.8\textwidth]{%
    \begin{overpic}[width=1.0\textwidth]{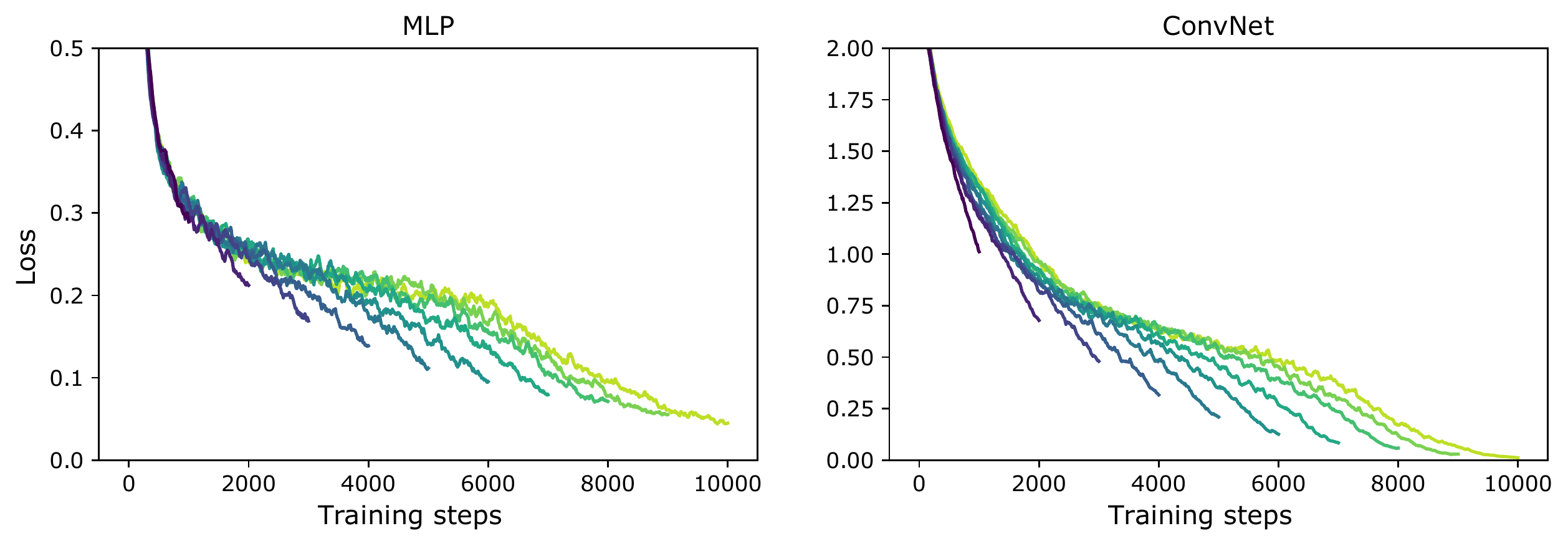}
    \end{overpic}
    }
    \caption{\textbf{Learned optimizers take into account the target training length.} We vary the length of inner training from 1K to 10K steps (shown in different colors) for an MLP and ConvNet.
    For both problems, near the end of training the loss decreases rapidly.
    \label{fig:num_step_aware}
    }
\end{figure}

Learning rate decay is a simple yet powerful technique to increase performance near some pre-specified end of training, and is commonly used in hand-designed optimizers in a variety of problems.
Motivated by this, our optimizer has access to an embedding of the fraction through training of the current iteration.
To probe how VeLO makes use of this feature, we train two tasks using VeLO for different lengths of time (Figure~\ref{fig:num_step_aware}).
We observe that VeLO intelligently makes use of this feature and drops the loss dramatically just before the end of training.

\subsubsection{VeLO Learns an Implicit Step Size Schedule}\label{sec:different_sized_steps}
\begin{figure}
    \centering
    \makebox[1.0\textwidth]{%
    \begin{overpic}[width=1.0\textwidth]{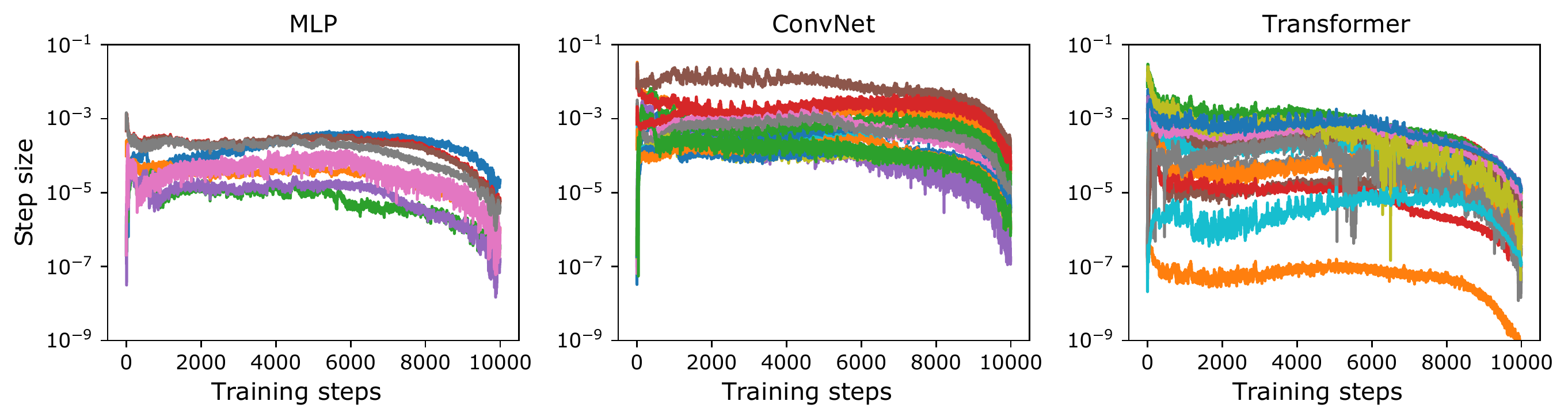}
    \put (8,6.5) {\textbf{\small(a)}}
    \put (40.5,6.5) {\textbf{\small(b)}}
    \put (73,6.5) {\textbf{\small(c)}}
    \end{overpic}
    }
    \caption{\textbf{VeLO adapts update step length to task, parameter type, and training iteration.}
    We monitor the step size of each tensor (shown in different colors) for 3 different models---a 3 hidden layer MLP trained on CIFAR10, a 3 hidden layer ConvNet with batch norm~\citep{Ioffe15} trained on CIFAR10, and a small Transformer trained on LM1B.
    We find the learned optimizer takes different sized steps across each model type, as well as different sized steps for different tensors.
    Additionally VeLO learns schedules which are shared across tasks including a rapid step size increase, and a gradual step size decay.
    \label{fig:3step_size}
    }
\end{figure}
One way in which VeLO uses information about the fraction through training is by adjusting its parameter update steps on an implicit schedule. 
To illustrate this, we train 3 models, a small MLP, a ConvNet, and a Transformer, and monitor the size of step taken for each tensor over the course of training (Figure~\ref{fig:3step_size}).
First, we note there is large variation in step size not only between different tasks, but also between different parameter tensors within the same task---differences can be as large as 6 orders of magnitude!
This level of variation is not generated by any of the hand-designed optimizers we examine.
Second, we note the implicit schedule learned by VeLO. We see signs of step size warm-ups, as well as a step size decay.
These features where not encoded by us, and result entirely from the meta-training process.
For further details and experiments showing a comparison of loss values, and different baseline optimizers, see \ref{app:additional_step_size}.

\subsection{VeLO Can Have a Larger Critical Batch Size than Baseline Optimizers} 
\label{sec:bs} 
Training on large batches is extremely important for large scale distributed training, as it lowers the communication cost between chips and enables increased utilization of hardware.
Prior work~\citep{shallue2018measuring, mccandlish2018empirical, zhang2019algorithmic} has shown that one can increase the batch size while proportionally decreasing the number of weight updates (maintaining a fixed number of total gradient evaluations) up until a point where performance starts to fall of, which has been referred to as the critical batch size.
It has been shown that optimizers that make use of momentum, and/or more sophisticated preconditioners can be used to increase this critical batch size~\citep{zhang2019algorithmic}. 
We explore whether VeLO can make effective use of batches that are larger than the critical batch size for hand-designed optimizers.

\begin{figure}
    \centering
    \makebox[\textwidth]{%
    \begin{overpic}[width=0.8\textwidth]{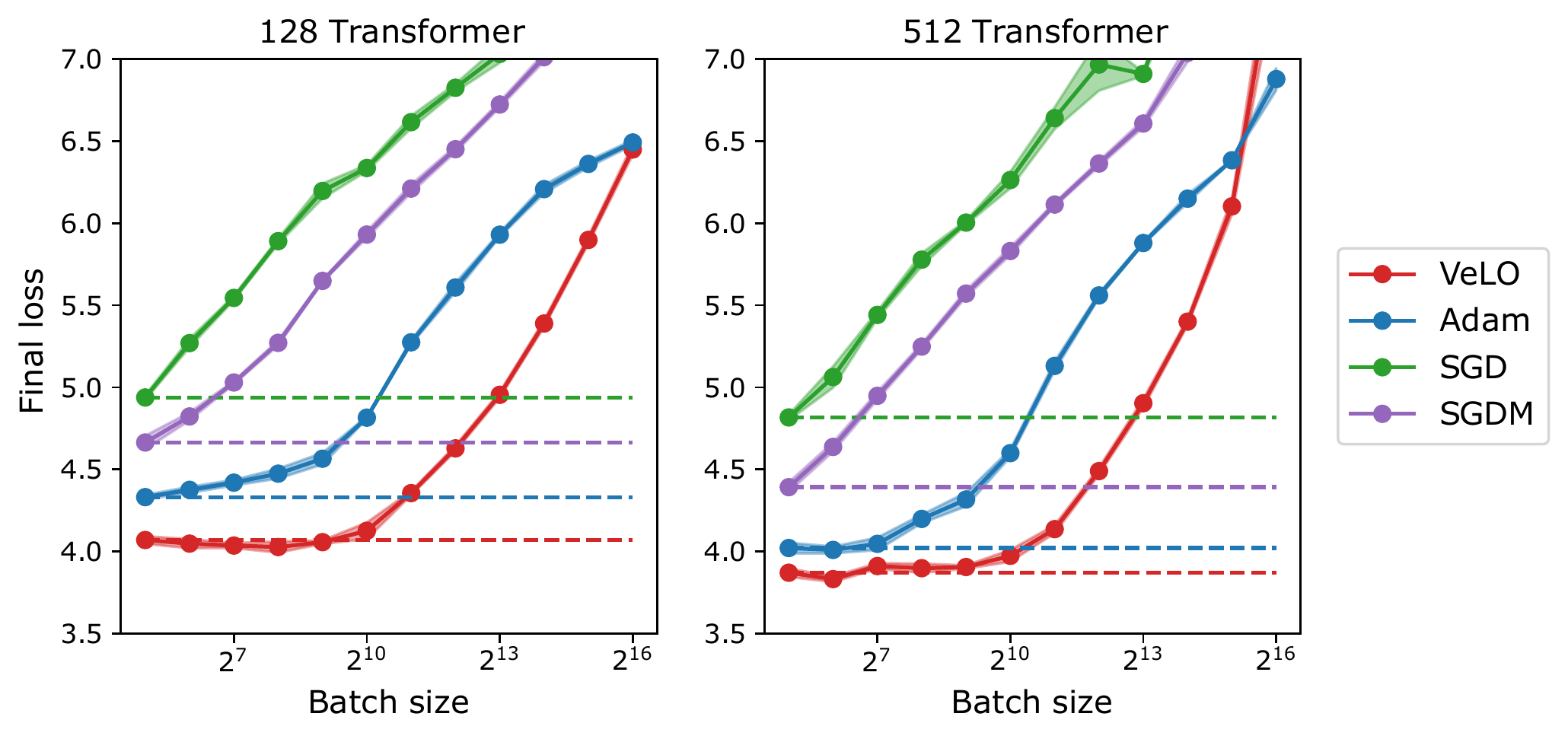}
    \end{overpic}
    }
    \caption{\textbf{VeLO makes efficient use of larger larger batch sizes than hand-designed optimizer baselines.} We show the final loss achieved by VeLO, and learning rate-tuned Adam, SGD, or SGDM after a fixed number of examples seen for different batch sizes.
    For all baselines each point represents the minimum over a learning rate search.
    In dashed lines, we show what optimal batch size scaling would look like (no decrease in performace when increasing batch size).
    VeLO can make use of batch sizes up to 10x as large as Adam before seeing performance degradation.
    \label{fig:bs_scaling}
    }
\end{figure}

We take two 5 layer Transformers with 128 and 512 dimensions and sweep the batch size while simultaneously decreasing the number of steps, keeping the total number of examples seen the same (Figure~\ref{fig:bs_scaling}).
For all models, we train for $2^{19}$ examples.
Thus, for a batch size of $2^{16}$, we only make 8 training steps.
For each batch size, we tune learning rates of Adam, SGDM, and SGD, selecting the best one. We only use a single trial of VeLO.
In addition to doing considerably better than the baseline optimizers, VeLO makes use of significantly larger batch sizes.
In the case of the two Transformers, VeLO has a critical batch size around 10x larger than baseline methods. See Appendix~\ref{app:experiment_more_bs} for this result on 4 additional models.

We note that this increased critical batch size does not appear to hold for larger models than those investigated in this subsection; for example, Figure\ref{fig:100m_transformer} does not show meaningful changes to critical batch size. This is likely due to the these 100M parameter Transformer models being far out of the meta-training distribution, compared to those investigated in this subsection. 
See also Figure~\ref{fig:init2winit_bs} which explores even larger sized problems on the MLCommons set of tasks.

\section{Related Work}
The idea of meta-learning update rules for optimization dates back to \citet{bengio1992optimization, runarsson2000evolution} which both learn simple update rules on simple neural networks.
More recently, \citet{andrychowicz2016learning} revived the topic by meta-training an RNN-parameterized learned optimizer on deep learning tasks by backpropogating through the optimization procedure~\citep{maclaurin2015gradient}.
Since then, there has been a flurry of new techniques, ranging from learned optimizer architectures to meta-training algorithms.

Closest to our work is the line of work on hyperparameter-free, neural-network parameterized learned optimizers trained on large distributions of tasks.
\citet{wichrowska2017learned} introduced hierarchical learned optimizers, similar to our work, and meta-trained them on a large distribution of synthetic tasks.
\citet{metz2020tasks} train on a more realistic task distribution~\citep{metz2020using}, with an improved learned optimizer architecture.

In this work, we leverage ES for meta-training~\citep{rechenberg1973evolutionsstrategie, nesterov2011random, salimans2017evolution}.
There has been extensive work studying different meta-training techniques ranging from ES improvements~\citep{maheswaranathan2019guided, metz2019understanding, pmlr-v139-vicol21a}, to reinforcement learning~\citep{li2016learning, li2017learning}, to techniques designed specifically to train learned optimizers~\citep{lv2017learning,chen2020training}.

In contrast to general-purpose learned optimizers, task specific learned optimizers have been proposed in many settings, including chemistry \citep{learn2hop}, robustness \citep{metz2019using}, adversarial training \citep{xiong2020improved},
few-shot learning \citep{ravi2016optimization},
min-max optimization \citep{shen2021learning},
human motion reconstruction \citep{gartner2022prep},
unsupervised learning~\citep{metz2018learning},
swarm optimization \citep{cao2019learning},
black box optimization \citep{chen2016learning},
and MCMC sampling \citep{levy2017generalizing,wang2017meta,gong2018meta}.

Improvements to the LSTM learned optimizer architecture in~\citet{andrychowicz2016learning} have also been proposed. \citet{lv2017learning} modify the inputs to improve training; \citet{metz2019understanding} swap out the LSTM with an MLP; \citet{premont2022simple} introduce the ability to fall back to a hand-designed optimizer to ensure convergence.
Hyperparameter controllers---neural networks which dynamically set the hyperparameters of existing optimizers---have also been explored \citep{daniel2016learning, xu2019learning, almeida2021generalizable}. 
Finally, work has been done to meta-learn symbolic parameter update rules.
\citet{Bello17} use reinforcement learning to learn a policy which produces symbolic optimizers.
\citet{zheng2022symbolic} first meta-train a neural network, and subsequently distill it to a symbolic form.
\citet{real2020automl} explores learning not just an optimizer, but an entire symbolic learning algorithm.

\section{Discussion and Outlook} 
In this work, we demonstrated dramatic improvements in the generality and performance of learned optimizers, by scaling up meta-training compute and dataset size, and by making architectural improvements.
The resulting learned optimizer, VeLO, has no hyperparameters, yet outperforms heavily hyperparameter-tuned baselines across more than 80 diverse optimization tasks, including large published machine learning models.

\subsection{Open Questions}
Despite the performance of this optimizer, there are many open questions and clear directions for improvement.
These include: improving the learned optimizer architecture, for instance to leverage second order or quasi-second order information; using more available information about the target task, for instance by conditioning on an embedding of the target task's computation graph;
more control to target both validation and training loss; reverse-engineering the techniques used by the learned optimizer, e.g.~as in \citet{maheswaranathan2020reverse}; and improving the computational efficiency of meta-training, for instance by using analytic gradients or partial unrolls.
See Appendix~\ref{app:open_questions} for an extended discussion of these and other open questions.

\subsection{Meta-Learned Algorithms are the Future}
One of the core lessons from the modern machine learning revolution is that, given enough compute and training data, learned algorithms can outperform even the most well-motivated hand-designed heuristics.
In this paper, we show that this lesson applies to the parameter update function used to train a neural network, though the required compute scales are far larger than for most supervised learning tasks (by a factor of around 100 million).
Similar benefits from meta-learning have been demonstrated in neural architecture search \citep{zoph2017, tan2019mnasnet} and data augmentation \citep{cubuk2018autoaugment}.

Almost every part of a typical machine learning pipeline is still built out of hand-designed heuristics, from defining the loss function, to choosing regularizers, to designing training curricula, to specifying transfer learning procedures.
As compute and data continue to grow, we expect that meta-learned algorithms will replace all of these hand-designed components.

\section*{Acknowledgements}
First, we thank the entire JAX team, in particular Matt Johnson, Roy Frostig, Peter Hawkins, Blake Hechtman, and Qiao Zhang, for answering our numerous questions and supporting our unusual use cases, as well as Sharad Vikram, for supporting our use cases in Oryx.
We thank Rohan Anil and Zack Nado for assistance with PAX as well as providing guidance on model configurations.
We thank the init2winit~\citep{gilmerinit2winit} team including Zack Nado, Justin Gilmer, George Dahl, Sourabh Medapati for the wonderful library making our MLCommons experiments possible.
We thank Erik Gärtner for writing the initial version of the JaxNerf tasks.
We would like to thank Erica Moreira for support with computing resources.
We thank Paul Vicol and Kelvin Xu for detailed feedback on a draft version of this paper.
We thank Chip Huyen for their support and feedback. 
Finally, we would like to thank Doug Eck, Zoubin Ghahramani, Jeff Dean and the rest of the Brain team for building a supportive research environment.

\bibliography{references}
\bibliographystyle{plainnat}

\appendix

\section{Author Contributions}

\paragraph{Luke Metz:}
Led the project, wrote the majority of infrastructure, and designed VeLOdrome.

\paragraph{James Harrison:} 
Led the final stages of the project (after Luke's departure).
Contributed to the design and implementation of VeLO, ran experiments, and contributed to writing and project organization.

\paragraph{C. Daniel Freeman:} Contributed to VeLO design and infrastructure, ran RL experiments, and contributed to writing.

\paragraph{Amil Merchant:} Contributed to infrastructure, ran detection and graph network experiments.

\paragraph{Lucas Beyer:} Contributed vision transformer and distillation experiments. 

\paragraph{James Bradbury:} Contributed large language model experiments. 

\paragraph{Naman Agarwal:} Implemented (on top of existing code) and ran the MLCommons Training Algorithms benchmark Adam baselines for different batch sizes.

\paragraph{Ben Poole:} Contributed to the design of VeLO and contributed to paper writing. 

\paragraph{Igor Mordatch:} Ran Decision Transformer experiments and contributed to paper writing.

\paragraph{Adam Roberts:} Proposed methods for extending VeLO training and analyzed their performance.

\paragraph{Jascha Sohl-Dickstein:} Supervised the project, contributed to learned optimizer and infrastructure design and theory, and contributed to paper writing.

\section{Learned Optimizer Architecture} \label{app:lopt_architecture}
In this section we present the architecture of VeLO.
We first present an overview of the hierarchical structure.
We then discuss each component and their input features, as well as discussing connections to hyperparameter-controller optimizer architectures and discussing optimizer complexity with respect to the complexity of the underlying model. 

\subsection{Extended Architecture Overview}
Most hand-designed optimizers consist of a handful of element-wise operations and are therefore relatively inexpensive to compute.
Adam \citep{kingma2014adam}, for example, is typically written as three (coupled) update equations applied independently to each parameter of the underlying model.
Learned optimizers, on the other hand, can make use of more complex functional forms, possibly parameterized by neural network.
While this greater expressivity results in greater representational power, it can can also be considerably more costly to compute.
For learned optimizers to be useful, this extra compute per-step must be balanced with the improvements in per-step optimization efficiency.
Recently, \citet{metz2022practical} showed that per-parameter learned optimizers can be parameterized by a extremely small neural network, and thus efficient to compute, while still outperforming tuned hand-designed optimizers.
While these small models are capable of strong performance on single tasks, we found that they lack the representational capacity needed to perform well across many tasks (See Figure~\ref{fig:capacity_perf}, in particular the MLP LOpt and AdaFac LOpt).

Past work has introduced hierarchy in learned optimizer parameterizations to address this limitation \citep{wichrowska2017learned, metz2020tasks}.
This hierarchy enables more expressive representations, while allowing for shared computation across either a full tensor, layer, or the entire network. 
Results of the upper level computation are then routed down and treated as conditioning inputs to the lower levels of the hierarchy.
For example, per-tensor computations are fed into the per-parameter network.
While this does greatly increase the capacity (and thus performance) of the learned optimizer without requiring excessive additional compute, it does not fix the core expressivity restriction of the per-parameter optimizers, as this per-parameter network would still need to perform a different computation depending on the conditioning thus potentially requiring a more expressive, and thus more expensive function.

To address the computational bottleneck at the parameter optimizer level, we propose a new class of learned optimizer architecture based on HyperNetworks~\citep{ha2016hypernetworks}.
Our optimizer consists of a two-layer hierarchy. 
The upper level acts at the tensor level, and is a recurrent network.
Instead of this tensor-level network passing information to the per-parameter layer as additional inputs, we instead have the network produce the entire weight matrix of the per-parameter network. 
This decouples the computational requirements of the upper and lower layer networks, and allows for increased expressivity without additional compute cost at the lower level.
To make this prediction easier, we parameterize it as a linear combination of some larger set of per-parameter network weights.

In the following sections we describe all the details of this model starting with what state it accumulates iteration to iteration, followed by a description of the per-tensor and per-parameter networks.

\subsection{Optimizer State: Non-Learned Accumulators}
VeLO keeps track of the following information from iteration to iteration. These features are used by both the per-parameter, and per-tensor network and are updated with each new gradient and loss value.

\begin{itemize}
    \item \textbf{Iteration number.} The number of iterations the learned optimizer has been applied.
    \item \textbf{Momentum at different timescales.} We keep track of an exponential moving average of gradient values at multiple timescales. We use 3 timescales (0.9, 0.99, 0.999).
    \item \textbf{Squared gradients.} We track the exponential moving average of the gradient squared, similarly to what is accumulated in Adam~\citep{kingma2014adam}. We track these squared gradients using a single EMA coefficient of 0.999.
    \item \textbf{Adafactor-style accumulators.} As in~\citet{metz2022practical} we also track a factorized variant of the squared gradients~\citep{shazeer2018adafactor}. The size of these accumulators is sub-linear in parameter count, so we track 3 different exponential moving averages of these values: (0.9, 0.99, 0.999).
    \item \textbf{Loss features.} We additionally track details about the loss values. This is parameterized as an exponential moving average of each training loss value at multiple timescales. To be precise, we construct 10 values between $1$ and $ \log(\text{num\_steps})$, and then to determine decay we use $\exp(-1/x)$ where $x$ is each one of these 10 values. Comparing the results from different timescales will allow our optimizer to react to loss changes over time. We also keep track of a running minimum of each of the different exponential timescales. See the following section for how these values are used. 
\end{itemize}

\subsection{The Tensor-Level Recurrent Network}

\begin{figure}
    \centering
    \makebox[\textwidth]{%
    \begin{overpic}[width=0.5\textwidth]{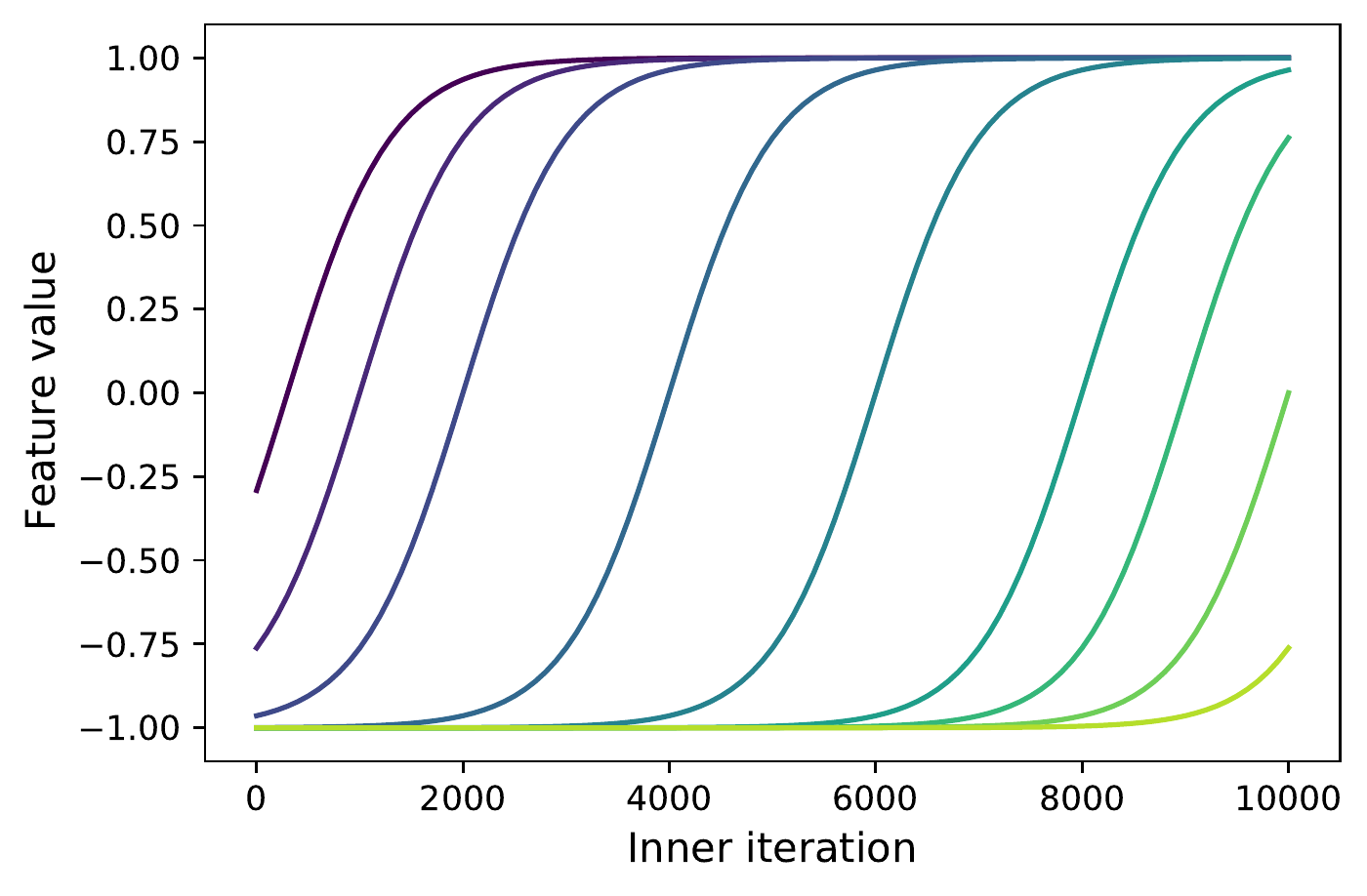}
    \end{overpic}
    }
    \caption{\textbf{Visualization of fraction of time left features.} These values are fed into the per-tensor network and are used by the learned optimizer to change behavior in a time dependent manner.  \label{app:fig:fraction_left}
    }
\end{figure}

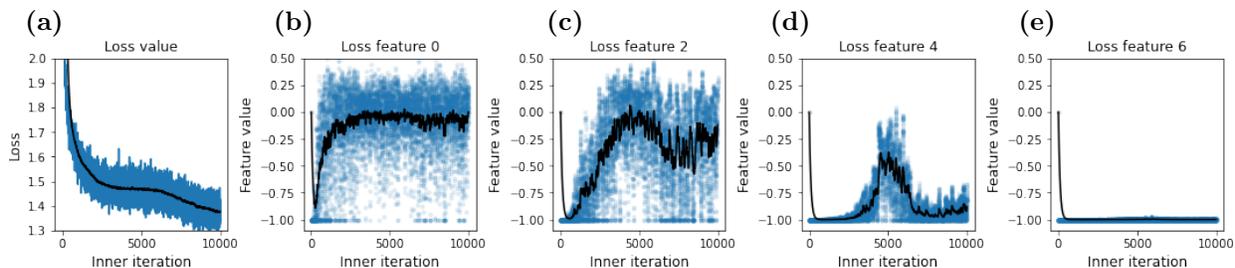
\begin{figure}
    \centering
    \makebox[\textwidth]{%
    \begin{overpic}[width=1.0\textwidth]{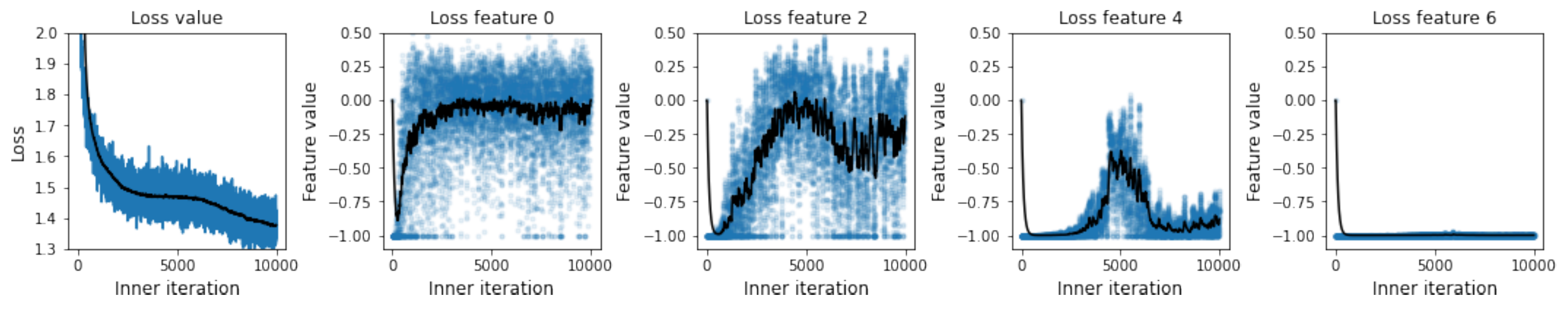}
    \put (2,20.0) {\textbf{\small(a)}}
    \put (22,20.0) {\textbf{\small(b)}}
    \put (44,20.0) {\textbf{\small(c)}}
    \put (62 ,20.0) {\textbf{\small(d)}}
    \put (82,20.0) {\textbf{\small(e)}}
    \end{overpic}
    }
    \caption{\textbf{Visualization of the loss features.} Results computed when applying a learned optimizer to the \texttt{RNNLM\_lm1bbytes\_Patch32\_LSTM128\_Embed64} task. \textbf{(a)} We show the training loss. In blue we show the raw values, which are passed into the learned optimizer. In black we show an exponential smoothed version with 0.99 decay. \textbf{(b-e)} We show different length scales of the loss features again where blue is raw values and black is smoothed by a 0.99 decay. First, in (b) we look at the shortest timescale of loss features. Because there is noise in the underlying loss values, this value is also quite noisy around 0 meaning in this timescale it is hard to tell if the loss is decreasing or not. In (c) we see a larger timescale, which is negative indicating the loss lowering until around 5K iterations at which point the loss plateaus (as seen in figure (a)). In (d) we see an even more smoothed version of this, with a lower maximum value. Finally in (e), one of the longer timescales, we see a near constant value of -1 indicating that the loss is always going down. \label{app:fig:loss_features}
    }
\end{figure}

The tensor-level network is an LSTM~\citep{hochreiter1997long}, with 512 hidden units which operates on features which are broadly reflective of overall training status and bulk statistics of per-parameter features. These are described below.
\begin{itemize}
    \item \textbf{Fraction of training remaining.} We use as an input the current training iteration, as well as the number of target steps to produce a set of values representing a soft progress through training. This is implemented by computing $\text{tanh}(10*(t/T - s))$ where $t$ is the current iteration, $T$ is the total number of steps, and $s$ is the fraction through training. We used $s \in [0.03, 0.1, 0.2, 0.4, 0.6, 0.8, 0.9, 1.0, 1.1]$ but did not ablate these values. See Figure~\ref{app:fig:fraction_left} for a visualization of these features.
    \item \textbf{Loss features.} These features enable the optimizer to tell if optimization is converging or diverging.
    We leverage both the exponential moving averages of loss, and the running minimum to construct a loss magnitude-invariant featurization.
    Values of -1 roughly correspond to a loss going down, values of 1 mean diverging, and values of 0 mean no change in loss.
    We have 10 loss timescales, and use pairs of these timescales to construct these features.
    We visualize 4 of these features along with the loss values which generated them in Figure~\ref{app:fig:loss_features}.
    For the exact implementation see the \href{https://github.com/google/learned_optimization/blob/687e72e7b5596dfb80c5196fd51f43058899edd9/learned_optimization/research/general_lopt/hyper_v2.py#L80}{code}.
    \item \textbf{First and second moment features.} We use the variance of each of the momentum timescales as an input feature, and both the mean and variance of the second moment features. Each feature $x$ is processed as follows: $\text{clip}(\text{log}(10^{-8} + \text{abs}(10x)), -5, 5)$. This ensures the inputs to the tensor-level network remain on a consistent scale and appropriate dynamic range.
    \item \textbf{Tensor rank.} We use a one hot of the rank of the tensor as an additional feature.
\end{itemize}

For each tensor, we compute these features, which are fixed length, and form a rank 2 array with a leading "batch" dimension being the number of tensors.
Before feeding these values into the per-tensor LSTM, we seek to ``mix'' information across the tensors.
To do this, we first apply a small neural network which operates on the feature dimension and consists of 2 linear layers with 512 units with ReLU activations.
Inspired by~\citet{zaheer2017deep}, we additionally mix this information by taking the max across the number of tensors dimension and adding this value to the original features projected linearly to 512 dimensions.
In code this operation looks like: $F_0(x) + \text{max}(\sigma(F_1(\sigma(F_2(x)))), \text{axis}=0, \text{keep\_dims}=True)$, where $F_i$ is a linear layer, and $\sigma$ is a ReLU activation.

The tensor-level model, given the mixed features described above, acts independently across each tensor. 
From this recurrent model, we output a scalar $c_{\text{lr}}$ which is used to modulate the step size for all tensor elements and a vector, $c_\text{hyper}$ which is used to linearly interpolate between a meta-learned collection of weights of the same shape as the per-parameter MLP network.

\subsection{The Parameter-Level Network}
The per-parameter optimizer we use is based on the optimizer investigated in~\citet{metz2022practical}.
The per-parameter weights for a given tensor are computed by taking an un-normalized weighted average (with weights $c_\text{hyper}$) of a bank of differently initialized, and meta-learned per-parameter weights.
The output of this is then multiplied by 100, a scaling factor to keep the MLP weights in a reasonable range at initialization.
Because these weights are computed per tensor, each tensor now has a different set of per-parameter weights.

The per-parameter features are normalized by the second moment across the entire tensor (each feature is normalized independently) and passed into the weights produced by the tensor-level LSTM hypernetwork.
This produces 2 values: $d$, $m$, which are combined (along with the scalar learning rate $c_{\text{lr}}$ also produced by the RNN), and 2 scaling hyperparameters, both 0.001, chosen to keep the outputs of this network to a reasonable range, as well as the current parameter norm.
In math, weight update to parameter vector $p$ is: $\Delta p = 0.001 * d * \text{exp}(0.001 * c_{\text{lr}}) ||p||_2$

\subsection{Comparing our Architecture to Hyperparameter Controllers} \label{app:compare_architecture_capacity}
Hyperparameter controller-based learned optimizers are another class of learned optimizer architecture, which aim to automate the tedious tuning inherent to common optimizers~\citet{daniel2016learning, xu2019learning, almeida2021generalizable}.
These optimizers map from training features to the hyperparameters of chosen optimizers (such as tuning Adam learning rates). 
One can view our work as a more complex version of this parameterization.
Our HyperNetwork LSTM is producing a small number of of weights, which are then used throughout the target network much in the same way hyperparameters act. The main difference, however, is instead of these small number of weights controlling a hand-designed optimizer, they instead control the weights of a neural network.

\subsection{Experimental Validation of Our HyperNet Compared to Past Work}

\begin{figure}
    \centering
    \makebox[\textwidth]{%
    \begin{overpic}[width=0.7\textwidth]{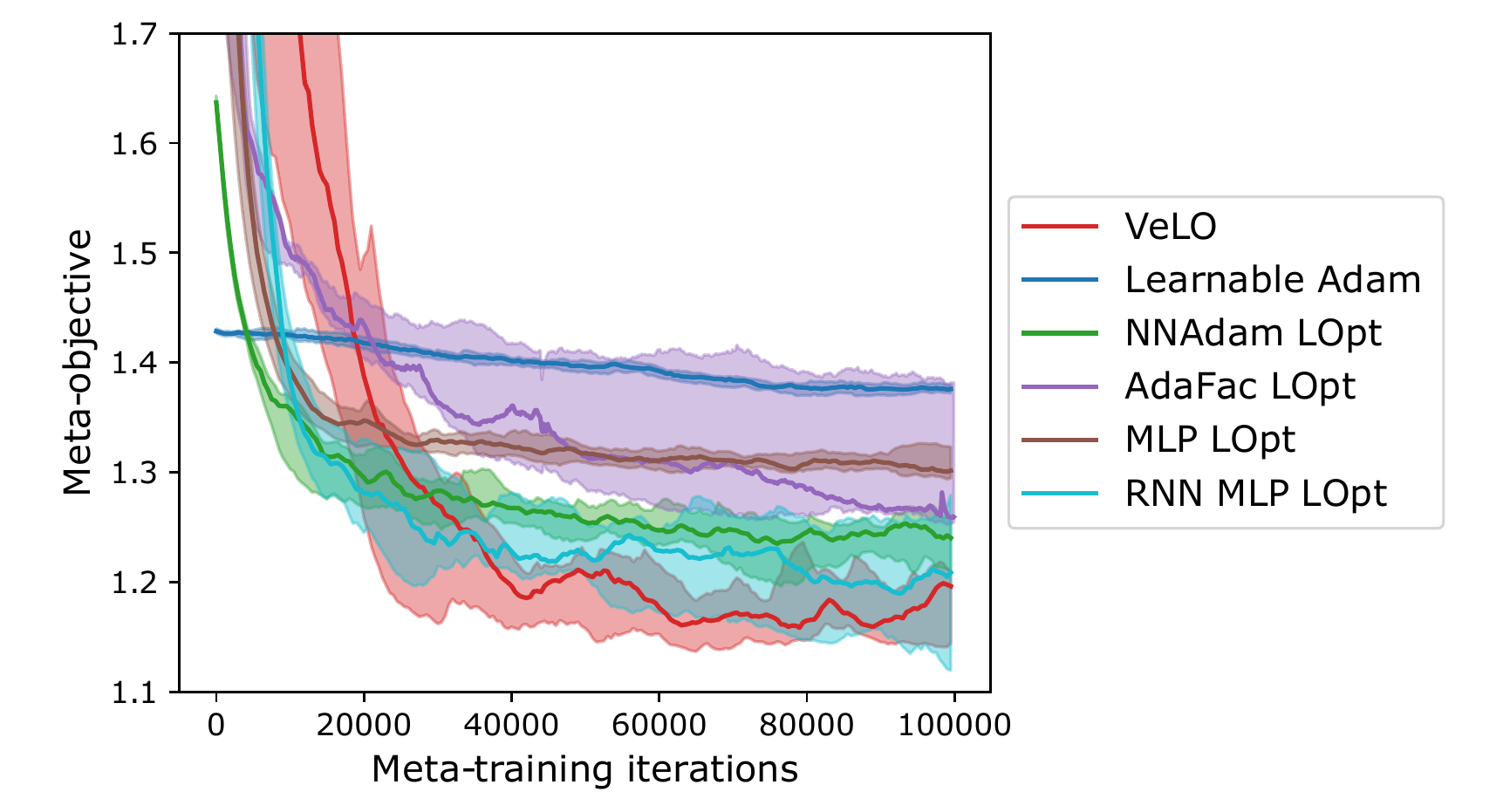}
    \end{overpic}
    }
    \vspace{-1em}
    \caption{\textbf{Compare capacity of different LOpts.} We meta-train 6 different learned optimizers on a mixture of small MLP tasks. We find the hierarchical networks (NNAdam -- a hyperparameter controller for Adam, RNN MLP LOpt and HyperNet LOpt) outperform the non-hierarchical (MLP -- AdaFac LOpt, and MLP LOpt) with our proposed learned optimizer performing the best of all methods. In the shaded regions we show the min and max performance across 3 random seeds, and the median in solid.
    \label{fig:capacity_perf}
    }
\end{figure}

To demonstrate this, we explore training different learned optimizers on a small scale, multi-task distribution of problems consisting of training a 1 hidden layer, 32 unit MLP on 4 different 8x8 datasets. See the \href{https://github.com/google/learned_optimization/blob/c8a4d4dac6987d70020e9894d871a493d9c21e72/learned_optimization/research/general_lopt/tasks/fast_mlp_diff_data.py#L58}{code} for more info on these tasks.
We show the per parameter optimizers from \citet{metz2019understanding} (MLPLOpt) and \citet{metz2022practical} (AdaFac LOpt), hand-designed optimizers (Learnable Adam---Adam where we meta-learn the hyperparameters), hyperparameter controller learned optimizers (NNAdam~\citep{metz2022practical}), the hierarchical optimizer from ~\citet{metz2020tasks} (RNN MLP LOpt), and our HyperNetwork based learned optimizer (HyperNet LOpt).
We show meta-training learning curves in Figure~\ref{fig:capacity_perf} for each optimizer.
We find our HyperNetwork based optimizer has low meta-loss, implying that this architecture has a higher capacity and thus can learn a function which performs optimization better. Previously proposed hierarchical learned optimizers also perform well (RNN MLP LOpt), but in practice are more expensive to compute. For each model we due a small learning rate search testing half powers of 10 and show performance over 3 random seeds.

\subsection{Understanding Computational Costs of VeLO} \label{app:computational_costs_of_lopt}
\begin{figure}
    \centering
    \makebox[\textwidth]{%
    \begin{overpic}[width=1.0\textwidth]{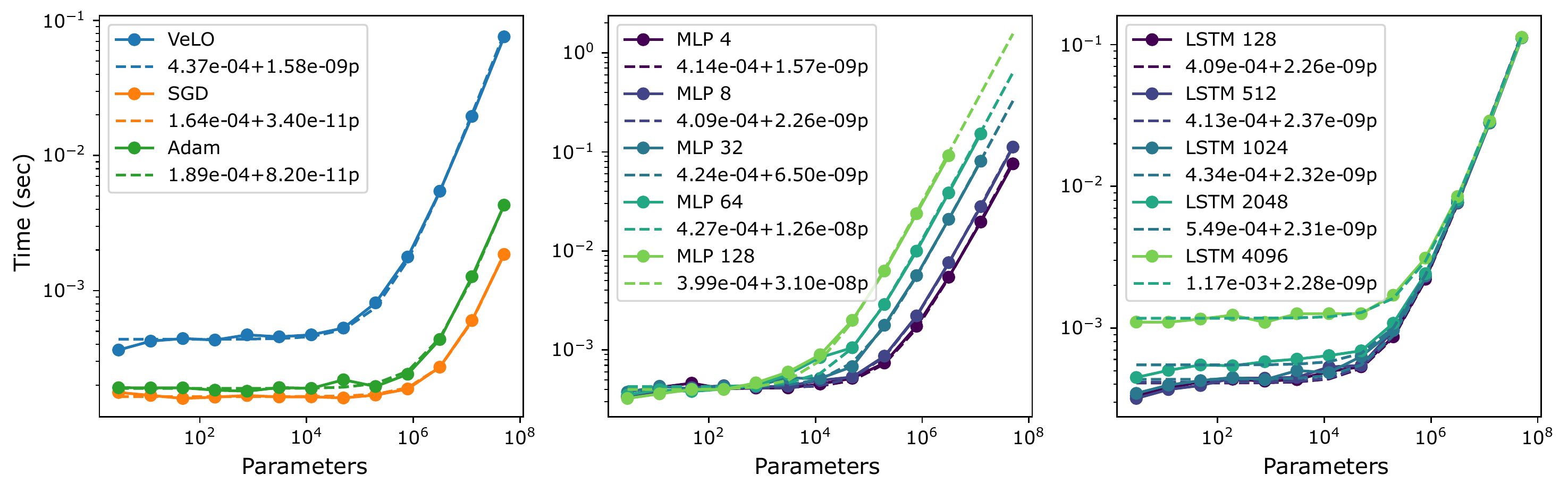}
    \put (29,7.0) {\textbf{\small(a)}}
    \put (61,7.0) {\textbf{\small(b)}}
    \put (94,7.0) {\textbf{\small(c)}}
    \end{overpic}
    }
    \vspace{-1em}
    \caption{\textbf{HyperNetwork computational costs.} We demonstrate the computational costs and overhead when applying our HyperNetwork based optimizer.
    For each configuration we show both empirical measurements computed on a TPUv3 (in solid) and a simple model fit to this data (in dashed).
    \textbf{(a)} We show a comparison between VeLO and 2 baseline optimizers: SGD and Adam. The learned optimizer has a ~3x higher fixed cost, and costs about about ~50x more per parameter. Despite this extra cost, however, these optimizers can still be used efficiently in many models. See Section~\ref{app:computational_costs_of_lopt}. \textbf{(b)} We visualize the costs of our learned HyperNetwork when varying the number of parameters the per-parameter MLP has.
    We find adding additional parameters in this way does not increase the fixed costs, but does predictably increase the cost per parameter.
    \textbf{(c)} We visualize the costs of varying the hidden size of our per-tensor LSTM in the learned optimizer.
    Because we are leveraging a hierarchy, this only increases the fixed costs, but does not effect the per parameter costs.
    \label{fig:timing}
    }
\end{figure}

Predicting exact run times of any deep learning system, especially one that leverages a compiler, is complex.
Instead, we designed a simple model for performance based on 3 different components: a constant execution overhead, per-tensor scaling cost, and per-parameter scaling cost. For further simplicity we assume a constant tensor count combining both the constant overhead and the per-tensor scaling. This leaves us with the following simplified model:
\begin{equation}
    \text{time} = \lambda_{\text{overhead}} + \lambda_{\text{params}} \cdot \text{params}.
\end{equation}

To test this model, we run a set of different optimization algorithm (learned and hand-designed) on a simple set of parameters---three square matrices, each of the same number of rows and columns.
We fit the parameters of this model ($\lambda_{\text{overhead}}$ and $\lambda_{\text{params}}$) using via gradient descent in log space. Despite its simplicity, our model is well-aligned with the data and strongly predictive.

In Figure~\ref{fig:timing}a we plot our predicted time-per-step for the learned optimizer, SGD, and Adam, as well as the data.
The learned optimizer has both higher overhead, and significantly larger cost per parameter, as expected.
SGD takes 3.4\e-11 seconds per parameter, whereas our model is 1.58\e-9 seconds per parameter ($46\times$ longer).
In terms of the constant overhead time, our optimizer has about 2.6 higher overhead.
This overhead region dominates below approximately one million parameters.
We expect a better implementation (instead of naively relying upon XLA) will shift this overhead region.
In Figure~\ref{fig:timing}b we look at the effects of changing the size of the per-parameter MLP.
As expected, the cost per parameter grows roughly linearly with the size of the per-parameter MLP.
Finally, in Figure~\ref{fig:timing}c we look at the size of the per-tensor LSTM.
As we increase the size, the per-parameter cost remains constant, though the overhead grows.
The impact of this overhead goes away, however, as the parameter count grows again saturating after around a million parameters. This means the per-tensor network can use more and more compute without additional overhead as the model it is training grows.

\paragraph{Distributed optimization.}
The computational costs can further be reduced by distributing the optimizer computation.
If one is performing model or pipeline parallelism optimizer computation only happens on the devices where the corresponding weights are located, and thus a factor of number of devices will be saved.
If using data parallel training, it is feasible to distribute the optimizer computation via Zero style data parallelism~\citep{rajbhandari2020zero}.
As machine learning models continue to grow in parameter count, the hardware to train these models also is growing.
As a result, the number of parameters on each device is not exploding -- in fact it might even be shinking making the overhead of learned optimizers less and less and thus a more appealing choice compared to hand-designed methods.

\textbf{Extrapolating optimizer overhead.} Finally, we extrapolate our results to explore the feasibility of training extremely large models. In particular, we will discuss the PaLM 540B parameter model~\citep{chowdhery2022palm}.

PaLM is trained with 540\e9 parameters. Training is done over 2 TPUv4 pods each with $3072$ chips.
Data parallelism is used across pods, and within pods 12-way model parallelism and 256-way data parallelism is used leveraging Zero style data parallel parameter sharding meaning weights are spread across all 3072 chips.
This means there are 540\e9$/ 3072$ = 176\e6 parameters per device.
Using our model, we can estimate the cost to run the optimizer per device would be approximately 0.3 seconds (On TPUv3).
The 540\e9 Palm model takes 17.6 seconds to perform one step on TPUv4\footnote{From personal correspondence with authors.} and thus using a learned optimizer will cost roughly 2\% extra compute.

\textbf{Overhead in practice.}
In practice, we find optimizer overhead ranges from minimal, to 2x the cost of training depending heavily on model and implementation.
Most of the poor performing models are from sub-optimal implementations.
Many multi device training implementations do not leverage any parameter shading whatsoever.
Instead, all parameters are stored on each device, and optimizer computation is duplicated on each device.
To make matters worse, as these models grow, the batch size of some of our tasks shrinks making the optimizer overhead even higher.
As model's grow and deep learning software matures, we expect these issues to go away.

In the near-term, however, if wall-time performance is bad, one can always explore using a larger batch size. As shown in Section~\ref{sec:bs}, our optimizers continue to work well with increased batch sizes even when hand-designed methods falter.

\textbf{More efficient implementations.}
Based on early profiles of VeLO, there is still plenty of room to optimize.
Because our per-parameter are extremely small, often it is faster to not leverage dense matrix multiplication hardware in modern accelerators.
This was briefly explored in \citet{metz2022practical} where they showed considerably faster run times for smaller sized per-parameter models.
When manually writing out matrix multiplications in this way, we also can benefit from unstructured sparsity in the weights of the per-parameter model.

Based on early profiling, we found XLA does not generate as efficient as of a kernel as possible requiring at least 2x the amount of memory reads as required. Further improvement there will lower our costs.

Finally, all of our experiments make use of float32 accumulators when past work has already demonstrated effective use of lower precision gradients, momentum, and second moment accumulators.
We expect we can also perform all the computation inside the learned optimizer in lower precision as well.
Because we are using ES to train, rather than backprop gradients, meta-training with these lower precision types should be trivial.

\section{Data: A Large, Diverse Distribution of Meta-Training Tasks}
\label{app:opt_benchmark}
Training effective deep learning models requires large and diverse datasets. 
For example, large-scale language modeling networks leverage data from the entire internet for training. 
When training a learned optimizer there is no such ready-made distribution of tasks. 
To make matters worse, we cannot possibly cover all possible tasks, as some tasks are far to computationally intensive to be used for meta-training. 
In this work, our strategy is to build a procedural generative process for machine learning tasks, as well as a method for sampling only tasks that are not prohibitively expensive to meta-train on.

Following both past work in meta-learning~\citep{wichrowska2017learned, metz2020tasks}, we train VeLO on a diverse distribution of data in the hopes to generalize to unseen tasks.
This distribution is composed of a mixture of parametric tasks definitions covering:
\begin{enumerate}
    \item \textbf{Image classification:} MLPs, convolutional networks, residual networks, and Vision Transformers.
    \item \textbf{Image generative modeling:} MLP-based auto-encoders, and MLP-based variational auto-encoders.
    \item \textbf{Language modeling:} recurrent networks and Transformers.
    \item \textbf{Training learned optimizers.}
\end{enumerate}
These base tasks are combined with randomized generation of elements such as depths, widths, datasets, training losses, and many other elements of the training problem.
We specify these elements through a task configuration language, which we discuss in the next subsection. 
In the remainder of this section we describe the components which are configured starting with different parametric task distributions, followed by a form of data-augmentation (task-augmentation) to further increase diversity.
Throughout the rest of this section, {\color{blue} \textbf{blue text}} denotes links to the open source implementation.

\subsection{A Configuration Language for Tasks} \label{app:config}
To define tasks, we make use of a \href{https://github.com/google/learned_optimization/blob/4447601601f04e3dc7b1aecd7854d44526593752/learned_optimization/tasks/parametric/cfgobject.py}{configuration language} built on top of \href{https://github.com/google/gin-config}{Gin}.
This configuration language lets us instantiate instances of objects and specify parameters values without actually constructing these objects.
These object specifications can also be nested.
As an example, the following is a configuration of a MLP with 2 hidden layers, trained on 8x8 images.
\begin{python}
CFGObject(obj="ParametricImageMLP",
  kwargs={"hidden_layers": [128, 128],
          "num_classes": 10,
          "datasets": CFGObject(obj="fashion_mnist_datasets",
                      kwargs={"image_size": (8, 8), "batch_size": 64})})
\end{python}

Here, both \texttt{ParametricImageMLP} and \texttt{fashion\_mnist\_datasets} both refer to python functions which are called with the provided kwargs upon construction.

Leveraging a configuration language provides a number of benefits.
First, it enables decoupling the code which creates tasks from how the task is configured.
This means we can sample configurations, rather than sampling entire tasks.
Second, these configurations themselves can be used as inputs to other machine learning systems.
In this work, we train machine learning models on these configurations to produce an estimate of the run time for rejection sampling (see Section~\ref{app:cfg_to_nn}).
We also experimented using these configs to learn value functions---functions which map from configuration to entire training curves -- though we could not make this reliable enough for use in this work. 

\paragraph{Task vectorization.}
For tractable meta-learning, it is necessary for each individual training problem to be efficient.
These problems are trained over and over again with a learned optimizer to estimate meta-gradients and thus this is the main computational cost.
While this is accomplished in part by focusing on small tasks, naive execution of task sampling still results in far too much computational overhead, especially on ML accelerators which excel at performing large linear algerbra operations.
To combat this we leverage JAX's vectorization and compilation features.
To vectorize training of tasks, the computations performed in each task must match, so we are only able to vectorize certain elements of task configurations.
We refer to the configuration elements that are vectorized over as \textit{dynamic configuration} elements, and those not vectorized over as \textit{static configuration} elements.
Thus, we have a two types of task components: configuration elements which can be vectorized such as network initialization and activation functions\footnote{Non-linearities in neural network are cheap and thus to vectorize we compute multiple activations for all tasks and only select one to be used.}, and configurations which cannot be vectorized such as number of hidden layers or units.
Each task family defines variations in tasks covering both kinds of configurations.

\subsection{Shared Components Used to Construct Task Distributions}
Before describing each different kind of task we construct, we describe several shared components which are used across multiple task kinds. The code for these shared components is available \href{https://github.com/google/learned_optimization/blob/4447601601f04e3dc7b1aecd7854d44526593752/learned_optimization/tasks/parametric/parametric_utils.py}{here}.

\subsubsection{{Initializers}}
For some tasks, we sample one initialization function for all tensors in a network. The different kinds initializers we sample from are:
\begin{enumerate}
    \item Orthogonal initialization.
    \item Uniform distribution with min and max of $\sqrt{3/\texttt{fan\_in}}$.
    \item Normal initalization based on $\sqrt{1/\texttt{fan\_in}}$.
\end{enumerate}
Here, $\texttt{fan\_in}$ denotes the number of incoming connections to a nonlinearity (corresponding to, for example, the width of a linear layer).
Additionally, we rescale the initialization by a random value sampled uniformly between [0.5, 2.0]. 
\label{app:tasks:init}

\subsubsection{{Activation Functions}}
Activation functions are sampled from ReLU, ReLU6, SELU~\citep{klambauer2017self}, tanh, sigmoid, SiLU~\citep{hendrycks2016gaussian}, swish~\citep{ramachandran2017searching}, GELU~\citep{hendrycks2016gaussian}, and leaky ReLU~\citep{maas2013rectifier}.
In addition to these more standard activation functions we also include $\cos$ and $\sin$ as activation functions~\citep{meronen2021periodic}.
Because activation functions are cheap to run, we select which activation to run dynamically "in-graph" with XLA conditionals via \texttt{jax.lax.cond}.
\label{app:tasks:act}

\subsection{Base Task Families} \label{app:base_task_families}
In this subsection we describe the base tasks used in task construction, and their static and dynamic configuration elements.
These configurations serve as a base which we then apply task-augmentations to.

\subsubsection{\href{https://github.com/google/learned_optimization/blob/4447601601f04e3dc7b1aecd7854d44526593752/learned_optimization/tasks/parametric/image_mlp.py}{Fully Connected Networks}} \label{mlp_task_sample_distribution}
These tasks consist of MLP classifiers trained on flattened image datasets.

\paragraph{Static configuration:}
\begin{enumerate}
    \item Hidden size: log uniform, between 8 and 128.
    \item Number of layers: uniform, between 0 and 4, inclusive.
    \item Image size: log uniform, between 4 and 32.
    \item Batch size: log uniform, between 4 and 512.
    \item Dataset: uniform, between MNIST~\citep{lecun1998mnist}, Fashion MNIST~\citep{xiao2017/online}, CIFAR10~\citep{Krizhevsky09learningmultiple}, CIFAR100~\citep{Krizhevsky09learningmultiple}, or 16x16 Imagenet~\citep{ILSVRC15}.
\end{enumerate}

\paragraph{Dynamic configuration:}
\begin{enumerate}
    \item Initialization function, sampled from those in Section \ref{app:tasks:init}.
    \item Activation function, sampled from those in Section \ref{app:tasks:act}.
\end{enumerate}

\subsubsection{\href{https://github.com/google/learned_optimization/blob/4447601601f04e3dc7b1aecd7854d44526593752/learned_optimization/tasks/parametric/image_conv.py}{Convolutional Networks}}
These tasks consist of image classification problems trained on convolutional neural networks.

\paragraph{Static configuration:}
\begin{enumerate}
    \item Hidden size: log uniform, between 4 and 64.
    \item Number of layers: uniform, between 0 and 3, inclusive.
    \item Number of layers with a stride of 2: uniform from 0, 1, or 2.
    \item Image size: log uniform, between 4 and 32.
    \item Batch size: log uniform, between 4 and 512.
    \item Dataset: uniform, between MNIST, Fashion MNIST, CIFAR10, CIFAR100, or 16x16 Imagenet.
\end{enumerate}

\paragraph{Dynamic configuration:}
\begin{enumerate}
    \item Initialization function, sampled from those in Section \ref{app:tasks:init}.
    \item Activation function, sampled from those in Section \ref{app:tasks:act}.
\end{enumerate}

\subsubsection{\href{https://github.com/google/learned_optimization/blob/4447601601f04e3dc7b1aecd7854d44526593752/learned_optimization/tasks/parametric/image_mlp_ae.py}{Auto-Encoders}}
These tasks consist of deterministic auto-encoders parameterized by MLP and trained with mean squared error.

\paragraph{Static configuration:}
\begin{enumerate}
    \item Hidden size: log uniform, between 8 and 128.
    \item Number of layers: log uniform, between 0 and 4, inclusive.
    \item Image size: log uniform, between 4 and 32.
    \item Batch size: log uniform, between 4 and 512.
    \item Dataset: uniform, between MNIST, Fashion MNIST, CIFAR10, CIFAR100, or 16x16 Imagenet.
\end{enumerate}

\paragraph{Dynamic configuration:}
\begin{enumerate}
    \item Initialization function, sampled from those in Section \ref{app:tasks:init}.
    \item Activation function, sampled from those in Section \ref{app:tasks:act}.
    \item Log loss: whether or not to train with log loss.
    \item Center data: whether or not to rescale data to lie in -1 to 1, rather than 0 to 1.
    \item Constrain output: whether or not to clamp the output with a sigmoid or tanh function, or to directly use the output of the final linear layer.
\end{enumerate}

\subsubsection{\href{https://github.com/google/learned_optimization/blob/4447601601f04e3dc7b1aecd7854d44526593752/learned_optimization/tasks/parametric/image_mlp_vae.py}{Variational Auto-Encoders}}
These tasks are based on variational auto-encoders~\citep{kingma2013auto}.
They construct a generative model with a quantized normal observation model and are trained with the negative evidence lower bound (ELBO).

\paragraph{Static configuration:}
\begin{enumerate}
    \item Encoder hidden size: log uniform, between 4 and 128.
    \item Decoder hidden size: log uniform, between 4 and 128.
    \item Number of encoder hidden layers: uniform, between 0 and 4, inclusive.
    \item Number of decoder hidden layers: uniform, between 0 and 4, inclusive.
    \item Latent dimension: log uniform, between 2 and 128.
    \item Image size: log uniform, between 4 and 32.
    \item Batch size: log uniform, between 4 and 512.
    \item Dataset: uniform, between MNIST, Fashion MNIST, CIFAR10, CIFAR100, or 16x16 Imagenet.
\end{enumerate}

\paragraph{Dynamic configuration:}
\begin{enumerate}
    \item Initialization function, sampled from those in Section \ref{app:tasks:init}.
    \item Activation function, sampled from those in Section \ref{app:tasks:act}.
    \item A boolean choice to center input data, or not.
    \item Per-dimension loss: whether or not to use a per-dimension loss, or the sum over all dimension loss.
\end{enumerate}

\subsubsection{\href{https://github.com/google/learned_optimization/blob/4447601601f04e3dc7b1aecd7854d44526593752/learned_optimization/tasks/parametric/image_resnet.py}{Residual Networks}}
These tasks consist of small residual network-style \citep{he2016deep} models for image classification. In practice, these models are considerably smaller than even ResNet18.

\paragraph{Static configuration:}
\begin{enumerate}
    \item Max blocks per group: The ResNet structure is divided into 4 stages---between each stage the image is downsampled. We (log uniformly) sample the maximum number of ResNet blocks for each group between 1 and 10.
    \item Blocks per group: Using the max value above, we uniformly sample a number of blocks for each group for each of the 4 blocks.
    \item Width pattern: We sample the multiplier on how large each block is. In standard ResNet architectures, the hidden size doubles each block. We uniformly sample one of the following scaling factors: $[(1, 1, 1, 1), (1, 2, 4, 8), (1, 2, 2, 4), (1, 2, 2, 2), (1, 2, 4, 4)]$.
    \item Base hidden size: We sample the base hidden size (multiplied by the above width pattern for each group) log uniformly between 8 and 256.
    \item Initial convolution kernel size: Either 3, 5, or 7.
    \item Initial convolution channels: log uniform, between 8 and 64.
    \item Include initial max pool: uniform, either true or false.
    \item Image size: log uniform, between 8 and 64.
    \item Batch size: log uniform, between 4 and 256.
    \item Dataset: uniform, between MNIST, Fashion MNIST, CIFAR10, CIFAR100, or 16x16 Imagenet.
\end{enumerate}

\paragraph{Dynamic configuration:}
\begin{enumerate}
    \item Activation function, sampled from those in Section \ref{app:tasks:act}.
\end{enumerate}

\subsubsection{\href{https://github.com/google/learned_optimization/blob/4447601601f04e3dc7b1aecd7854d44526593752/learned_optimization/tasks/parametric/lm_rnn.py}{Recurrent Networks}}
These consist of language models parameterized by an RNN. 

\paragraph{Static configuration:}
\begin{enumerate}
    \item RNN size: log uniform, between 8 and 256.
    \item Embedding size: log uniform, between 8 and 256.
    \item Batch size: log uniform, between 4 and 512.
    \item Sequence length: log uniform, between 4 and 512.
    \item Dataset: either LM1B~\citep{DBLP:journals/corr/ChelbaMSGBK13} or Wikipedia english~\citep{wikidump} with either bytes or 32K vocab size.
    \item Recurrent cell type: uniform, between a vanilla RNN, LSTM, or GRU~\citep{chung2014empirical}.
    \item Vocab size: If the bytes dataset is not being used, we take the 32K vocab size and truncate it to a smaller size sampled log uniformly between 100 and 10K.
\end{enumerate}

\paragraph{Dynamic configuration:}
\begin{enumerate}
    \item Initialization function, sampled from those in Section \ref{app:tasks:init}.
\end{enumerate}

\subsubsection{\href{https://github.com/google/learned_optimization/blob/4447601601f04e3dc7b1aecd7854d44526593752/learned_optimization/tasks/parametric/lm_transformer.py}{Transformers}}
These consist of language models parameterized by a Transformer \citep{vaswani2017attention}. 

\paragraph{Static configuration:}
\begin{enumerate}
    \item Model dimension divided by head dimension: log uniform, between 8 to 128.
    \item Number of heads: log uniform, between 8 and 128.
    \item Number of layers: log uniform, between 1 and 8.
    \item Batch size: log uniform, between 4 and 512.
    \item Sequence length: log uniform, between 4 and 512.
    \item Dataset: either LM1B or Wikipedia english with either bytes or 32K vocab size.
    \item Vocab size: If the bytes dataset is not being used, we take the 32K vocab size and truncate it to a smaller size sampled log uniformly between 100 and 10K.
\end{enumerate}

\paragraph{Dynamic configuration:}
No dynamic configurations are specified for this task distribution.

\subsubsection{\href{https://github.com/google/learned_optimization/blob/4447601601f04e3dc7b1aecd7854d44526593752/learned_optimization/tasks/parametric/vit.py}{Vision Transformers}}
We use Vision Transformers built from the \href{https://github.com/google-research/vision_transformer}{\texttt{vision\_transformer} package} . 

\paragraph{Static configuration:}
\begin{enumerate}
    \item Number of layers: log uniform, between 1 and 16.
    \item Number of heads: log uniform, between 1 and 16.
    \item Hidden size per head: if number of heads is <4, we uniformly sample between 8 and 64, otherwise we uniformly sample 8 through 32.
    \item MLP dimension: we sample on a log scale between 32 and 512.
    \item Image size: log uniform, between 4 and 64.
    \item Batch size: log uniform, between 4 and 256.
    \item Dataset: uniform, between MNIST, Fashion MNIST, CIFAR10, CIFAR100, or 16x16 Imagenet.
    \item Patch size: uniform, between 2, 4, 8, 12, or 16.
    \item Dropout: used with 70\% probability.
    \item Dropout scale: uniform, between 0 and 0.8.
    \item Dropout on attention: uniform, between 0 and 0.8 on a log scale.
\end{enumerate}

\paragraph{Dynamic configuration:}

No dynamic configuration for this task distribution is used.

\subsubsection{\href{https://github.com/google/learned_optimization/blob/4447601601f04e3dc7b1aecd7854d44526593752/learned_optimization/tasks/parametric/lopt.py}{Learned Optimizer Training}}
Finally, we have a distribution of tasks that resembles training learned optimizers.
This task distribution was introduced in the hopes that we could use our learned optimizer to train themselves. 

\textbf{Static configuration:}
\begin{enumerate}
    \item Learned optimizer architecture: Either meta-learning the Adam hyperparameters, the SGD hyperparameters with and without momentum, the MLP based learned optimizer from ~\citep{metz2019understanding}, and the MLP based learned optimizer from ~\citep{metz2022practical}.
    \item LOpt hidden size: If using an MLP LOPT, the hidden size is sampled on a log scale between 2 and 512.
    \item LOpt hidden layers: If using an MLP LOPT, the number of layers is sampled on a log scale between 1 and 4 on a log scale.
    \item Two step scaling multiplier hyperparameters: sampled between 1\e-5 and 1 on a log scale.
    \item Number of inner training steps: sampled from 1 to 100 on a log scale
    \item Outer batch size: between 1 and 8 on a log scale.
    \item Target task distribution: We sample a task from the MLP image classification distribution of tasks described in Section~\ref{mlp_task_sample_distribution}.
\end{enumerate}

\paragraph{Dynamic configuration:}
\begin{enumerate}
    \item The dynamic data corresponding to the task sampled from~\ref{mlp_task_sample_distribution}.
\end{enumerate}

\subsection{Controlling Task Size} \label{app:cfg_to_nn}
\begin{figure}
    \centering
    \makebox[\textwidth]{%
    \begin{overpic}[width=0.6\textwidth]{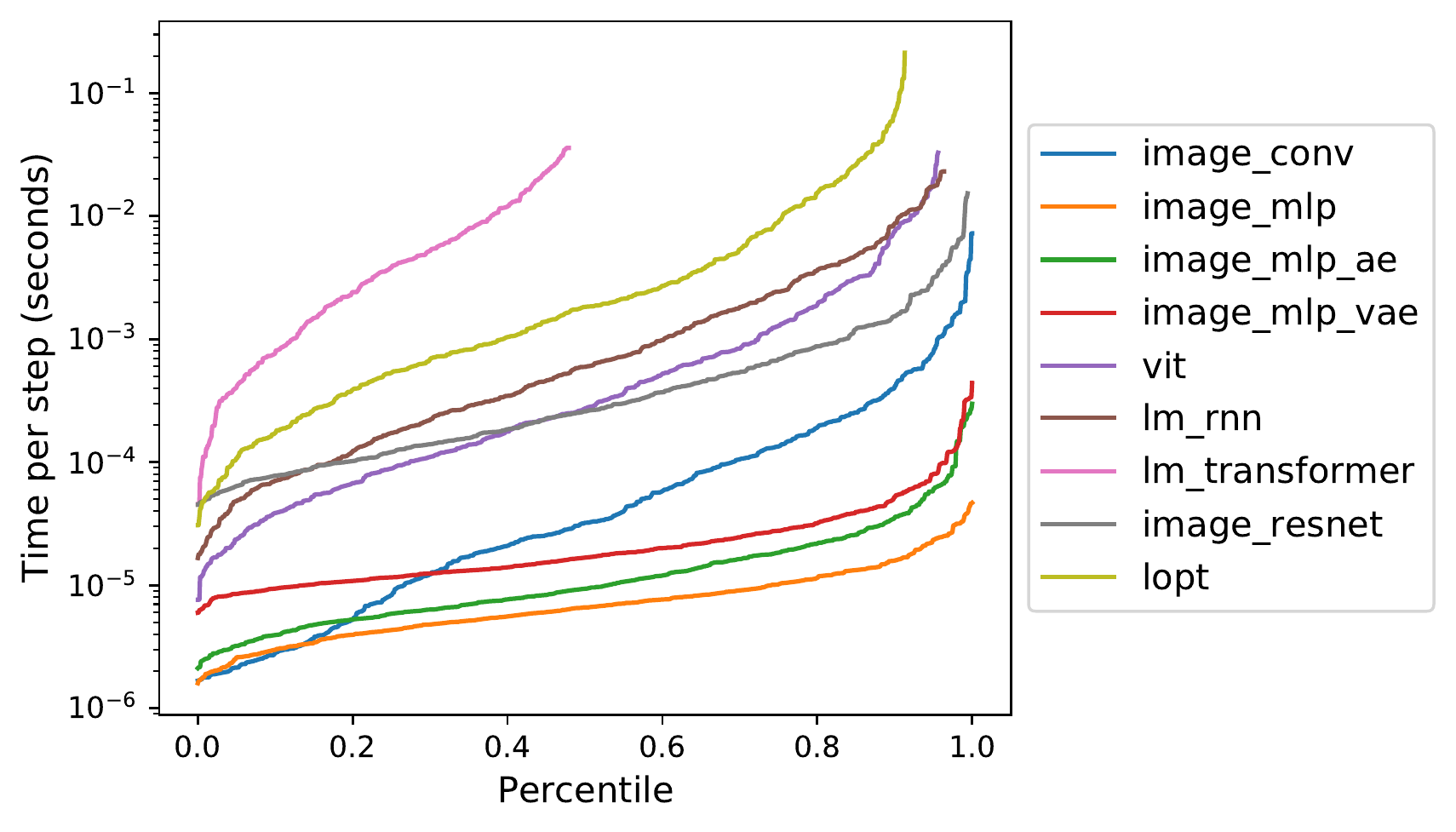}
    \end{overpic}
    }
    \caption{\textbf{Time per step of sampled models.} In each color, we train and time 1024 different configurations and measure the time per iteration.
    The gaps / empty space denote if a model ran out of memory, or was otherwise invalid. We find a huge variability in run times across the sampled task distributions. We see both large variability both between different kinds of tasks (different colors) and within each task family.
    \label{fig:filter_times}
    }
\end{figure}

The speed of meta-training is fundamentally limited by the size of the underlying tasks being trained to compute gradient estimates. 
As such, it is crucial to be able to sample tasks based on the run time of the task.
To demonstrate the impact of this, we plot the measured run times of 1024 tasks per base task family in Figure~\ref{fig:filter_times}. One can see up to 4 orders of magnitude variability in the run times of tasks, which would correspond to dramatically slower meta-training.

To sample only tasks of appropriate duration for meta-training, we need to compute the run time of a particular task configuration.
Directly timing a given configuration, while simple, is computationally too expensive as the compilation of this task can take up to a couple minutes while the actual computation can take seconds.

Instead, we train a neural network model to approximate this computation for us.
This is done by first randomly sampling tasks (approximately 10,000 tasks), and training a network which maps from task configuration to measured run time. Once trained, we can use our learned models, to estimate timings in only a couple milliseconds. The structure of this time predicting network follows.

\paragraph{\href{https://github.com/google/learned_optimization/blob/4447601601f04e3dc7b1aecd7854d44526593752/learned_optimization/time_filter/time_model.py}{A network for run time prediction}.}
This network maps from a potentially nested configuration to a single floating point value representing the predicted run time.
We expect the specifics of this functional form to be relatively unimportant as long as the function has access to all the information from the configuration.

For our implementation, we flatten the configuration into a list of key, value pairs.
The keys, which contain a string name of the argument, are then hashed, and looked up in an embedding table to be converted to a floating point vector also contains a string, that is hashed and looked up as well.
If the value is a number, it is normalized based on the min and max of the values seen for that key.
This leaves us with a rank 2 array with one dimension being the number of configurations key value pairs, and a second dimension being the feature dimension of the embedding.

This data is then normalized based on a running min and max seen during training and fed into a neural network which does deep set-style operations~\citep{zaheer2017deep} along the key value pairs dimension, and then ultimately is reduced to a single, fixed length feature vector.
Another neural network is applied to this to create a final prediction.
We train this model with mean squared error in log space with Adam monitoring a validation set of tasks to ensure we do not over-fit.
We train one of these models for each task family defined in Section~\ref{app:base_task_families}. The pre-trained run time predicting models used in this work are also open source and can be found \href{https://github.com/google/learned_optimization/blob/687e72e7b5596dfb80c5196fd51f43058899edd9/learned_optimization/time_filter/model_paths.py#L19}{here}.

\paragraph{Learned models to predict if a task will run or not.}
Similar to the previous 2 sections, we also train a set of models mapping from configuration to a binary prediction of if the configuration would result in a valid task.
This catches potential issues ranging from invalid sizes or strides for a convolutional network, to running out of memory for all problems.

\subsection{Task Sampling and Augmentations}\label{app:meta_training_task_distribution}
The meta-training distribution is defined by a sequence of sampled tasks and task transformations.
First, we seek to sample a set of smaller run time tasks to make meta-training tractable.
To do this, we sample a max run time of either 2\e-5 seconds per step with probability 0.65, 1\e-4 with probability 0.3, 4\e-4 with probability 0.1 and 1\e-3 with probability 0.05. 
We use these thresholds, along with the learned run time models defined in Section~\ref{app:cfg_to_nn} to sample tasks that run within this amount of time.
Not all task families have tasks which run in some of the smaller amount of times.
If the time is less than 1\e-4 we sample a task uniformly from the image MLP tasks, image convolution tasks, image auto-encoding, image VAE, and the VIT models.
If the run time is greater than or equal to 1\e-4 we additionally add in the option to sample from the RNN language models, the Transformer language models, image ResNet models, and the learned optimizer tasks. See Figure \ref{fig:filter_times} for a visualization of these run times, and how these cutoffs were determined.

\paragraph{Task augmentation.} Data augmentation has proven incredibly successful at aiding generalization of supervised machine learning models~\citep{perez2017effectiveness}
as well as to encourage generalization in robotics tasks~\citep{sadeghi2016cad2rl,james20163d,tobin2017domain}.
To further increase diversity of our distribution of tasks apply a similar idea and construct \textit{task augmentations} which generate new, transformed tasks which have different optimization properties. As discussed in Section~\ref{app:config} these augmentations are configured with both static and dynamic configurations. We use the following augmentations when meta-training.
\vspace{1em}

\begin{enumerate}
    \item \href{https://github.com/google/learned_optimization/blob/a42b207b9ef5395fdf7e64978c26379ccba4e264/learned_optimization/tasks/task_augmentation.py#L103}{\textbf{Re-parameterization of weights.}} We scale the task initialization by some constant, $c$. Before using the weights, we multiply by $1/c$. This leaves the function expressed by the initialization the same, but changes the optimization dynamics with adaptive optimizers.
    This is similar to what was done in~\citep{lv2017learning}.
    We sample the form of re-parameterization we use uniformly from: [None, None, global, global, tensor, tensor, parameter]. If None, we use no re-parameterization. Otherwise, we re-parameterize the weights of the target model either globally, different scaling for each tensor, or different scaling for each parameter.
    The size of the scaling is also sampled. First we sample the ranges: either (0.001,1000), (0.01, 100), or (0.1, 10), and then use this single range to sample each of the different ranges for each tensor or parameter.
    
    \item \href{https://github.com/google/learned_optimization/blob/a42b207b9ef5395fdf7e64978c26379ccba4e264/learned_optimization/tasks/es_wrapper.py#L227}{\textbf{Evolution gradient estimation.}} With 8\% probability we estimate gradients of the inner task with ES~\citep{salimans2017evolution}. We log uniformly sample the standard deviation of sampling between 0.001 and 0.1, and the number of antithetic pairs to be either 1, 2, 4, 8, or 16.
    
    \item \href{https://github.com/google/learned_optimization/blob/a42b207b9ef5395fdf7e64978c26379ccba4e264/learned_optimization/tasks/task_augmentation.py#L238}{\textbf{Batch size reduction.}} We add a task augmentation to reduce batch size with 20\% probability, and reduce the batch size by a sampled value between 0.01 and 1.0.
    
    \item \href{https://github.com/google/learned_optimization/blob/a42b207b9ef5395fdf7e64978c26379ccba4e264/learned_optimization/tasks/task_augmentation.py#L330}\textbf{{Lower precision training.}} With 20\% probability we lower the floating point precision of the model to \texttt{bfloat16}.
    
    \item \href{https://github.com/google/learned_optimization/blob/a42b207b9ef5395fdf7e64978c26379ccba4e264/learned_optimization/tasks/task_augmentation.py#L393}{\textbf{Gradient normalization.}} With 5\% probability we normalize the gradients of the inner-problem.
    
    \item \href{https://github.com/google/learned_optimization/blob/a42b207b9ef5395fdf7e64978c26379ccba4e264/learned_optimization/tasks/task_augmentation.py#L445}{\textbf{Directional gradient subspace.}} With 8\% probability we use the gradient to compute some number of directional derivatives and average these. We log uniformly sample the number of directional derivatives to be between 1 and 1024. Additionally, before averaging, with 50\% probability we ignore the magnitude of this directional gradient and only use the normalized direction. This modification is meant to resemble the types of gradient estimates computed with ES---but are much cheaper to compute.
    
    \item \href{https://github.com/google/learned_optimization/blob/a42b207b9ef5395fdf7e64978c26379ccba4e264/learned_optimization/tasks/task_augmentation.py#L533}{\textbf{Asynchronous gradients.}} With 5\% probability we add an artificial delay between when gradients are computed, and when they are applied to emulate asynchronous training. We log uniformly sample the delay between 1 and 8 steps.
\end{enumerate}

\section{Meta-Training}
In this section we describe our meta-training procedure.
First, we discuss our meta-objective and gradient estimation strategy, followed by detailing our curriculum strategy for training, and a discussion on multi-task training.

\subsection{Meta-Objective}
The meta-objective is the objective which we seek to minimize in meta-training, and defines the performance of our learned optimizers.
In our work, our meta-training objective is the training loss computed at the end of inner-training. 
In an effort to increase generality, this objective is computed in expectation over several sampled quantities:
\begin{enumerate}
    \item \textbf{Task static configuration.} A sample from the distribution of tasks defined in \ref{app:meta_training_task_distribution}.
    \item \textbf{Task dynamic configuration.} A sample of configurations which can be vectorized over.
    \item \textbf{Task initialization.} Initialization of the inner-parameters for a meta-training run.
    \item \textbf{Randomness in task.} Any randomness involved when training the problem (e.g. batches of training data, or dropout).
    \item \textbf{Inner problem length.} The number of steps the inner problem is trained for which is sampled from a log uniform distribution between 200 and 20K early in training, and 200 and 200K later in training.
    \item \textbf{Data used to calculate meta-loss.} We use additional batches of data at the end of inner-training to estimate the meta-loss.
\end{enumerate}

\subsection{Meta-Gradient Estimation}
To estimate gradients of the meta-objective, we leverage ES~\citep{salimans2017evolution} with antithetic samples~\citep{mcbook}, and share as much randomness between pairs as possible to further reduce variance.
This includes initializing the target tasks with the same parameter values, as well as using the same batches of data for each antithetic pair and the same batches to evaluate performance at the end of training.

We opt for ES rather than a more sophisticated method such as Truncated ES~\citep{metz2019understanding}, Persistent ES~\citep{pmlr-v139-vicol21a} or hybrid ES/Backprop methods~\citep{metz2019understanding} for simplicity, and lower communication overhead between machines in the cluster.
Additionally, non-truncated methods are also more amenable to target the loss at the end of training, rather than average loss.

\subsection{Curriculum and Meta-Generalization}
Meta-training on larger scale problems is prohibitively expensive.
To save training time, we make use of both curricula over various meta-training parameters, and the fact that our optimizer can generalize from smaller to larger sized inner tasks.
This property is crucial as it saves orders of magnitude of compute.
To demonstrate this, we show the effect of curriculum over the length of inner problem in the following section.

\begin{figure}
    \centering
    \makebox[\textwidth]{%
    \begin{overpic}[width=0.6\textwidth]{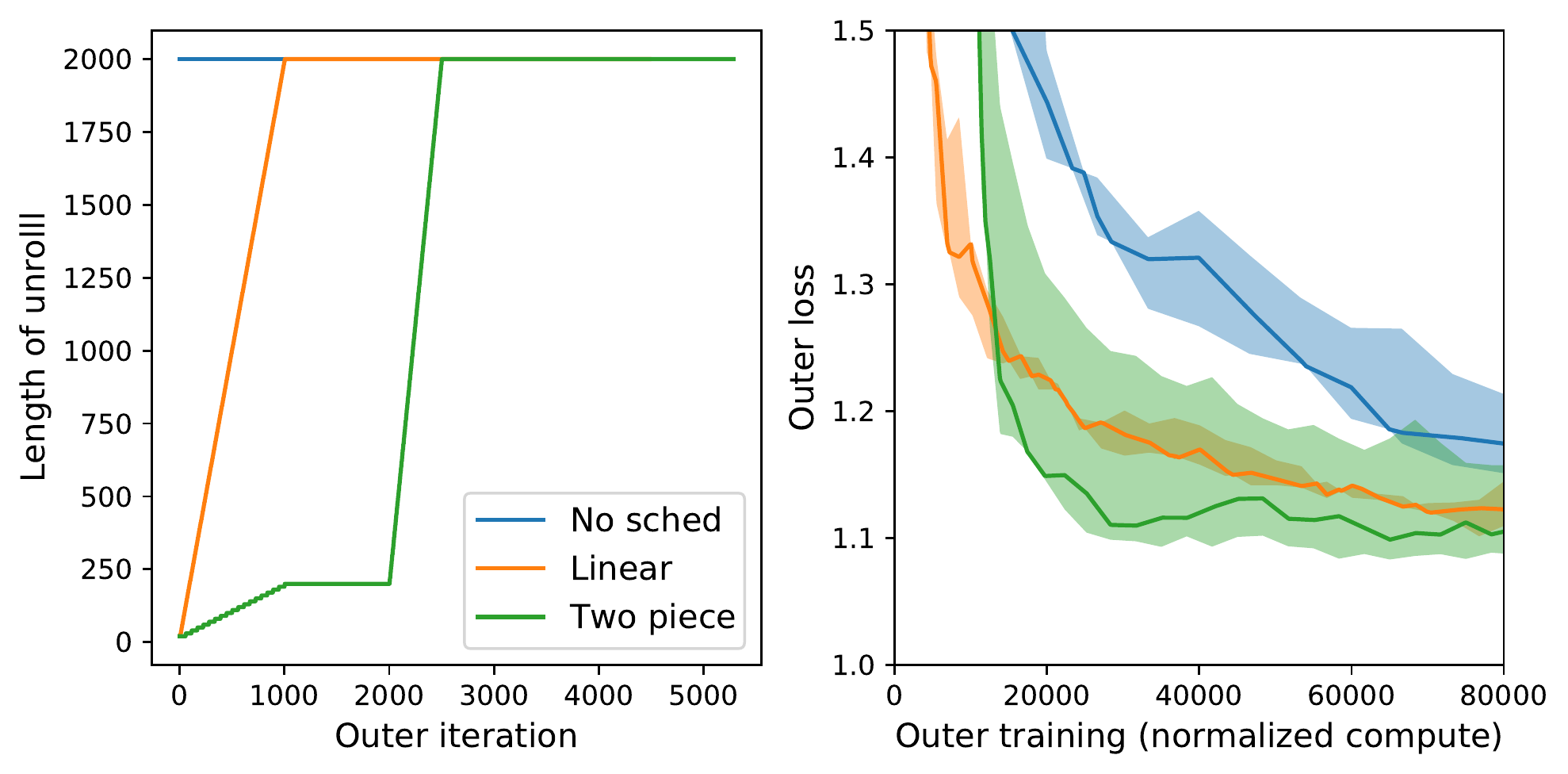}
    \put (3,50.0) {\textbf{\small(a)}}
    \put (49,50.0) {\textbf{\small(b)}}
    \end{overpic}
    }
    \vspace{-1em}
    \caption{\textbf{Schedules over unroll length speed up meta-training.} \textbf{(a)} we show 3 different kinds of schedules used when meta-training: A constant schedule of the desired length, a linear increase for the first 1K outer-iterations, and a 3 part piece wise linear schedule. \textbf{(b}) we show the resulting outer-learning curves when meta-trained, showing our curriculum choice improves efficiency of meta-training. On the x-axis we show a compute normalized measure of outer-training performance. 
    \label{fig:sched_unroll_length}
    }
\end{figure}

\paragraph{Experiment: Curriculum over unroll length.} In this experiment, our goal is to train a learned optimizer to train a distribution of tasks for 2,000 inner iterations.
Because computational cost of computing meta-gradients is proportional to the inner problem length, we can compute meta-gradients faster with shorter inner-problems at the cost of bias (doing well on shorter problems doesn't guarantee performance on longer inner-training time).
This bias can be mitigated by eventually meta-training with problems run for the desired number of iterations. In Figure~\ref{fig:sched_unroll_length} we show the effects of two curriculum over problem length where we find faster training. In this example, we only train targeting 2K steps. This curriculum only becomes more important as we increase this length. For our learned optimizers, we ultimately target 200K iterations---2 orders of magnitude longer.

\subsection{Multi-Task Training} \label{app:hand_designed_normalizer}
As meta-training a learned optimizer is inherently a multi-task optimization problem, we must balance the gradient contributions from each task.
Unlike many other multi-task problems, there is both no natural or expected scale for the losses from each task which makes targeting performance across all tasks difficult. 
We opt for a simple solution to this problem---simply to normalize the length of each meta-gradient independently per task making each gradient constant norm.
As a result, the meta-training procedure only depends on the ratios of how tasks are sampled, and is invariant to the scale of losses for each task. 
Using this normalization, however, comes at the cost of losing a well-defined meta-objective, and can even produce non-descent directions when the meta-loss is stochastic.
Nevertheless, we found this method works quite well in practice. 

We also experimented with normalizers designed by hand to roughly re-scale losses into a standardized range. For each task family, we manually define a transformation via trial and error to keep the loss between -10 and 10. Examples of such a transformation include normalizing by the number of classes predicted for classification tasks, as well as clipping the max loss. Implementations of these normalizations can be found in the \texttt{normalizer} function for all task families.
Despite the fact that meta-training with these resulted in worse performance, we find these aggregated normalized loss values to be useful to monitor the rough performance of our learned optimizer. 

Finally, we explored normalizers based on the performance of a set of baseline optimizers on a given task. Because our task distribution does not contain a finite number of tasks, we cannot simply run these baselines on all problems.
Instead, we attempted to train a neural network model to map from task configuration and baseline hyperparameters to learning curves using an architecture similar to that described in~\ref{app:cfg_to_nn}.
While initially promising, this method substantially increases complexity, and did not outperform the much simpler normalization of gradient and thus we chose not to explore this direction further even though we believe this is closer to correct solution in the long run.

\section{Meta-Training Infrastructure}
\label{sec:infra}
In this section we discuss practical aspects of our distributed meta-training infrastructure. 
Both meta-training and evaluating the learned optimizer during meta-training requires significant compute.
Our distributed training infrastructure is built inside Google's infrastructure.
Despite this, all the code and components we used to meta-train are open source and can be adapted to any distributed computing engine.
One needs a distributed file system (we use Colossus~\citep{dean2010evolution}), and some way to perform Remote Procedure Calls (RPC) (we use Courier~\citep{yang2021launchpad} which uses the GRPC library).

\subsection{One Learner, Many Workers} \label{app:learner_worker}
The main meta-training set of machines consists of a single learner process which provides an RPC interface fetching the current learned optimizer weight values, and an RPC interface to receive meta-gradients calculated with these weights. This interface is defined \href{https://github.com/google/learned_optimization/blob/10574578689feabc8677ac4aab100f7bb4e777c2/learned_optimization/distributed.py}{here}.
Each worker process has a collection of different tasks on which VeLO is trained. 
The worker first fetches the weights of the learned optimizer, estimates a meta-gradient by training a model, and sends the gradients back to the learner process.
The learner and each worker process run on a single TPU chip.
To prevent wasted compute, the learner process is always run in a more reliable manner (i.e. dedicated hardware) so that it will not be preempted.

\subsection{Evaluation Clusters}
While meta-training, the learner process saves weights of the learned optimizer to the distributed file system.
An "evaluation chief" process monitors this file system, and when a new checkpoint is found, it enqueues some set of different evaluation configurations on a distributed task queue. These evaluations are defined by a list of configurations which specify some target model to evaluate, the target length to train that model, and other information required to for the evaluation.
Evaluation workers (each with a single TPU chip) fetch these tasks, load weights of the learned optimizer, train this models, and report back the results to the evaluation chief.
By distributing our evaluation across many TPU chips in this way we are able to continuously monitor the performance of our learned optimizers on hundreds of tasks, greatly increasing our confidence that meta-training is performing as expected.

In practice, we use multiple evaluation clusters, each of which is running at a different frequency.
These evaluation clusters monitor the performance on the meta-training distribution of tasks for different length unrolls (1K inner training steps, 10K inner training steps), as well as a subset of the evaluation tasks described in \ref{app:opt_benchmark}.
For a full list of these tasks see the \href{https://github.com/google/learned_optimization/blob/b1d8267c5e513a4112e7422b98bacc16e1f0e844/learned_optimization/continuous_eval/evaluation_set.py#L90}{code}.
When we shift from meta-training 200-20K iterations to 200-200K iterations, we additionally monitor performance after 100K inner-iterations.
In total, we reserve around 100 accelerators specifically for evaluation.

\subsection{Task Selection and Staleness of Meta-Gradients}\label{app:distributed:staleness}
The most straightforward implementation of our training infrastructure as we have described it so far would consist of each worker sampling a new task (both static and dynamic configurations) i.i.d.~from the task distribution after the completion of the previous task.
Doing this, however, would result in a new compilation of the computation graph which takes multiple minutes.
This would be extremely wasteful as the computation itself often only takes a few seconds to run.
As such, we compute multiple gradients for a given static task configuration before sampling a new task\footnote{We also explored building a large precompiled dataset of tasks, but were unable to load these fast enough from disk, leading to a similar problem.}.
Because dynamic configurations leverage the same compilation, we are able to use different settings for each computation of these gradient estimates, along with different dynamic task configurations for each of the vectorized problems we run in parallel.
While re-sampling dynamic tasks increases variation beyond simply evaluating multiple gradients on the exact same task, our sampling scheme does induce auto-correlation of gradients in time. 

Reusing tasks in this way also results in a different problem: machines which sample a fast task would produce significantly more meta-gradient estimates than machines which sampled a slow task.
This further breaks i.i.d.~assumptions and causes yet more gradient correlation resulting in unstable meta-training.
To compensate for this issue, each machine samples more than one task (we use 8 for the majority of our experiments), and when computing meta-gradients cycle through each to mitigate gradient correlation.
As we increase the number of tasks each machine has we will approach a uniform rate of gradients from each machine by the law of large numbers, though this comes at the cost of increased compile times.
We found 8 tasks balances startup costs (which take approximately 40 minutes) while sufficiently lowering gradient auto-correlation.
We additionally resample elements of these tasks probabilistically so as to ensure coverage of the entire task distribution.

These meta-gradients are all sent to a centralized learner which aggregates them in batches, and applies weight updates by passing them through Adam.
To combat staleness (meta-gradients computed with old learned optimizer parameter values), we throw out any gradients which have been computed too far in the past.
We find a gradient staleness of approximately 10 to be a good balance between not throwing away to many of the slower tasks, while also ensuring the gradients are still useful and not too stale.

\paragraph{Globally-Distributed Workers.}
The computational load required for meta-learning workloads is quite different than those of large supervised models and as such we can make use of considerably cheaper hardware than a dedicated supercomputer.
The bulk of our compute infrastructure consists of TPU chips designed originally for inference scattered across the globe.
Thus, we are leveraging idle compute resources in Google's fleet, and which are connected via commodity networking (by data center standards) rather than specialized ICI links in the usual TPU deployments~\citep{jouppi2020domain}.
As we perform data-parallel training over this cluster (Section~\ref{app:learner_worker}), each machine can operate largely independently.
Leveraging this amount of machines is possible due to extremely low networking requirements when transmitting meta-gradients compared to the costly compute requirements (training some sampled task with a learned optimizer)\footnote{\citet{salimans2017evolution} took this further and only transmitted random number seeds and loss values. This is possible with our implementation, but we found this was actually not needed.}.
To account for the large number of machines as well as to control gradient staleness, we also leverage an extremely large outer-batch size (up to $\sim$100K) in our largest models.
At peak capacity, we use a bit over 4K accelerators spanning 3 generations of TPU hardware.
Meta-training training took approximately one month.

\subsection{Interactive Hyperparameter Tuning} \label{app:interactive_hparams}
During meta-training, we frequently made modification to the running job.
These changes include increasing the batch size, lowering the learning rate, changing the distribution of inner-problems to include larger problems, and modifying the maximum staleness.
This was inspired by the reported success of online hyperparameter modification in OpenAI Five~\citep{berner2019dota}.

The exact meta-training was relatively ad-hoc and not easily replicated.
Due to the computational cost, we could only afford to train one model at full scale.
Our previous largest model used a fraction of the total compute.
Our training was divided into 4 phases, each using the previous weights as the starting point.
We monitored a variety of losses, as well as qualitatively test our trained learned optimizers to help us determine if any modifications are needed.

\begin{figure}[t!]
    \centering
    \makebox[\textwidth]{%
    \begin{overpic}[width=1.0\textwidth]{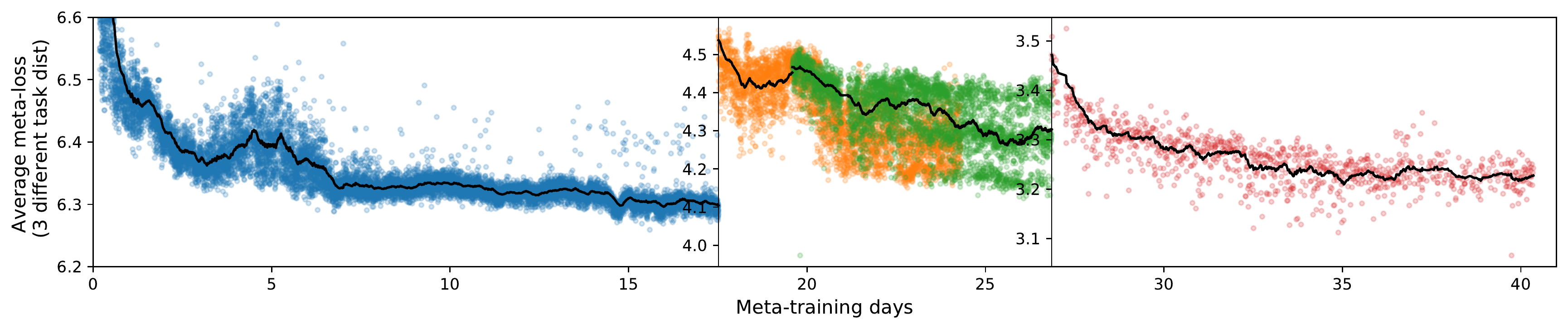}
    \end{overpic}
    }
    \caption{\textbf{Meta-training losses based on the hand-designed normalized task losses.} Each phase of meta-training is denoted with a different color. After phase 1 and phase 3, we switch meta-training distribution so show a new y-axis. As we describe in Section~\ref{app:hand_designed_normalizer} there is no single meta-loss to plot. Instead, in these figures we measure the performance of our learned optimizer applied to an i.i.d. sample of tasks from the meta-training distribution, applied for 10K steps for the first 3 phases, and 100K steps for the last phase. Losses are first normalized with the hand-designed normalizers described in Section~\ref{app:hand_designed_normalizer}, and then aggregated with a mean. Each dot denotes the average over 20 static configurations, with 2 dynamic configurations per static.
    \label{fig:meta_training_curve_loss}
    }
\end{figure}

\textbf{Phase 1}: Phase one consisted of the smallest sized problems. We use a total outer-batch size of 131,072 tasks which is made up of 2048 batches of outer-gradients, each of which being the average of 8 different static configurations, and vectorized over 8 dynamic configurations. We use a learning rate of 3\e-4.
This phase lasted for approximately 17 days.

\textbf{Phase 2}: Next, we switch to the mid-sized problem distribution. This was in an effort to better align the training distribution with more realistic problems as we started to notice meta-overfitting in evaluations. We additionally switch how gradients are aggregated, resulting in a smaller total batch size of 40,960, computed with 5120 batches of outer-gradients, where each outer-gradients only comes from one (possibly duplicated) static configuration, and again is averaged over 8 dynamic samples.

\textbf{Phase 3}: Upon noticing divergence in the evaluation tasks, we increased the batch size by a factor of 2 to 81,920 and reset training to an earlier checkpoint.

\textbf{Phase 4}: Finally, we noticed our learned optimizers performed poorly on long horizon problems. To solve this, we shifted the max problem length from 20K to 200K. This slowed down meta-training considerably, but dramatically improved performance in these settings. In this phase we also shifted the distribution of tasks to even larger problems.

\begin{figure}[t!]
    \centering
    \makebox[\textwidth]{%
    \begin{overpic}[width=1.0\textwidth]{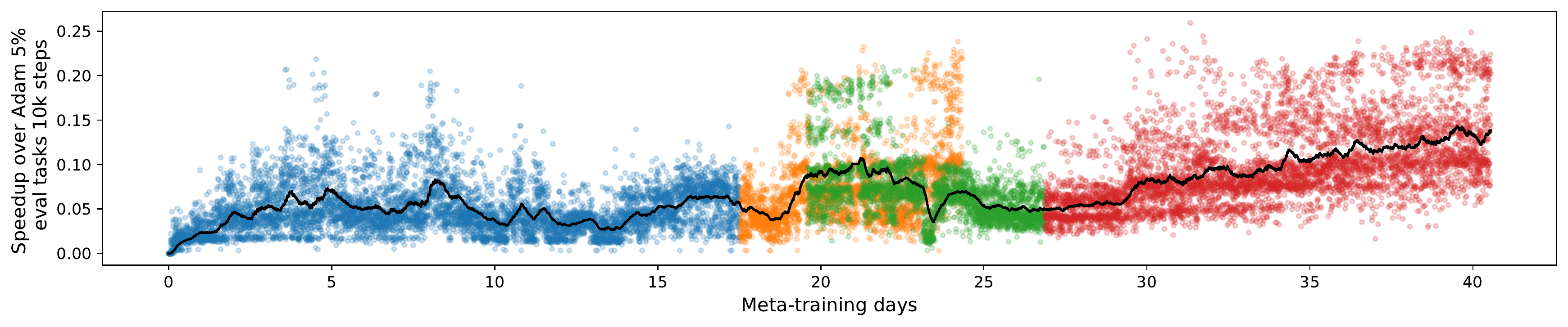}
    \end{overpic}
    }
    \makebox[\textwidth]{%
    \begin{overpic}[width=1.0\textwidth]{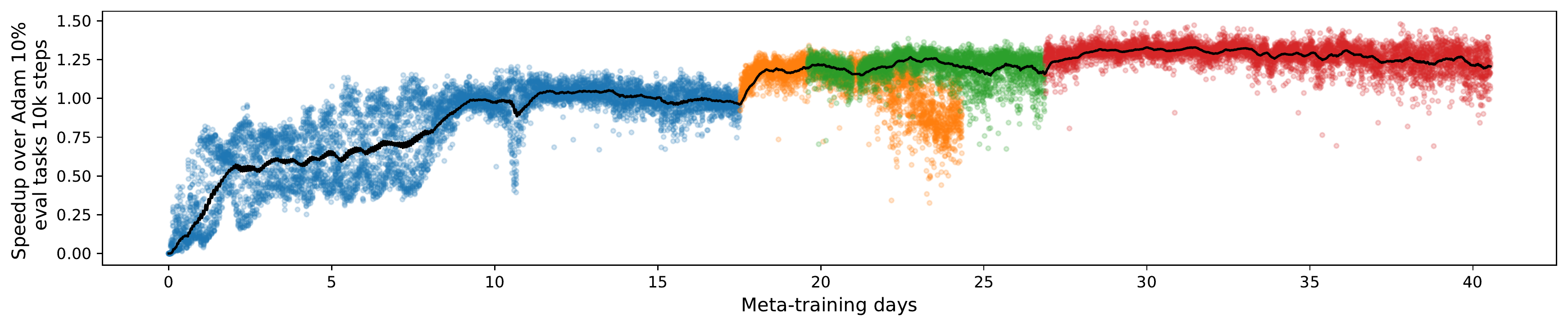}
    \end{overpic}
    }
    \makebox[\textwidth]{%
    \begin{overpic}[width=1.0\textwidth]{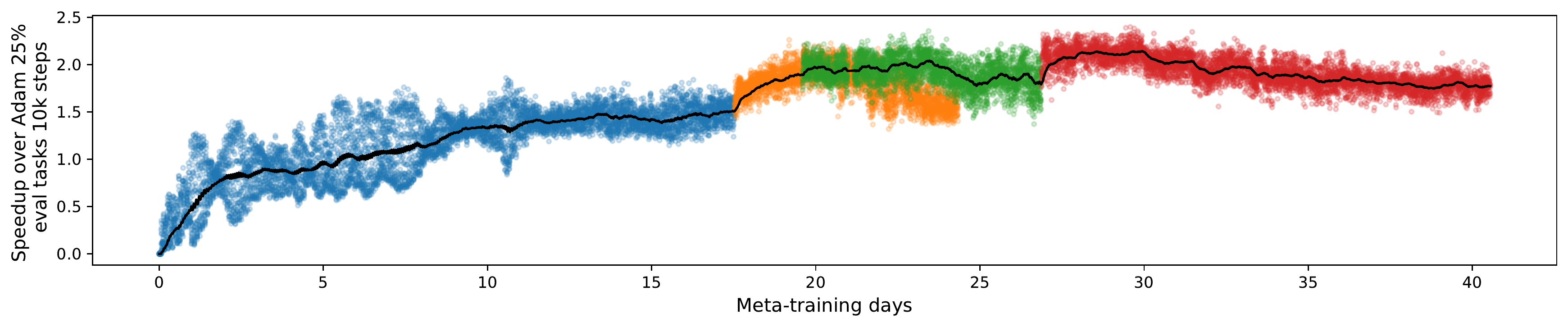}
    \end{overpic}
    }
    \makebox[\textwidth]{%
    \begin{overpic}[width=1.0\textwidth]{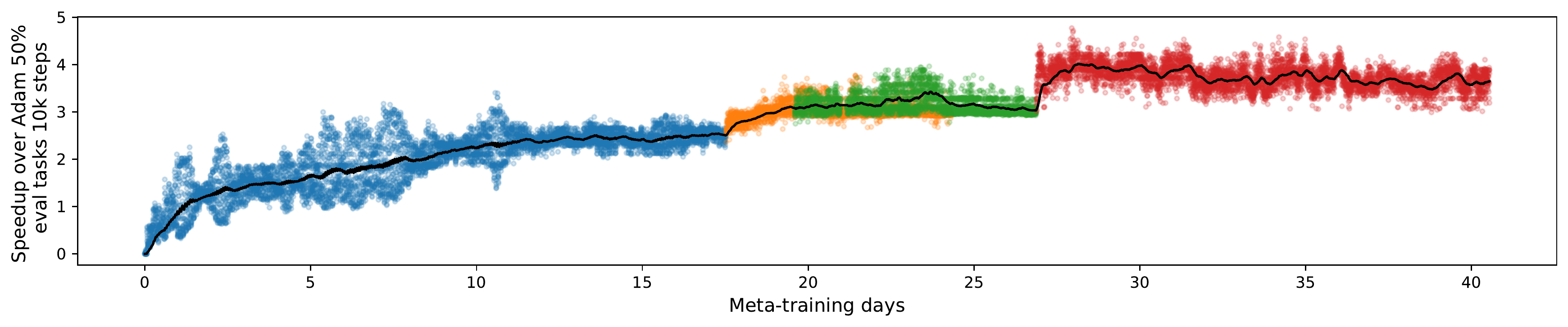}
    \end{overpic}
    }
    \makebox[\textwidth]{%
    \begin{overpic}[width=1.0\textwidth]{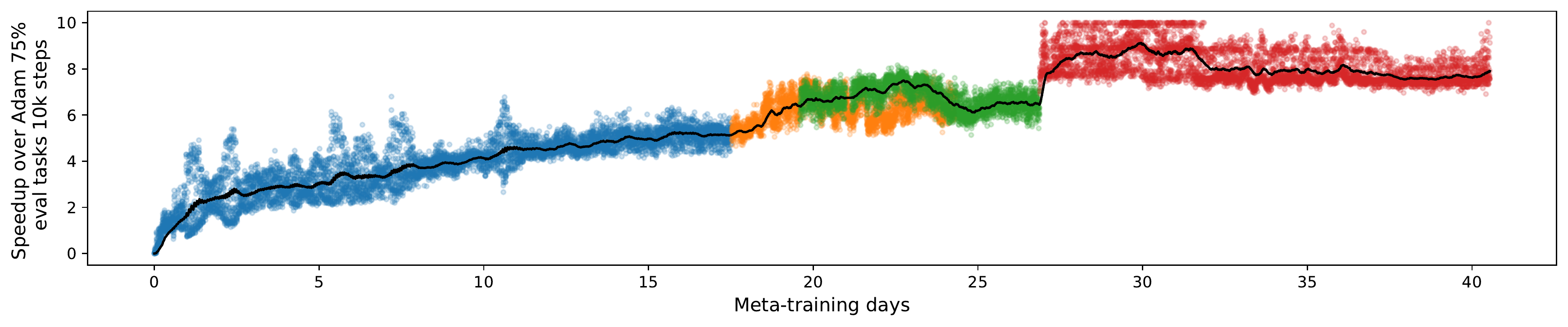}
    \end{overpic}
    }
    \caption{\textbf{Meta-training performance on fast evaluation set of 67 tasks.} Each row denotes a different percentile from top to bottom, 5\%, 10\%, 25\%, 50\%, 75\%. The settings used to train these models were changed 3 times, and different hyperparameter value training segments are denoted by different colors. In black we show an exponential moving average. We can see in phase 2 (orange) diverging in the lower percentile. As a fix for this we roll back and meta-train with a different configuration for phase 3.
    \label{fig:meta_training_eval_curves}
    }
\end{figure}

\begin{figure}[t!]
    \centering
    \makebox[\textwidth]{%
    \begin{overpic}[width=1.0\textwidth]{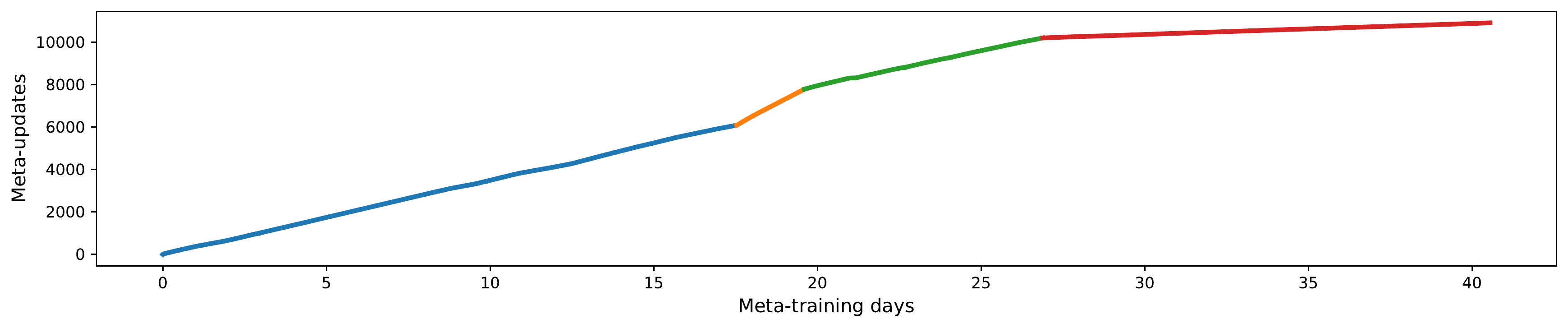}
    \end{overpic}
    }
    \makebox[\textwidth]{%
    \begin{overpic}[width=1.0\textwidth]{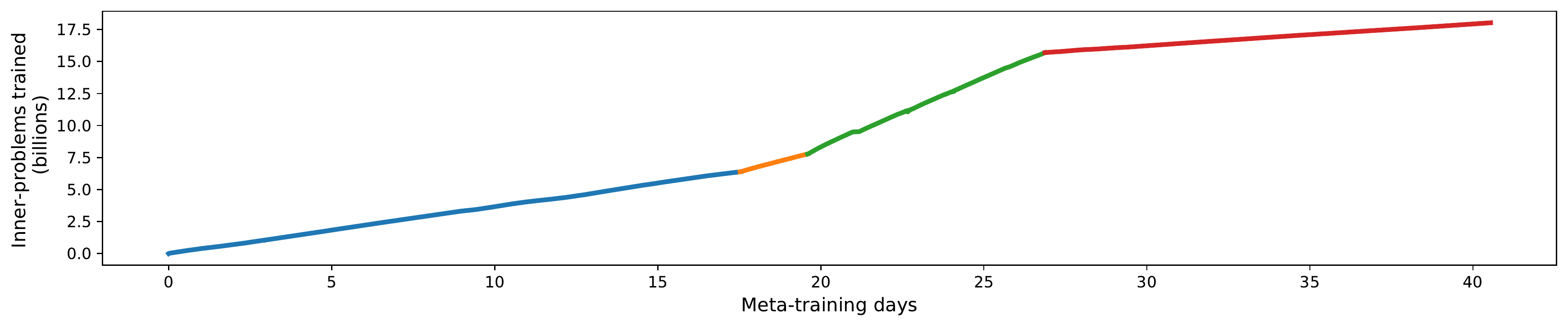}
    \end{overpic}
    }
    \makebox[\textwidth]{%
    \begin{overpic}[width=1.0\textwidth]{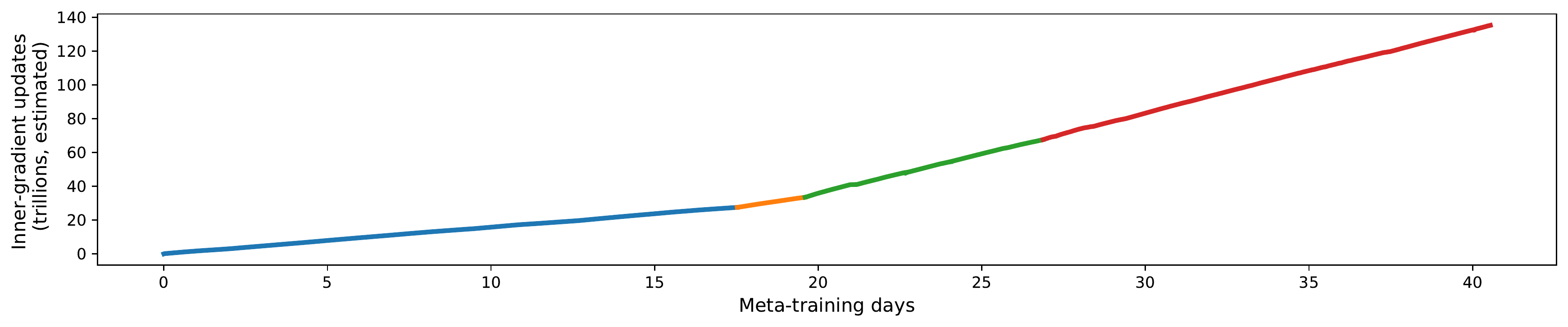}
    \end{overpic}
    }
    \caption{\textbf{Meta-updates, problems trained, and updates computed over meta-training time.} On the top row, we show the number of updates we apply to the learned optimizer weights. As we modify the batch size, and increase compute amounts the slope of this line changes. Next, we show the number of inner problems trained. When we increase the horizon length from 20K to 200K (phase 3, red) we see a dramatic slowdown as each task takes longer to train. Finally we show the number of inner-gradients computed, and the number of times our learned optimizer has been applied.
    \label{fig:meta_training_stats}
    }
\end{figure}

We visualize these different phases in a variety of ways.
First, in Figure~\ref{fig:meta_training_curve_loss} we show the closest thing to a meta-loss for each phase of training. In Figure~\ref{fig:meta_training_eval_curves} we visualize the evaluation performance of our learned optimizer on a set of hand-designed tasks which are faster to compute. We visualize performance at different percentiles of performance. These figures we often used the 10th percentile to guide much of our development as we found good performance on all tasks was more important than extremely fast training on a small number of tasks and often generalized better to larger problems.
Finally, in Figure~\ref{fig:meta_training_stats} we show statistics from meta-training including the number of outer-iterations, number of problems trained, and the number of times our learned optimizer has been applied.

\subsection{Areas of Improvement}
Our infrastructure is still in its infancy. There are several directions for improvement, which we detail below.

\begin{itemize}
    \item \textbf{TPU utilization.} Despite our effort, the efficiency of the matrix multiplication units (MXU) on the TPU across our cluster is relatively low (<10\%).
    This is due to the mismatch in hardware design: TPUs are designed for large computations, not small, highly iterative operations. Often our computations are limited by memory throughput.
    Counter-intuitively, GPUs are actually worse as kernel execution overhead dominates, even with automated kernel fusion provided by XLA.
    We expect custom kernels, with more operations fused, on GPU would be more efficient but to our knowledge there is no automated compiler stack that supports this, let alone that is integrated with JAX.
    Unlike traditional machine learning, compilers are almost a necessity given the sheer number of different tasks on which we meta-train.
    \item \textbf{Compile time overhead.} Each TPU machine compiles its own computation graph. This means while the compilation is happening no computation is occurring.
    We explored using an additional set of CPU only machines which cross-compiled JAX programs to run on TPU, but were unable to create a reliable or performant enough system at scale to be useful to our work.
    \item \textbf{Sensitivity to cluster status.} Because our clusters make use of preemptable hardware and asynchronous training, the training dynamics can change depending on the state of the cluster.
    We tried to mitigate this with larger outer batch sizes and lower staleness tolerance with mixed success.
    To make matters worse, these issues only becomes apparent after a long period of meta-training.
    We expect further iterations of this infrastructure is needed to be more synchronous in nature.
    Doing this, however, will require care, as unlike standard synchronous training, tasks vary greatly across machines.
\end{itemize}

\section{Experimental Details From Main Text}
In this section we describe the experimental details of the models presented in the main text.

\subsection{Evaluation Set Normalization} \label{app:fixed_task_normalizer}
Our normalization scheme is based on the calculation of how long it would have taken to reach the same loss using a set of baseline optimizers.
In the paper body, these baseline optimizers are learning rate-tuned Adam with learning rates picked every half power of 10. The baseline optimizers are run for 300K iterations which allows us to measure speedups up to 30x faster when the target model is run for 10K iterations.

In Section~\ref{app:extended_opt_benchmark} we also make use of a more sophisticated set of baseline optimizers: learning rate-tuned Adam, learning rate-tuned Adam with a learning rate warm up of 1K steps followed by a $1/\sqrt{t}$ decay (where $t$ is training iteration), Adam with an exponential decay (of rate 7\e-5, chosen so that the learning rate drops 3 orders of magnitude over 100K iterations), and fixed learning rate RAdam. Each optimizer family is run over 14 learning rates, with 5 random seeds apiece.

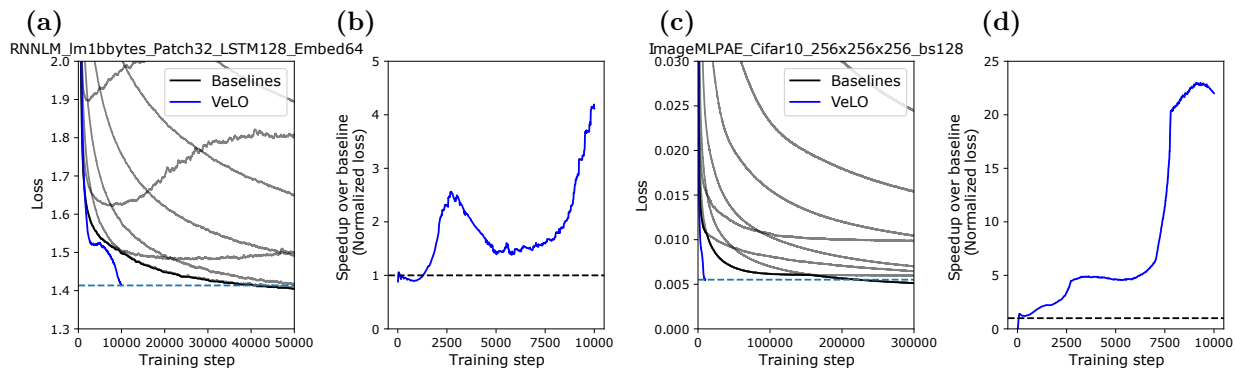
\begin{figure}
    \centering
    \makebox[\textwidth]{%
    \begin{overpic}[width=1.0\textwidth]{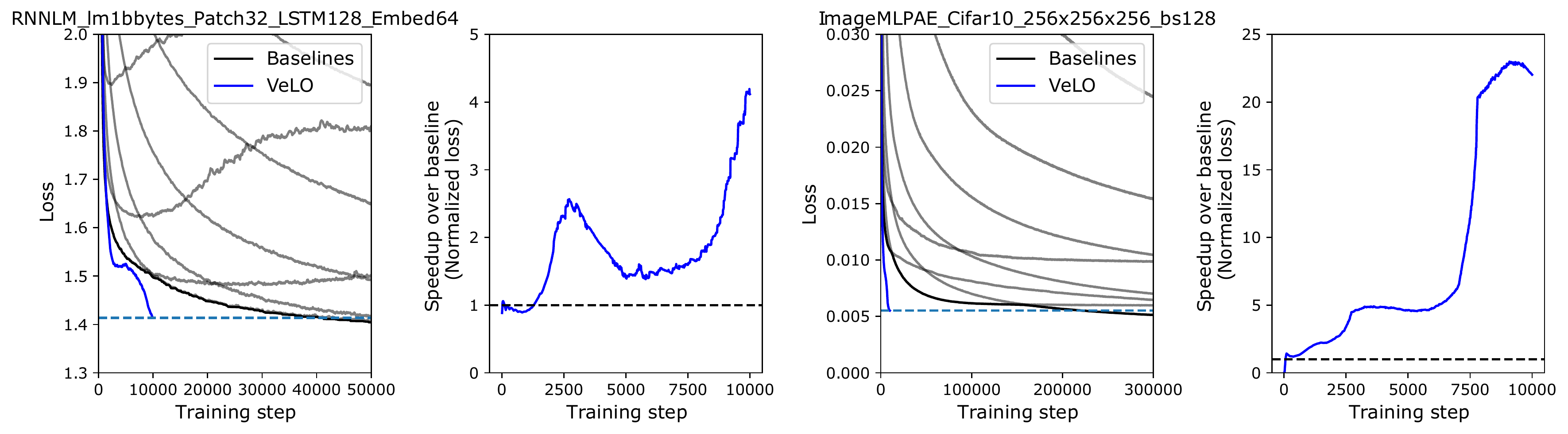}
    \put (2,28.0) {\textbf{\small(a)}}
    \put (27,28.0) {\textbf{\small(b)}}
    \put (53,28.0) {\textbf{\small(c)}}
    \put (79,28.0) {\textbf{\small(d)}}
    \end{overpic}
    }
    \vspace{-1em}
    \caption{\textbf{Visualization for how loss normalization works.} We show normalized loss values of our learned optimizer on two tasks using the Adam based normalizers. \textbf{(a,c)} we show unnormalized loss values. \textbf{(b,d)} we show normalized training curves. In gray we show each baseline learning curve, and in black we show the minimum over all baselines. In blue we see our target optimizer. In panel (a), it would take approximately 40K iterations for Adam to match the performance of VeLO at 10K iterations.
    This corresponds to a normalized loss value (panel (b)) of approximately 4.
    A value of 1.0 in this normalized view means optimization takes the same amount of time as the best baseline.
    \label{fig:normalize_vis}
    }
\end{figure}

We demonstrate how this normalization works in Figure~\ref{fig:normalize_vis} for 2 different tasks. First we load all baselines, average over the random seeds run, and compute an exponential moving averages to reduce noise.
All these curves are shown in gray in Figure~\ref{fig:normalize_vis}ac. Next, we compute the minimum value at every timepoint (shown in black). The process of normalization is then an interpolation to this curve.

An implementation of these normalizations can be found \href{https://github.com/google/learned_optimization/blob/main/learned_optimization/baselines/normalizers.py}{here}.

\subsection{Details on VeLOdrome Evaluation Set}
Our full set of tasks has 10 task families:

\begin{enumerate}
    \item \href{https://github.com/google/learned_optimization/blob/84dcc4aedca079e5ace851a2839d60663821e2ea/learned_optimization/tasks/fixed/image_mlp.py}{ImageMLP}: Image classification with MLP.
    \item \href{https://github.com/google/learned_optimization/blob/b1d8267c5e513a4112e7422b98bacc16e1f0e844/learned_optimization/tasks/fixed/conv.py}{Conv}: Convolutional neural networks.
    \item \href{https://github.com/google/learned_optimization/blob/b1d8267c5e513a4112e7422b98bacc16e1f0e844/learned_optimization/tasks/fixed/image_mlp_ae.py}{ImageMLPAE}: auto-encoders with MLP on images.
    \item \href{https://github.com/google/learned_optimization/blob/b1d8267c5e513a4112e7422b98bacc16e1f0e844/learned_optimization/tasks/fixed/mlp_mixer.py}{MLP-Mixer}: MLP-Mixer based image classification tasks.
    \item \href{https://github.com/google/learned_optimization/blob/b1d8267c5e513a4112e7422b98bacc16e1f0e844/learned_optimization/tasks/fixed/resnet.py}{ResNet}: ResNet based tasks.
    \item \href{https://github.com/google/learned_optimization/blob/b1d8267c5e513a4112e7422b98bacc16e1f0e844/learned_optimization/tasks/fixed/rnn_lm.py}{RNNLM} RNN based language modeling tasks.
    \item \href{https://github.com/google/learned_optimization/blob/b1d8267c5e513a4112e7422b98bacc16e1f0e844/learned_optimization/tasks/fixed/transformer_lm.py}{TransformerLM}: Transformer language modeling tasks.
    \item \href{https://github.com/google/learned_optimization/blob/b1d8267c5e513a4112e7422b98bacc16e1f0e844/learned_optimization/tasks/fixed/vit.py}{VIT}: Tasks which leverage Vision Transformers for image classification.
    \item \href{https://github.com/google/learned_optimization/blob/b1d8267c5e513a4112e7422b98bacc16e1f0e844/learned_optimization/tasks/fixed/lopt.py}{LOpt}: Tasks which train learned optimizers.
    \item \href{https://github.com/google/learned_optimization/blob/b1d8267c5e513a4112e7422b98bacc16e1f0e844/learned_optimization/tasks/fixed/es_wrapped.py}{ESWrapped} Other tasks where instead of using gradients with backprop, we use ES.
\end{enumerate}

The test set of problems we evaluate in the paper body (which we refer to as VeLOdrome-83) consist of 83 tasks from the ImageMLP, Conv, ImageMLPAE, ResNet, RNNLM, TransformerLM, and VIT tasks described above---it excludes the learned optimization and ES-based tasks, which are not common training tasks. We also evaluate performance on a full set of 109 tasks (which includes learned optimization and ES tasks, and which we refer to as VeLOdrome-109), and the results are plotted in the next section. 

\paragraph{Details on best and worst case problems.}
In Figure~\ref{fig:two_good_two_bad} we show show results for the 
\begin{center} 
\texttt{ImageMLP\_Cifar10\_128x128\_Dropout08\_Relu\_MSE}, \texttt{ImageMLPAE\_Mnist\_128x32x128\_bs128},\\ \texttt{RNNLM\_lm1bbytes\_Patch32\_LSTM128\_Embed64}, \texttt{RNNLM\_lm1b32k\_Patch32\_LSTM256\_Embed128}
\end{center}
tasks from left to right.
These problems were selected by first sorting the normalized performance and looking at the best and worst performance.

\subsection{MLCommons Tasks}

Below, we include details on the MLCommons tasks. 

\paragraph{Vision Transformer on ImageNet.}
We test a Vision Transformer trained on ImageNet with batch size 1024. The implementation and Adam baseline closely follow \citet{beyer2022better}. We find VeLO minimizes training loss more effectively than all of the 20 random trials, and performs better on validation loss as well (see Appendix~\ref{app:extended_mlcommons}).

\paragraph{ResNet50 on ImageNet.}
We test a ResNet-50 trained on ImageNet with batch size of 1024.
The model definition follows that in the MLPerf Training Benchmark \citep{mattson2020mlperf}. 
We find VeLO minimizes training loss as effectively as the best of the 20 random trials, though it doesn't generalize as well on the validation set of data.

\paragraph{Machine translation WMT17 German-English with a Transformer.}
We evaluate a sequence-to-sequence, Transformer-based translation model trained on German to English from WMT17~\citep{bojar-EtAl:2017:WMT1} with a batch size of 256. VeLO outperforms all baseline trials, 
despite their being no sequence-to-sequence models nor translation tasks in the meta-training task distribution.

\paragraph{LibriSpeech DeepSpeech.}
We test a DeepSpeech model~\citep{hannun2014deep} trained on LibriSpeech~\citep{panayotov2015librispeech} with batch size 256 using a ctc loss~\citep{graves2006connectionist}. The DeepSpeech model is a bi-directional RNN based architecture with gated recurrent units (GRUs) incorporating batch normalization. Despite no speech data being used for meta-training, VeLO outperforms all of the Adam trials.

\paragraph{LibriSpeech Conformer.}
We investigate a conformer model~\citep{gulati2020conformer} trained on LibriSpeech with a batch size of 512 using a CTC loss\footnote{Our results are based on an earlier version of the model which uses a batch size of 512, a smaller encoder dimension of 256, and uses a word piece model tokenizer instead of a sentence piece tokenizer.}). 
Despite no speech data being used for meta-training, VeLO performs comparably to the best Adam trial.

\paragraph{Graph neural network (GNN) on molecular property prediction.}
\label{sec mlc gnn}
Finally, we test on a graph property prediction task, ogbg\_molpcba~\citep{hu2020open}, with a batch size of 512.
Here we find VeLO performs at roughly the 50th percentile of the Adam search space -- qualitatively worse than it performed on all other MLCommons tasks. 
We hypothesize that this is due to the task being qualitatively very different than any in the meta-training distribution.
VeLO was not trained to optimize GNNs, nor on any problems involving scientific data or graph data. See Section \ref{sec chem gnns} for further exploration of GNN tasks. 

\subsection{Details on Out of Distribution Problems}

\subsubsection{NERF} The NERF problems we tested were based off of the \href{https://github.com/google-research/google-research/tree/master/jaxnerf}{JAXNerf package} \citep{jaxnerf2020github}. We use 2 datasets from the Blender family of data.
The code for these tasks is located \href{https://github.com/google/learned_optimization/blob/a42b207b9ef5395fdf7e64978c26379ccba4e264/learned_optimization/research/jaxnerf/jaxnerf.py}{here}. The tasks we test on are \texttt{JAXNeRF\_ShipBlenderTask} and \texttt{JAXNeRF\_LegoBlenderTask}. 

\subsubsection{MLP-Mixer} The MLP-Mixer tasks used are based on \href{https://github.com/google-research/vision_transformer}{code} released with \citet{tolstikhin2021mlp}. We test on 2 smaller mixer variants both with patch sizes of 16x16.
The code and configurations for our models can be found \href{https://github.com/google/learned_optimization/blob/4447601601f04e3dc7b1aecd7854d44526593752/learned_optimization/tasks/fixed/mlp_mixer.py}{here}. The tasks we test on are \texttt{MLPMixer\_ImageNet64\_tiny16} and \texttt{MLPMixer\_ImageNet64\_small16}.

\subsubsection{Object Detection}

For the results on object detection, we utilize a re-implementation of Faster R-CNN \citep{ren2015faster} and evaluate using the COCO dataset \citep{lin2014microsoft}. By default, we train for 22500 steps. After 500 steps of linear warmup, the learning is decreased by a factor of 10x after 15000 and 20000 steps. The default optimizer is SGD with a momentum of 0.9. Weight decay of 4\e-5 is also utilized. Models are trained with a batch size of 64.

\subsubsection{Large Language Models}
\label{sec:llm}

The language models are decoder-only Transformers \citep{vaswani2017attention} trained on the C4 dataset \citep{raffel2020exploring} using Adafactor \citep{shazeer2018adafactor}. We train using the Pax framework (a JAX version of \cite{shen2019lingvo}). The 100M model consisted of 16 layers with model size 768, hidden size 3072, and 8 heads, while the 8B model consisted of 32 layers with model size 4096, hidden size 24576, and 32 heads. All models used a vocab size of 32000, ``pre''-LayerNorm, and ReLU activations.

The Adafactor optimizer used in the 100M baselines performed linear warmup to a peak learning rate of 6.4\e-3 over the first 4000 steps (2000 for batch 4M and 1000 for batch 8M) followed by cosine decay (this schedule was tuned); the optimizer in the 8B baseline performed linear warmup to a peak learning rate of 1\e-4 over the first 4000 steps followed by exponential decay. All Adafactor baselines used beta1 of 0.9, Adam decay of 0.99, and weight decay of 1\e-3, and clipped gradients to a norm of 5.0.

Due to Pax's use of ``scan over layers'', VeLO was forced to optimize each layer independently (i.e. with no signal from other layers other than the global loss).

\paragraph{Adding weight decay to VeLO.} For larger experiments, parameter magnitudes grew extremely large, leading to numerical issues. This was resolved by adding in a small amount of weight decay. As opposed to directly decaying parameter values (as in e.g. AdamW \citep{loshchilov2017decoupled}), we added an auxiliary loss function penalizing the L2 norm of the parameter values. This auxiliary loss thereby modified the gradients fed into VeLO. 

\subsubsection{Graph NN} \label{app:gnn_materials}
The graph neural networks are 3-layer message passing neural networks. Each node represents an element; prior to any message-passing, all one-hot nodes are embedded into vectors of size 128.
All edges are embedded versions of the inter-atomic distance with size 128.
By default, 2 layer MLPs with swish nonlinearities process messages, and segment sums are used to aggregate the varying lengths. To get appropriately-scaled activations after message passing, we also divide messages by the average length across the dataset. The output arises from a global representation, which is connected to all edges and nodes and updated at every layer. 
The Adam baseline uses a constant learning rate, tuned at every half integer of powers of 10, from 1\e-3 to 1\e-5.

\paragraph{Generalization performance.}
\begin{figure}
    \centering
    \makebox[\textwidth]{%
    \begin{overpic}[width=0.6\textwidth]{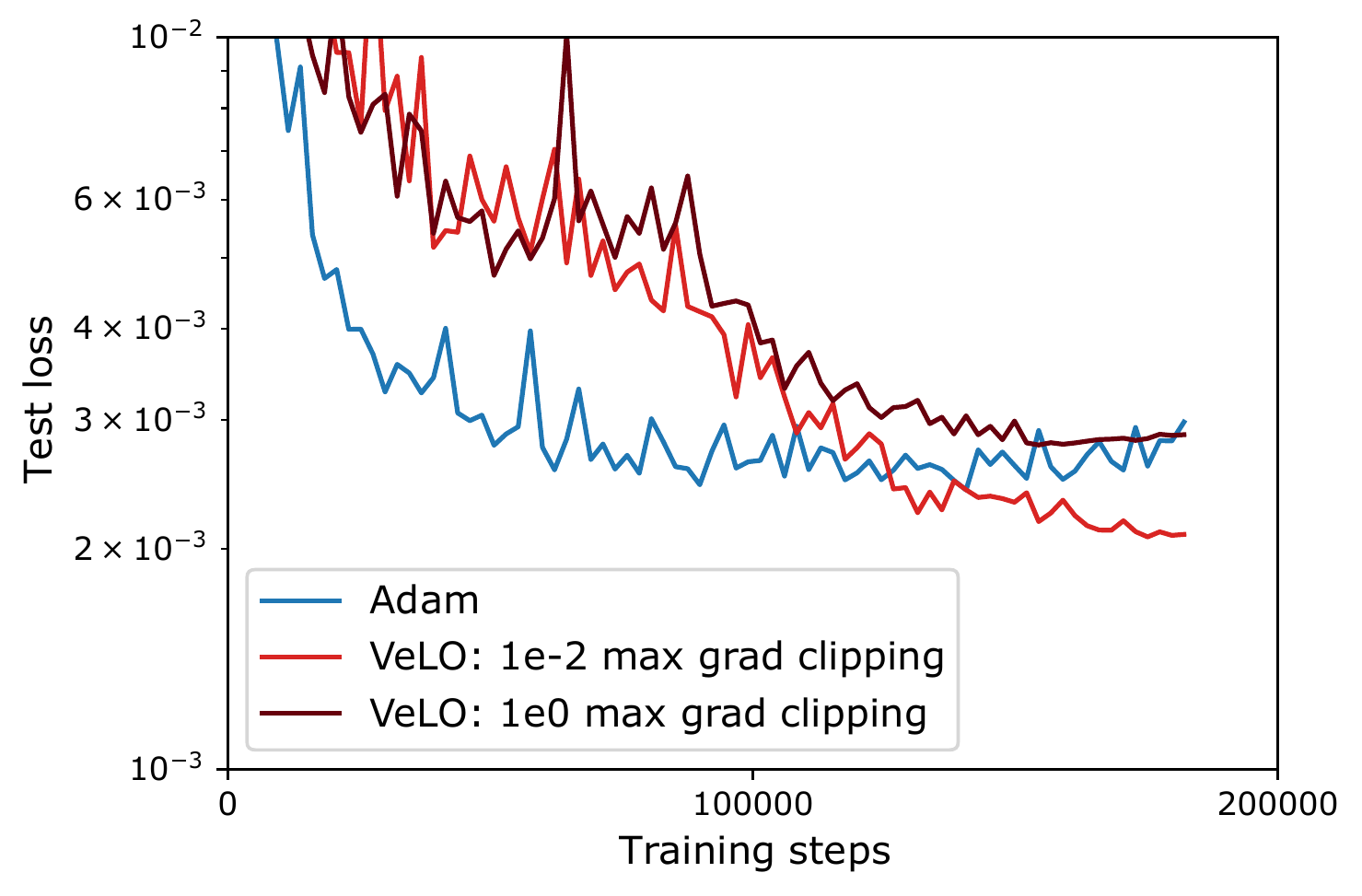}
    \end{overpic}
    }
    \caption{\textbf{Training Graph NN on chemical material prediction.} Naively applying our learned optimizers, while minimizing training loss well (See Figure~\ref{fig:physics_graph_nets}) do not generalize well. When leveraging additional gradient clipping, our learned optimizers also generalize better than Adam.
    \label{fig:pax_300m}
    }
\end{figure}

Graph neural networks also present a difficult test-bed for the learned optimizers, as these models are out-of-distribution for meta-training and often do not see monotonic improvements in train-loss corresponding to improvements in generalization performance.
Figure \ref{fig:physics_graph_nets} showed that these models are able to perform comparably to Adam on train performance for these GNNs; however, test set performance lags behind.
This can be fixed by clipping gradient norms such that the maximum norm is 1\e-2 before input to the learned optimizer.

\begin{figure}
    \centering
    \begin{overpic}[width=0.75\textwidth]{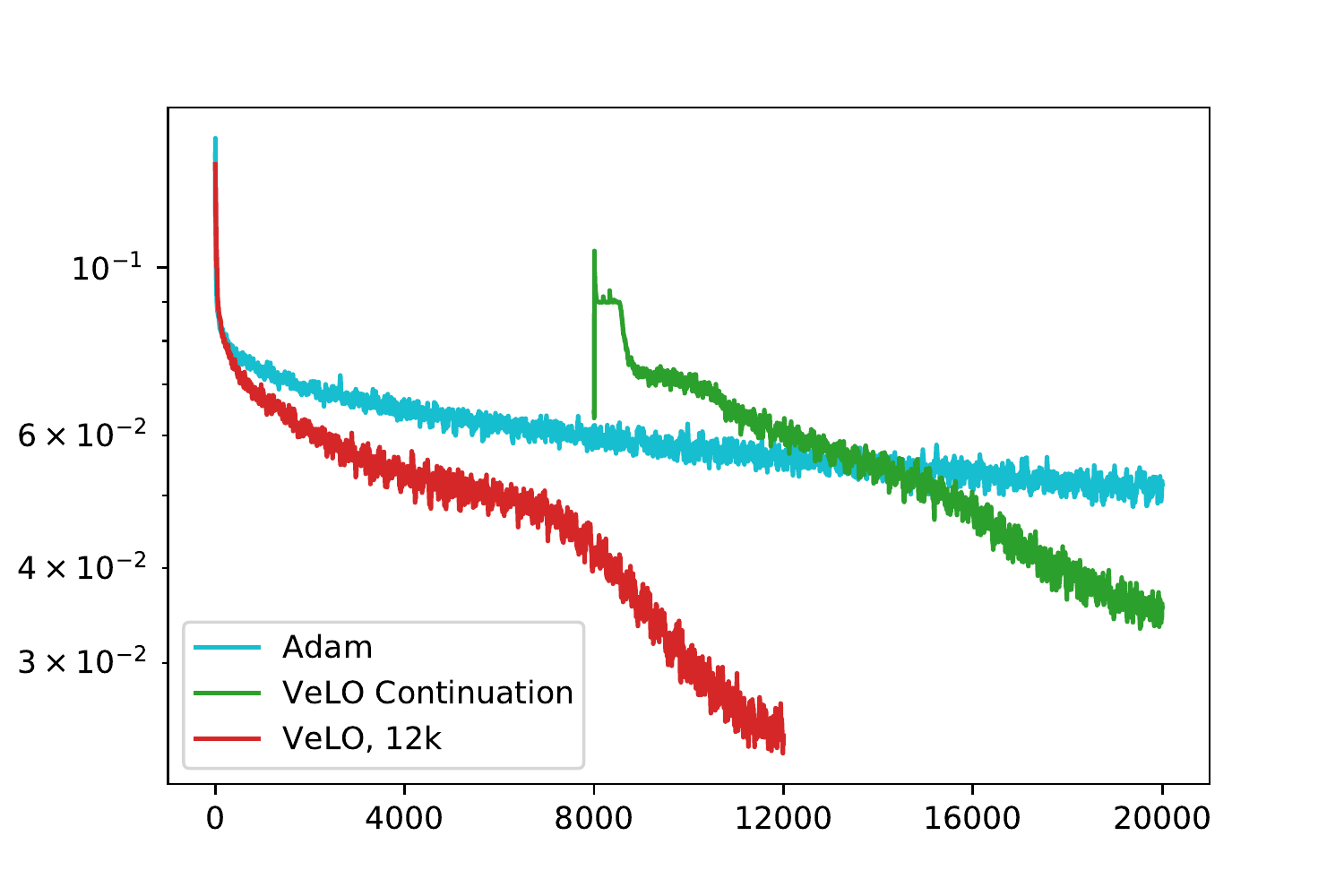}
    \end{overpic}
    \caption{\textbf{Extending an Adam training run with VeLO.} We train an image MLP model on CIFAR10 for 8,000 steps with Adam with a learning rate of 0.001 ("Adam") and then extend training for an additional 12,000 steps with VeLO ("VeLO Continuation"). The loss immediately spikes demonstrating VeLO's lack of robustness to begin training from a non-random state. While the continuation ultimately performs better than continuing with Adam, it performs worse than training using VeLO from a random initialization for the same number of steps ("VeLO, 12K").}
    \label{fig:adam_continuation}
\end{figure}

\subsection{Details on Optimizers Learn to Make Use of Finite Training Time}
A Colab to reproduce these experiments is available \href{https://colab.research.google.com/drive/1qaLWUUnAN_aQdp_fdkkydiDREsW5oGa9?usp=sharing}{here}.
The MLP and convolutional network tasks used were
\texttt{ImageMLP\_FashionMnist\_Relu128x128} and \texttt{Conv\_Cifar10\_32x64x64}.

\section{Extended Experimental Results}
\label{sec:ext_exp_res}

\subsection{Learned Optimizers Take Smaller Steps for Larger Models}
\begin{figure}
    \centering
    \makebox[\textwidth]{%
    \begin{overpic}[width=0.6\textwidth]{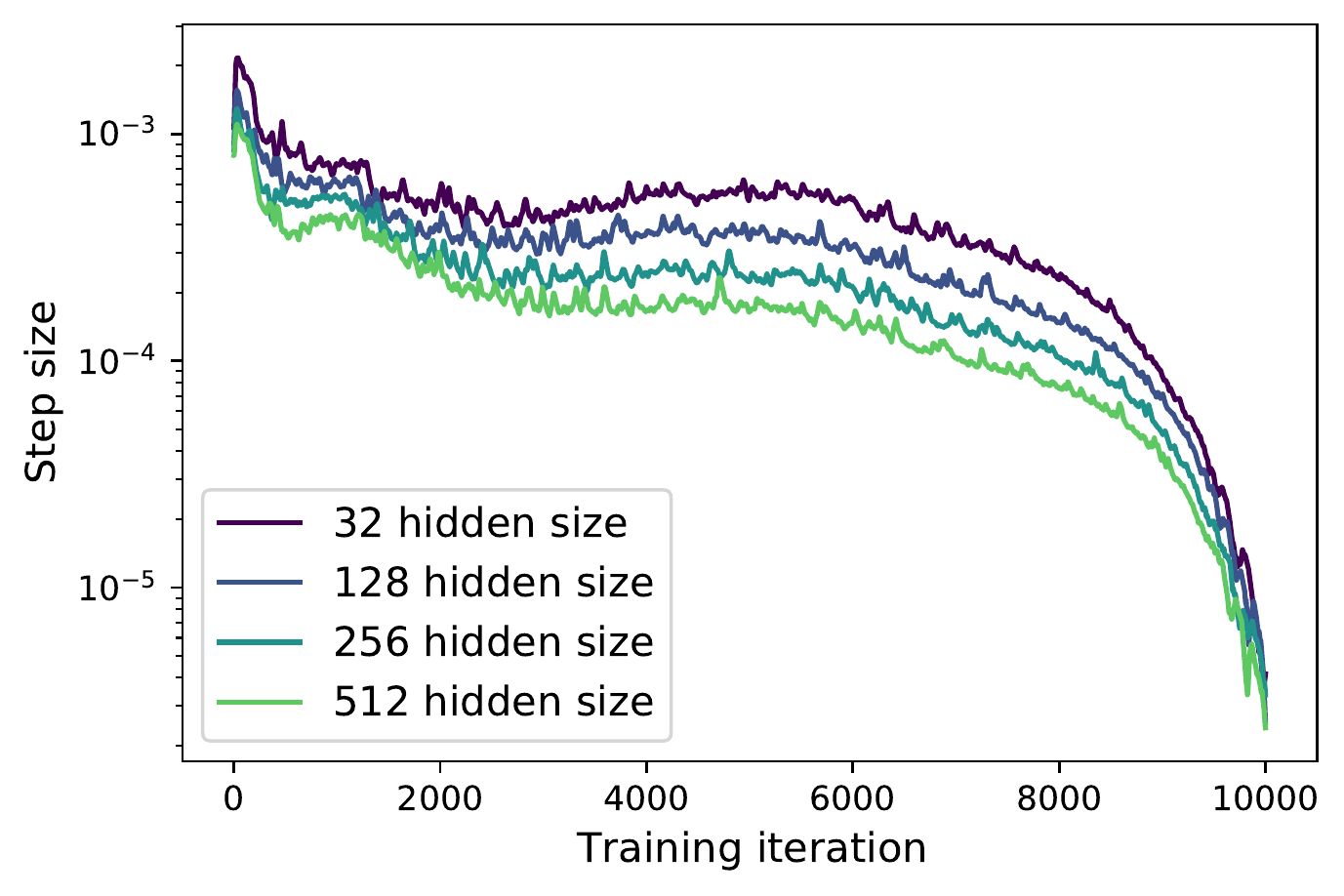}
    \end{overpic}
    }
    \caption{\textbf{Step sizes of learned optimizer for different sized Transformers.} We train 4 Transformers of varying sizes and monitor the step sizes throughout training. We find the learned optimizer takes smaller steps as the model size grows larger.
    \label{app:fig:diff_model_size}
    }
\end{figure}

When training models of increasing sizes with hand-designed optimizers one must decrease learning rate accordingly~\citep{krizhevsky2014one, goyal2017accurate}.
We sought to see if VeLO exhibits this property.
To this end, we train a 5 layer Transformer with 32, 128, 256, and 512 (corresponding to the \texttt{TransformerLM\_LM1B\_5layer\_<size>width} tasks) units with our learned optimizer and monitor the size of the step taken by the learned optimizer (Figure~\ref{app:fig:diff_model_size}).
We find that our learned optimizer shrinks the step size as a function of model size.
Somewhat unexpectedly, we find the gap between the smallest and largest step sizes grows in the middle of training.
This is unlike any hand-designed optimizers we are aware of.

\subsection{Learned Optimizers Leverage Loss Features}
\begin{figure}[t!]
    \centering
    \makebox[\textwidth]{%
    \begin{overpic}[width=0.7\textwidth]{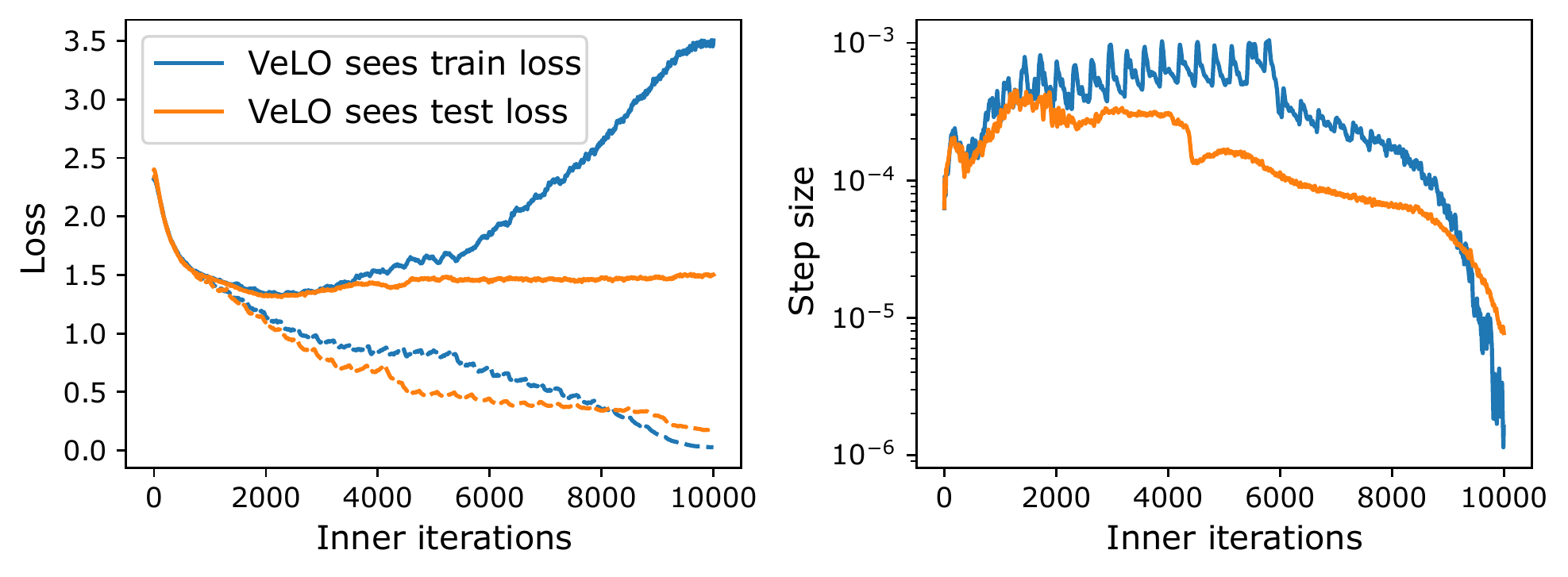}
    \put (9,9.0) {\textbf{(a)}}
    \put (59.5,9.0) {\textbf{(b)}}
    \end{overpic}
    }
    \caption{\textbf{VeLO uses loss features in sensible way.} We train an MLP with 2 hidden layers of size with 1024 hidden units and either feed in training loss, or validation loss into the learned optimizer.
    In \textbf{(a)} we show the training (dashed lines) and validation loss (solid lines) over the course of training.
    In \textbf{(b)} we show the average step size taken.
    When passing in validation loss the learned optimizer takes smaller steps which we believe slow down learning and thus slow down overfitting.
    \label{fig:loss_features_generalize}
    }
\end{figure}

Unlike standard hand-designed optimizers, our learned optimizer takes the current training loss value as a feature.
We hypothesized that this would allow the optimizer to detect if it was diverging and adjust step sizes accordingly.
To test this hypothesis, we modify the loss value fed into the learned optimizer from train loss to validation loss (Figure~\ref{fig:loss_features_generalize}).
When the optimizer sees train loss, it continues to decrease the training loss (overfitting on validation loss).
When the optimizer sees validation loss, however, the loss values at the start of training look similar until overfitting starts occurring at which point the learned optimizer adjusts it's step size and seemingly doesn't overfit as much.
This figure surprised us, as during meta-training there was no notion of validation loss, but the ability to prevent overfitting seemingly emerges for free -- simply by swapping what the optimizer sees.
This experiment was performed with the \texttt{ImageMLP\_Cifar10\_1024x1024\_Relu} task.

\subsection{Learned Optimizers Learn Implicit Step Size Schedule} \label{app:additional_step_size}
\begin{figure}[t!]
    \centering
    \makebox[\textwidth]{%
    \begin{overpic}[width=1.0\textwidth]{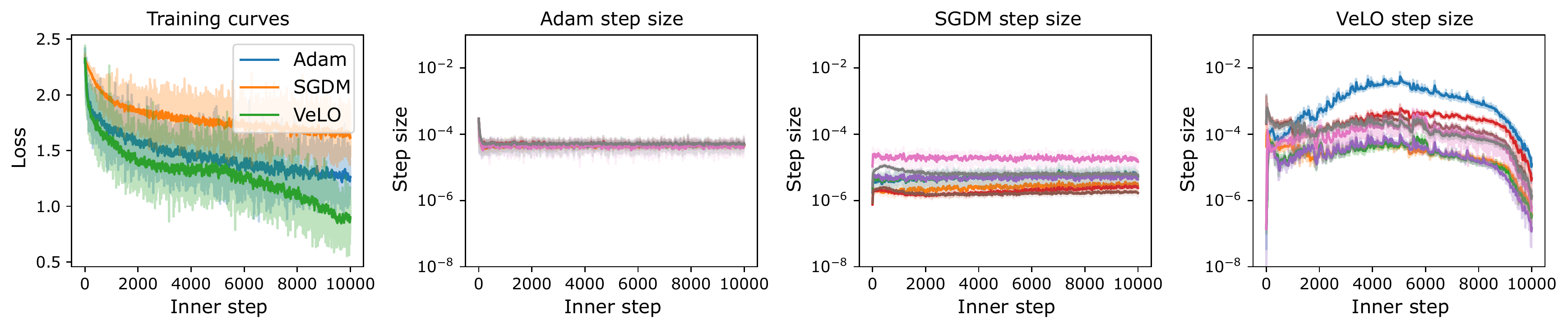}
    \end{overpic}
    }
    \makebox[\textwidth]{%
    \begin{overpic}[width=1.0\textwidth]{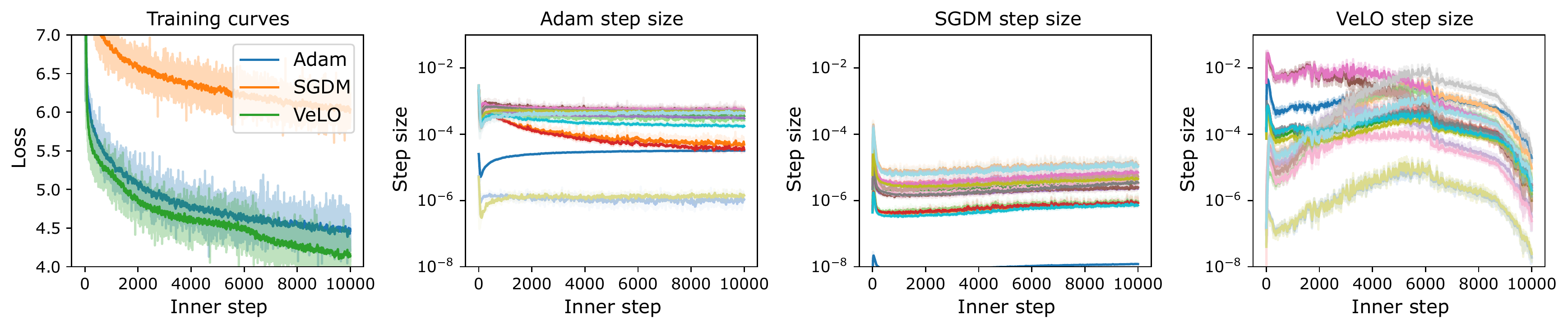}
    \end{overpic}
    }
    \caption{\textbf{Step size schedule of learned optimizer.} We show training curves, followed by the mean absolute step sizes for each optimizer and each tensor in the target problem. In the top row we show a 3 layer MLP with Tanh activations trained on CIFAR100, in the bottom row we show a 5 layer Transformer.
    \label{fig:step_size}
    }
\end{figure}

In the main text in Section~\ref{sec:different_sized_steps}, we show how the step sizes change for the 
\begin{center}
\texttt{ImageMLP\_Cifar10\_128x128x128\_Tanh\_bs64}, \texttt{Conv\_Cifar10\_32x64x64\_batchnorm}, and \texttt{TransformerLM\_LM1B\_MultiRuntime\_5} 
\end{center}
tasks. In this section, we additionally show training curves, and the step sizes taken by Adam and SGDM for reference. Over the course of training, we monitor step size on two problems: a 3 layer MLP with tanh activations trained on CIFAR100 (the \texttt{ImageMLP\_Cifar10\_128x128x128\_Tanh\_bs64} task), and a 5 layer Transformer with 128 hidden units (the \texttt{TransformerLM\_LM1B\_5layer\_128width} task).
For each optimizer, we monitor the mean absolute step size for each layer independently every step (Figure~\ref{fig:step_size}).
We find our learned optimizers first learn a rapid step size increase, followed by rapid decay, followed by a slow increase until near the end of training where the step size decays again.
This resembles some common practices in modern machine learning---learning rate warm ups and decays---but the slow increase in step size through the first half of training is something we have not seen before.
The step sizes learned by our learned optimizer are extremely different than both Adam and SGD with momentum exhibiting much more variability across tensors and in time.
In contrast, the step sizes of Adam and SGDM are relatively consistant both in time, and across the different tensors.

\begin{figure}[t!]
    \centering
    \makebox[\textwidth]{%
    \begin{overpic}[width=1.0\textwidth]{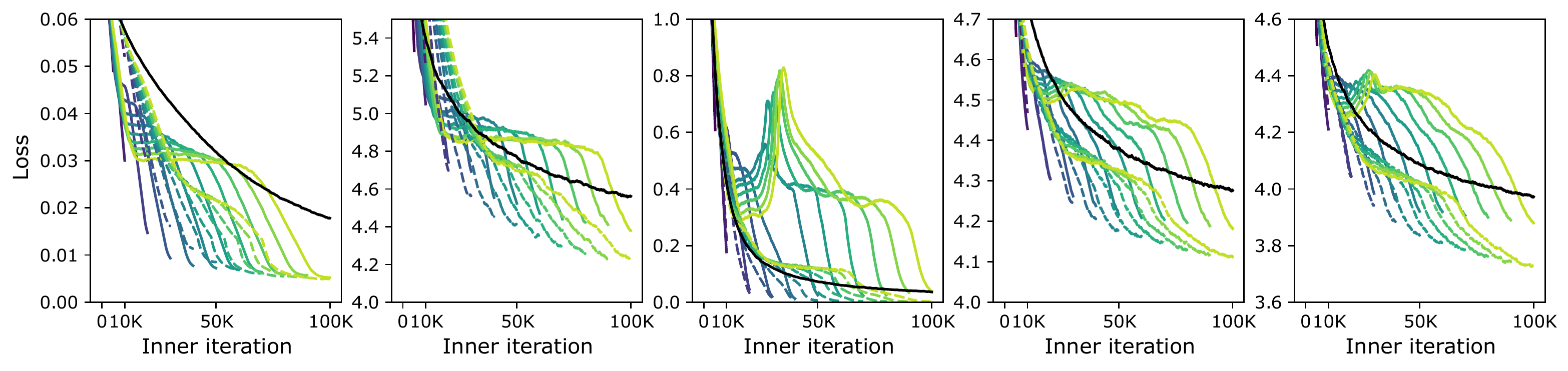}
    \put (18,20.0) {\textbf{\small(a)}}
    \put (37,20.0) {\textbf{\small(b)}}
    \put (56.5,20.0) {\textbf{\small(c)}}
    \put (75.5,20.0) {\textbf{\small(d)}}
    \put (95,20.0) {\textbf{\small(e)}}
    \end{overpic}
    }
    \makebox[\textwidth]{%
    \begin{overpic}[width=1.0\textwidth]{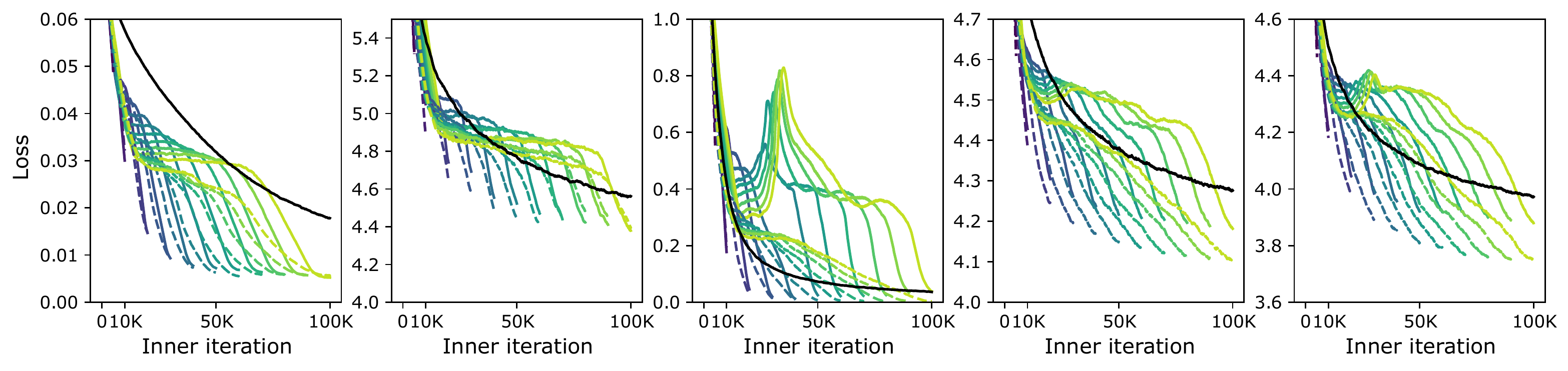}
    \end{overpic}
    }
    \caption{\textbf{Longer horizon meta-training} We show training curves for different lengths of inner training beyond the meta-training distribution.
    In solid color, we show learned optimizer (color denoting different length of training), and in solid black we show the best of learning rate-tuned Adam. Here the learned optimizer is applied out of distribution as it was only meta-trained for at max 20K steps).
    In dashed we show interventions which make the learned optimizer work for longer horizons.
    \textbf{Top:} We show the same learned optimizer, but only applying the learned optimizer every 10 iterations with gradients accumulated from the previous 10 iterations.
    \textbf{Bottom:} We show a new learned optimizer, finetuned to work well up to 200K iterations. This finetuning removes the instability we see in (c) and dramatically improves performance in (d) and (e).
    These results are shown (from left to right) on the following models:
    A MLP trained on CIFAR10 (\texttt{ImageMLP\_Cifar10\_128x128x128\_Relu\_MSE}),
    a convnet trained on 32x32 ImageNet (\texttt{Conv\_imagenet32\_16\_32x64x64}),
    a MLP trained on CIFAR10 with batch norm (\texttt{ImageMLP\_Cifar10\_128x128x128\_BatchNorm\_Relu}),
    a smaller Transformer on LM1B (\texttt{TransformerLM\_LM1B\_MultiRuntime\_8}),
    and a different Transformer on LM1B (\texttt{TransformerLM\_LM1B\_5layer\_128width}).
    \label{fig:num_step_aware_longer}
    }
\end{figure}
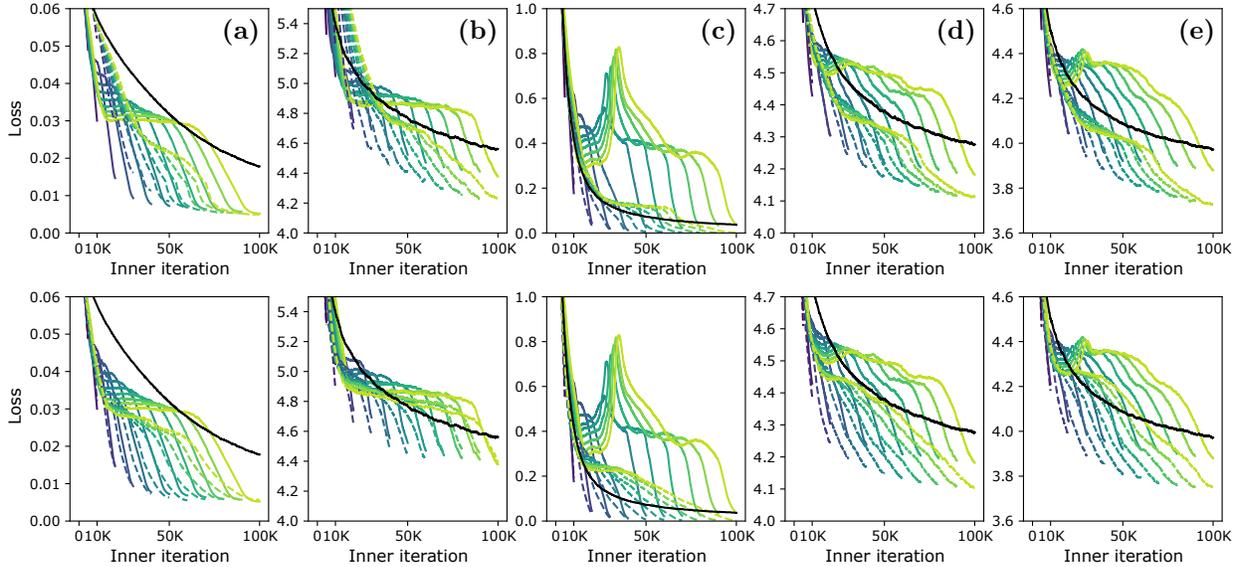

\subsection{Sensitivity to Number of Training Iterations With Longer Horizons}
For a large portion of our meta-training procedure we used a sampled inner-problem training lengths between 200-20,000 iterations.
To further test the impact of horizon, we test our learned optimizers out of distribution on a range of considerably longer horizons up to 100K iterations to see if this behavior extrapolates (Figure~\ref{fig:num_step_aware_longer}, solid lines).
For some target tasks, we find our learned optimizers generalize well to this extended training, while for others we see unexpected behavior and non-monotonic learning curves suggesting some instability. We discuss two potential workarounds to this.

First, one can employ gradient accumulation, in particular accumulate 10 iterations worth of gradients before each step by the learned optimizer.
This lowers the max target length the optimizer sees to 10K iterations (which is inside the meta-training regime) and results in more stable (and more performant) meta-training curves (Figure~\ref{fig:num_step_aware_longer}, top).
Second, we can simply fine tune the learned optimizer on longer horizon problems, 200-200K iterations. While this is quite expensive, it does fix this problem (results in Figure~\ref{fig:num_step_aware_longer}, bottom).

\begin{figure}[t!]
    \centering
    \makebox[\textwidth]{%
    \begin{overpic}[width=1.0\textwidth]{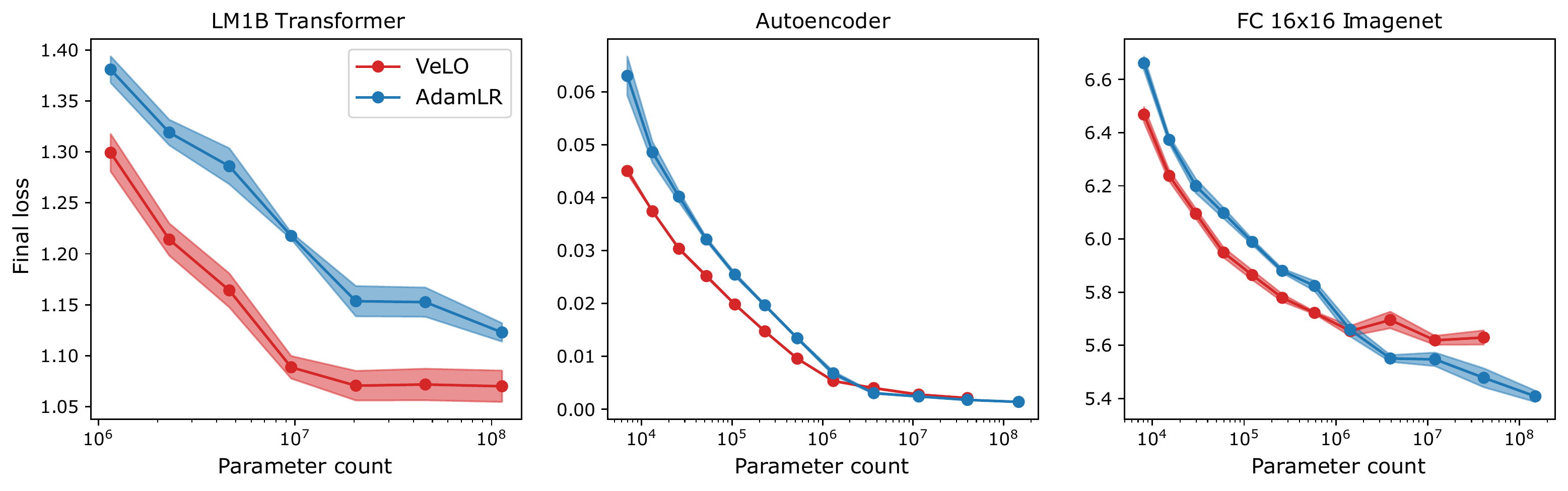}
    \put (7.5,7.0) {\textbf{\small(a)}}
    \put (40,7.0) {\textbf{\small(b)}}
    \put (73.5,7.0) {\textbf{\small(c)}}
    \end{overpic}
    }
    \caption{\textbf{Optimizer performance as a function of inner problem scale.} In the best case, our optimizers provide a constant offset in optimization performance. In the worse case, our optimizer underperforms learning rate Adam. Error bars denote standard error across 5 random seeds.
    \label{fig:param_scaling}
    }
\end{figure}

\subsection{Effects of Scaling Model Size}
In this section we explore how performant VeLO is as a function of the hidden size of a variety of different models.
We test a 5 layer Transformer with hidden sizes between 16 and 16,384, an auto-encoder with hidden sizes between 4 and 65,536, and a fully connected network doing image classification on 16x16 ImageNet with power of 2 hidden sizes between 4 and 8192 (Figure \ref{fig:param_scaling}).
For each model family, at each model size, we train with learning rate-tuned Adam, taking the best performing learning rate for each model scale, as well with a single trial of our learned optimizer.
All models are trained for 10K iterations.

For the Transformer problem (Figure~\ref{fig:param_scaling}a), we see a similar slope of performance improvement scale, but with a constant shift.
At the 10 million parameter scale, VeLO optimizes better than the 100 million parameter Transformer with tuned Adam.
For the auto-encoder and image classification network we see a similar shift for smaller models.
As the model grows, however, learning rate-tuned Adam starts to match VeLO or outperform.
We suspect this is due to these large models being increasingly far from the meta-training distribution.

\subsection{Exploration Into Outputs of Per-Tensor LSTM}
In this section we probe how the per-tensor LSTM behaves.
To do this, for $\sim$100 different, smaller scale tasks we record the output of the per-tensor LSTM (the components used to linearly interpolate the per-parameter models) for each tensor, at every iteration.
This yields 35 million data points.

\begin{figure}[t!]
    \centering
    \makebox[\textwidth]{%
    \begin{overpic}[width=1.\textwidth]{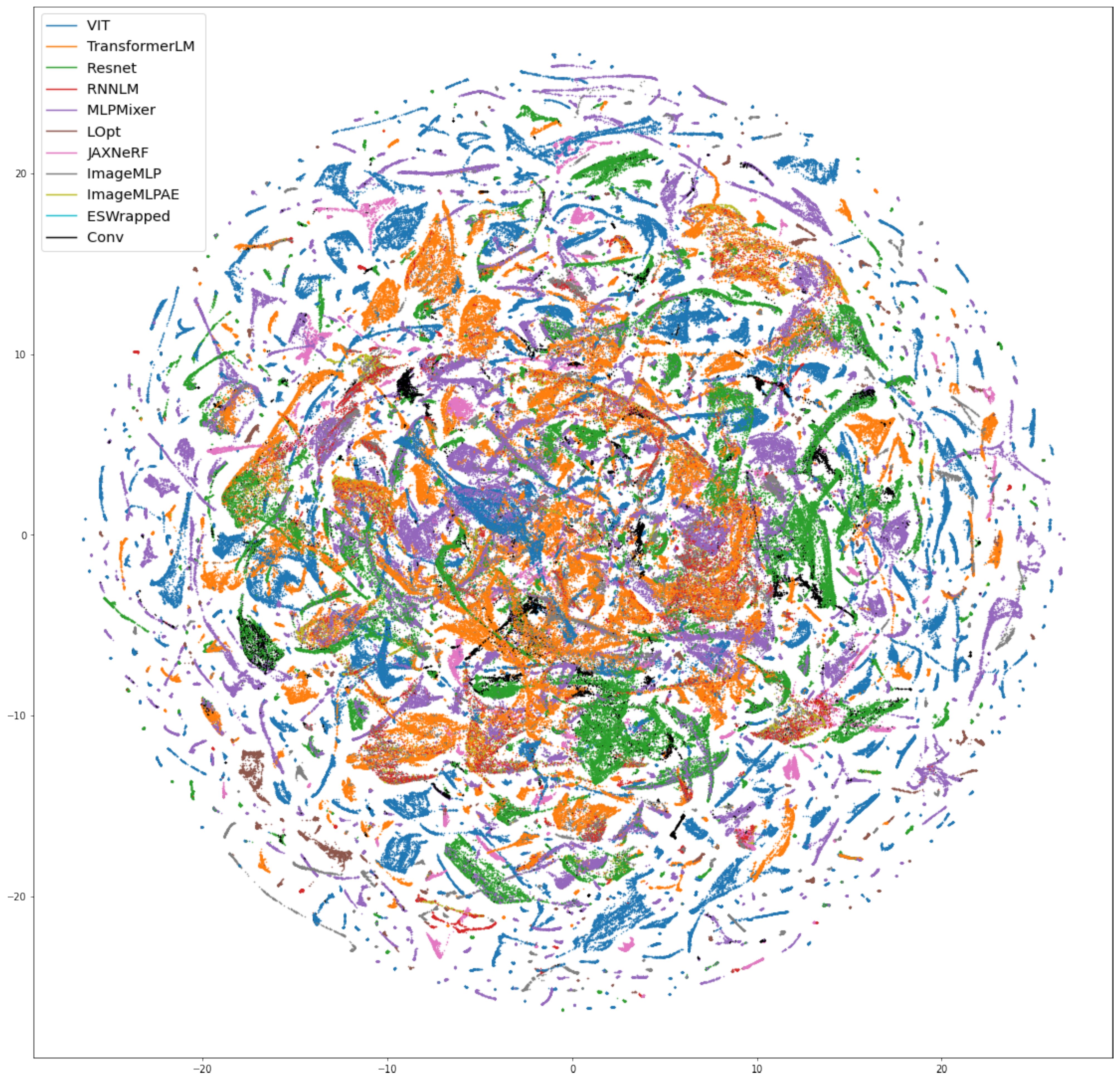}
    \end{overpic}
    }
    \caption{\textbf{t-SNE of per-tensor LSTM outputs. Colored by task kind.} We find similar tasks are grouped together suggesting similar outputs of the per-tensor LSTM. For a larger size see \url{https://storage.googleapis.com/gresearch/learned_optimization/figs/tsne_tasks.png}
    \label{fig:tsne_task}
    }
\end{figure}

\begin{figure}[t!]
    \centering
    \makebox[\textwidth]{%
    \begin{overpic}[width=1.\textwidth]{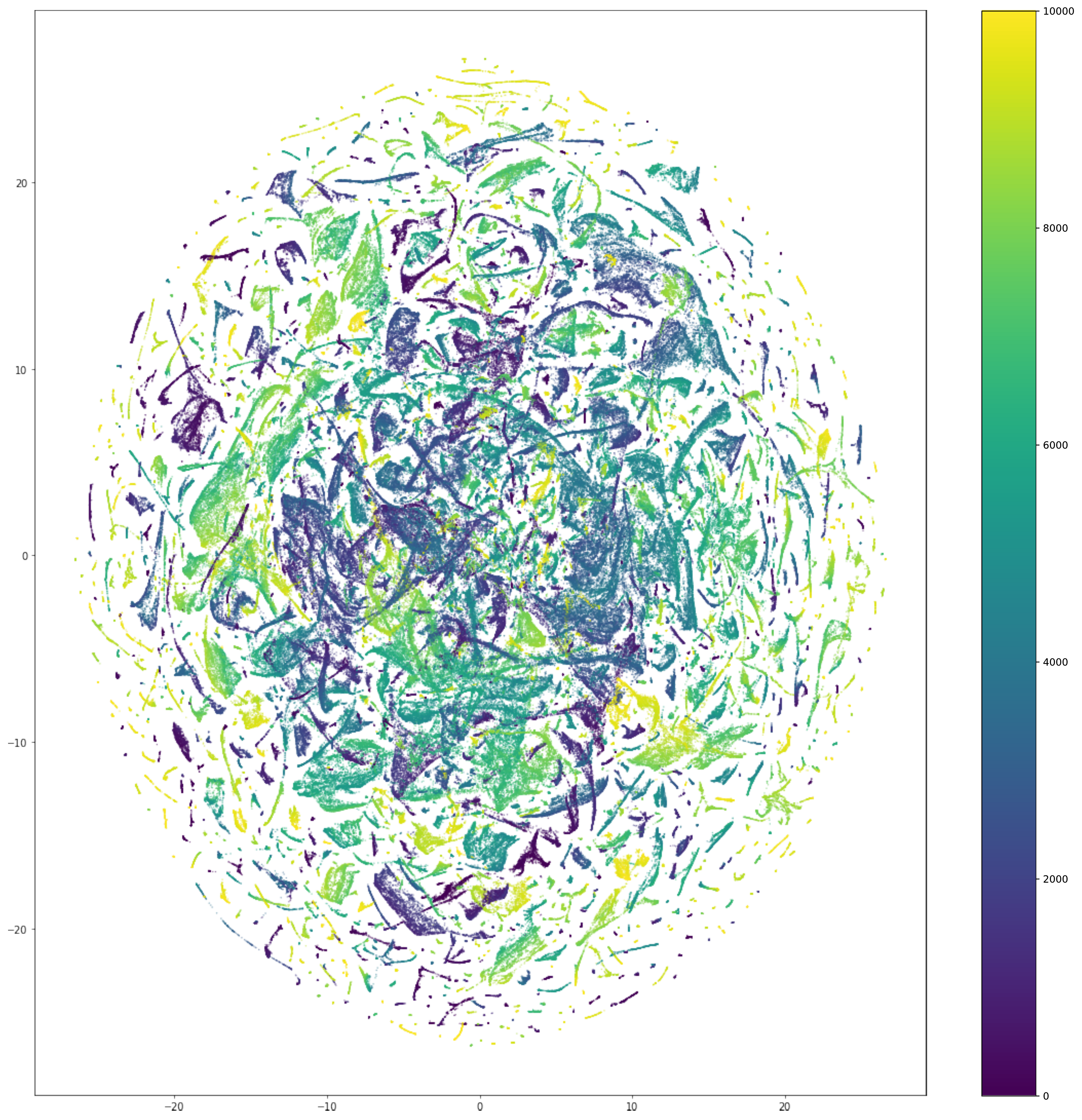}
    \end{overpic}
    }
    \caption{\textbf{TSNE of per-tensor LSTM outputs. Colored by training iteration.} We find similar times are grouped together suggesting the outputs of the per-tensor LSTM are similar. For a larger size see \url{https://storage.googleapis.com/gresearch/learned_optimization/figs/tsne_steps.png}
    \label{fig:tsne_step}
    }
\end{figure}

First, we subsample these datapoints to 800K for compute reasons, and apply t-SNE~\citep{maaten2008visualizing}. Once transformed, we plot the results coloring both by task kind (Figure~\ref{fig:tsne_task}) and by the iteration through training (Figure~\ref{fig:tsne_step}).
These figures suggest that this per-tensor LSTM performs differently for different kinds of problems, and that the behavior of the learned optimizer (as measured by these per-tensor outputs) varies as a function of time.

\subsection{MLCommons: Validation Performance and Hyperparameter Sensitivity} \label{app:extended_mlcommons}
\begin{figure}[t!]
    \centering
    \makebox[\textwidth]{%
    \begin{overpic}[width=\textwidth]{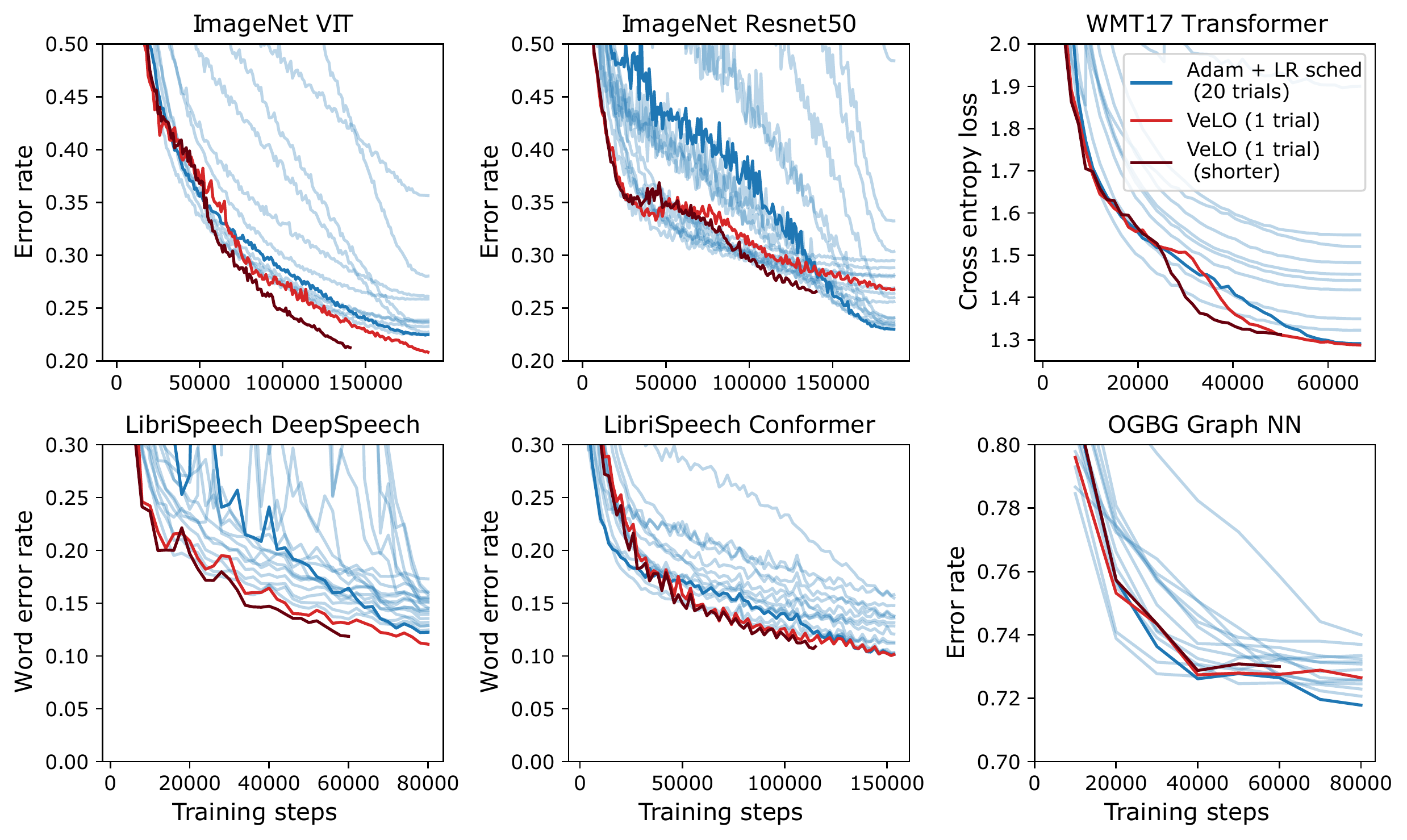}
    \put (8.5,36.0) {\textbf{\small(a)}}
    \put (42,36.0) {\textbf{\small(b)}}
    \put (75,36.0) {\textbf{\small(c)}}
    \put (8.5,7.5) {\textbf{\small(d)}}
    \put (42,7.5) {\textbf{\small(e)}}
    \put (75,7.5) {\textbf{\small(f)}}
    \end{overpic}
    }
    \caption{\textbf{Learned optimizer validation performance on six MLCommons workloads.} Our learned optimizers generalize well on most tasks with the exception of ResNet50 where we severely over fit. 
    \label{fig:init2winit_valid}
    }
\end{figure}

In this section we additionally show validation performance of our learned optimizers on six MLCommons workloads (Figure~\ref{fig:init2winit_valid}).
Despite only being meta-trained targeting train loss, our optimizers also generalize quite well.
We substantially outperform the baseline on ViT and DeepSpeech, match on WMT17 and Conformer, and perform substantially worse on ResNet50.
We hypothesize this poor performance is due to the limited data-augmentation used in this model.

\begin{figure}[t!]
    \centering
    \makebox[\textwidth]{%
    \begin{overpic}[width=\textwidth]{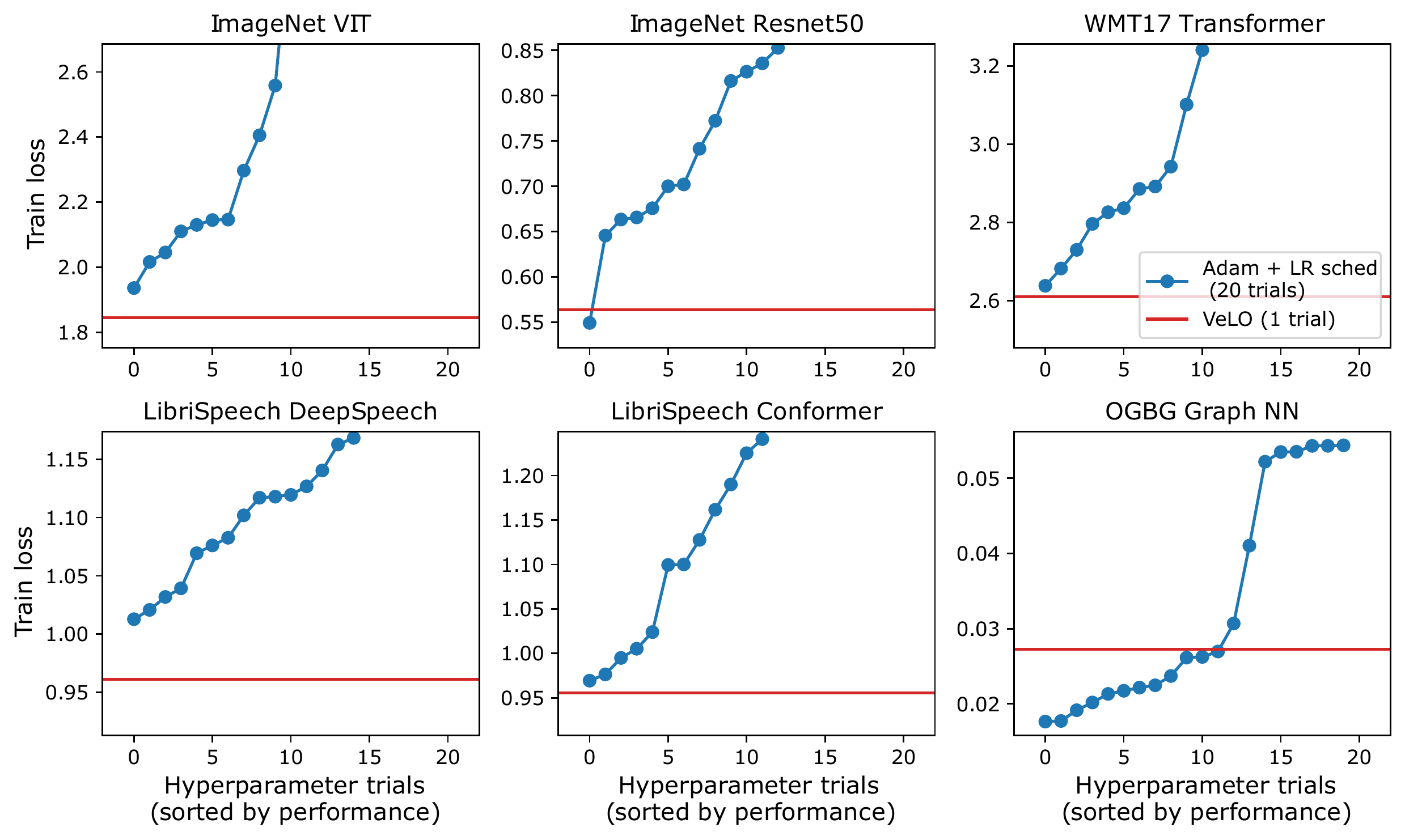}
    \put (8.5,54.0) {\textbf{\small(a)}}
    \put (41,54.0) {\textbf{\small(b)}}
    \put (73.5,54.0) {\textbf{\small(c)}}
    \put (8.5,26) {\textbf{\small(d)}}
    \put (41,26) {\textbf{\small(e)}}
    \put (73.5,26) {\textbf{\small(f)}}
    \end{overpic}
    }
    \caption{\textbf{Hyperparameter sensitivity on six MLCommons workloads.} We show the performance achieved at the end of training with each of the 20 Adam trials along with the performance of our learned optimizer. In all but OGBG our optimizer is competitive with or outperforms this baseline.
    \label{fig:init2winit_train_hparams}
    }
\end{figure}

Next, we show the final train loss of our learned optimizers versus the final loss achieved by each of the random Adam trials (Figure~\ref{fig:init2winit_train_hparams}).
This view allows a better understanding of the sensitivity to hyperparameter values and the relative strength of the hyperparameter-tuning free VeLO optimizer.
VeLO outperforms all of the 20 hyperparameter trials for all but the Graph NN task and a single ResNet50 task.

For all MLCommons figures we make use of \href{https://github.com/google/learned_optimization/blob/b1d8267c5e513a4112e7422b98bacc16e1f0e844/learned_optimization/research/general_lopt/pretrained_optimizers.py#L108}{this optimizer checkpoint}.

\subsection{MLCommons: Larger Batch Sizes}
\begin{figure}[t!]
    \centering
    \makebox[\textwidth]{%
    \begin{overpic}[width=0.9\textwidth]{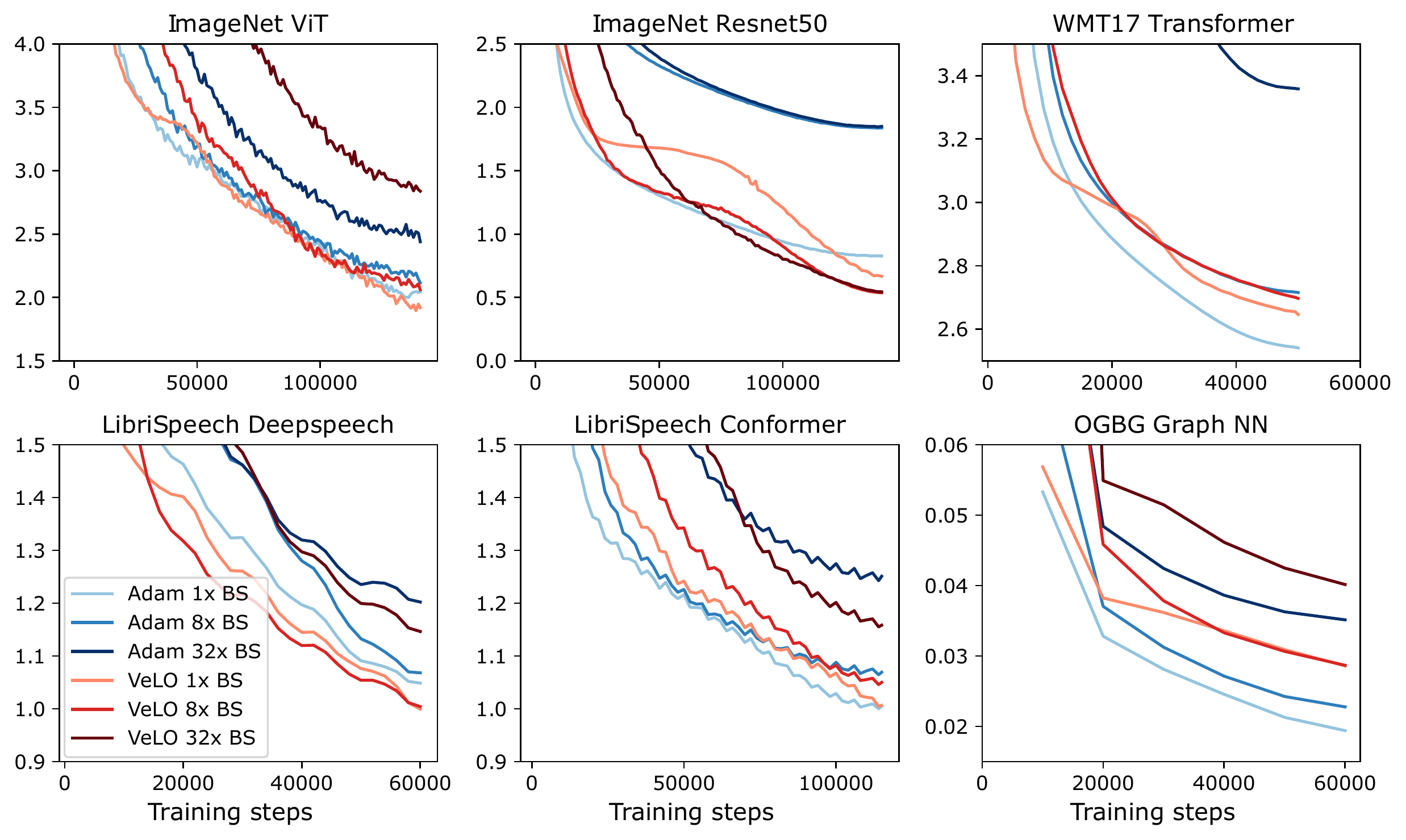}
    \put (26,53.5) {\textbf{\small(a)}}
    \put (59,53.5) {\textbf{\small(b)}}
    \put (92,53.5) {\textbf{\small(c)}}
    \put (26,25.0) {\textbf{\small(d)}}
    \put (59,25.0) {\textbf{\small(e)}}
    \put (92,25.0) {\textbf{\small(f)}}
    \end{overpic}
    }
    \caption{\textbf{MLCommons problems at different batch sizes.} We train both our learned optimizer and Adam with 1x, 8x, and 32x larger batch sizes plotted so that each curve uses the same amount of examples.
    In WMT17, the largest batch size learned optimizer scores around 4.7 and thus is not shown on this figure.
    \label{fig:init2winit_bs}
    }
\end{figure}

Next we probe how VeLO behaves at different batch sizes on the MLCommons problems.
We take the same 6 tasks, and train both our learned optimizers, and a tuned Adam baseline at different batch sizes implemented via gradient accumulation.
We test an 8x and 32x batch size. For each Adam baseline we select the best model from a random grid search of 100 points. The search space searches for the learning rate in the range $[10^{-5}, 10^{-1}]$ on a logarithmic scale, weight decay in the range $[10^{-5}, 1.0]$ on a logarithmic scale, $(1-\beta_1,1-\beta_2)$ in the range $[10^{-3}, 1]$ on a logarithmic scale and the number of warm up steps as a fraction of the total training steps in the set $\{0.05, 0.1, 0.15, 0.2, 0.25\}$.
Results are shown in Figure~\ref{fig:init2winit_bs}.

We find in all non cases other than the Graph NN, our 8x batch size outperforms this Adam baseline.
In the DeepSpeech and ResNet models, there is no change in performance suggesting that our optimizers have a higher critical batch size for these tasks.
In 1x, we note our WMT17 performance is below baseline where as in the previous MLCommons experiments we outperform.
We believe this is due to the increased number of trials (100, rather than 20).
For the 32x batch size, both Adam and our learned optimizer performs poorly.
Our learned optimizers perform better than Adam on DeepSpeech, Conformer, and ResNet50, but worse on VIT, WMT17, and the Graph NN.

\subsection{Additional Tasks for Batch Size Scaling} \label{app:experiment_more_bs}
\begin{figure}[t!]
    \centering
    \makebox[\textwidth]{%
    \begin{overpic}[width=0.9\textwidth]{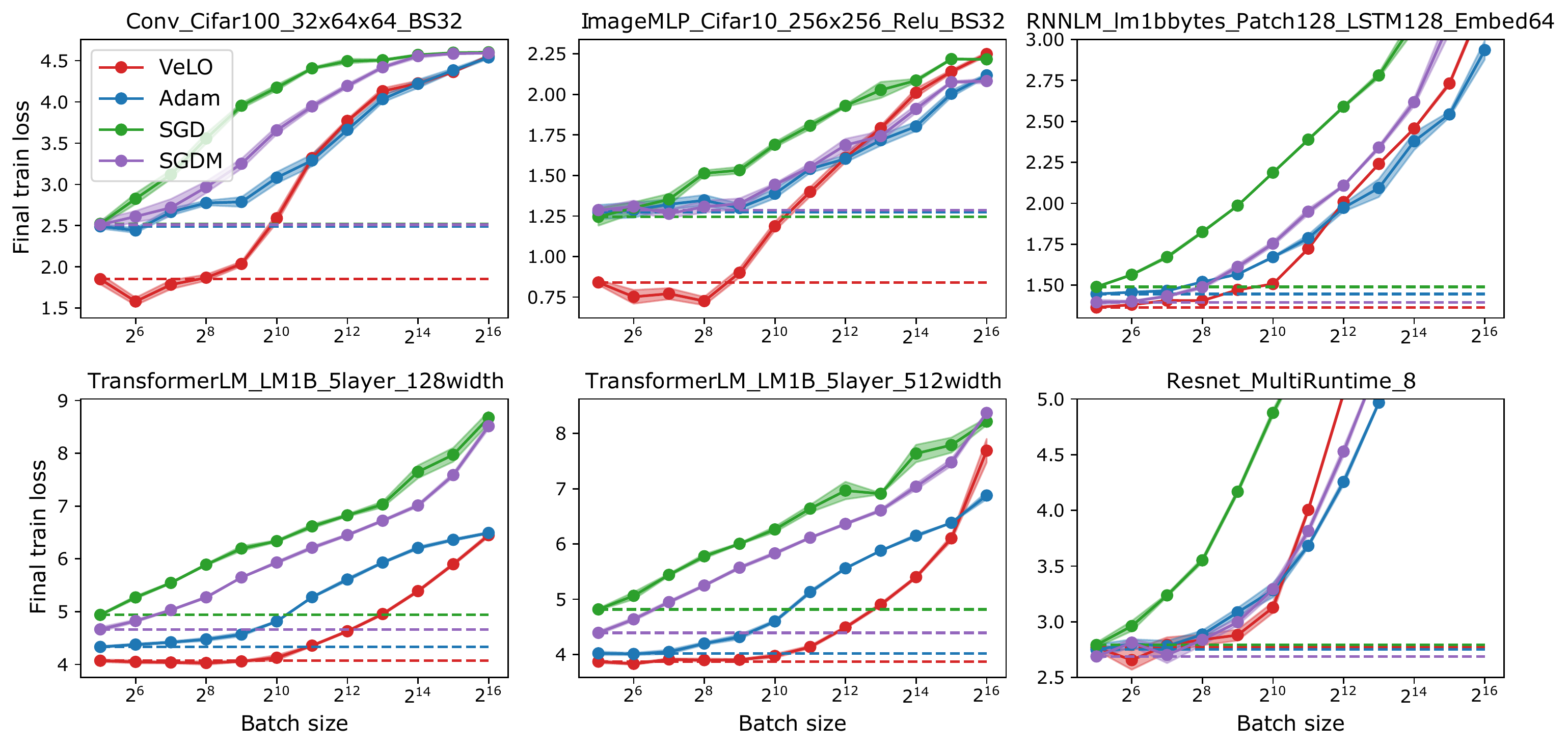}
    \end{overpic}
    }
    \caption{\textbf{Performance after training for a fixed number of examples with varying batch size.} We test 6 different tasks. Our learned optimizers perform well on small to medium batch sizes, but underperform on larger batch sizes.
    \label{fig:full_bs_scaling}
    }
\end{figure}

In this section we provide additional batch size scaling figures to those already shown in Section~\ref{sec:bs}.
Here we test 4 additional models: a ConvNet, an MLP image classifier, an RNN language model, the 2 Transformers already shown, and a small ResNet (Figure~\ref{fig:full_bs_scaling}).
For each model we train for a fixed number of examples while varying the batch size.
We find our learned optimizer always outperforms, or matches the baselines at small to medium batch sizes. In the extremely large batch size setting, however we find learning rate-tuned Adam often performs best, but again we speculate that this is a consequence of distributional shift between meta-training and evaluation.

\subsection{300M Language Model Trained on 64 Accelerators}
We apply our learned optimizers to training decoder only language model.
We leverage a pre-existing 300M parameter, decoder only language model designed after Primer~\citep{so2021searching} trained on a data mixture similar to that used in PALM~\citep{chowdhery2022palm} implemented in PAX\footnote{\url{https://github.com/google/paxml/}}.
It consists of 16 layers, 8 heads per layer, has a model dimension of 768, with a 4x width expansion for the MLP. We train with Rotary embeddings~\citep{su2021roformer}, and with a 256K vocab size.
This model was trained with a 4x4x4 TPUv4 slice with a batch size of 1,024 sequences of length 1,280, or approximately 1.3 million tokens with 16-way data parallel and 4-way model parallel training.
For our baseline optimizer, we use AdaFactor\footnote{We use AdaFactor without factorization, the same optimizer used in PALM~\citep{chowdhery2022palm}.} with a learning rate warm up and decay, and with learning rate tuned across half powers of 10.
Due to implementation details\footnote{PAX, leverages "scan over layers" trick to reduce compile times of large models. This adds a restriction that optimizers be independent per layer.}, for these experiments our learned optimizer optimizes each layer independently, and ignores all communication across layers.
Despite this, our optimizer performs comparably to the best trial of the learning rate-tuned baseline. 

\begin{figure}[t!]
    \centering
    \makebox[\textwidth]{%
    \begin{overpic}[width=0.8\textwidth]{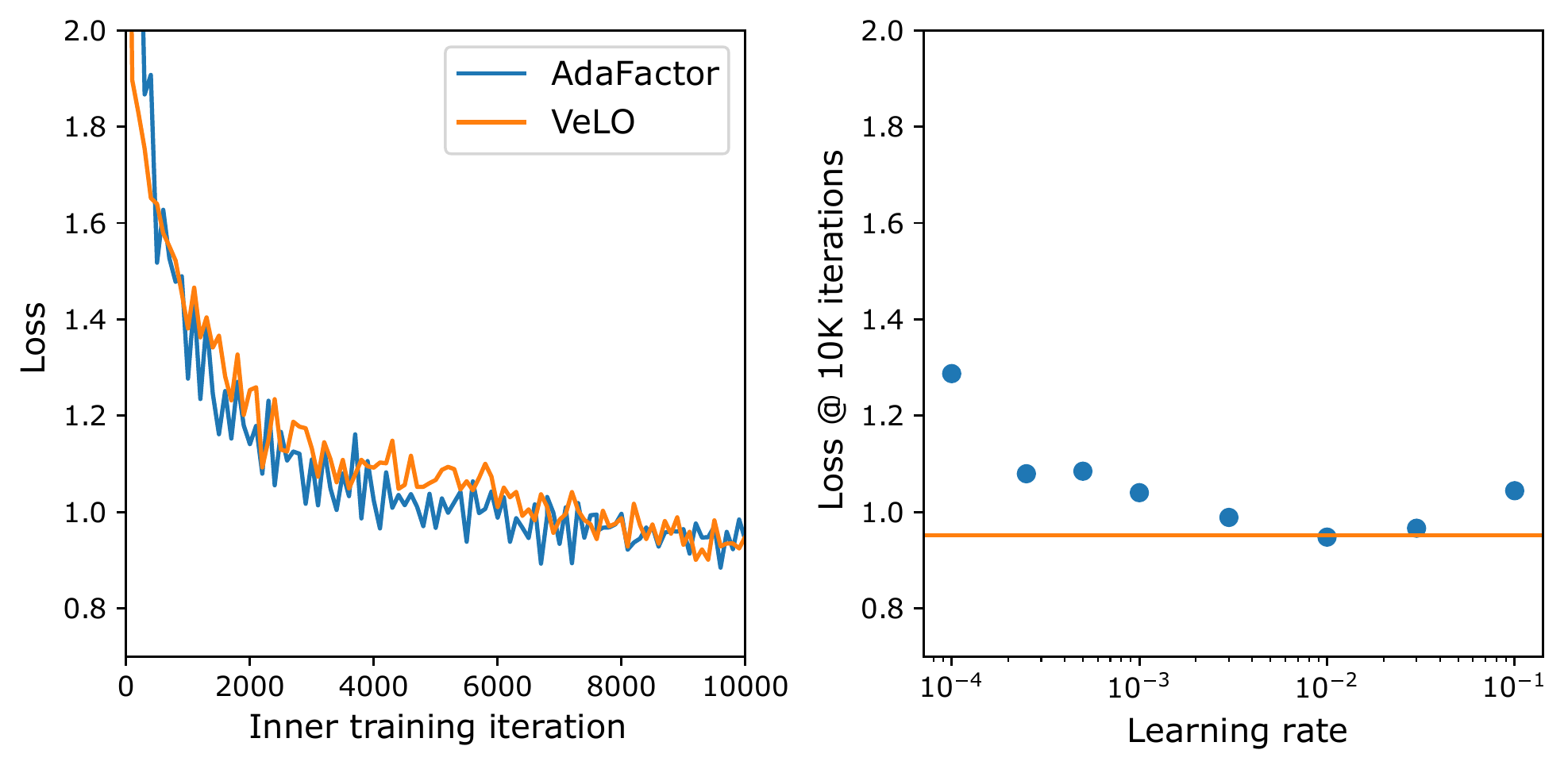}
    \end{overpic}
    }
    \caption{\textbf{300M parameter, large batch transformer.} Our learned optimizer can be used to train a 300M parameter Transformer on par with learning rate-tuned AdaFactor.
    \label{fig:pax_300m}
    }
\end{figure}

\subsection{Finetuning of the Learned Optimizer on a Single Problem} \label{app:per_task_finetune}
\begin{figure}[t!]
    \centering
    \makebox[\textwidth]{%
    \begin{overpic}[width=1.0\textwidth]{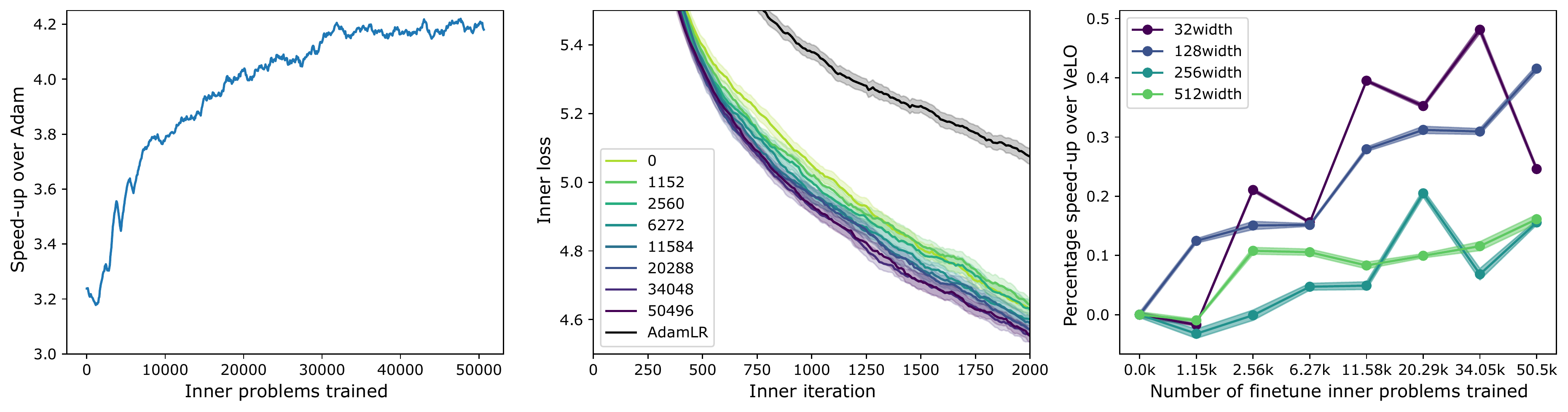}
    \put (2,27.0) {\textbf{\small(a)}}
    \put (37,27.0) {\textbf{\small(b)}}
    \put (70,27.0) {\textbf{\small(c)}}
    \end{overpic}
    }
    \caption{\textbf{Finetuning a pretrained learned optimizer on a single problem.} We explore finetuning as a way to specialize a pretrained learned optimizer to a particular task. \textbf{(a)}: We show meta-training curves measured in speedup over Adam. \textbf{(b)}: We show inner-training curves from different points in meta-training. \textbf{(c)}: We show the percentage speedup for 4 different sized models. For (a) and (b), confidence bounds indicate standard error across 5 random seeds.
    \label{fig:finetune_trans}
    }
\end{figure}

In some applications, it is desirable to repeatedly train the same model with minor variations---for example, in continual or online learning in which data is continuously varying.
With standard optimizers, training efficiency may be improved by finding good hyperparameters; in our case, we can directly finetune VeLO to a particular task. 
We finetuned VeLO on a 5 layer, 128 hidden size Transformer (\texttt{TransformerLM\_LM1B\_5layer\_128width}) training for 2K inner-iterations.
We performed this finetuning using ES with antithetic samples on 16 accelerators for $\sim$5 days.
We used Adam as our outer-optimizer, and tuned learning rates with half power of 10 resolution and picked the best value. 

In Figure~\ref{fig:finetune_trans}a, we show the meta-training curves measured in units of how much much time it would take Adam to reach this loss value (see Section~\ref{fig:normalize_vis} for more details on this normalization).
Our optimizers start out 3.2 times faster than Adam, and after finetuning reach 4.2 times faster.
In Figure~\ref{fig:finetune_trans}b, we show the learning curves for different points in outer-training.
Finally, in Figure~\ref{fig:finetune_trans}c we test if this finetuned optimizer will transfer to similar problems.
While we do see more variability, our finetuning generally improves the performance of a model which has a 4x larger hidden size by about 10\%. This suggests one might be able to finetune on a similar, but smaller task which would be considerably cheaper.

\subsection{Extended Evaluation on Optimizer Benchmark} \label{app:extended_opt_benchmark}
\begin{figure}[t!]
    \centering
    \makebox[\textwidth]{%
    \begin{overpic}[width=0.8\textwidth]{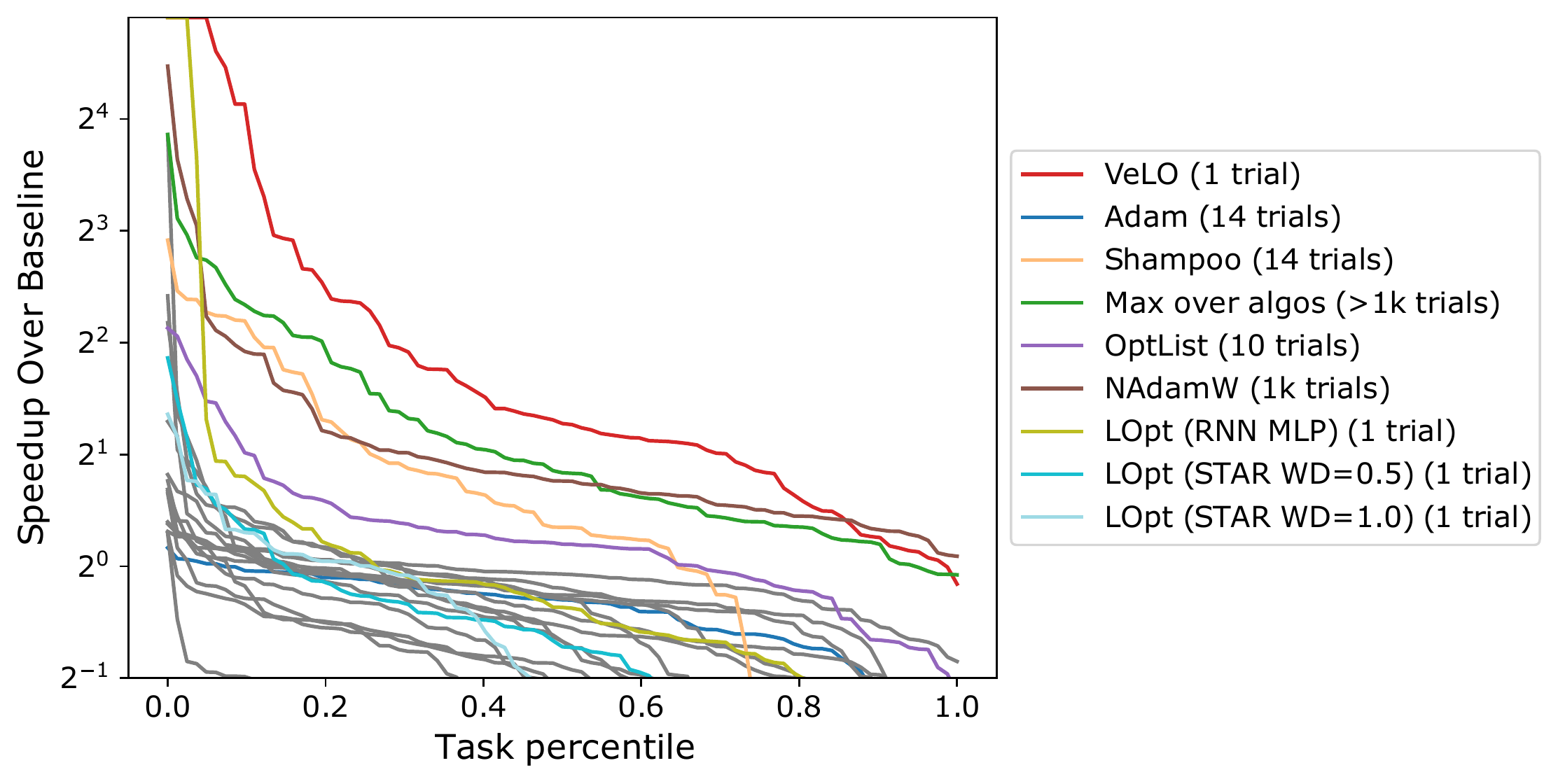}
    \end{overpic}
     }
    \caption{\textbf{Optimizer performance on 83 tasks with improved baseline.} The $y$-axis shows the relative number of steps it takes the best of learning rate-tuned Adam, RAdam, and Adam with 2 kinds of learning rate decay to achieve the same loss each optimizer reaches after 10K training steps.
    The $x$-axis shows the fraction of tasks for which the optimizer achieves at least that large a speedup over the baselines.
    \label{fig:aggregate_multitask_better_norm}
    }
\end{figure}

\paragraph{83 tasks, different normalizer.}
To measure impact of the normalizer used in the aggregate task comparisons described in Section~\ref{sec:83}, we replot this result using a different set of baselines (Figure~\ref{fig:aggregate_multitask_better_norm}).
As described in Sec~\ref{app:fixed_task_normalizer}, we normalize performance of our optimizer with respect to learning rate-tuned Adam, Adam with inverse square root learning rate decay, Adam exponential learning rate decay decay, and RAdam. 
As expected, all values shift down as the speedup shrinks due to this better baseline. Nevertheless, we find VeLO is still superior over even extensively tuned methods. 

\begin{figure}[t!]
    \centering
    \makebox[\textwidth]{%
    \begin{overpic}[width=0.8\textwidth]{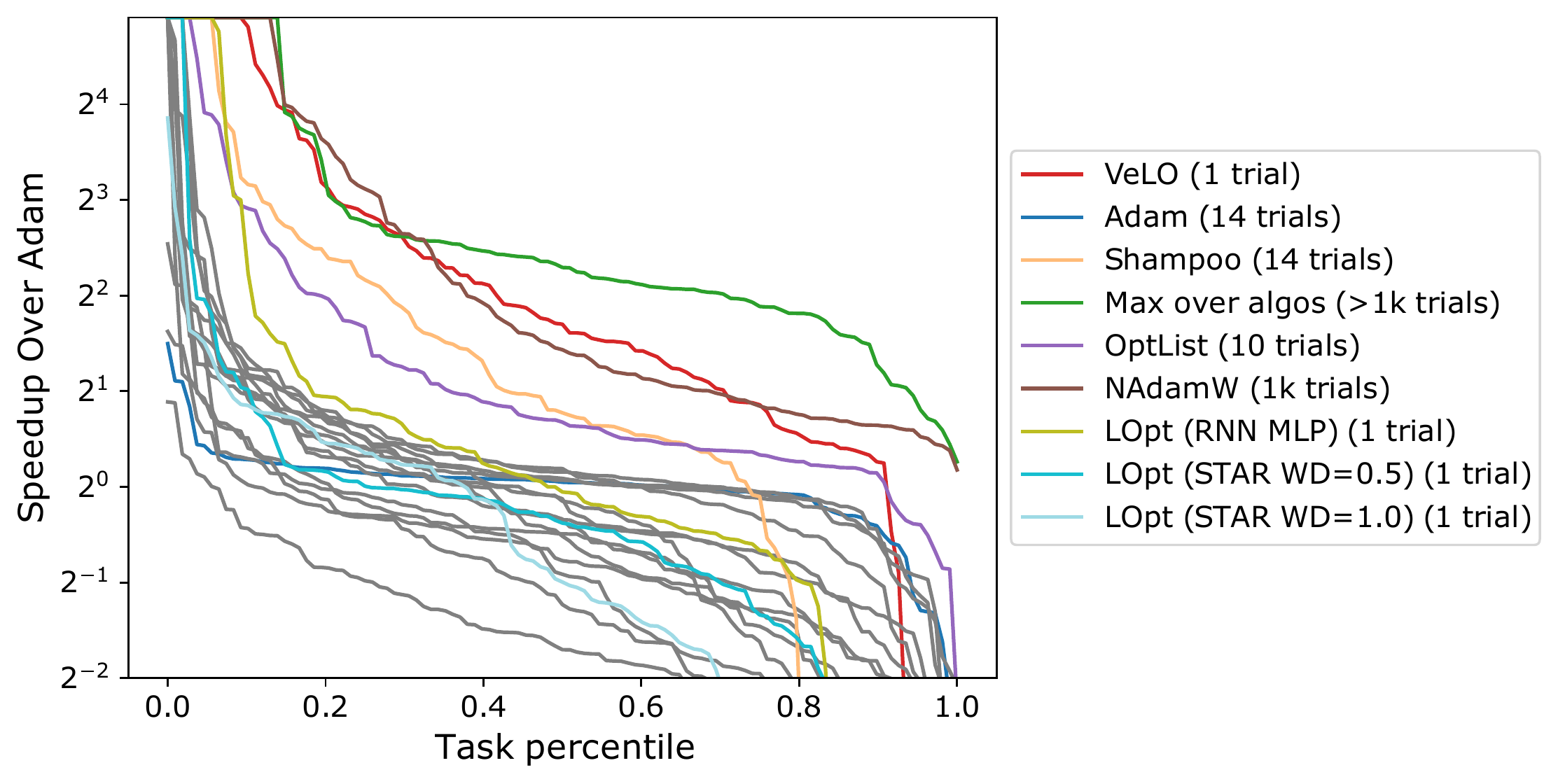}
    \end{overpic}
     }
    \caption{\textbf{Optimizer performance on 109 tasks. Adam normalized.} The $y$-axis shows the relative number of steps it takes the best of learning rate-tuned Adam to achieve the same loss each optimizer reaches after 10K training steps.
    The $x$-axis shows the fraction of tasks for which the optimizer achieves at least that large a speedup over the baselines.
    \label{fig:aggregate_100_adam}
    }
\end{figure}

\begin{figure}[t!]
    \centering
    \makebox[\textwidth]{%
    \begin{overpic}[width=0.8\textwidth]{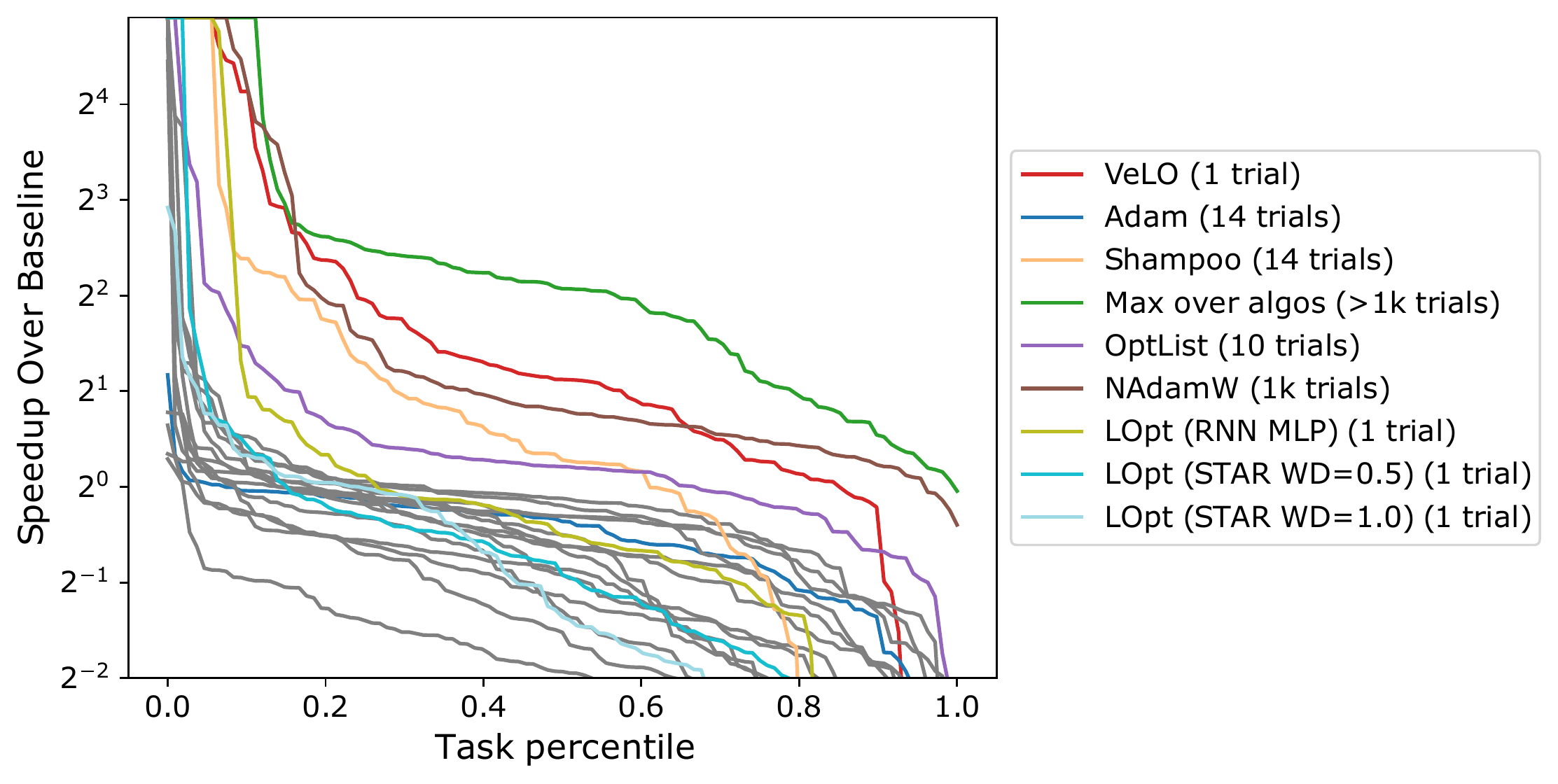}
    \end{overpic}
     }
    \caption{\textbf{Optimizer performance on 109 tasks. Adam with LR schedule + RAdam normalized.} The $y$-axis shows the relative number of steps it takes the best of learning rate-tuned Adam, RAdam, and Adam with 2 learning rate decays to achieve the same loss each optimizer reaches after 10K training steps.
    The $x$-axis shows the fraction of tasks for which the optimizer achieves at least that large a speedup over the baselines.
    \label{fig:aggregate_100_baseline}
    }
\end{figure}

\paragraph{An extended task set of 109 tasks.}
Next, we shift our attention to looking at \emph{all} 109 tasks in our evaluation set.
This includes the original 83 tasks presented before in addition to the MLP-Mixer, NERF, Evolution strategies, and learned optimizer training tasks. We plot both speedup over Adam, and speedup over the improved baseline in Figures~\ref{fig:aggregate_100_adam} and \ref{fig:aggregate_100_baseline} respectively. While our learned optimizer performance remains strong, we find there exist some tasks for which we do not perform well---in particular, VeLO struggles on ES and learned optimizer training tasks.

\paragraph{Tables of aggregated performance.}
An alternative view of performance metrics for baselines is a tabular view of aggregated performance measures. 
For a given kind of normalizer (either with Adam, or the extended set of baselines, see Section~\ref{app:fixed_task_normalizer}) we show the mean speedup over this normalizer ($\sum^N_i \text{norm}(x_i)/N$, where $x_i$ is the performance for a particular task, and $N$ is the total number of tasks), and the mean of the reciprocals ($(1/N)\sum^N_i 1/\text{norm}(x_i)$).
We see this value analogously to the ``cost'' to train these models. If we are 2x faster than the baseline, it would cost 0.5 the compute.
Averaging in this space puts more weight on poor performing tasks.
To prevent infinities/extreme sensitivity to outliers both of these means are clipped per task. In the speedup variant (in which we average per-task ``speedups''), we clip to at most 30x faster than Adam.
In the cost variant, we clip the max cost to be 10.
Next we include a variety of percentile based measurements reporting the speedup for a given percentile of tasks. We include 5\%, 10\% 25\%, 50\%, and 75\%; note that the 50\% quantile has better outlier robustness than the mean. 

We include 4 tables in total containing this data with different normalizers, and different number of hyperparameter trials considered.
For normalizers, in Table~\ref{table:single_adam} and Table~\ref{table:multi_adam} we use LR-tuned Adam to normalize our runs, and in Table~\ref{table:single_baseline} and Table~\ref{table:multi_baseline} we use the extended set of baseline optimizers for normalization purposes (LR-tuned Adam, RAdam, Adam with reciprocal sqrt LR decay, and Adam with exponential LR decay).

In Table~\ref{table:single_adam} and Table~\ref{table:single_baseline} we only compare to baselines which use a single trial.
For LR-tuned baselines, this means a single learning rate is used per row. We do not include all 15 learning rates for clutter and instead opt to show 2 values---one which performs best in the mean speedup, and a second which performs best on the median performance. If these are the same learning rate, we also include the second best learning rate as measured by the median. For the OptList baseline, we simply use the first hyperparameter in the list of hyperparameters. For NAdamW we do a similar procedure to that of learning rate-tuned optimizers to select 2 different hyperparameter configurations.

In Table~\ref{table:multi_adam} and Table~\ref{table:multi_baseline} we show baselines consisting of taking the best performance from multiple optimizer trials.
This is analogous to hyperparameter tuning.
We show the best performance of learning rate-tuned Adam, the first 5 and 10 values from OptList, and all 1K trials from NAdamW.
Next, we show even more aggregation including the best of all shampoo models, all non-shampoo learning rate-tuned models, all learning rate-tuned optimizers, and finally a massive aggregation over all baselines we tried.
Next, we include ensembles of learned optimizers.
These are learned optimizers trained throughout this work including different meta-training configurations and different checkpoints throughout meta-training.
See \href{https://github.com/google/learned_optimization/blob/b1d8267c5e513a4112e7422b98bacc16e1f0e844/learned_optimization/research/general_lopt/pretrained_optimizers.py#L32}{here} for the full list we evaluate with. 
And finally, to get a sense of the best possible performance, we include the best performance over all optimizers we evaluate.

There are several takeaways from this view of the data: 
\begin{enumerate}
    \item On average, VeLO is more than 7x faster than learning rate-tuned Adam. The best of an ensemble of learned optimizers is more than 11x faster than learning rate-tuned Adam.
    \item Our single trial learned optimizer performs well on average. It outperforms every other single model by a large margin in every category except for fifth percentile performance, suggesting that there are a handful of problems on which VeLO doesn't do well.
    \item Our single trial learned optimizer does well compared to multiple trial baselines though the gap between the baselines shrinks. On average, we do not yet outperform the extensive hyperparameter tuning with thousands of trials, although we emphasize that this is an extremely high bar. 
    \item Ensembles of learned optimizers work extremely well, suggesting that different learned optimizers are better at different tasks. We interpret this result as a demonstration that further gains with learned optimizers are possible.
    \item Shampoo, in general, performs quite well overall compared to all other Adam-like optimizers.
    \item Levering both learned optimizers and hand-designed optimizers (Best Everything) results in an additional improvement bump suggesting that further improvement can be made.
\end{enumerate}

\begin{table}[h!]
    \centering
    \caption{\textbf{Single trial optimizers. Normalizer computed with LR-tuned Adam.}}
    \input{app_table_adamnorm_single_trial.tex}
    \label{table:single_adam}
\end{table}

\begin{table}[h!]
    \centering
    \caption{\textbf{Multi trial optimizers. Normalizer computed with LR-tuned Adam.}}
    \input{app_table_adamnorm_multi_trial.tex}
    \label{table:multi_adam}
\end{table}

\begin{table}[h!]
    \centering
    \caption{\textbf{Single trial optimizers. Normalizer computed with extended data.}}
    \input{app_table_baselinenorm_single_trial.tex}
    \label{table:single_baseline}
\end{table}

\begin{table}[h!]
    \centering
    \caption{\textbf{Multi trial optimizers. Normalizer computed with extended data.}}
    \input{app_table_baselinenorm_multi_trial.tex}
    \label{table:multi_baseline}
\end{table}

\begin{figure}[h!]
    \centering
    \input{app_page1}
    \makebox[\textwidth]{%
    \begin{overpic}[width=1.16\textwidth]{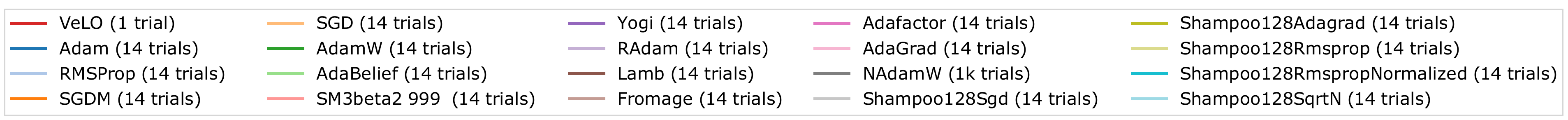}\end{overpic}
    }
    \caption{\textbf{Page 1 of all evaluation tasks.} \label{fig:learning_curves_1}
    }
\end{figure}

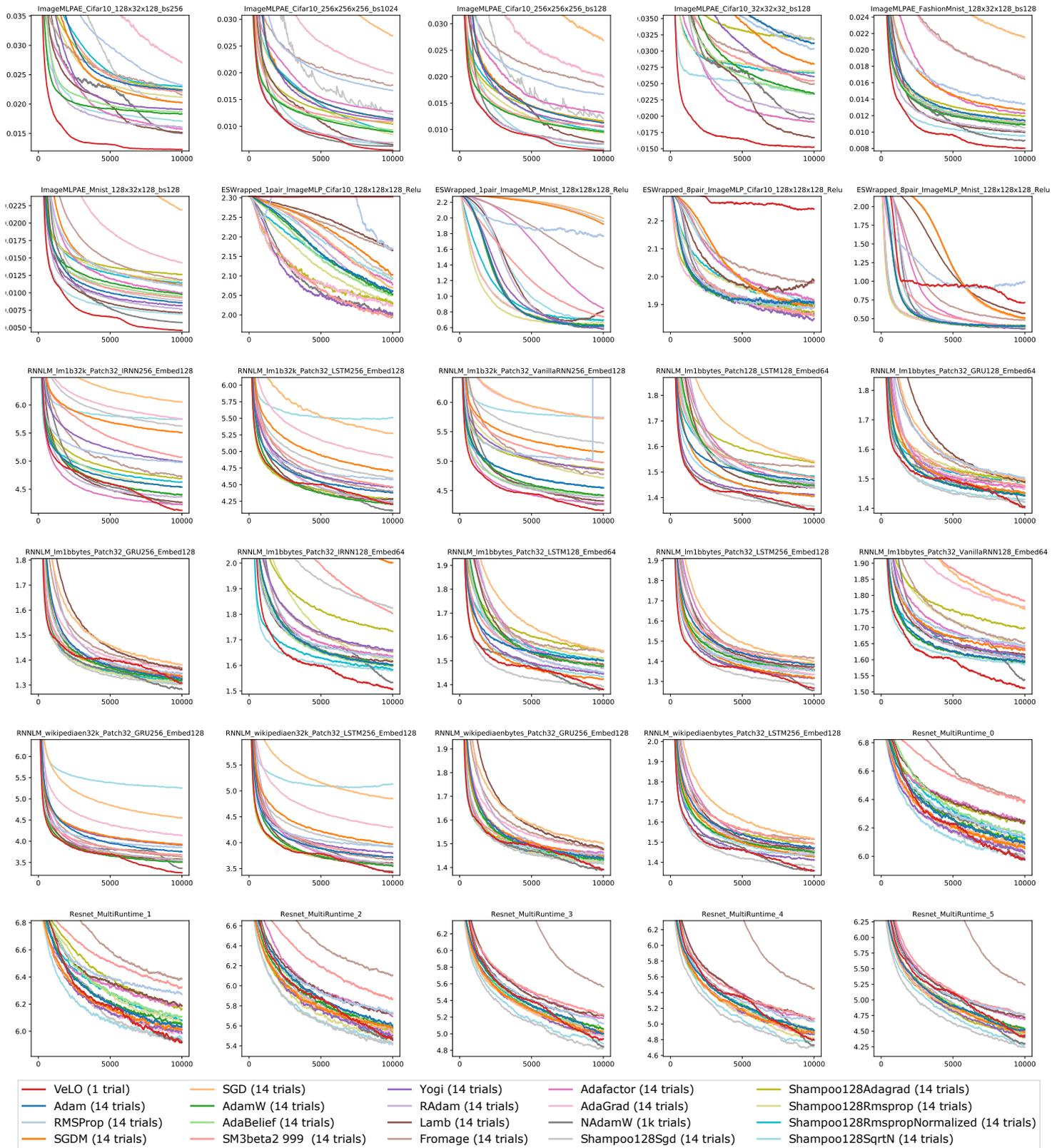
\begin{figure}[h!]
    \centering
    \input{app_page2.tex}
    \makebox[\textwidth]{%
    \begin{overpic}[width=1.16\textwidth]{figs/legend_bot_all.pdf}\end{overpic}
    }
    \caption{\textbf{Page 2 of all evaluation tasks.} \label{fig:learning_curves_2}
    }
\end{figure}

\begin{figure}[h!]
    \centering
    \input{app_page4.tex}
    \makebox[\textwidth]{%
    \begin{overpic}[width=1.16\textwidth]{figs/legend_bot_all.pdf}\end{overpic}
    }
    \caption{\textbf{Page 4 of all evaluation tasks.} \label{fig:learning_curves_4}
    }
\end{figure}
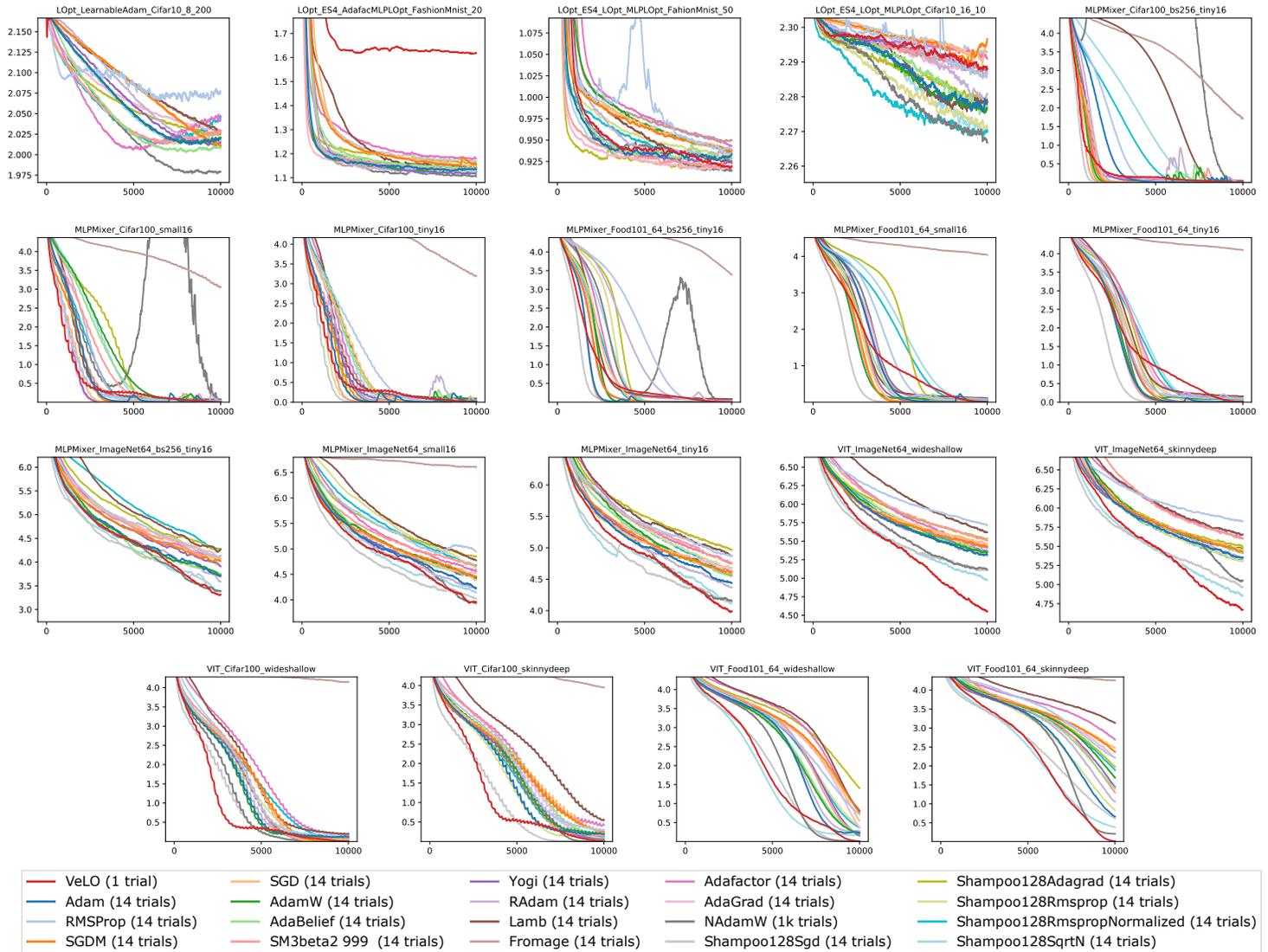

\paragraph{Learning curves for all tasks.}
Finally, we include learning curves for all tasks and many baseline optimizers (Figures \ref{fig:learning_curves_1}-\ref{fig:learning_curves_4}). We do not include SM3 or Shampoo with AdagradNormalized grafting due to difficulty finding a legible color scheme with more than 20 values.
All models are averaged over 5 random seeds with only the best performing learning rate/hyperparameter setting per model.
In some cases---e.g. MLP-Mixer and ViT tasks on smaller datasets---many optimizers reach near zero training loss. 
In these settings, the learned optimizer appears to be slow to train.
This is a side effect of targeting last loss -- our optimizer has no incentive to reach low performance early in training and thus waits until the very end to do so.

\begin{figure}[h!]
    \centering
    \input{app_page3.tex}
    \makebox[\textwidth]{%
    \begin{overpic}[width=1.16\textwidth]{figs/legend_bot_all.pdf}\end{overpic}
    }
    \caption{\textbf{Page 3 of all evaluation tasks.} \label{fig:learning_curves_3}
    }
\end{figure}
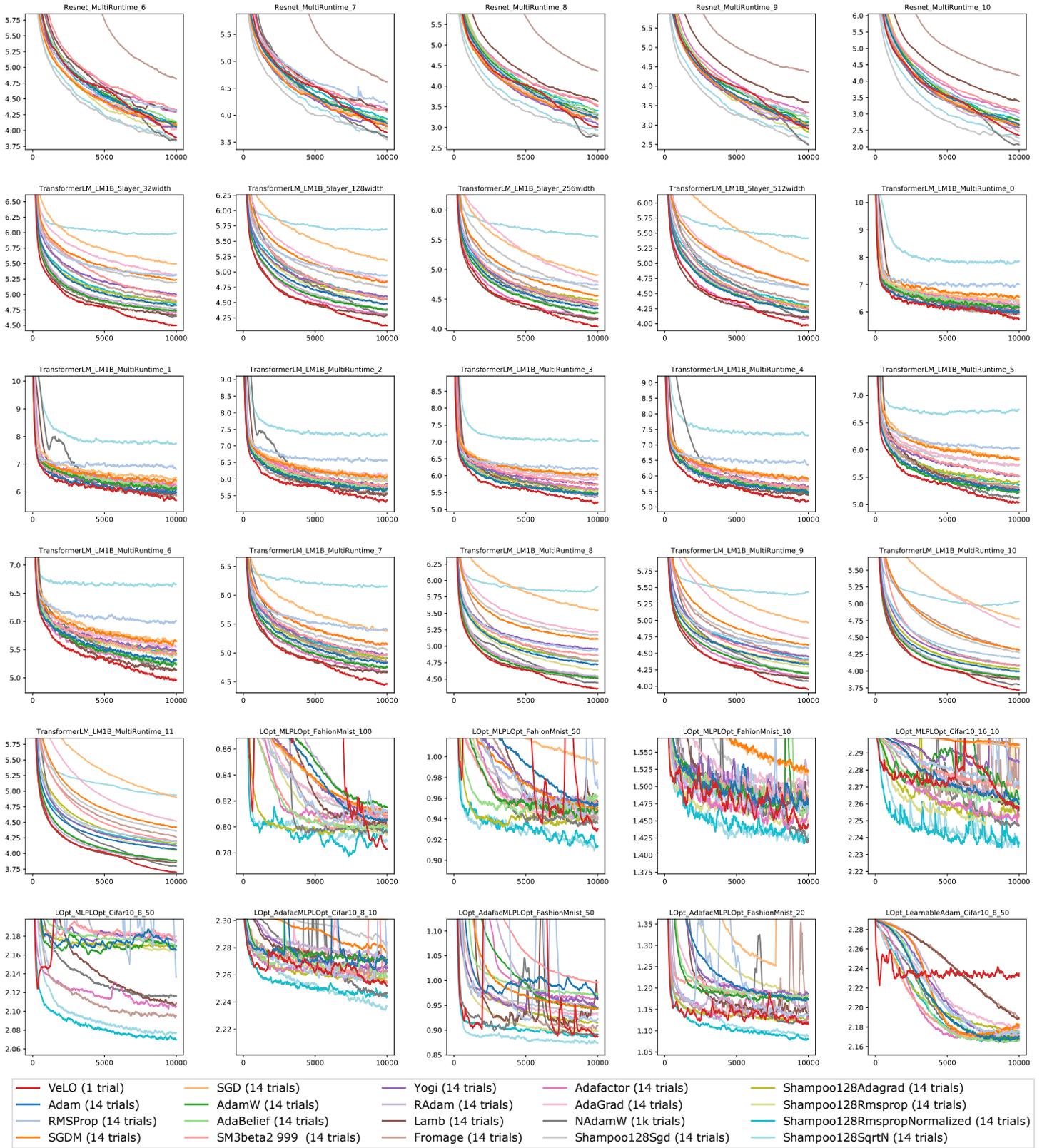

\paragraph{Limitations.}
Optimizer comparison is fraught with difficulties. This work is no exception.

First, in our comparisons, we are averaging over a \emph{specific} distribution of tasks. This distribution of tasks could be changed to make any optimizer appear to be the strongest\footnote{We note that this issue of fair baselining is not unique to learned optimization.}. Our goal here is to present as fair a measure of performance as possible. This distribution of 109 tasks was slowly built as a tool for our team to measure the performance of our optimizers. Tasks were never deleted, only added, although we emphasize the paper body contains a reduced view of these tasks which excludes ES and learned optimizer training tasks. While far from perfect, we have found performance on this evaluation set predictive of the performance we have seen on larger models.

Second, in a number of figures we perform minimization or maximization to find the best performing optimizer configurations from a particular set. Because each one of these losses is itself stochastic, this aggregation will potentially result in better performance than possible. We have mitigated this to some extent by computing final losses with many batches, as well as always reporting numbers averaged over 5 random seeds for \emph{every} hyperparameter configuration.

\section{Learned Optimizers Training Other Learned Optimizers}
In order to accelerate our research on learned optimization, we have also been attempting to leverage our pretrained learned optimizers to train other learned optimizers.
The field of compilers has been utilizing this self-improvement for some time now with ``self-hosting'' compilers \citep{hart1962ai}, but at this point we are unaware of any successful attempts in machine learning.
In learned optimizers this has been attempted in ~\citep{metz2020tasks, metz2021training} though the resulting optimizers are not reliable enough to exclusively be used for meta-training.
We test our optimizer on a variety of different learned optimizer setups ranging from small scale distributions all the way up to training another version of the optimizer used in this work, and this is discussed in the remainder of this section. 
While we are able to make use of learned optimizers to train learned optimizers to a limited degree---and even outperform the baseline methods we use in some cases---our results show that the goal of self-improving optimizers is currently out of reach.
The task to train an optimizer is too far out of distribution and the resulting learned optimizers have the tendency to go unstable.
We include the results in this section to motivate future work on this problem. 

\subsection{Learned Optimizers Which Train a Single Problem} \label{app:learned_optimizer:single_problem}
\begin{figure}[t!]
    \centering
    \makebox[\textwidth]{%
    \begin{overpic}[width=0.23\textwidth]{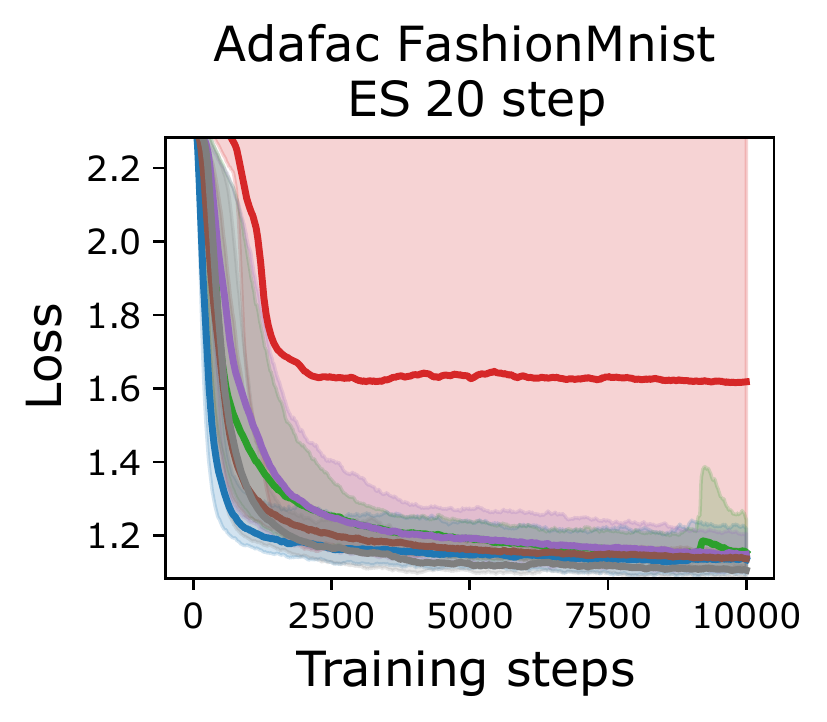}
    \put (2,84.0) {\textbf{\small(a)}}
    \end{overpic}
    \begin{overpic}[width=0.23\textwidth]{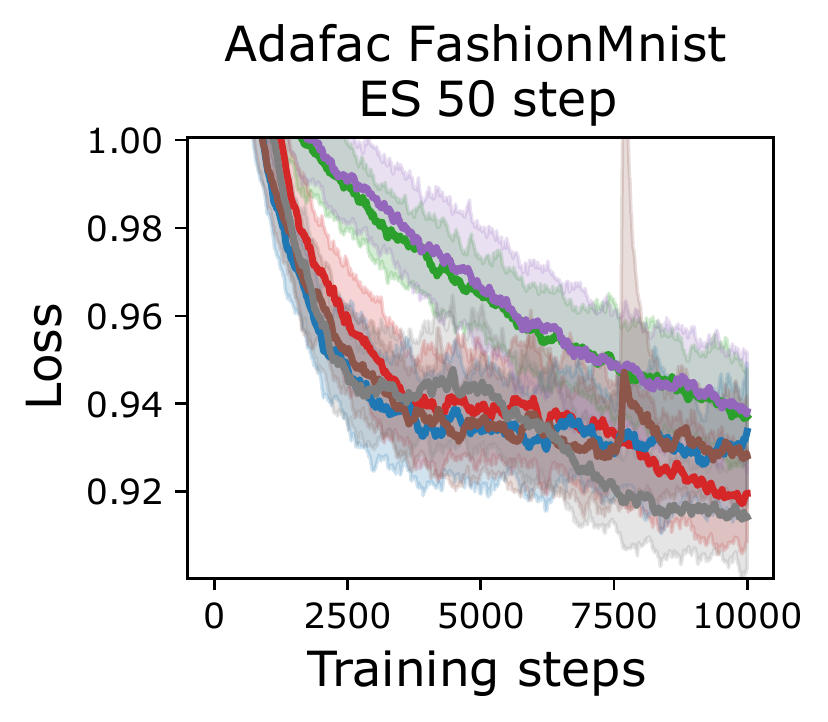}
    \put (2,84.0) {\textbf{\small(b)}}
    \end{overpic}
    \begin{overpic}[width=0.23\textwidth]{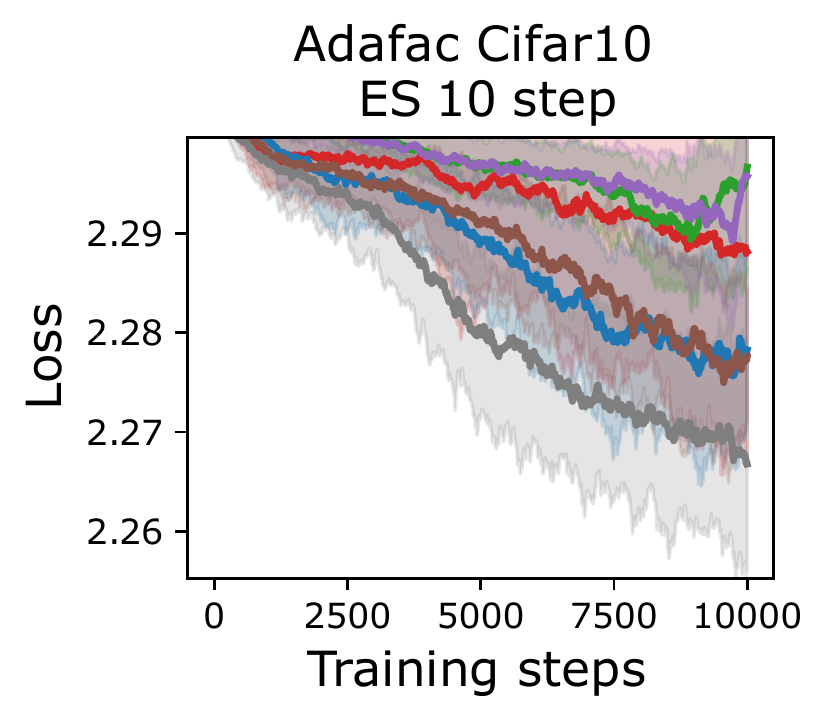}
    \put (2,84.0) {\textbf{\small(c)}}
    \end{overpic}
    }
    \begin{overpic}[width=0.5\textwidth]{figs/fig_pc/legend_bot.pdf}
    \end{overpic}
    \caption{\textbf{Learned optimizers training other learned optimizers with ES.} We find the AdaFactor LOpt trained on FashionMnist for both 20 ((a)) and 50 ((b)) iterations optimize better than learning rate-tuned Adam -- the technique used to train these optimizers. On CIFAR10, we find our learned optimizer diverges.
    \label{fig:selfopt_single_es}
    }
\end{figure}

We test the ability of our pretrained learned optimizers to train other learned optimizers which train a single task (Figure~\ref{fig:selfopt_single_es}).
We use the learned optimizers from \citet{metz2019understanding} (called MLPOpt) and \citet{metz2022practical} (called AdaFactor, due to AdaFactor-style input features) and train optimizers to train MLPs on CIFAR10 or FashionMNIST.
First, we test meta-training optimizers with gradients estimated via ES.
This problem corresponds to the following tasks: \href{https://github.com/google/learned_optimization/blob/a42b207b9ef5395fdf7e64978c26379ccba4e264/learned_optimization/tasks/fixed/lopt.py#L225}{LOpt\_ES4\_AdafacMLPLOpt\_FashionMnist\_20}, 
\href{https://github.com/google/learned_optimization/blob/a42b207b9ef5395fdf7e64978c26379ccba4e264/learned_optimization/tasks/fixed/lopt.py#L230}{LOpt\_ES4\_AdafacMLPLOpt\_FashionMnist\_50},
and \href{https://github.com/google/learned_optimization/blob/a42b207b9ef5395fdf7e64978c26379ccba4e264/learned_optimization/tasks/fixed/lopt.py#L235}{LOpt\_ES4\_LOpt\_MLPLOpt\_Cifar10\_16\_10}.
In 2 of the 3 tested architectures our learned optimizers match or outperform Adam (what has been used to train these learned optimizers in past work), but do not outperform an extensive hyperparameter tuning (NAdamW).
In the final task, our learned optimizer diverges.

\begin{figure}[t!]
    \centering
    \makebox[\textwidth]{%
    \begin{overpic}[width=0.23\textwidth]{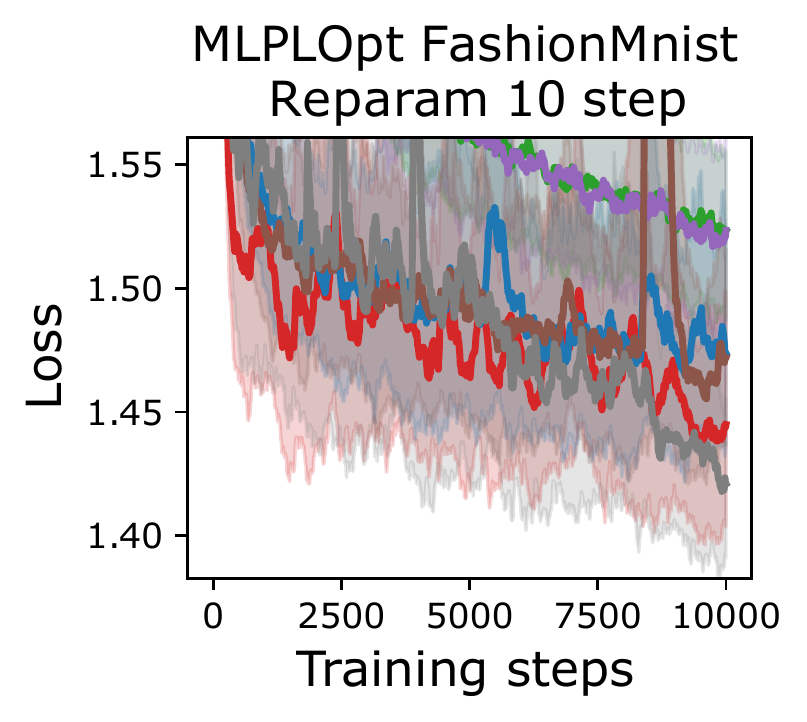}
    \put (2,84.0) {\textbf{\small(a)}}
    \end{overpic}
    \begin{overpic}[width=0.23\textwidth]{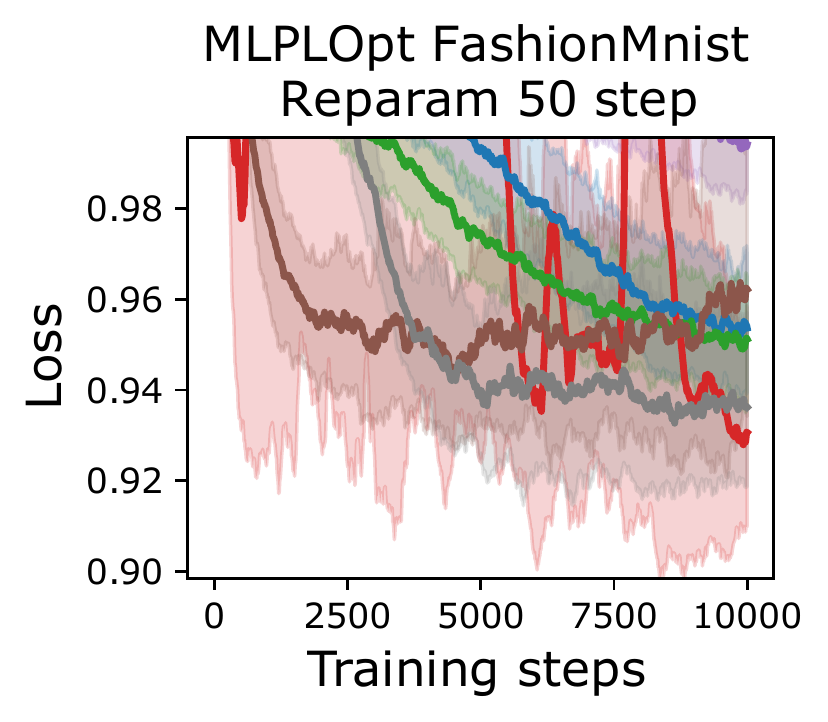}
    \put (2,84.0) {\textbf{\small(b)}}
    \end{overpic}
    \begin{overpic}[width=0.23\textwidth]{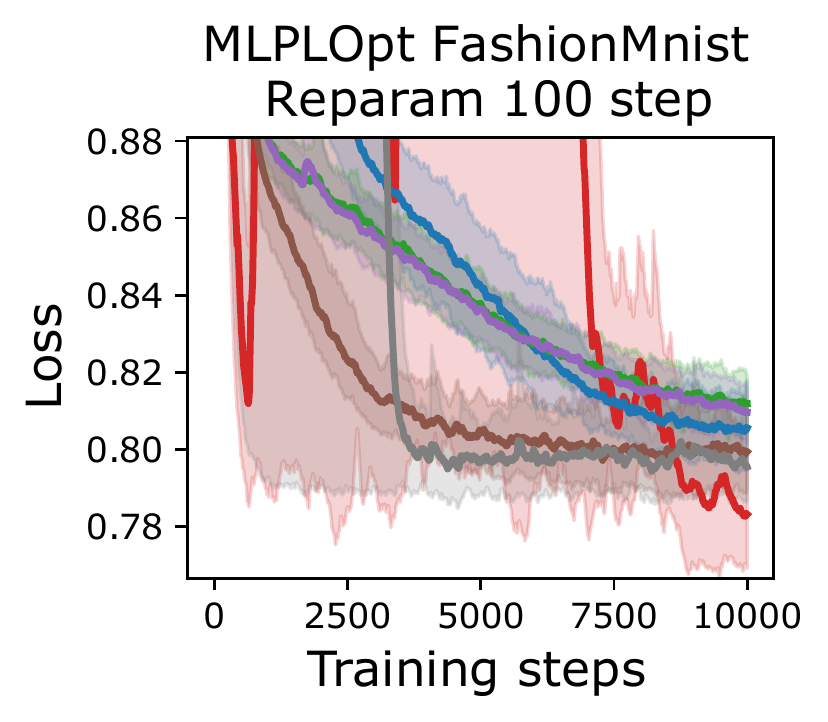}
    \put (2,84.0) {\textbf{\small(c)}}
    \end{overpic}
    \begin{overpic}[width=0.23\textwidth]{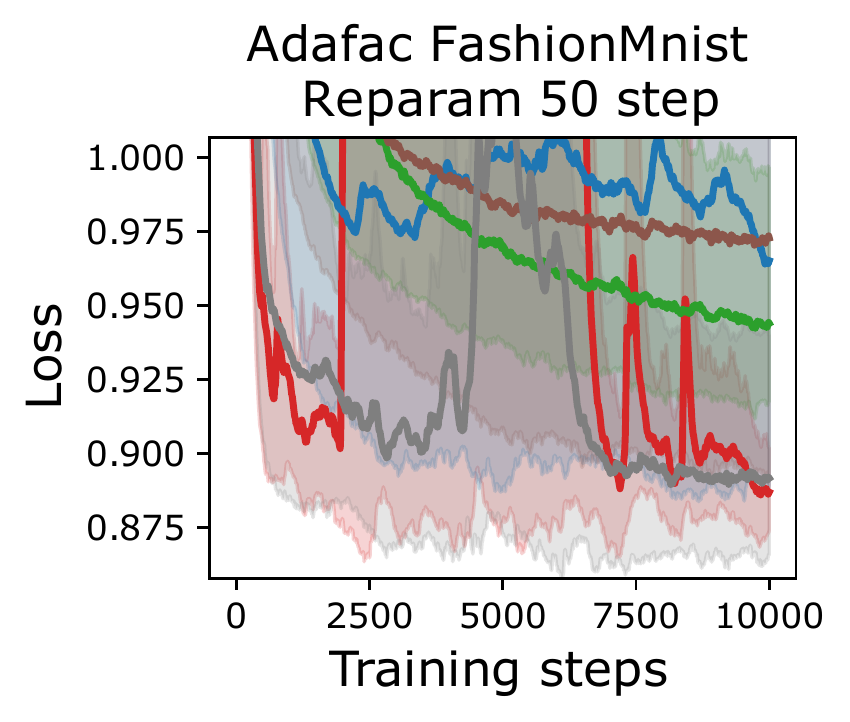}
    \put (2,84.0) {\textbf{\small(d)}}
    \end{overpic}
    \begin{overpic}[width=0.23\textwidth]{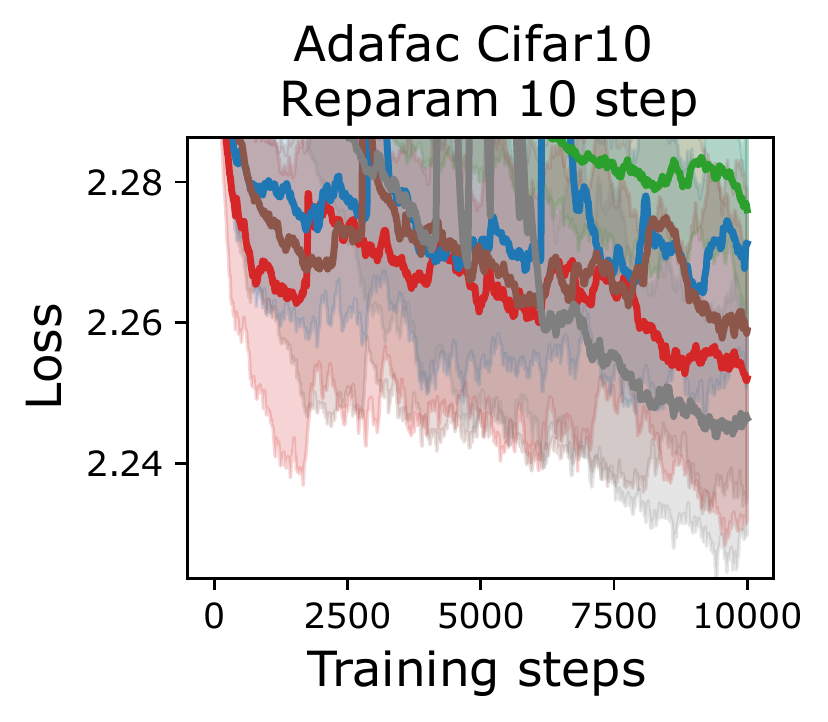}
    \put (2,84.0) {\textbf{\small(e)}}
    \end{overpic}
    }
    \begin{overpic}[width=0.5\textwidth]{figs/fig_pc/legend_bot.pdf}
    \end{overpic}
    \caption{\textbf{Learned optimizers training other learned optimizers with gradients estimated with backprop.} On all of the 5 tasks shown we match or outperform learning rate-tuned Adam -- what is commonly used to train these optimizes. See section~\ref{app:learned_optimizer:single_problem} for details.
    \label{fig:selfopt_single_backprop}
    }
\end{figure}

Next, we test our optimizers on gradients computed with backpropagation.
Past work~\citep{metz2019understanding} has shown this is particularly unstable due to exploding and vanishing gradients making this a particularly challenging problem to apply our learned optimizers to.
Results are shown in Figure~\ref{fig:selfopt_single_backprop}; we find in all cases again we match or outperform Adam though training curves are considerably more unstable, despite showing the median over 5 random initializations per optimizer configuration.
The problems we test on are 10, 50, and 100 step training on FashionMNIST with MLPOpt (\href{https://github.com/google/learned_optimization/blob/a42b207b9ef5395fdf7e64978c26379ccba4e264/learned_optimization/tasks/fixed/lopt.py#L169}{LOpt\_MLPLOpt\_FashionMnist\_10}, \href{https://github.com/google/learned_optimization/blob/a42b207b9ef5395fdf7e64978c26379ccba4e264/learned_optimization/tasks/fixed/lopt.py#L230}{LOpt\_MLPLOpt\_FashionMnist\_50}, \href{https://github.com/google/learned_optimization/blob/a42b207b9ef5395fdf7e64978c26379ccba4e264/learned_optimization/tasks/fixed/lopt.py#L155}{LOpt\_MLPLOpt\_FashionMnist\_100}), 50 step FashionMNIST  with the AdaFactor MLPLOpt (\href{https://github.com/google/learned_optimization/blob/a42b207b9ef5395fdf7e64978c26379ccba4e264/learned_optimization/tasks/fixed/lopt.py#L197}{LOpt\_AdafacMLPLOpt\_FashionMnist\_50}), as well as 10 step AdaFactor MLPLopt on CIFAR10 (\href{https://github.com/google/learned_optimization/blob/a42b207b9ef5395fdf7e64978c26379ccba4e264/learned_optimization/tasks/fixed/lopt.py#L190}{LOpt\_AdafacMLPLOpt\_Cifar10\_8\_10}).

\subsection{Multi-Task, Small Variation Across Tasks}
Next we explore meta-training on a small mixture of MLP tasks specifically designed to be fast to meta-train.
Unlike the previous section, this optimization problem has more diversity in training problems, and is both a harder to optimize and more representative of the general-purpose learned optimizers trained in this work.
The task distribution consists of training single hidden layer MLPs with 32 hidden units trained on a mixture of CIFAR10, FashionMNIST, MNIST, and SVHN~\citep{netzer2011reading} each of which is resized to 8x8 and converted to black and white.

As done by this work, meta-gradients are computed with ES. Unlike the rest of this work, and because all the tasks have the same computational costs, we use synchronous training.

\begin{figure}[t!]
    \centering
    \makebox[\textwidth]{%
    \begin{overpic}[width=0.7\textwidth]{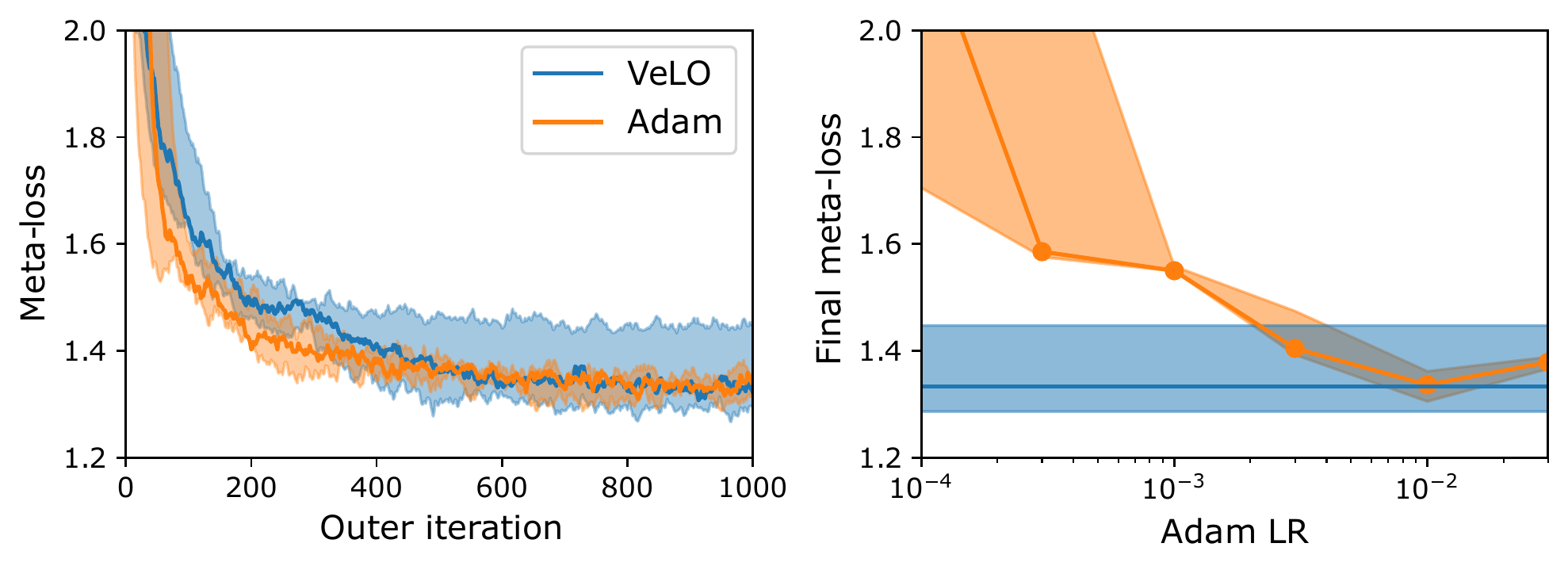}
    \put (8,36.0) {\textbf{\small(a)}}
    \put (56,36.0) {\textbf{\small(b)}}
    \end{overpic}
    }
    \caption{\textbf{Learned optimizers used to train other learned optimizers (AdaFac LOpt).} \textbf{(a)} We show learning curves of the best learning rate for learning rate-tuned Adam, and our learned optimizer. We find our learned optimizer matches performance without any tuning. \textbf{(b)} We show different learning rates as well as our learned optimizer performance. In both figures, the shaded region denotes min and max with the solid value showing median performance across 3 random seeds for LR-tuned Adam, and 5 random seeds for our learned optimizer.
    \label{fig:selfopt_adafac}
    }
\end{figure}

First, we meta-train with the learned optimizer from \citet{metz2022practical}---a small per-parameter MLP (Figure~\ref{fig:selfopt_adafac}).
We find the learned optimizer is able to match learning rate-tuned Adam (the optimizer used to train these in the original work).

\begin{figure}[t!]
    \centering
    \makebox[\textwidth]{%
    \begin{overpic}[width=1.0\textwidth]{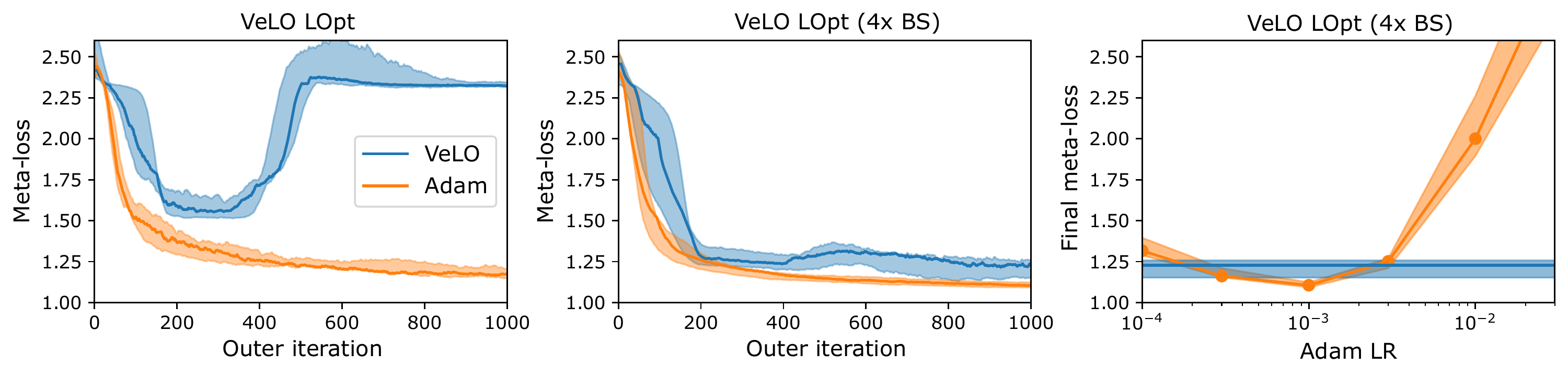}
    \put (4,23.0) {\textbf{\small(a)}}
    \put (36,23.0) {\textbf{\small(b)}}
    \put (72,23.0) {\textbf{\small(c)}}
    \end{overpic}
    }
    \caption{\textbf{Learned optimizers used to train other learned optimizers (Hyper LOpt).} \textbf{(a)} Meta training the learned optimizer with an outer batch size of 32. \textbf{(b)} Meta-training with an outer batch size of 128 improves the stability of the learned optimizer. \textbf{(c)} Outer-learning rate sensitivity of Adam. We find our learned optimizers are close to the best learning rate though do not outperform Adam. Across all plots we show median with min and max shown in the shaded region with 3 trials for Adam and 5 used for the learned optimizer.
    \label{fig:selfopt_hyperlopt}
    }
\end{figure}

Next, we meta-train the learned optimizer used in this work, the hierarchical hypernetwork-based model described in Section \ref{app:lopt_architecture} on this smaller task.  
When meta-training with a small outer-batch size, we find the learned optimizer goes unstable and diverges (Figure~\ref{fig:selfopt_hyperlopt}a).
We hypothesized this was due to the high noise in the gradients, which we tested by increasing the batch size by a factor of 4.
We found this significantly improved meta-training stability though the learned optimizers still lag behind well-tuned Adam.

\subsection{Large scale: Self-Hosting Optimizer, Replicating Work with a Learned Optimizer}
\begin{figure}[t!]
    \centering
    \makebox[\textwidth]{%
    \begin{overpic}[width=1.0\textwidth]{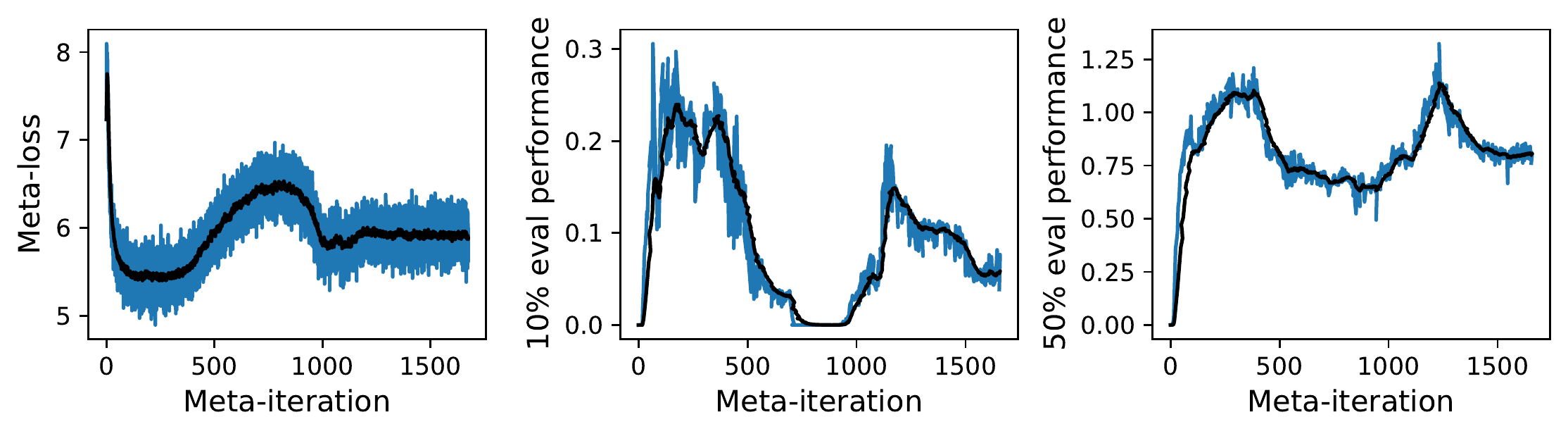}
    \put (27,23.0) {\textbf{\small(a)}}
    \put (60.5,23.0) {\textbf{\small(b)}}
    \put (95,23.0) {\textbf{\small(c)}}
    \end{overpic}
    }
    \caption{\textbf{Pre-trained learned optimizer re-training itself.} \textbf{(a)} We show the meta-loss, the loss fed into the LOpt for training. Early in meta-training our loss decreases nicely, but diverges after 500 inner iterations. \textbf{(b,c)} We show the 10th and 50th percentile evaluations. While our optimizer is able to match learning rate-tuned Adam on >50\% of the tasks after a modest amount of training, we are unable to reach better performance.
    Training occurred over the course of 1.2 days.
    \label{fig:selfopt_big_hyperlopt}
    }
\end{figure}

Finally, we meta-train the same meta-training setup used to train the optimizer used in this work. Due to the computational expense we were unable to perform careful studies of this setting. We construct a cluster of approximately 4K machines and meta-train with an outer-batch size of 20,480---5,120 batches of 4 vectorized tasks. First we test meta-training the exact same hypernetwork learned optimizer as used in VeLO (Figure~\ref{fig:selfopt_big_hyperlopt}). We found VeLO was not capable of this task, and quickly diverged.

\begin{figure}[t!]
    \centering
    \makebox[\textwidth]{%
    \begin{overpic}[width=1.0\textwidth]{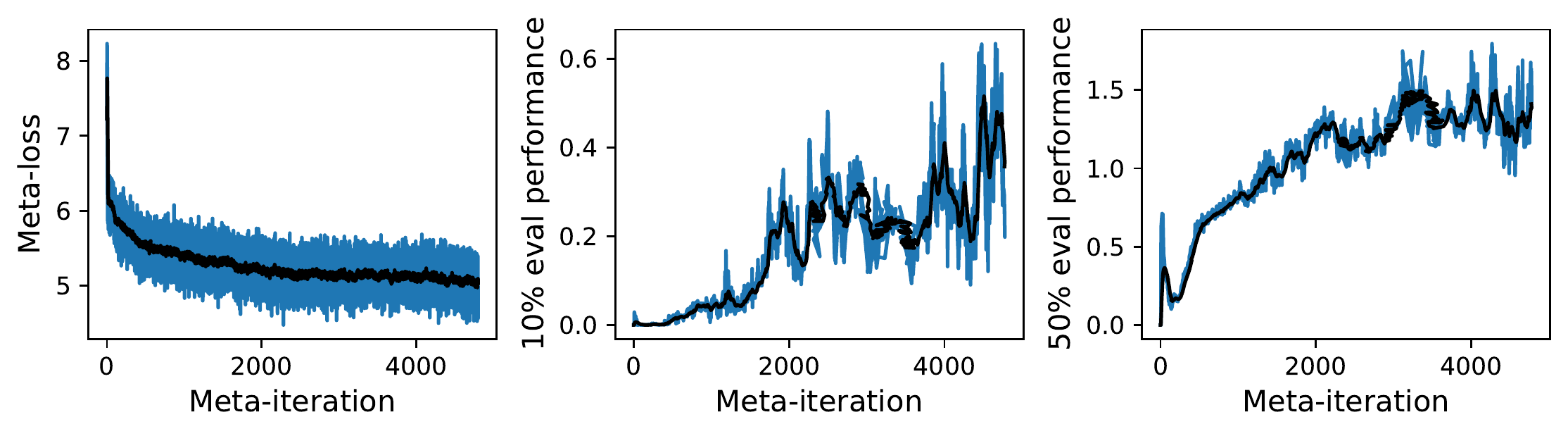}
    \put (7,23.0) {\textbf{\small(a)}}
    \put (40,23.0) {\textbf{\small(b)}}
    \put (74,23.0) {\textbf{\small(c)}}
    \end{overpic}
    }
    \caption{\textbf{Pre-trained learned optimizer re-training a hierarchical RNN LOpt.} \textbf{(a)} We show the meta-loss, the loss fed into the LOpt for training. \textbf{(b,c)} We show the 10th and 50th percentile evaluations. While our optimizer is able to match learning rate-tuned Adam on >50\% of the tasks after a modest amount of training, we are unable to reach better performance.
    Training occurred over the course of 3 days.
    \label{fig:selfopt_big_mlprnn}
    }
\end{figure}

Based on our results in the previous section, we initially believed this poor performance was due to the fact that architectures such as VeLO's---including, for example, hypernetwork multiplicative interactions---are far out of distribution. 
To test this, we take our same learned optimizer meta-training setup, but instead train the hierarchical learned optimizer from \citet{metz2020tasks} (Figure~\ref{fig:selfopt_big_mlprnn}).
We find our learned optimizer is capable of optimizing this model and continually making improvements. We see this as a promising initial result, but further study is needed.

\section{Open Source Details}
In this section we give a practical overview of how to use our learned optimizer. This includes known issues we have documented, as well as steps we have taken to work around them.

\subsection{Considerations When Trying VeLO}
While we believe our learned optimizer is vastly superior to prior work, it still has a couple of flaws and limitations.
\begin{enumerate}
    \item \textbf{Out of distribution tasks.} Models with optimization properties unlike those seen in meta-training (e.g. graph neural networks, reinforcement learning) sometimes have poor performance as discussed in Section \ref{subsec:limitations}. We expect broadening the meta-training distribution will help fix this.
    \item \textbf{Long training horizons.} The performance of VeLO is unreliable past 200K iterations. This is a limitation of the meta-training procedure. In practice, we found gradient accumulation can be used in place of longer horizon training.
    \item \textbf{Computational and memory overhead.} Depending on the task, our optimizers can require substantial overhead---about twice the memory requirements of Adam, as well as additional compute.
    While we believe in most cases this is acceptable, there will be cases where this is difficult or impossible to accommodate within a particular set of infrastructure choices. As outlined in Section~\ref{app:computational_costs_of_lopt}, however, our optimizers will work better given appropriate choices with modern distributed training techniques.
    \item \textbf{Longer compile times.} Our optimizers increase the number of operations greatly and this can cause JAX compile times to increase. In our experience, compilation can take up to four times as long as standard optimizers.
\end{enumerate}

\subsection{Released Optimizers and Learning Curves}
We release the following items:
\begin{itemize}
    \item \textbf{Learned optimizer implementations.} We release weights for a wide variety of pre-trained learned optimizers. The models can be loaded from \href{https://github.com/google/learned_optimization/blob/b1d8267c5e513a4112e7422b98bacc16e1f0e844/learned_optimization/research/general_lopt/pretrained_optimizers.py}{here}. We hope these different optimizers, meta-trained in different settings with different hyperparameters, will be useful to the study of optimization in general.
    \item \textbf{``Guard rail'' model.} In an effort to mitigate some of the previous issues, we release an optimizer wrapped in a set of ``guard rails''. First, to account for poor performance on longer training horizons, if the number of steps passed into the optimizer is greater than 150K, we will automatically do gradient accumulation to keep the number of steps the learned optimizer under 150K. Second, we provide an optional setting to add weight decay. In some cases, even a modest amount of weight decay (1\e-6) was able to stabilize a target model, and even improve performance. The code for these guard rails can be found \href{https://github.com/google/learned_optimization/blob/b1d8267c5e513a4112e7422b98bacc16e1f0e844/learned_optimization/research/general_lopt/prefab.py#L32}{here}.
    \item \textbf{Learning curves.} Finally, we release over a million training curves consisting of multiple different optimizer kinds, with different hyperparameters, evaluated on over 100 tasks. See our \href{https://learned-optimization.readthedocs.io/en/latest/optimizer_baselines.html}{docs} for more info. See this \href{https://colab.research.google.com/drive/1L2CuFrSvofW1L4BptHPjnfNGsBkFZ5-k#scrollTo=Iu5uuGqlGIQk}{colab} for an example of how to load the data and make plots similar to the ones presented in this work.
\end{itemize}

\section{Open Questions for Future Work} \label{app:open_questions}
The study of learned optimizers is relatively young. We have structured our research in two threads: exploratory work to better understand learned optimizers, and large scale training incorporating insights from the exploratory work.
While we recognize that the compute required for large scale meta-training is likely not available to all groups, there are still numerous areas open for research that have potential for high impact. 
We highlight a few such areas we think are interesting to study in follow up work.

\subsection{Learned Optimizer Architectures}
\paragraph{More features from inner-training.}
Currently, our learned optimizer takes in the mean gradient value.
This mirrors the standard interface of optimizers in deep learning.
By performing this reduction, however we are losing potentially useful information which could further accelerate training.
Additionally, in some distributed training setting, this information is already available for free.

\paragraph{Second, or approximate second order methods.}
Leveraging additional features such as those from approximate second order methods such as KFac~\citep{martens2015optimizing} or Shampoo~\citep{gupta2018shampoo, anil2020second} could greatly speed up training.
In addition to these approximations, many modern deep learning libraries have the ability to compute hessian-vector products which could also be useful for optimization. 

\paragraph{One model with many configurations.}
The models explored in this work provide only a single hyperparameter to the user---the number of steps one seeks to run the optimizer for.
In reality, researchers often have different use cases or goals for optimizers.
Creating learned optimizers that, at test time, can be configured to optimize targeting validation loss, optimize without a target length of time, or any other desired configuration is an open area of research. We attempted this, but had trouble meta-training such models.

\subsection{Learned Optimizer Introspection}
\paragraph{Learning about optimization by observing learned optimizers.}
Our learned optimizers have learned to optimize quite well, but understanding \textit{what} this optimizer is actually doing is currently difficult.
It is desirable to better understand specific mechanisms of action that arise in meta-training to study them theoretically or distill them into simpler optimizers. 

\subsection{Meta-Training}
\paragraph{Computational costs.}
The current computational cost of training learned optimizers on small problems is tractable---our small scale test problem in ~\ref{app:compare_architecture_capacity} can be trained on 4 TPU in a couple hours.
Our setup for training on more general distributions of tasks, however, uses vastly more compute requiring thousands of accelerators for over a month.
We believe that the compute requirements in meta-training will mimic those in neural architecture search, which originally required hundreds of accelerators~\citep{zoph2017}, whereas subsequent work was multiple orders of magnitude more efficient~\citep{pham2018efficient}.
We expect a similar level of speedup is possible by leveraging more, and better curricula, as well as leveraging more sophisticated meta-training techniques such as PES~\citep{pmlr-v139-vicol21a}, Truncated ES~\citep{metz2019understanding}, gradients computed via backprop~\citep{metz2019understanding}, or Guided ES~\citep{maheswaranathan2019guided}.

\paragraph{Data distributions.}
In this work, the data distribution on which we meta-trained is largely heuristic with minimal comparisons performed.
We expect this distribution is far too diverse in some areas and contains a number of models which are not representative of real world workloads, and likely deficient in other areas. 
The distribution of tasks we meta-train on also controls the computational cost of meta-training, so careful design of these distributions could also dramatically accelerate meta-training as well.

\paragraph{Systematic gradient staleness.}
When meta-training on diverse task distributions, meta-gradient estimation time can vary widely. For our longer horizons, we estimate a difference of 4 orders of magnitude across tasks in time taken.
When meta-training with asynchronous batched gradients the tasks which are slower to run will be computed on some weights from some previous point in meta-training.
This causes increased gradient staleness on these larger tasks.
As far as we are aware, this type of staleness is unique to this kind of meta-learning setup, and thus under-explored by the community.
We are unsure of the impact this has on meta-training, but have attempted to mitigate it by meta-training with very large outer-gradient batches.

\subsection{Usability of Learned Optimizers}
\paragraph{Controllability of the optimizer.}
While being hyperparameter-free makes experimentation with new architectures, datasets, or other training element considerably cheaper and easier, there are cases when having the absolute best model, or training procedure outweighs all compute considerations (for example, models being deployed to millions of users).
In these cases, it is desirable to be able to regain some level of specialization through tuning for a particular problem.
We explore this briefly in Appendix~\ref{app:per_task_finetune}, but found naively fine tuning with ES to be computationally costly.

\paragraph{Extended horizon training.}
One limitation we found with our optimizers is poor performance when extrapolating for much longer than the meta-training distribution was trained on. We anticipate that alternate meta-training strategies would mitigate this generalization in part, but more work is required in this direction. 

\paragraph{Better worst case performance.}
In our experiments, VeLO works well on around 90-95\% of the tasks we tried. In some cases, especially further away from the meta-training distribution, we get highly variable performance and even cases where our learned optimizer diverges.
Understanding why these optimizers explode and fixing this will be crucial for wider adoption. As an example of such a fix, \citet{premont2022simple} proposed an example fix to improve stability by dynamically switching to Adam.

\end{document}

%% file: app_table_adamnorm_single_trial.tex
\footnotesize
\begin{tabular}{||c c c c c c c c c||} 
     \hline
     Name & Trials & Avg Sp & Avg 1/Sp & 5\% & 10\% & 25\% & 50\% & 75\%  \\ [0.5ex] 
     \hline\hline
     
Adam LR=0.003 & 1 &0.64 & 3.24 & 0.01 & 0.03 & 0.24 & 0.72 & 1.00 \\
Adam LR=0.001 & 1 &0.67 & 2.68 & 0.09 & 0.15 & 0.35 & 0.64 & 1.01 \\
AdaBelief LR=0.001 & 1 &0.75 & 2.59 & 0.07 & 0.14 & 0.35 & 0.76 & 1.05 \\
AdaBelief LR=0.0003 & 1 &0.57 & 3.23 & 0.10 & 0.14 & 0.21 & 0.53 & 0.81 \\
SGD LR=0.3 & 1 &0.47 & 5.22 & 0.00 & 0.00 & 0.05 & 0.27 & 0.79 \\
SGD LR=0.1 & 1 &0.49 & 5.61 & 0.00 & 0.00 & 0.06 & 0.16 & 0.68 \\
SGDM LR=0.1 & 1 &0.51 & 4.68 & 0.00 & 0.00 & 0.07 & 0.36 & 0.83 \\
SGDM LR=0.03 & 1 &0.61 & 4.81 & 0.00 & 0.00 & 0.08 & 0.34 & 0.87 \\
RMSProp LR=0.001 & 1 &0.44 & 4.18 & 0.00 & 0.07 & 0.17 & 0.36 & 0.65 \\
RMSProp LR=0.0003 & 1 &0.65 & 4.48 & 0.06 & 0.08 & 0.15 & 0.31 & 0.60 \\
SM3 LR=0.03 & 1 &0.23 & 5.89 & 0.06 & 0.07 & 0.11 & 0.18 & 0.29 \\
SM3 LR=0.01 & 1 &0.24 & 6.44 & 0.06 & 0.07 & 0.09 & 0.15 & 0.32 \\
SM3beta2=999 LR=0.001 & 1 &0.49 & 3.52 & 0.05 & 0.12 & 0.21 & 0.37 & 0.72 \\
SM3beta2=999 LR=0.003 & 1 &0.43 & 4.55 & 0.00 & 0.02 & 0.10 & 0.36 & 0.64 \\
Yogi LR=0.01 & 1 &0.81 & 3.52 & 0.00 & 0.00 & 0.19 & 0.67 & 1.02 \\
Yogi LR=0.003 & 1 &0.73 & 3.38 & 0.02 & 0.08 & 0.21 & 0.62 & 0.98 \\
RAdam LR=0.003 & 1 &0.89 & 3.06 & 0.00 & 0.06 & 0.33 & 0.82 & 1.17 \\
RAdam LR=0.001 & 1 &1.17 & 2.32 & 0.10 & 0.20 & 0.36 & 0.80 & 1.09 \\
Lamb LR=0.003 & 1 &1.28 & 2.56 & 0.01 & 0.10 & 0.36 & 0.70 & 1.25 \\
Lamb LR=0.001 & 1 &1.43 & 2.52 & 0.05 & 0.19 & 0.32 & 0.62 & 1.01 \\
Lars LR=1 & 1 &0.85 & 2.75 & 0.01 & 0.13 & 0.31 & 0.52 & 1.09 \\
Lars LR=0.3 & 1 &0.94 & 3.83 & 0.07 & 0.10 & 0.20 & 0.39 & 0.67 \\
Fromage LR=0.01 & 1 &0.54 & 4.13 & 0.03 & 0.04 & 0.16 & 0.35 & 0.73 \\
Fromage LR=0.003 & 1 &0.32 & 6.00 & 0.00 & 0.02 & 0.07 & 0.16 & 0.50 \\
AdamW LR=0.001 & 1 &0.78 & 2.38 & 0.08 & 0.17 & 0.40 & 0.83 & 1.05 \\
AdamW LR=0.0003 & 1 &0.90 & 3.16 & 0.11 & 0.13 & 0.23 & 0.52 & 0.83 \\
Adafactor LR=0.01 & 1 &1.85 & 1.80 & 0.22 & 0.37 & 0.50 & 0.75 & 1.30 \\
Adafactor LR=0.003 & 1 &1.20 & 3.28 & 0.08 & 0.11 & 0.26 & 0.43 & 0.76 \\
AdaGrad LR=0.1 & 1 &0.58 & 4.57 & 0.00 & 0.01 & 0.10 & 0.35 & 0.84 \\
AdaGrad LR=0.3 & 1 &0.53 & 5.06 & 0.00 & 0.00 & 0.05 & 0.33 & 0.63 \\
ShampooSgd LR=0.03 & 1 &1.69 & 3.97 & 0.00 & 0.00 & 0.08 & 0.91 & 2.45 \\
ShampooSgd LR=0.1 & 1 &1.14 & 3.93 & 0.00 & 0.00 & 0.09 & 0.71 & 1.65 \\
ShampooAdagrad LR=0.003 & 1 &0.94 & 2.67 & 0.03 & 0.12 & 0.31 & 0.71 & 1.11 \\
ShampooAdagrad LR=0.001 & 1 &0.99 & 3.09 & 0.10 & 0.13 & 0.22 & 0.43 & 0.94 \\
ShampooRmsprop LR=0.0001 & 1 &1.22 & 2.49 & 0.03 & 0.11 & 0.39 & 0.91 & 1.23 \\
ShampooRmsprop LR=3e-05 & 1 &1.17 & 2.15 & 0.19 & 0.23 & 0.34 & 0.65 & 1.06 \\
ShampooRmspropNormalized LR=0.0001 & 1 &2.66 & 2.34 & 0.10 & 0.15 & 0.39 & 0.86 & 1.23 \\
ShampooRmspropNormalized LR=3e-05 & 1 &1.99 & 2.20 & 0.16 & 0.23 & 0.33 & 0.64 & 1.10 \\
ShampooSqrtN LR=0.0001 & 1 &3.49 & 3.09 & 0.04 & 0.06 & 0.21 & 1.30 & 2.25 \\
ShampooSqrtN LR=3e-05 & 1 &2.15 & 3.27 & 0.05 & 0.06 & 0.24 & 0.69 & 1.76 \\
ShampooAdagradNormalized LR=0.003 & 1 &2.22 & 2.57 & 0.12 & 0.14 & 0.34 & 0.57 & 0.98 \\
ShampooAdagradNormalized LR=0.001 & 1 &1.23 & 3.41 & 0.10 & 0.12 & 0.21 & 0.41 & 0.74 \\
\hline
OptList idx=0 & 1 &1.36 & 2.02 & 0.09 & 0.18 & 0.55 & 1.05 & 1.42 \\
NAdamW idx=37 & 1 &1.67 & 2.52 & 0.00 & 0.03 & 0.62 & 1.44 & 1.96 \\
NAdamW idx=0 & 1 &1.36 & 2.02 & 0.09 & 0.18 & 0.55 & 1.05 & 1.42 \\
\hline
RNN MLP LOpt & 1 &3.39 & 2.41 & 0.00 & 0.06 & 0.63 & 0.96 & 1.74 \\
STAR (wd=1.0) LOpt & 1 &1.00 & 3.42 & 0.01 & 0.06 & 0.20 & 0.50 & 1.28 \\
STAR (wd=0) LOpt & 1 &0.41 & 5.51 & 0.02 & 0.04 & 0.07 & 0.22 & 0.63 \\
STAR (wd=0, output\_norm=False) LOpt & 1 &0.53 & 5.09 & 0.00 & 0.00 & 0.05 & 0.35 & 0.68 \\
\hline
VeLO (Ours) & 1 &7.10 & 1.02 & 0.03 & 1.20 & 1.81 & 3.25 & 7.21 \\

\hline

\end{tabular}

%% file: app_table_adamnorm_multi_trial.tex
\footnotesize
\begin{tabular}{||c c c c c c c c c||} 
     \hline
     Name & Trials & Avg Sp & Avg 1/Sp & 5\% & 10\% & 25\% & 50\% & 75\%  \\ [0.5ex] 
     \hline\hline

VeLO (Ours) & 1 &7.10 & 1.02 & 0.03 & 1.20 & 1.81 & 3.25 & 7.21 \\
\hline
Adam & 14 &1.03 & 1.16 & 0.43 & 0.74 & 0.96 & 1.03 & 1.11 \\
AdaBelief & 14 &1.53 & 1.11 & 0.52 & 0.67 & 0.92 & 1.06 & 1.32 \\
SGD & 14 &1.10 & 3.67 & 0.05 & 0.07 & 0.16 & 0.53 & 0.94 \\
SGDM & 14 &1.32 & 2.30 & 0.09 & 0.18 & 0.36 & 0.90 & 1.16 \\
RMSProp & 14 &0.94 & 2.63 & 0.10 & 0.14 & 0.40 & 0.59 & 0.81 \\
SM3 & 14 &0.41 & 3.87 & 0.10 & 0.13 & 0.20 & 0.31 & 0.51 \\
SM3beta2=999 & 14 &0.78 & 1.99 & 0.20 & 0.28 & 0.47 & 0.71 & 0.88 \\
Yogi & 14 &1.74 & 1.24 & 0.25 & 0.50 & 0.79 & 1.11 & 1.55 \\
RAdam & 14 &1.78 & 0.99 & 0.44 & 0.68 & 0.94 & 1.14 & 1.53 \\
Lamb & 14 &2.00 & 1.50 & 0.24 & 0.32 & 0.62 & 0.92 & 1.43 \\
Lars & 14 &1.53 & 1.93 & 0.22 & 0.26 & 0.41 & 0.70 & 1.34 \\
Fromage & 14 &1.08 & 3.40 & 0.04 & 0.07 & 0.26 & 0.43 & 0.80 \\
AdamW & 14 &1.51 & 1.08 & 0.46 & 0.72 & 0.96 & 1.05 & 1.25 \\
Adafactor & 14 &2.06 & 1.55 & 0.31 & 0.40 & 0.54 & 0.79 & 1.41 \\
AdaGrad & 14 &0.94 & 2.50 & 0.11 & 0.15 & 0.33 & 0.79 & 1.11 \\
ShampooSgd & 14 &3.25 & 1.79 & 0.08 & 0.25 & 0.42 & 1.56 & 3.99 \\
ShampooAdagrad & 14 &1.48 & 1.29 & 0.30 & 0.48 & 0.66 & 0.95 & 1.65 \\
ShampooRmsprop & 14 &1.83 & 1.02 & 0.40 & 0.53 & 0.88 & 1.19 & 1.89 \\
ShampooRmspropNormalized & 14 &3.25 & 1.02 & 0.52 & 0.64 & 0.82 & 1.14 & 1.82 \\
ShampooSqrtN & 14 &4.45 & 2.42 & 0.05 & 0.06 & 0.93 & 1.71 & 4.48 \\
ShampooAdagradNormalized & 14 &3.53 & 1.25 & 0.41 & 0.46 & 0.58 & 0.83 & 1.67 \\
\hline
OptList5 & 5 &3.12 & 0.98 & 0.42 & 0.71 & 1.06 & 1.41 & 2.33 \\
OptList10 & 10 &3.76 & 0.64 & 0.76 & 1.09 & 1.26 & 1.60 & 3.18 \\
\hline
NAdamW & 1000 &7.74 & 0.35 & 1.47 & 1.56 & 1.80 & 2.70 & 8.66 \\
\hline
Best LR (Shampoo) & 90 &6.60 & 0.46 & 0.85 & 0.97 & 1.56 & 3.01 & 6.50 \\
Best LR (No Shampoo) & 225 &4.40 & 0.56 & 1.06 & 1.09 & 1.21 & 1.76 & 3.14 \\
Best LR (All) & 315 &7.70 & 0.34 & 1.40 & 1.56 & 1.76 & 3.49 & 6.83 \\
\hline
Best Baseline & 1315 &8.91 & 0.28 & 1.56 & 1.64 & 2.26 & 4.27 & 10.95 \\
\hline
Best LOpt Ensemble & 76 &11.42 & 0.41 & 1.46 & 1.65 & 2.89 & 6.28 & 20.28 \\
\hline
LOpt Ensemble 1 & 1 &7.37 & 1.13 & 0.06 & 0.79 & 1.63 & 3.32 & 7.61 \\
LOpt Ensemble 2 & 2 &8.55 & 0.88 & 0.11 & 1.38 & 2.02 & 4.24 & 8.96 \\
LOpt Ensemble 3 & 3 &9.54 & 0.72 & 0.41 & 1.41 & 2.42 & 4.59 & 14.96 \\
LOpt Ensemble 4 & 4 &9.99 & 0.71 & 0.42 & 1.43 & 2.47 & 4.94 & 15.59 \\
LOpt Ensemble 5 & 5 &10.30 & 0.70 & 0.42 & 1.45 & 2.64 & 4.94 & 15.59 \\
LOpt Ensemble 5 & 5 &10.30 & 0.70 & 0.42 & 1.45 & 2.64 & 4.94 & 15.59 \\
LOpt Ensemble 10 & 10 &11.06 & 0.68 & 0.42 & 1.50 & 2.64 & 5.62 & 20.14 \\
LOpt Ensemble 15 & 15 &11.33 & 0.42 & 1.44 & 1.60 & 2.84 & 6.28 & 20.27 \\
LOpt Ensemble 20 & 20 &11.40 & 0.42 & 1.44 & 1.62 & 2.84 & 6.28 & 20.27 \\
LOpt Ensemble 30 & 30 &11.42 & 0.41 & 1.46 & 1.65 & 2.89 & 6.28 & 20.28 \\
LOpt Ensemble 37 & 37 &11.42 & 0.41 & 1.46 & 1.65 & 2.89 & 6.28 & 20.28 \\
\hline
Best Everything & 1391 &12.63 & 0.21 & 1.67 & 1.82 & 3.12 & 6.95 & 24.17 \\

        \hline
\end{tabular}

%% file: app_table_baselinenorm_single_trial.tex
\footnotesize
\begin{tabular}{||c c c c c c c c c||} 
     \hline
     Name & Trials & Avg Sp & Avg 1/Sp & 5\% & 10\% & 25\% & 50\% & 75\%  \\ [0.5ex] 
     \hline\hline
Adam LR=0.003 & 1 &0.47 & 3.74 & 0.00 & 0.03 & 0.22 & 0.50 & 0.69 \\
Adam LR=0.001 & 1 &0.51 & 3.35 & 0.08 & 0.11 & 0.26 & 0.49 & 0.78 \\
AdaBelief LR=0.001 & 1 &0.54 & 3.11 & 0.06 & 0.09 & 0.28 & 0.53 & 0.78 \\
AdaBelief LR=0.0003 & 1 &0.41 & 3.97 & 0.09 & 0.12 & 0.18 & 0.32 & 0.62 \\
SGD LR=0.3 & 1 &0.34 & 5.47 & 0.00 & 0.00 & 0.05 & 0.24 & 0.65 \\
SGD LR=0.1 & 1 &0.34 & 5.91 & 0.00 & 0.00 & 0.06 & 0.15 & 0.60 \\
SGDM LR=0.03 & 1 &0.42 & 5.08 & 0.00 & 0.00 & 0.07 & 0.31 & 0.70 \\
SGDM LR=0.1 & 1 &0.39 & 5.02 & 0.00 & 0.00 & 0.06 & 0.31 & 0.71 \\
RMSProp LR=0.001 & 1 &0.32 & 4.94 & 0.00 & 0.06 & 0.11 & 0.29 & 0.49 \\
RMSProp LR=0.0003 & 1 &0.51 & 5.16 & 0.05 & 0.07 & 0.10 & 0.25 & 0.52 \\
SM3 LR=0.03 & 1 &0.19 & 6.54 & 0.06 & 0.07 & 0.11 & 0.16 & 0.25 \\
SM3 LR=0.01 & 1 &0.19 & 7.07 & 0.06 & 0.07 & 0.08 & 0.12 & 0.24 \\
SM3beta2=999 LR=0.001 & 1 &0.36 & 4.37 & 0.05 & 0.11 & 0.16 & 0.28 & 0.57 \\
SM3beta2=999 LR=0.003 & 1 &0.30 & 5.09 & 0.00 & 0.02 & 0.09 & 0.28 & 0.50 \\
Yogi LR=0.01 & 1 &0.53 & 4.10 & 0.00 & 0.00 & 0.12 & 0.53 & 0.80 \\
Yogi LR=0.003 & 1 &0.52 & 3.92 & 0.01 & 0.07 & 0.15 & 0.48 & 0.84 \\
RAdam LR=0.003 & 1 &0.57 & 3.47 & 0.00 & 0.06 & 0.28 & 0.61 & 0.91 \\
RAdam LR=0.001 & 1 &0.56 & 2.83 & 0.09 & 0.16 & 0.30 & 0.56 & 0.82 \\
Lamb LR=0.003 & 1 &0.59 & 2.86 & 0.01 & 0.10 & 0.33 & 0.58 & 0.78 \\
Lamb LR=0.001 & 1 &0.62 & 3.00 & 0.05 & 0.15 & 0.27 & 0.47 & 0.80 \\
Lars LR=1 & 1 &0.54 & 3.18 & 0.01 & 0.13 & 0.27 & 0.43 & 0.82 \\
Lars LR=0.3 & 1 &0.73 & 4.48 & 0.07 & 0.09 & 0.15 & 0.29 & 0.53 \\
Fromage LR=0.01 & 1 &0.38 & 4.72 & 0.03 & 0.04 & 0.10 & 0.27 & 0.59 \\
Fromage LR=0.003 & 1 &0.24 & 6.61 & 0.00 & 0.02 & 0.06 & 0.13 & 0.37 \\
AdamW LR=0.001 & 1 &0.54 & 2.99 & 0.06 & 0.12 & 0.30 & 0.53 & 0.80 \\
AdamW LR=0.0003 & 1 &0.68 & 3.86 & 0.09 & 0.11 & 0.19 & 0.33 & 0.63 \\
Adafactor LR=0.01 & 1 &0.94 & 2.22 & 0.18 & 0.25 & 0.41 & 0.63 & 0.90 \\
Adafactor LR=0.003 & 1 &0.58 & 3.84 & 0.08 & 0.09 & 0.21 & 0.37 & 0.59 \\
AdaGrad LR=0.3 & 1 &0.37 & 5.37 & 0.00 & 0.00 & 0.04 & 0.28 & 0.57 \\
AdaGrad LR=0.1 & 1 &0.42 & 4.87 & 0.00 & 0.00 & 0.08 & 0.28 & 0.71 \\
ShampooSgd LR=0.03 & 1 &1.07 & 4.18 & 0.00 & 0.00 & 0.07 & 0.68 & 1.60 \\
ShampooSgd LR=0.1 & 1 &0.81 & 4.37 & 0.00 & 0.00 & 0.06 & 0.56 & 1.30 \\
ShampooAdagrad LR=0.003 & 1 &0.63 & 3.20 & 0.03 & 0.10 & 0.28 & 0.50 & 0.85 \\
ShampooAdagrad LR=0.001 & 1 &0.57 & 3.82 & 0.08 & 0.12 & 0.18 & 0.35 & 0.75 \\
ShampooRmsprop LR=0.0001 & 1 &0.94 & 2.93 & 0.01 & 0.10 & 0.32 & 0.59 & 0.89 \\
ShampooRmsprop LR=3e-05 & 1 &0.90 & 2.87 & 0.14 & 0.17 & 0.27 & 0.41 & 0.81 \\
ShampooRmspropNormalized LR=0.0001 & 1 &2.12 & 2.76 & 0.07 & 0.13 & 0.32 & 0.57 & 0.92 \\
ShampooRmspropNormalized LR=3e-05 & 1 &1.69 & 2.84 & 0.13 & 0.17 & 0.27 & 0.45 & 0.82 \\
ShampooSqrtN LR=0.0001 & 1 &2.87 & 3.34 & 0.04 & 0.05 & 0.19 & 0.81 & 1.45 \\
ShampooSqrtN LR=3e-05 & 1 &1.52 & 3.70 & 0.05 & 0.06 & 0.22 & 0.45 & 1.18 \\
ShampooAdagradNormalized LR=0.003 & 1 &1.94 & 3.14 & 0.08 & 0.14 & 0.25 & 0.41 & 0.76 \\
ShampooAdagradNormalized LR=0.001 & 1 &0.80 & 4.10 & 0.08 & 0.11 & 0.16 & 0.29 & 0.61 \\
\hline
OptList idx=0 & 1 &0.86 & 2.52 & 0.07 & 0.12 & 0.40 & 0.67 & 1.01 \\
NAdamW idx=37 & 1 &0.93 & 2.81 & 0.00 & 0.03 & 0.43 & 1.04 & 1.31 \\
NAdamW idx=0 & 1 &0.86 & 2.52 & 0.07 & 0.12 & 0.40 & 0.67 & 1.01 \\
\hline
RNN MLP LOpt & 1 &2.99 & 2.83 & 0.00 & 0.06 & 0.44 & 0.70 & 1.09 \\
STAR (wd=1.0) LOpt & 1 &0.70 & 3.86 & 0.00 & 0.06 & 0.16 & 0.39 & 0.99 \\
STAR (wd=0) LOpt & 1 &0.31 & 5.91 & 0.02 & 0.04 & 0.07 & 0.20 & 0.47 \\
STAR (wd=0, output\_norm=False) LOpt & 1 &0.37 & 5.49 & 0.00 & 0.00 & 0.05 & 0.28 & 0.53 \\
\hline
VeLO (Ours) & 1 &5.05 & 1.19 & 0.02 & 0.79 & 1.20 & 2.18 & 3.87 \\

        \hline
\end{tabular}

%% file: app_table_baselinenorm_multi_trial.tex
\footnotesize
\begin{tabular}{||c c c c c c c c c||} 
     \hline
     Name & Trials & Avg Sp & Avg 1/Sp & 5\% & 10\% & 25\% & 50\% & 75\%  \\ [0.5ex] 
     \hline\hline

VeLO (Ours) & 1 &5.05 & 1.19 & 0.02 & 0.79 & 1.20 & 2.18 & 3.87 \\
\hline
Adam & 14 &0.72 & 1.83 & 0.22 & 0.37 & 0.57 & 0.78 & 0.91 \\
AdaBelief & 14 &1.06 & 1.57 & 0.40 & 0.43 & 0.64 & 0.81 & 0.94 \\
SGD & 14 &0.79 & 4.08 & 0.04 & 0.05 & 0.15 & 0.41 & 0.80 \\
SGDM & 14 &0.95 & 2.93 & 0.07 & 0.13 & 0.26 & 0.71 & 0.94 \\
RMSProp & 14 &0.67 & 3.42 & 0.10 & 0.11 & 0.24 & 0.46 & 0.66 \\
SM3 & 14 &0.30 & 4.71 & 0.08 & 0.11 & 0.15 & 0.26 & 0.38 \\
SM3beta2=999 & 14 &0.52 & 2.74 & 0.12 & 0.18 & 0.32 & 0.55 & 0.68 \\
Yogi & 14 &1.12 & 1.86 & 0.17 & 0.28 & 0.53 & 0.87 & 1.04 \\
RAdam & 14 &0.83 & 1.48 & 0.30 & 0.40 & 0.69 & 0.92 & 1.01 \\
Lamb & 14 &1.03 & 1.90 & 0.19 & 0.27 & 0.51 & 0.65 & 0.98 \\
Lars & 14 &1.18 & 2.44 & 0.16 & 0.20 & 0.33 & 0.53 & 0.89 \\
Fromage & 14 &0.71 & 4.01 & 0.04 & 0.07 & 0.19 & 0.35 & 0.61 \\
AdamW & 14 &1.04 & 1.64 & 0.30 & 0.42 & 0.67 & 0.81 & 0.92 \\
Adafactor & 14 &1.11 & 2.03 & 0.19 & 0.27 & 0.43 & 0.65 & 0.92 \\
AdaGrad & 14 &0.61 & 3.27 & 0.07 & 0.11 & 0.23 & 0.63 & 0.88 \\
ShampooSgd & 14 &2.16 & 2.39 & 0.07 & 0.13 & 0.30 & 1.22 & 2.27 \\
ShampooAdagrad & 14 &0.84 & 1.91 & 0.19 & 0.29 & 0.47 & 0.68 & 1.02 \\
ShampooRmsprop & 14 &1.26 & 1.58 & 0.24 & 0.42 & 0.56 & 0.80 & 1.15 \\
ShampooRmspropNormalized & 14 &2.61 & 1.51 & 0.33 & 0.44 & 0.55 & 0.79 & 1.14 \\
ShampooSqrtN & 14 &3.45 & 2.62 & 0.05 & 0.06 & 0.53 & 1.21 & 2.45 \\
ShampooAdagradNormalized & 14 &2.91 & 1.84 & 0.27 & 0.33 & 0.44 & 0.63 & 1.11 \\
\hline
OptList5 & 5 &1.70 & 1.35 & 0.32 & 0.42 & 0.78 & 1.00 & 1.27 \\
OptList10 & 10 &2.09 & 1.01 & 0.52 & 0.63 & 0.89 & 1.16 & 1.42 \\
\hline
NAdamW & 1000 &5.24 & 0.53 & 1.07 & 1.17 & 1.39 & 1.75 & 2.94 \\
\hline
Best LR (Shampoo) & 90 &5.08 & 0.71 & 0.58 & 0.64 & 1.01 & 1.61 & 4.13 \\
Best LR (No Shampoo) & 225 &2.76 & 0.80 & 0.83 & 0.92 & 1.00 & 1.12 & 1.50 \\
Best LR (All) & 315 &5.80 & 0.52 & 0.98 & 1.10 & 1.29 & 1.81 & 4.18 \\
\hline
Best Baseline & 1315 &6.68 & 0.43 & 1.17 & 1.28 & 1.57 & 2.24 & 4.73 \\
\hline
Best LOpt Ensemble & 76 &8.62 & 0.53 & 0.93 & 1.29 & 1.87 & 3.38 & 9.57 \\
\hline
LOpt Ensemble 1 & 1 &5.05 & 1.19 & 0.02 & 0.79 & 1.20 & 2.18 & 3.87 \\
LOpt Ensemble 2 & 2 &6.00 & 1.01 & 0.11 & 0.96 & 1.50 & 2.48 & 5.23 \\
LOpt Ensemble 3 & 3 &6.65 & 0.90 & 0.34 & 1.05 & 1.59 & 2.60 & 6.28 \\
LOpt Ensemble 4 & 4 &7.09 & 0.89 & 0.35 & 1.08 & 1.75 & 2.62 & 6.32 \\
LOpt Ensemble 5 & 5 &7.39 & 0.88 & 0.35 & 1.16 & 1.75 & 2.62 & 6.68 \\
LOpt Ensemble 5 & 5 &7.39 & 0.88 & 0.35 & 1.16 & 1.75 & 2.62 & 6.68 \\
LOpt Ensemble 10 & 10 &8.18 & 0.84 & 0.35 & 1.23 & 1.80 & 3.09 & 7.90 \\
LOpt Ensemble 15 & 15 &8.46 & 0.81 & 0.44 & 1.23 & 1.80 & 3.23 & 9.57 \\
LOpt Ensemble 20 & 20 &8.56 & 0.54 & 0.91 & 1.25 & 1.80 & 3.33 & 9.57 \\
LOpt Ensemble 30 & 30 &8.61 & 0.53 & 0.93 & 1.29 & 1.87 & 3.34 & 9.57 \\
LOpt Ensemble 37 & 37 &8.62 & 0.53 & 0.93 & 1.29 & 1.87 & 3.38 & 9.57 \\
\hline
Best Everything & 1391 &10.21 & 0.30 & 1.29 & 1.51 & 2.32 & 4.05 & 16.23 \\
\hline

\end{tabular}

%% file: app_page1.tex

    \makebox[\textwidth]{%
    \begin{overpic}[width=0.23\textwidth]{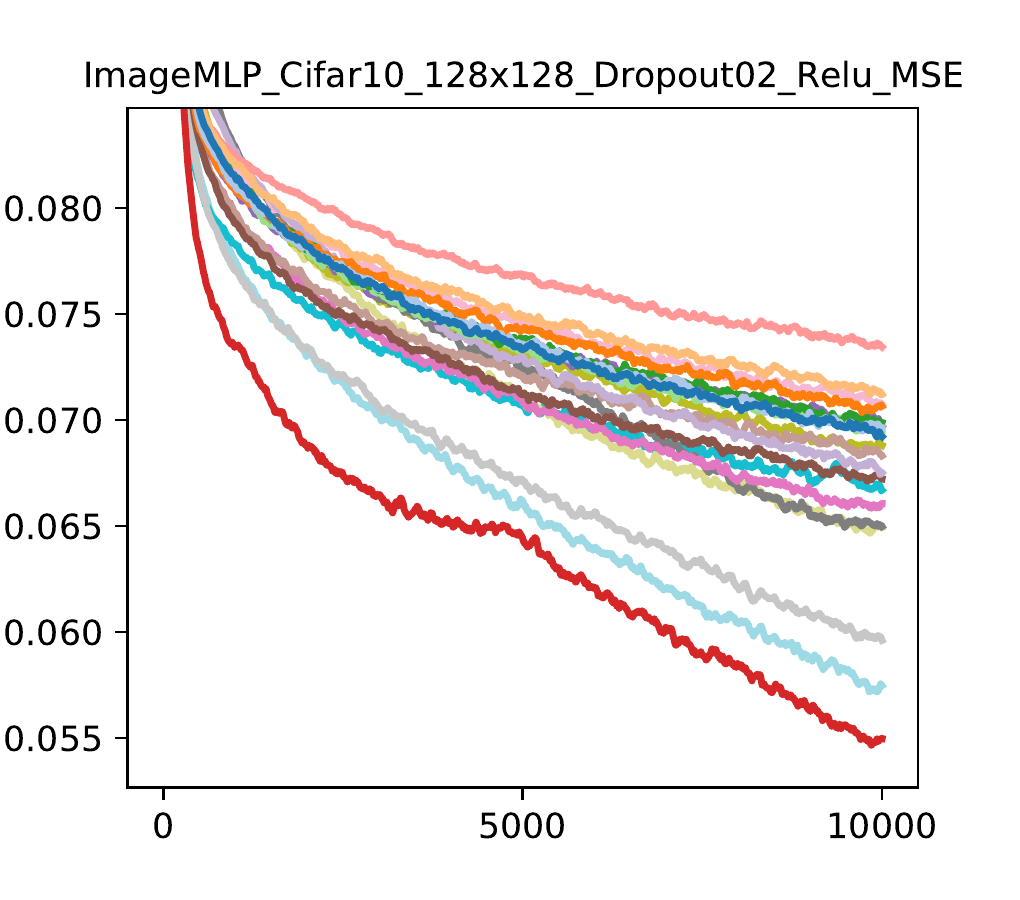}
    \end{overpic}
    \begin{overpic}[width=0.23\textwidth]{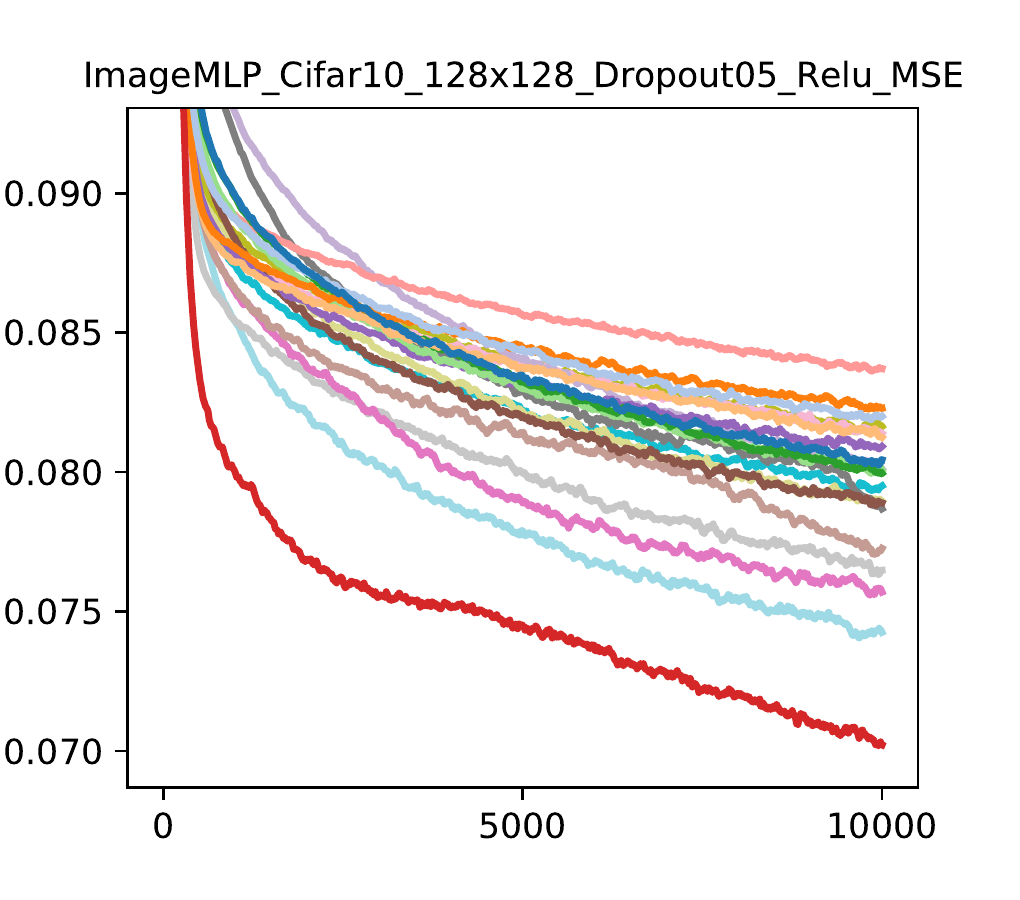}
    \end{overpic}
    \begin{overpic}[width=0.23\textwidth]{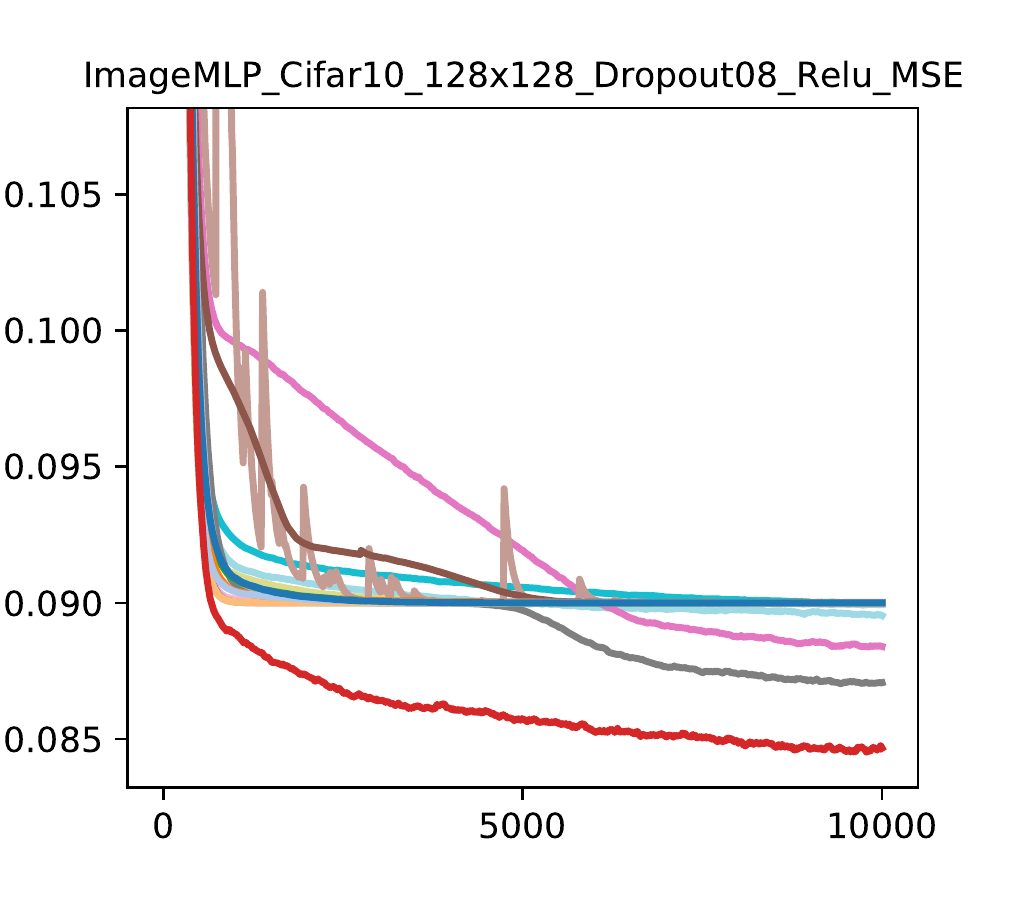}
    \end{overpic}
    \begin{overpic}[width=0.23\textwidth]{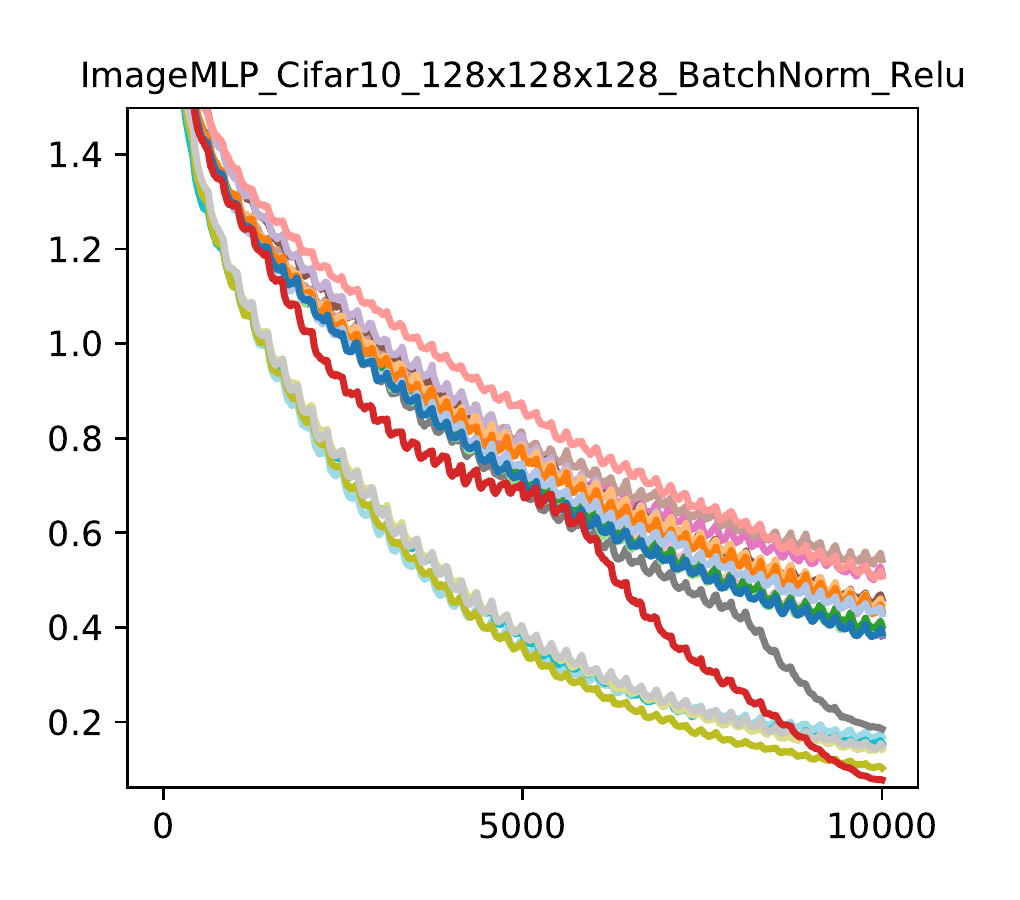}
    \end{overpic}
    \begin{overpic}[width=0.23\textwidth]{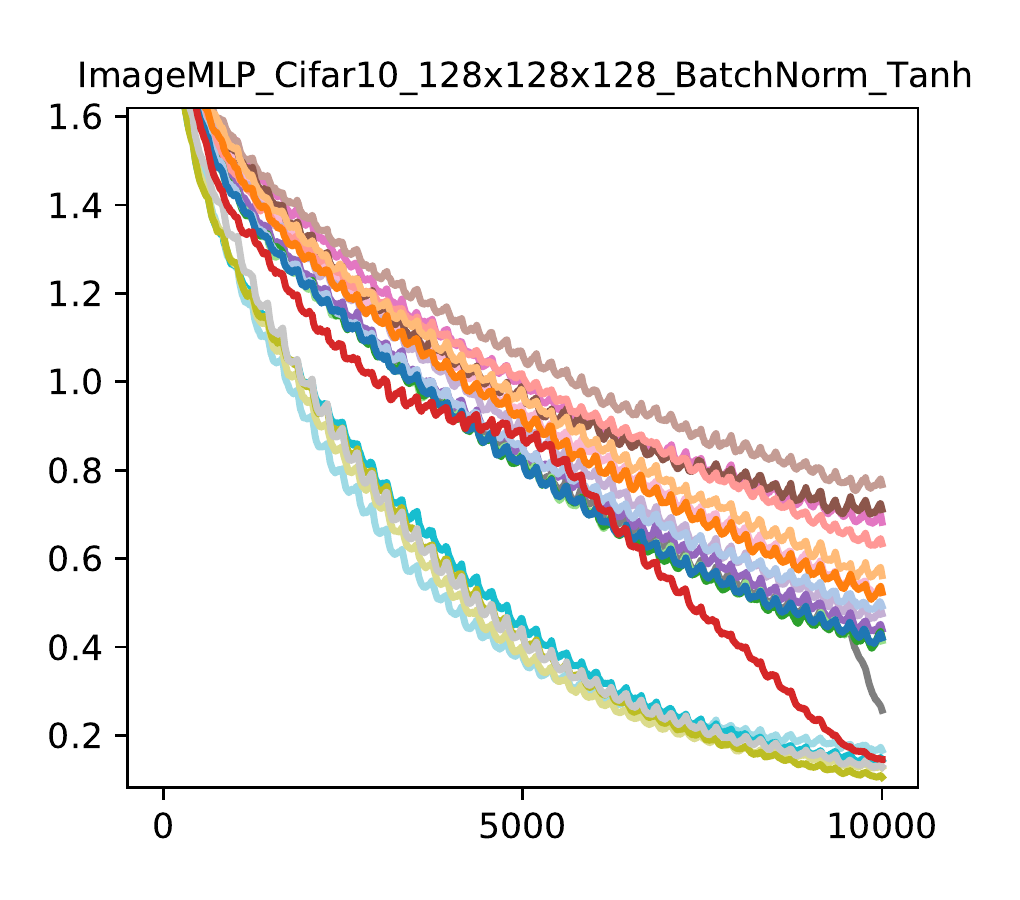}
    \end{overpic}
    }

    \makebox[\textwidth]{%
    \begin{overpic}[width=0.23\textwidth]{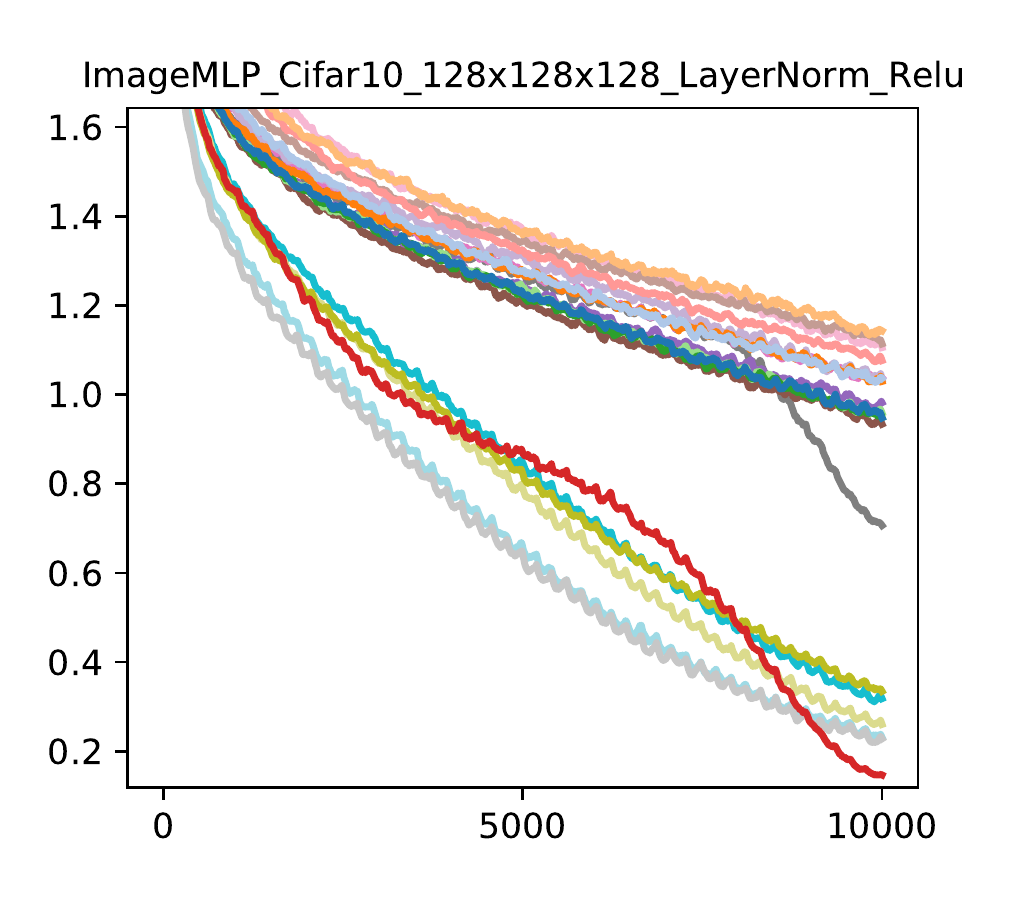}
    \end{overpic}
    \begin{overpic}[width=0.23\textwidth]{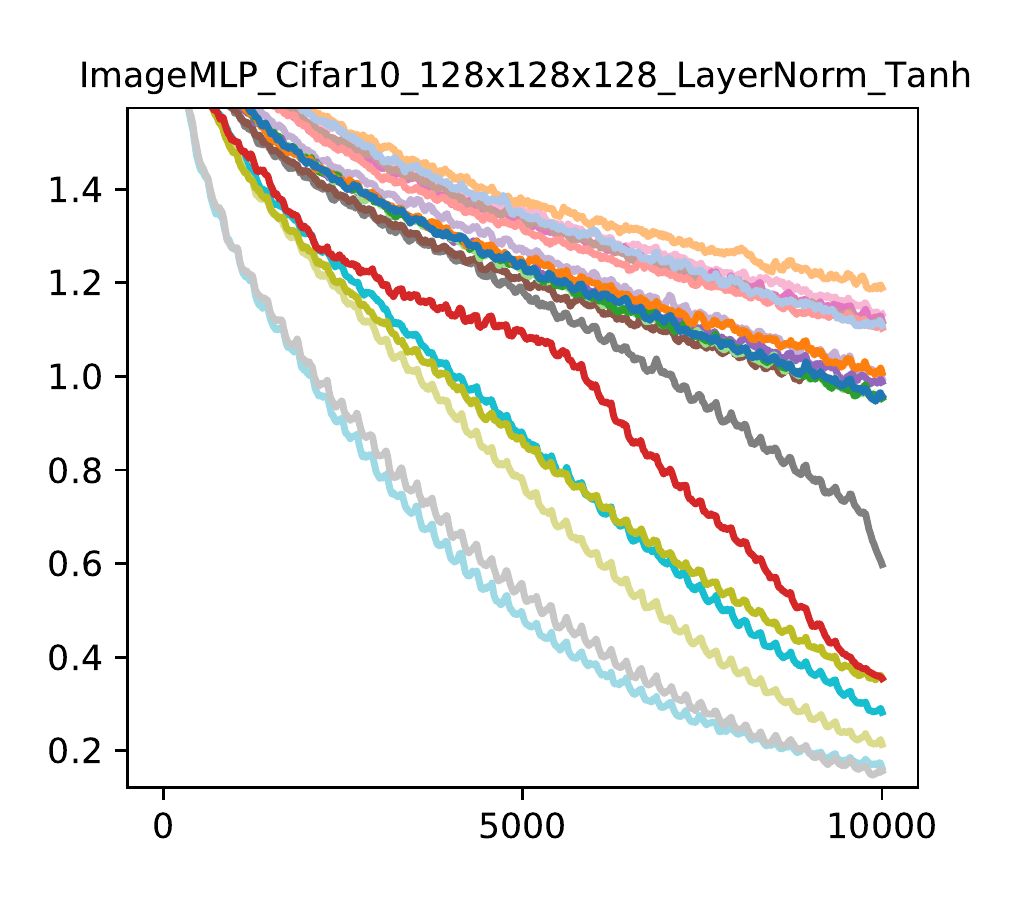}
    \end{overpic}
    \begin{overpic}[width=0.23\textwidth]{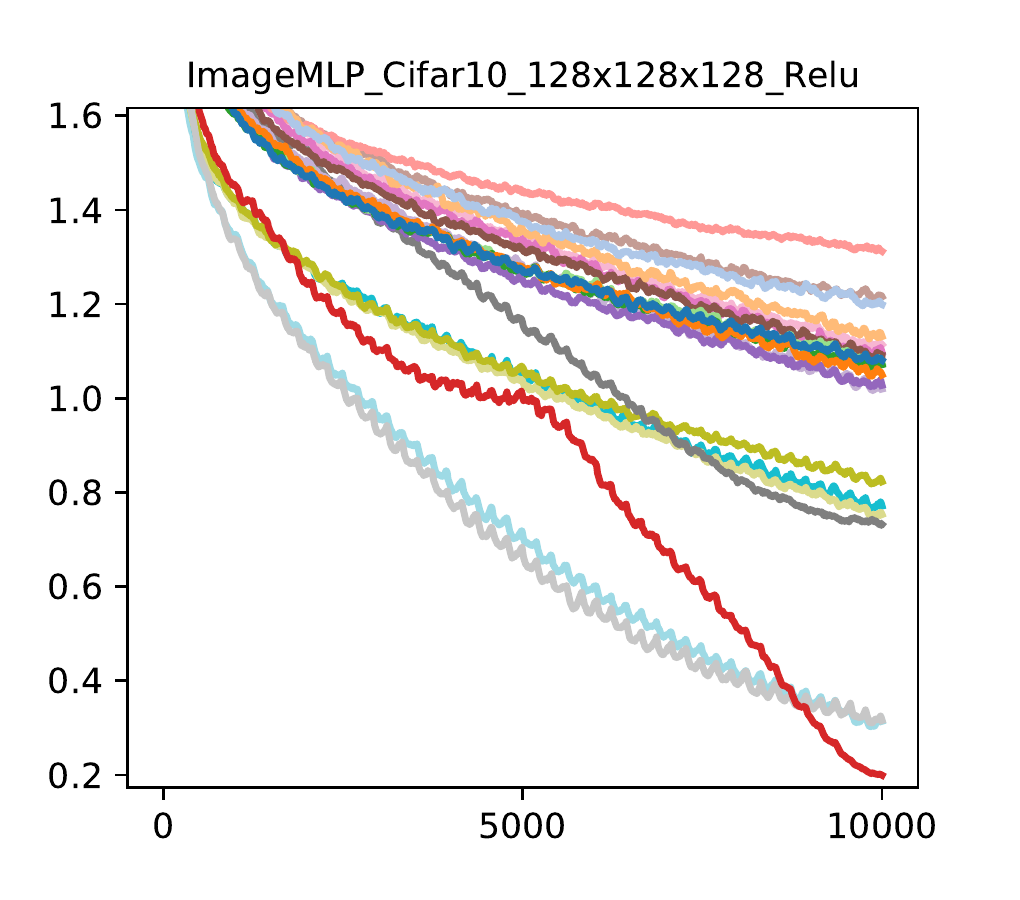}
    \end{overpic}
    \begin{overpic}[width=0.23\textwidth]{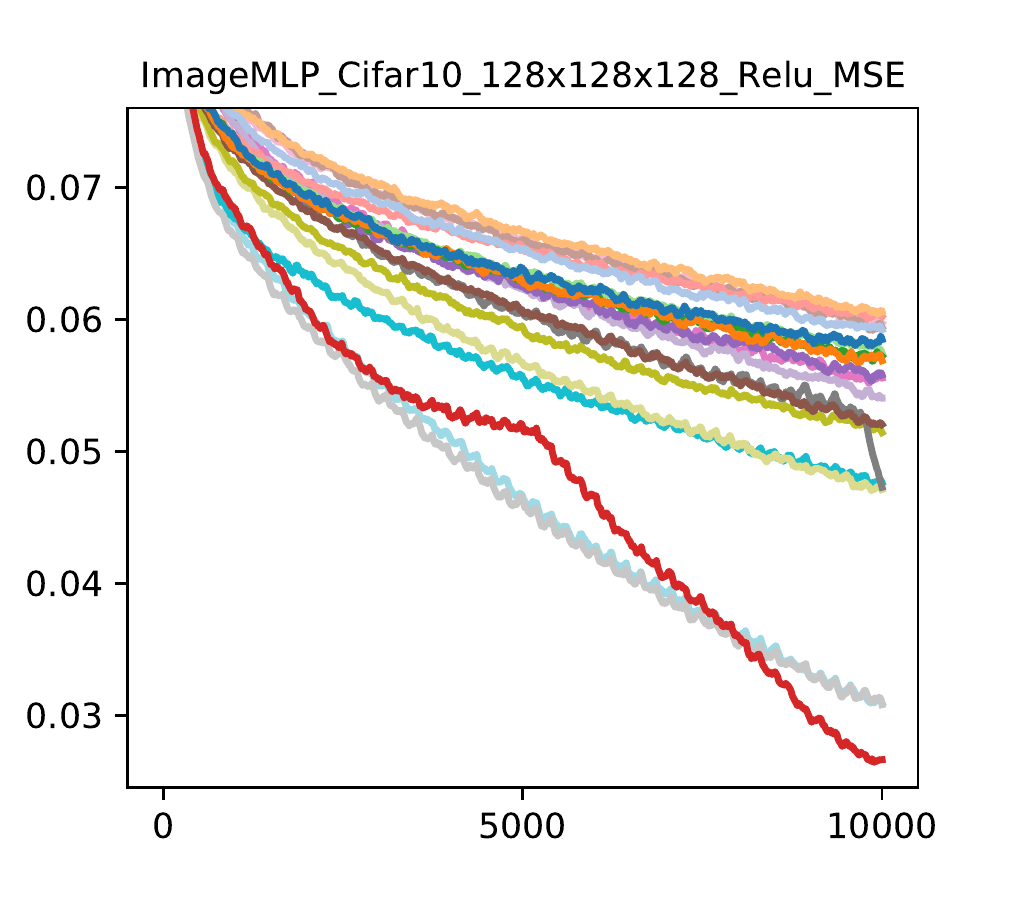}
    \end{overpic}
    \begin{overpic}[width=0.23\textwidth]{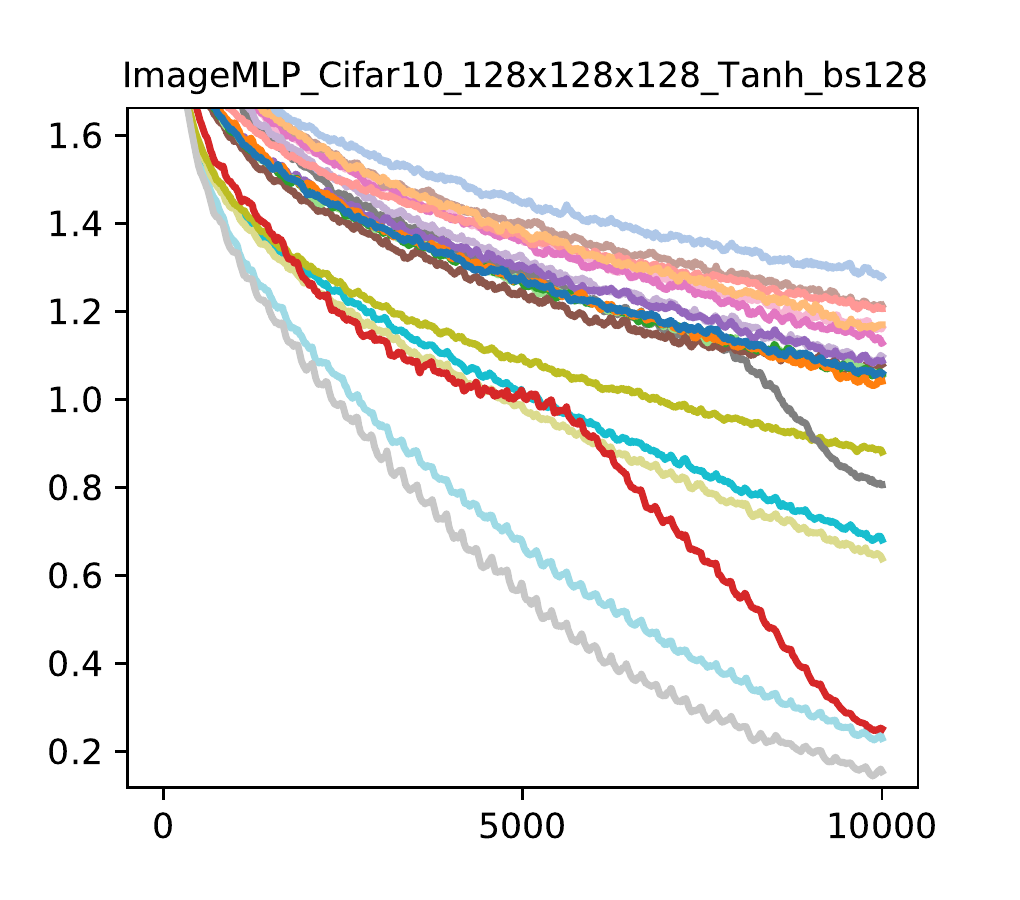}
    \end{overpic}
    }


    \makebox[\textwidth]{%
    \begin{overpic}[width=0.23\textwidth]{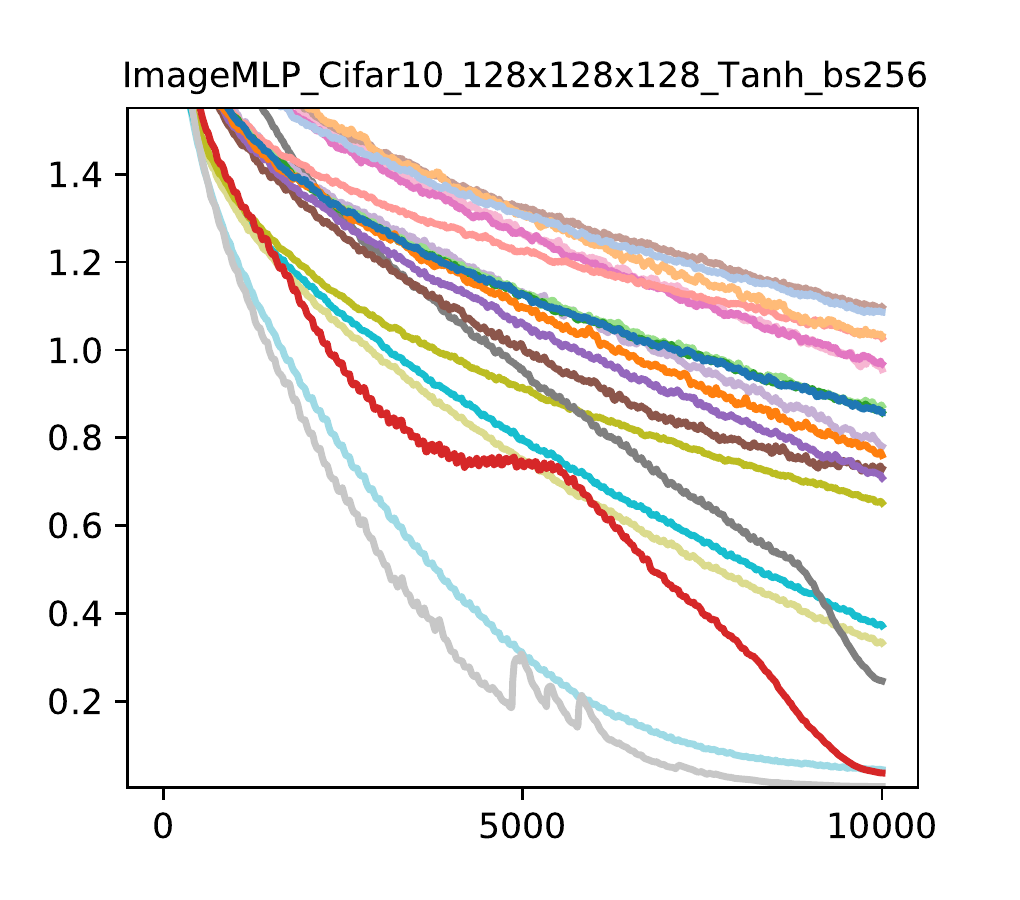}
    \end{overpic}
    \begin{overpic}[width=0.23\textwidth]{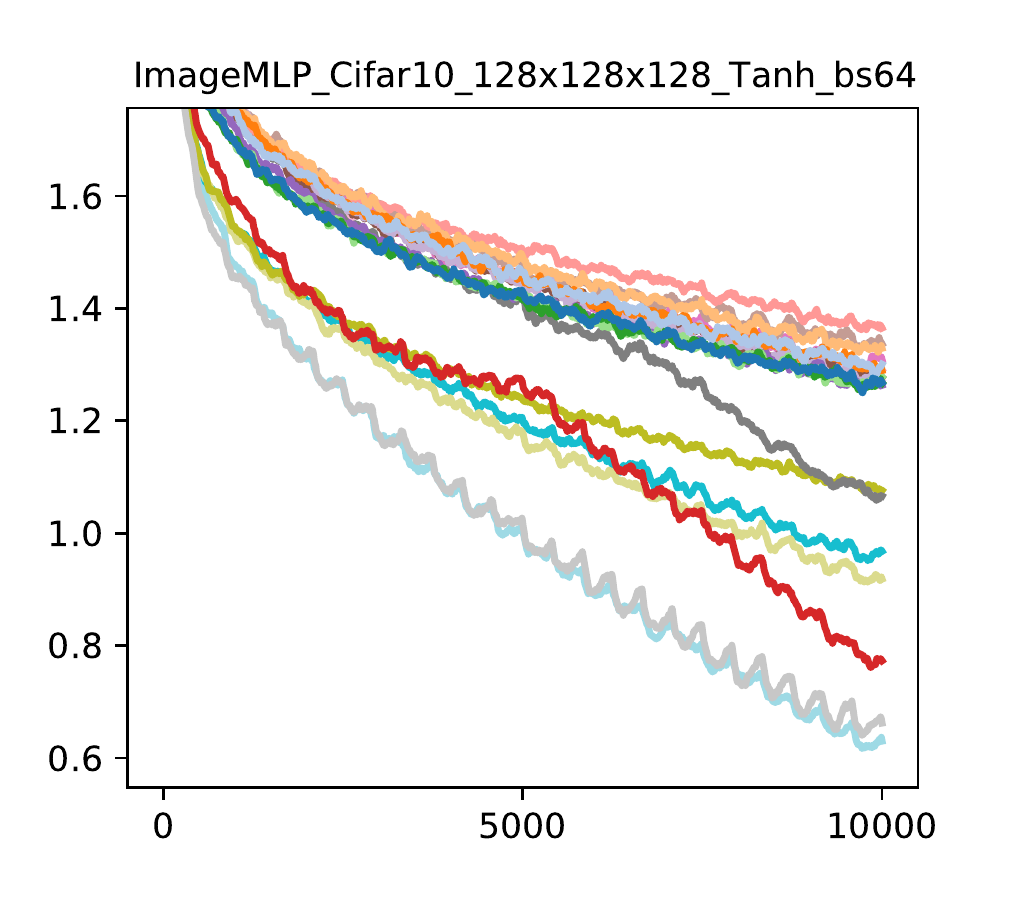}
    \end{overpic}
    \begin{overpic}[width=0.23\textwidth]{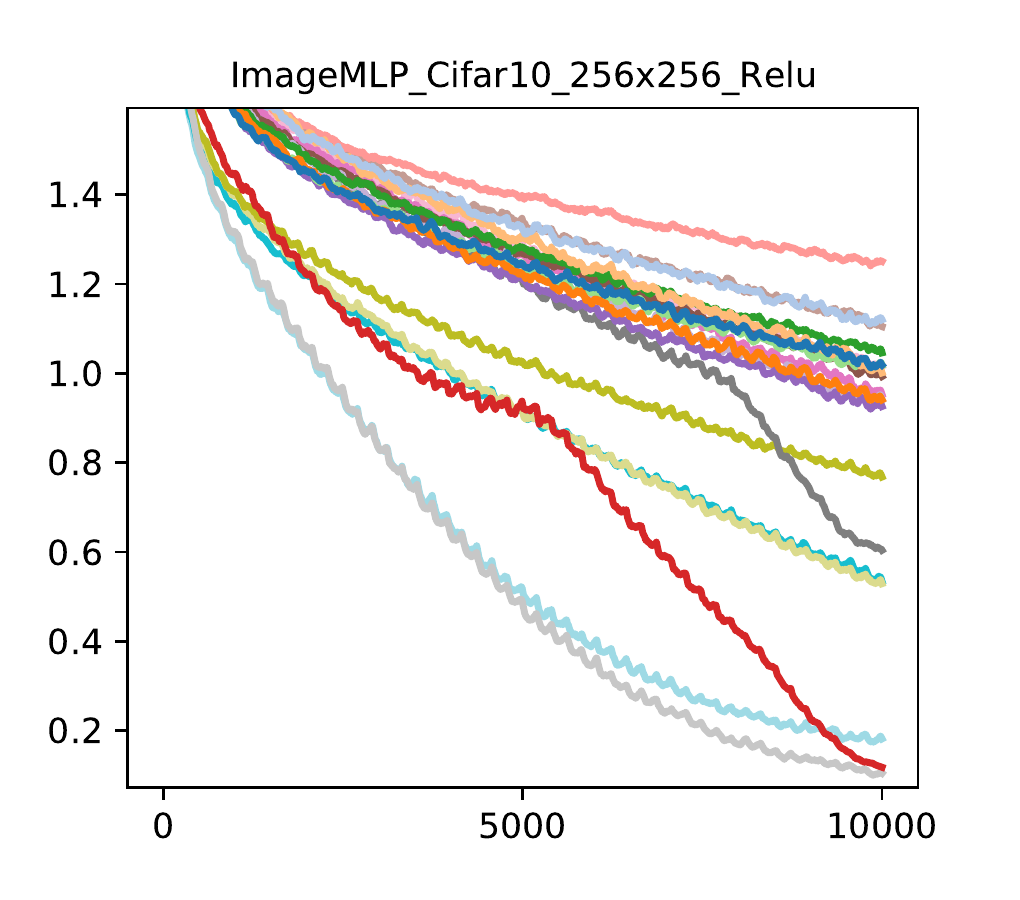}
    \end{overpic}
    \begin{overpic}[width=0.23\textwidth]{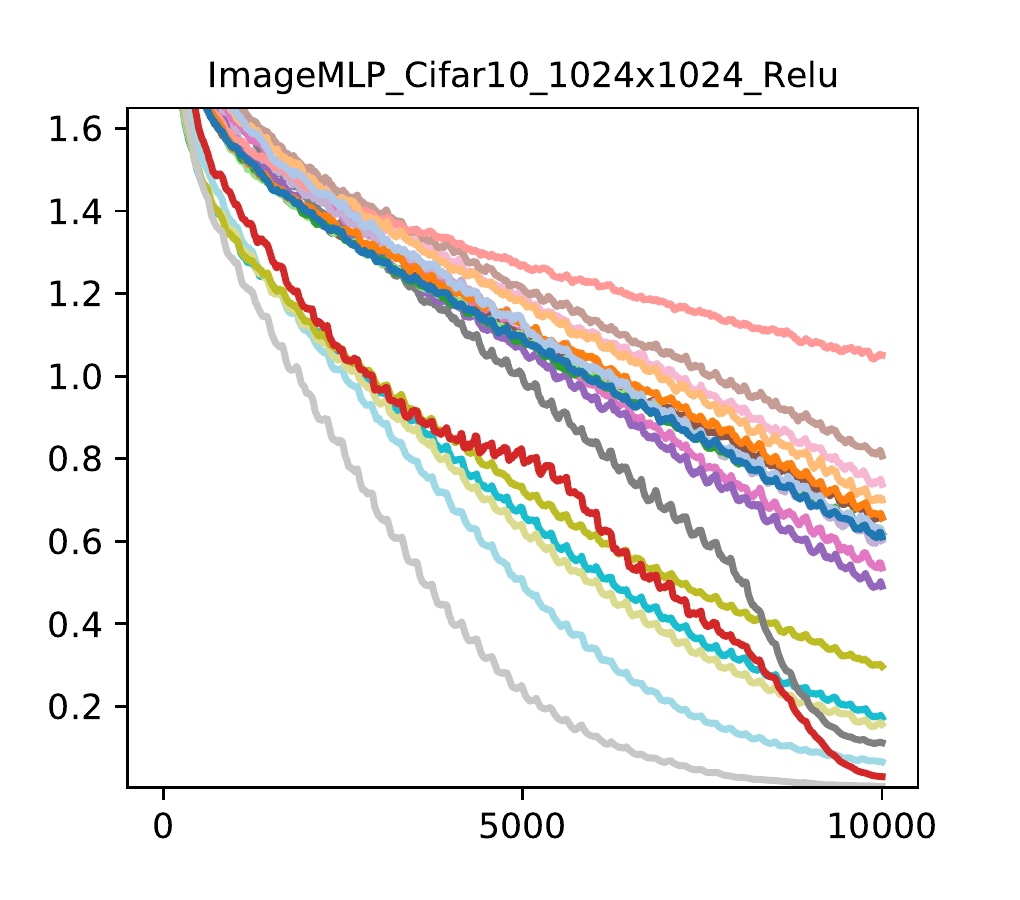}
    \end{overpic}
    \begin{overpic}[width=0.23\textwidth]{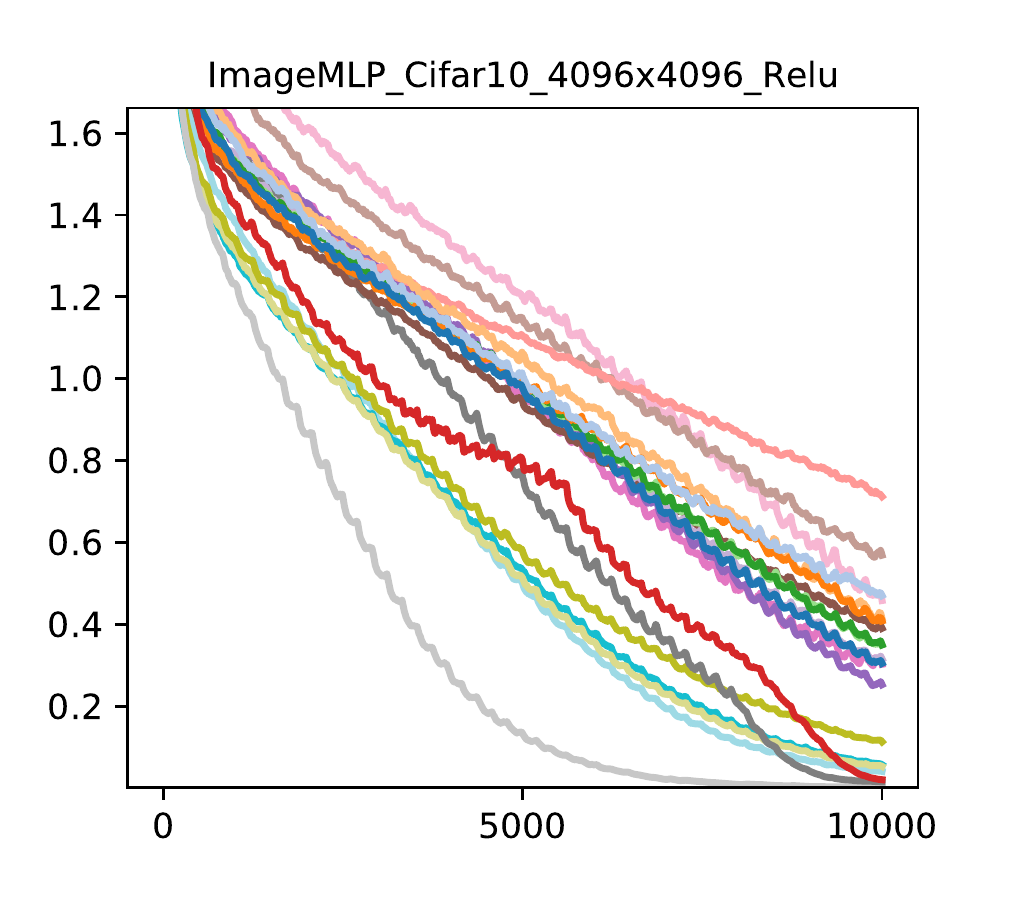}
    \end{overpic}
    }


    \makebox[\textwidth]{%
    \begin{overpic}[width=0.23\textwidth]{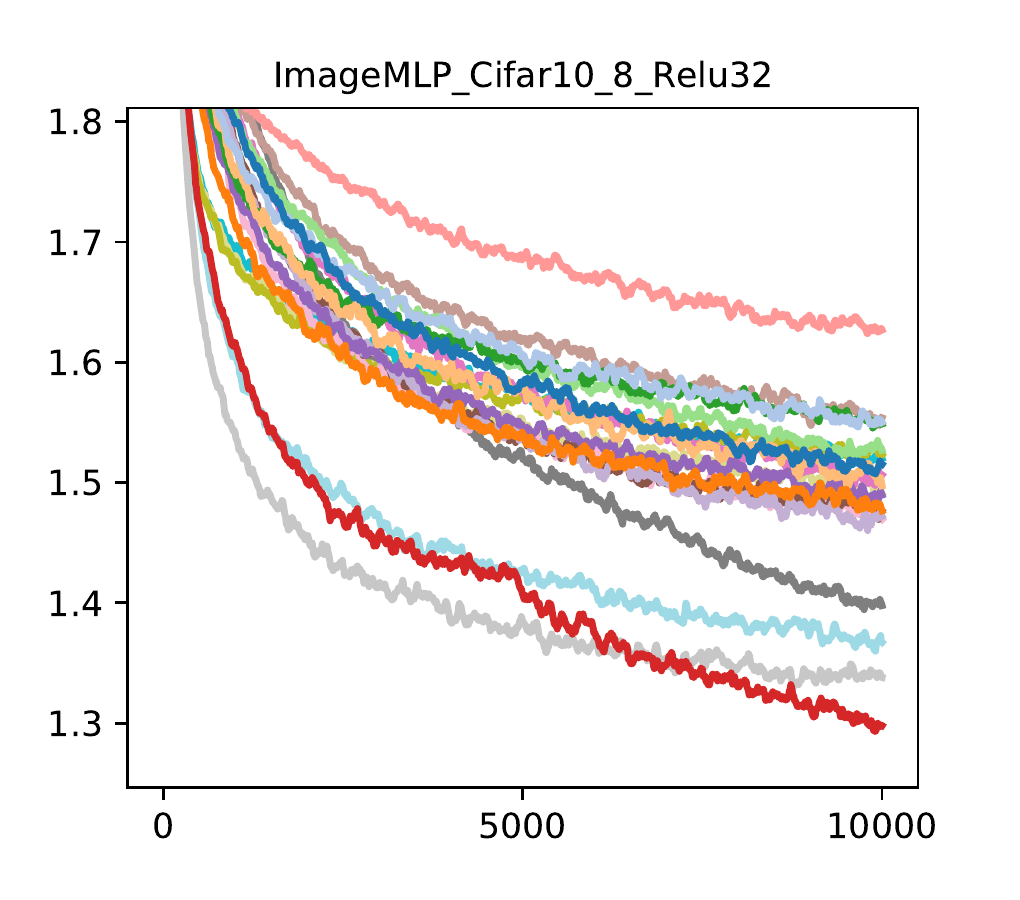}
    \end{overpic}
    \begin{overpic}[width=0.23\textwidth]{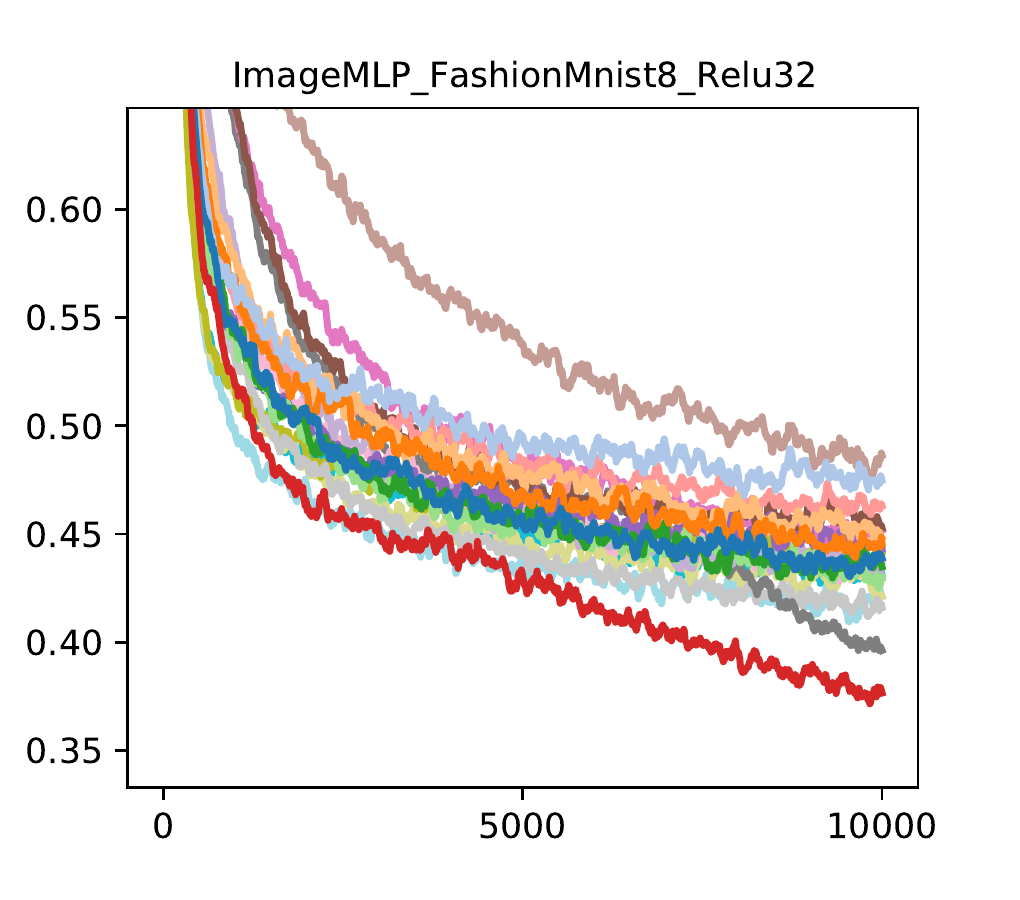}
    \end{overpic}
    \begin{overpic}[width=0.23\textwidth]{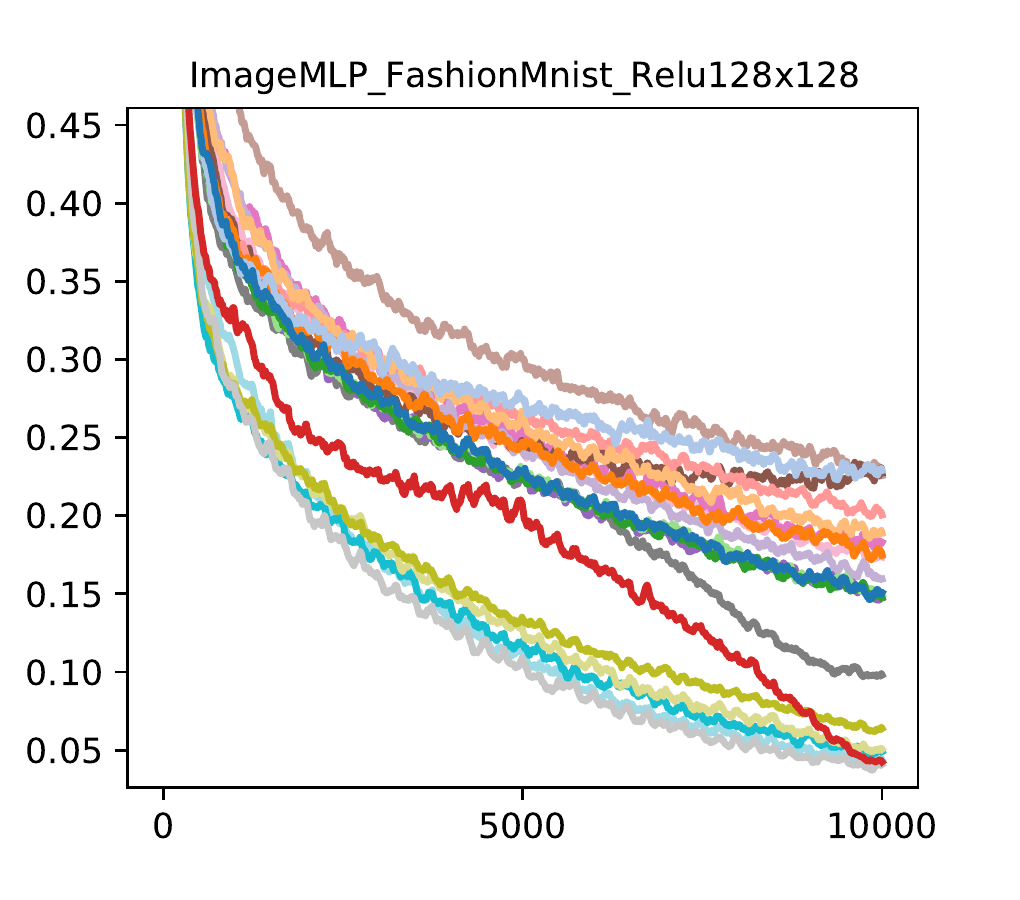}
    \end{overpic}
    \begin{overpic}[width=0.23\textwidth]{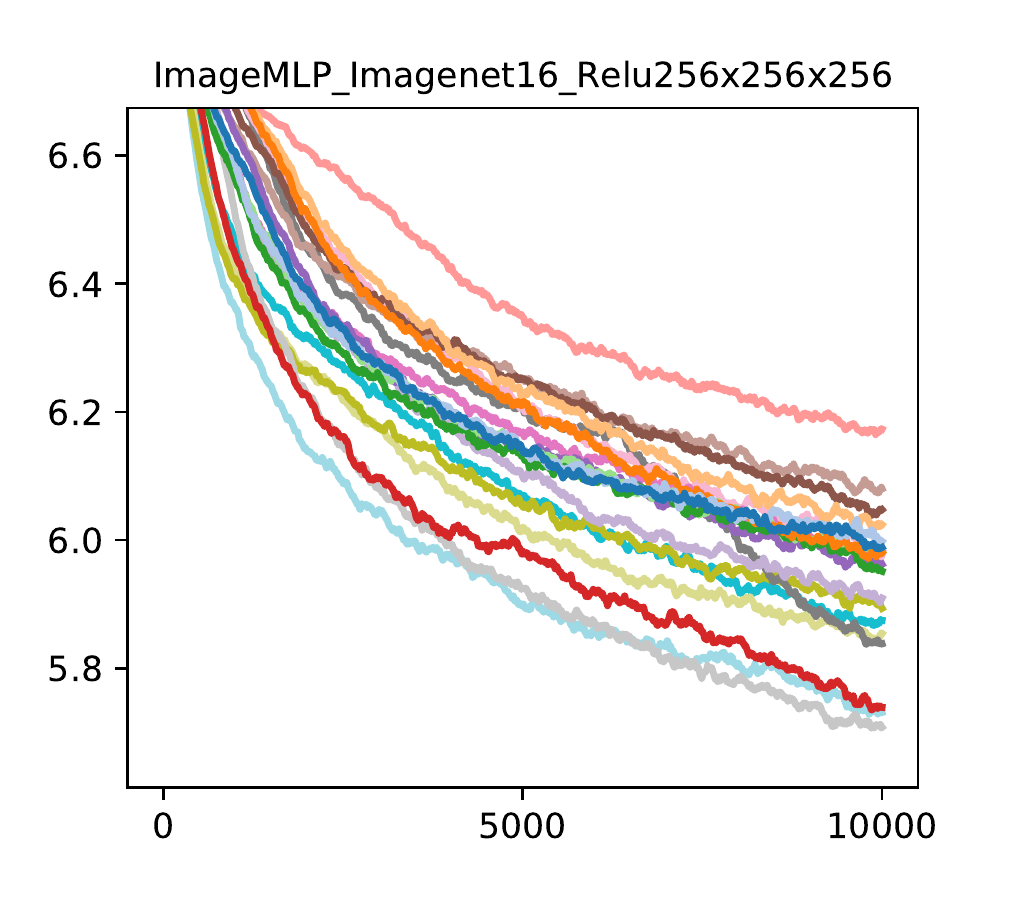}
    \end{overpic}
    \begin{overpic}[width=0.23\textwidth]{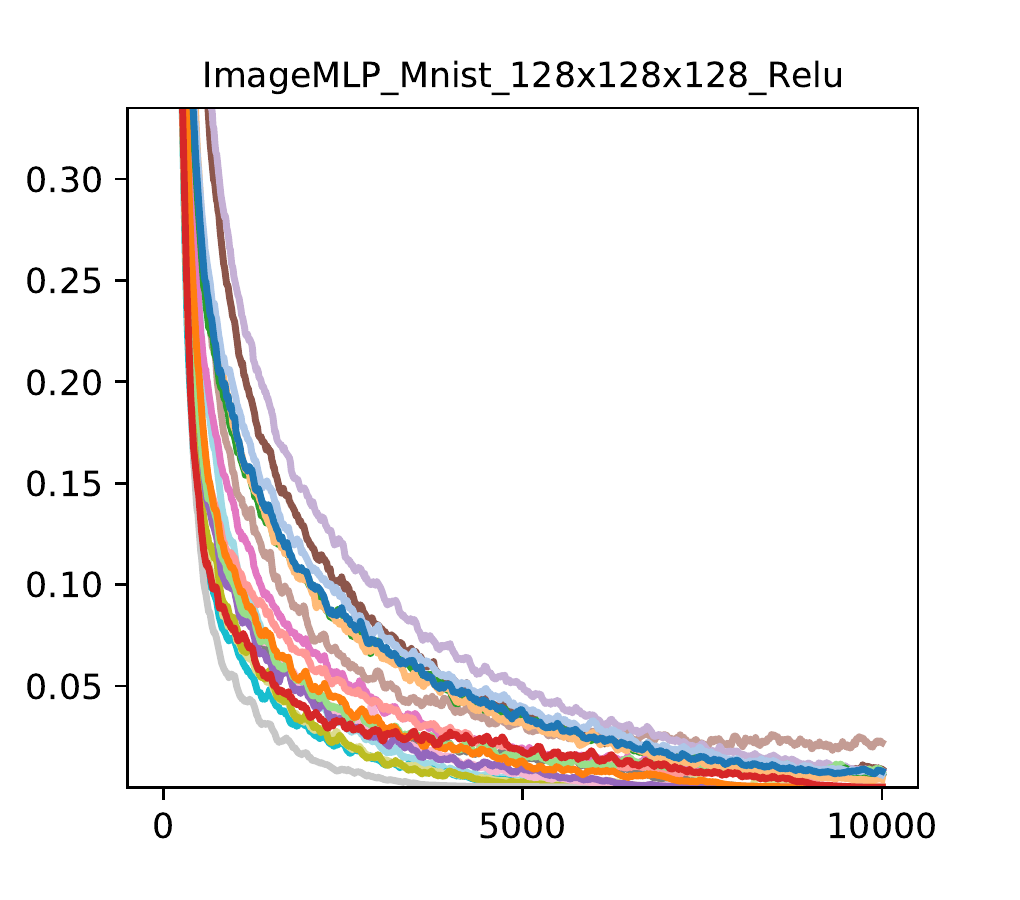}
    \end{overpic}
    }

    \makebox[\textwidth]{%
    \begin{overpic}[width=0.23\textwidth]{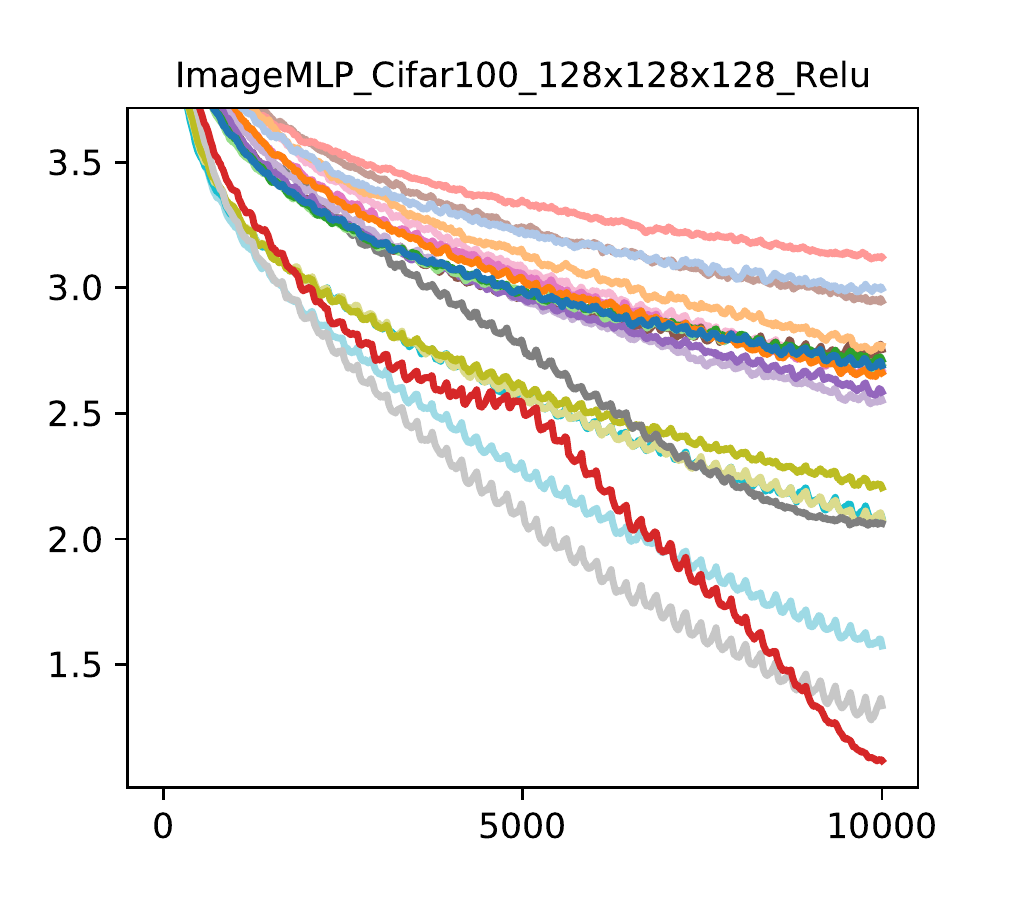}
    \end{overpic}
    \begin{overpic}[width=0.23\textwidth]{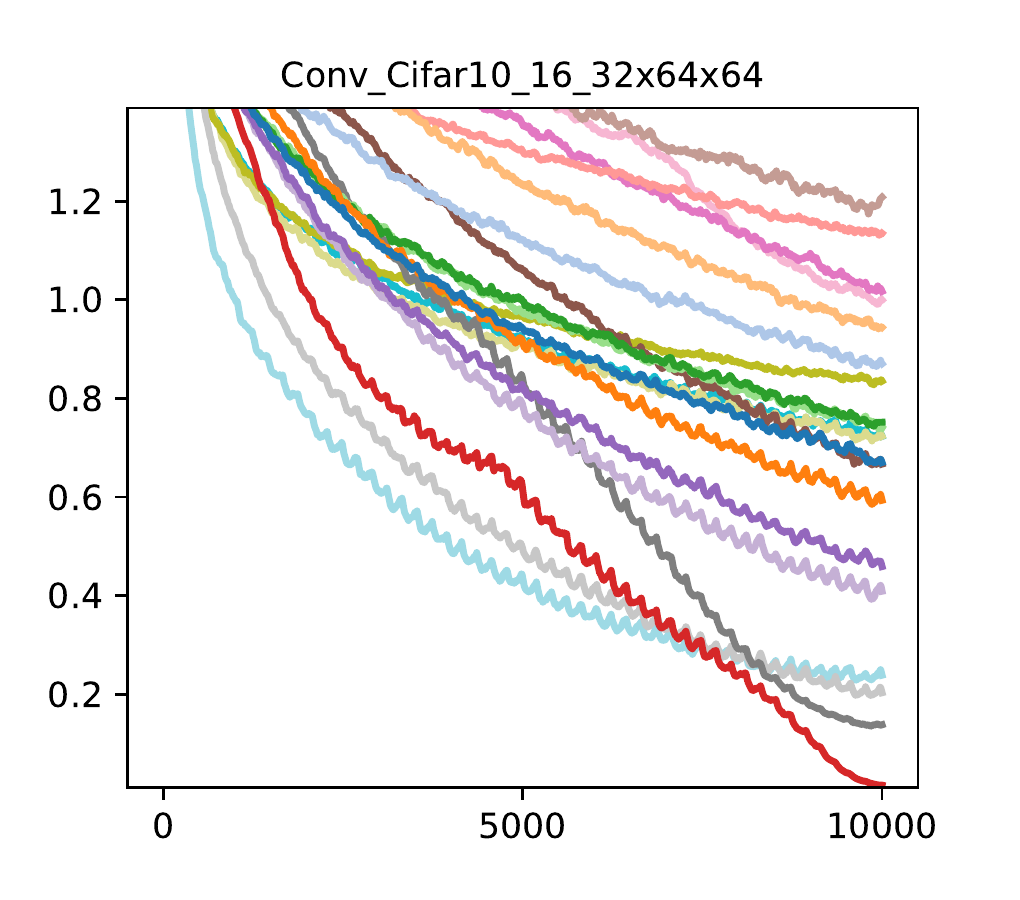}
    \end{overpic}
    \begin{overpic}[width=0.23\textwidth]{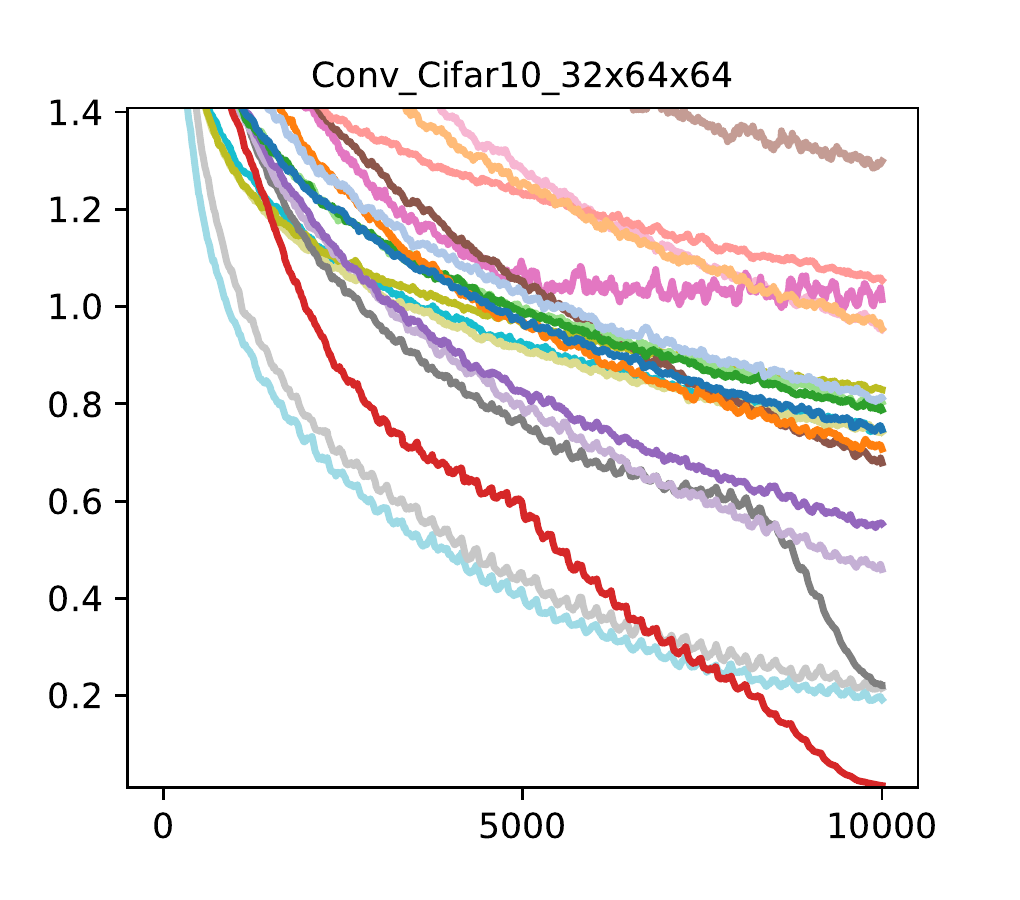}
    \end{overpic}
    \begin{overpic}[width=0.23\textwidth]{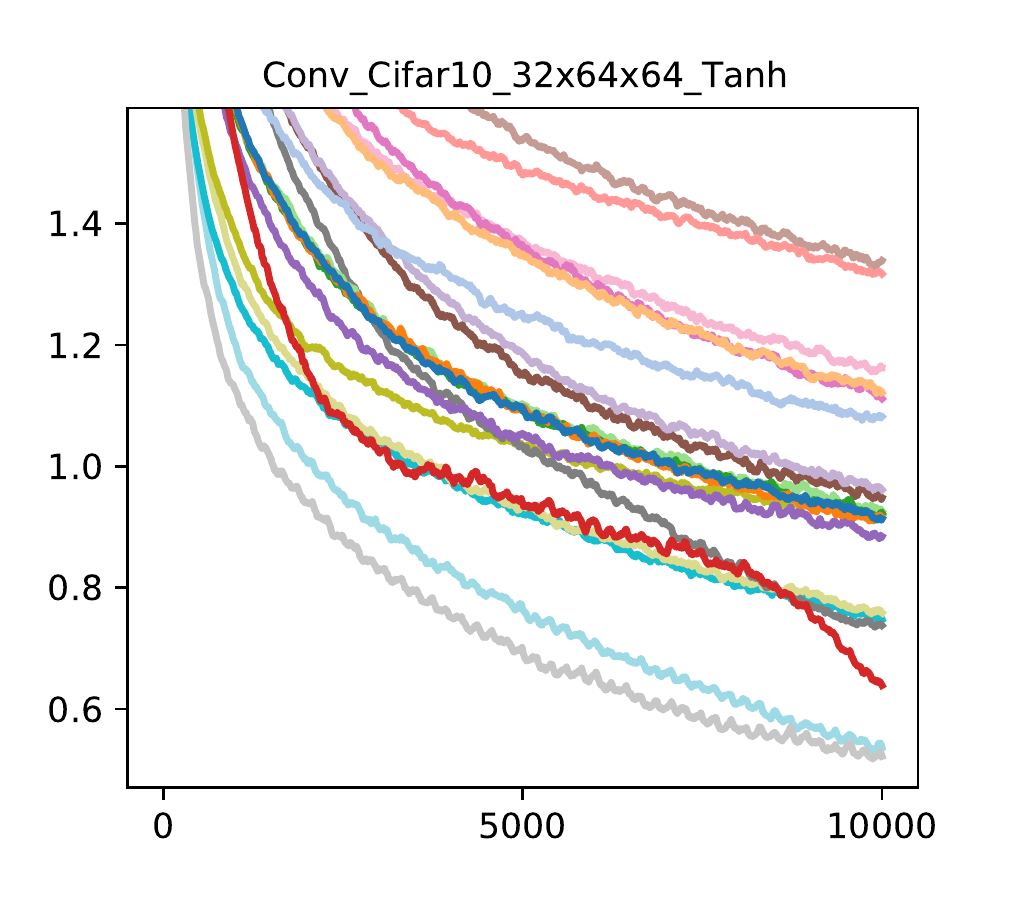}
    \end{overpic}
    \begin{overpic}[width=0.23\textwidth]{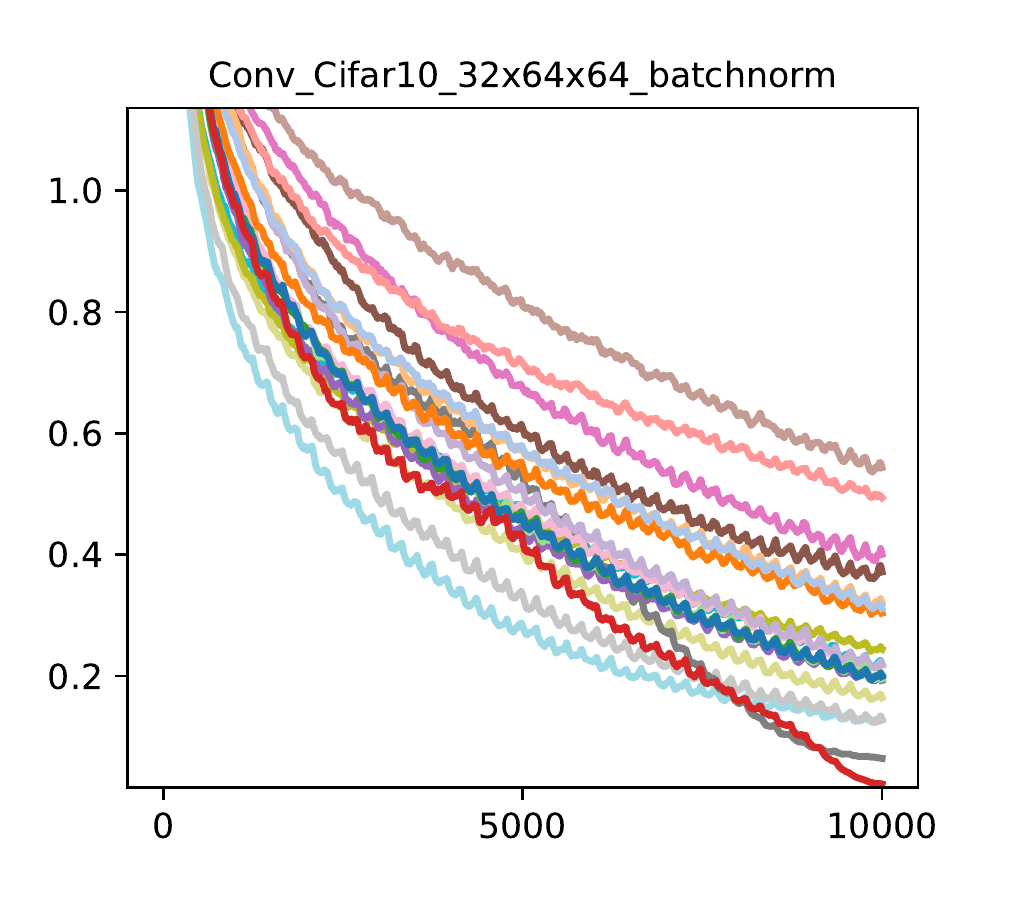}
    \end{overpic}
    }


    \makebox[\textwidth]{%
    \begin{overpic}[width=0.23\textwidth]{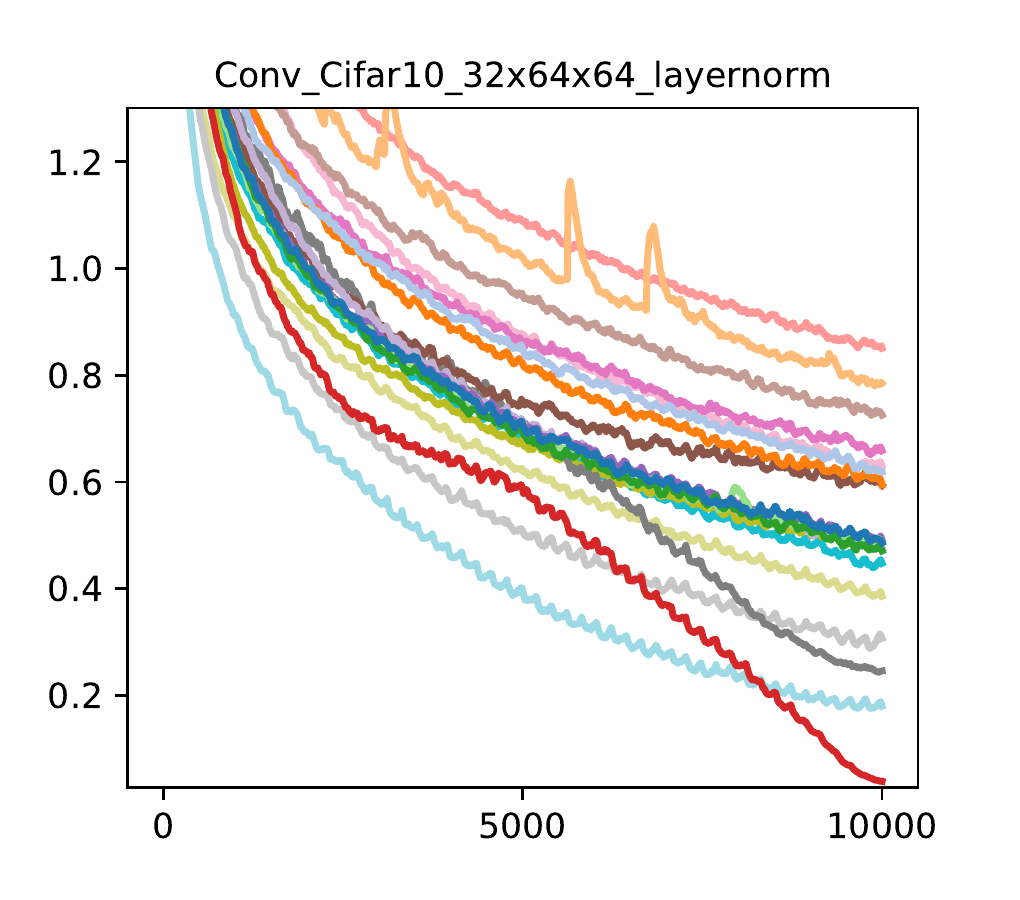}
    \end{overpic}
    \begin{overpic}[width=0.23\textwidth]{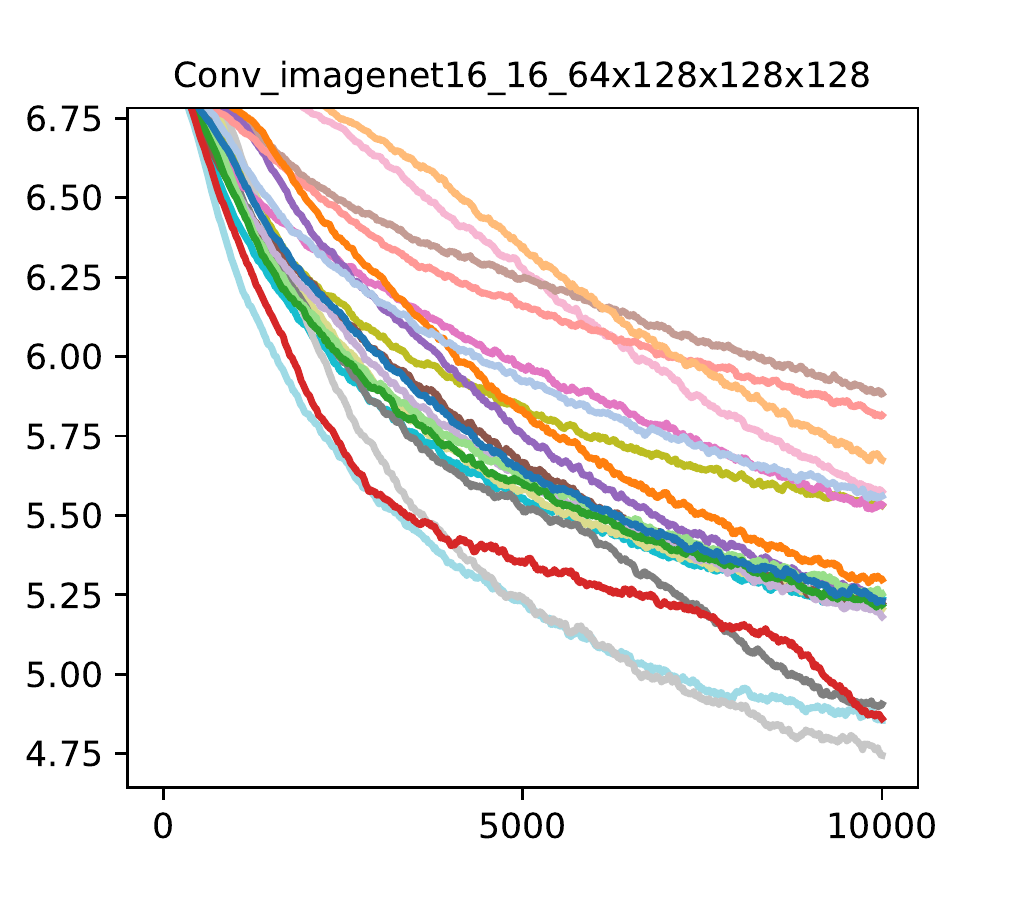}
    \end{overpic}
    \begin{overpic}[width=0.23\textwidth]{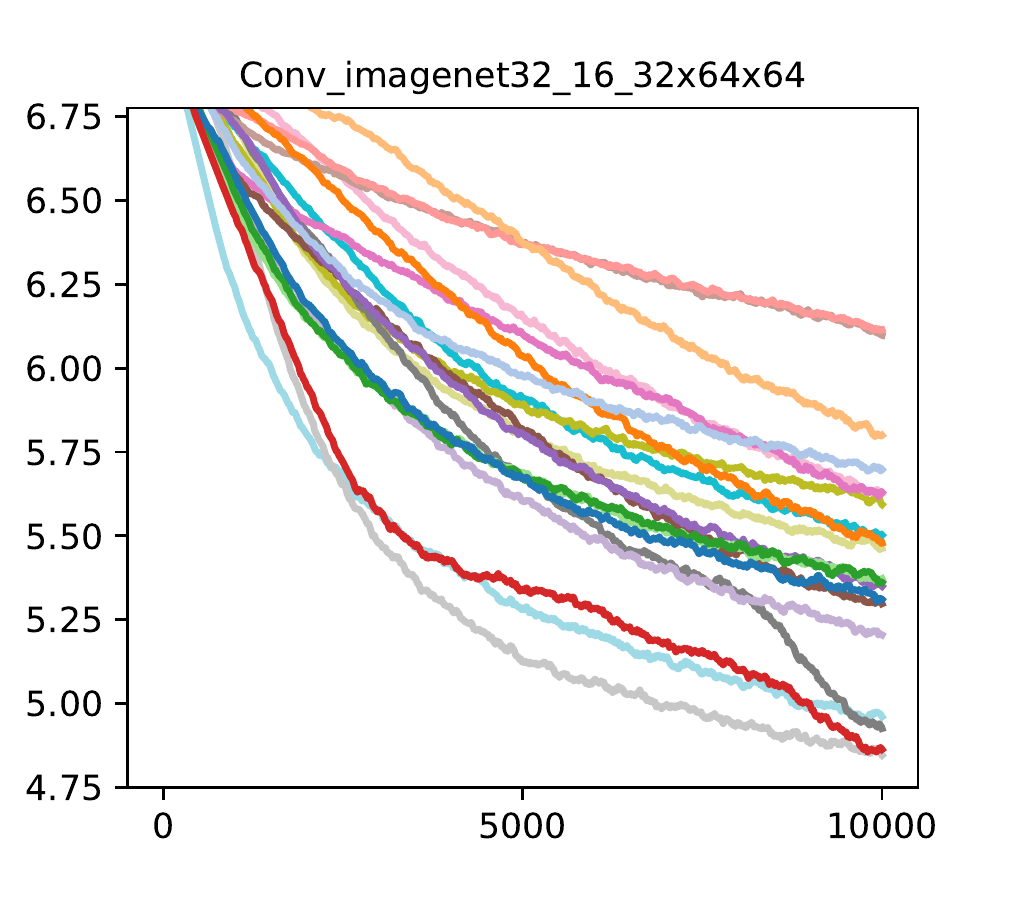}
    \end{overpic}
    \begin{overpic}[width=0.23\textwidth]{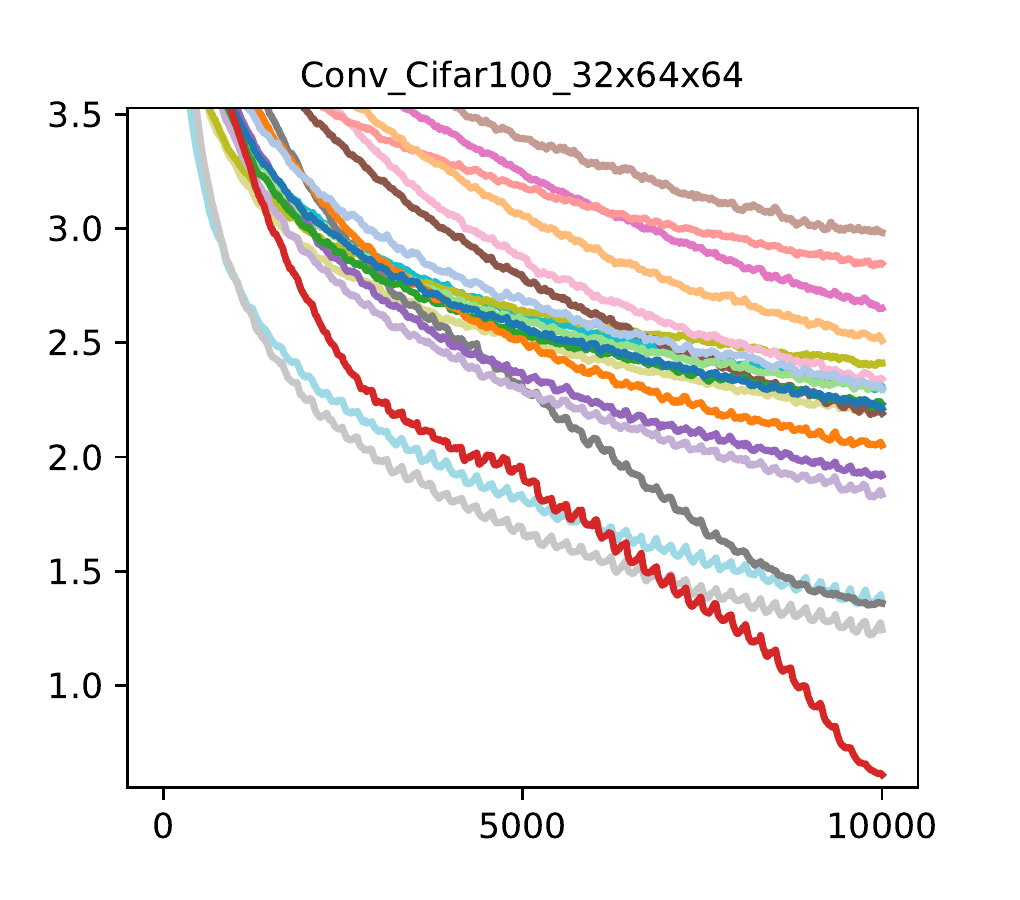}
    \end{overpic}
    \begin{overpic}[width=0.23\textwidth]{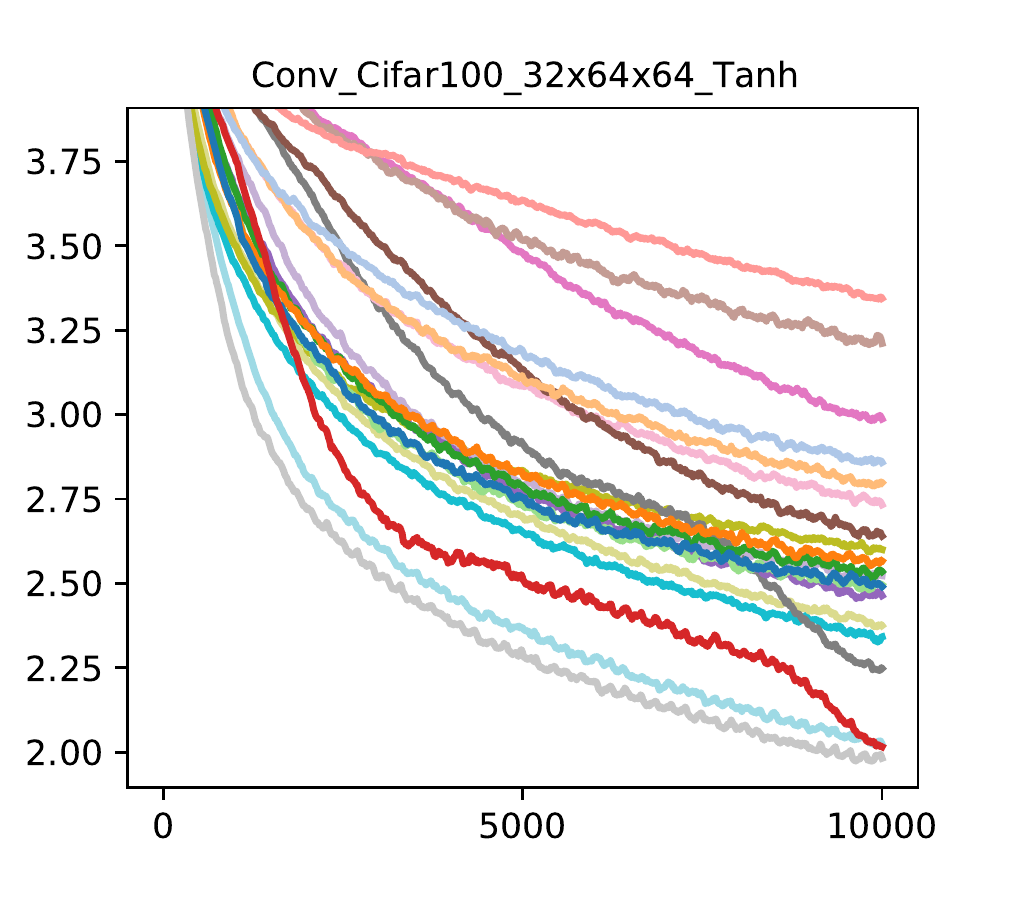}
    \end{overpic}
    }

%% file: app_page2.tex

    \makebox[\textwidth]{%
    \begin{overpic}[width=0.23\textwidth]{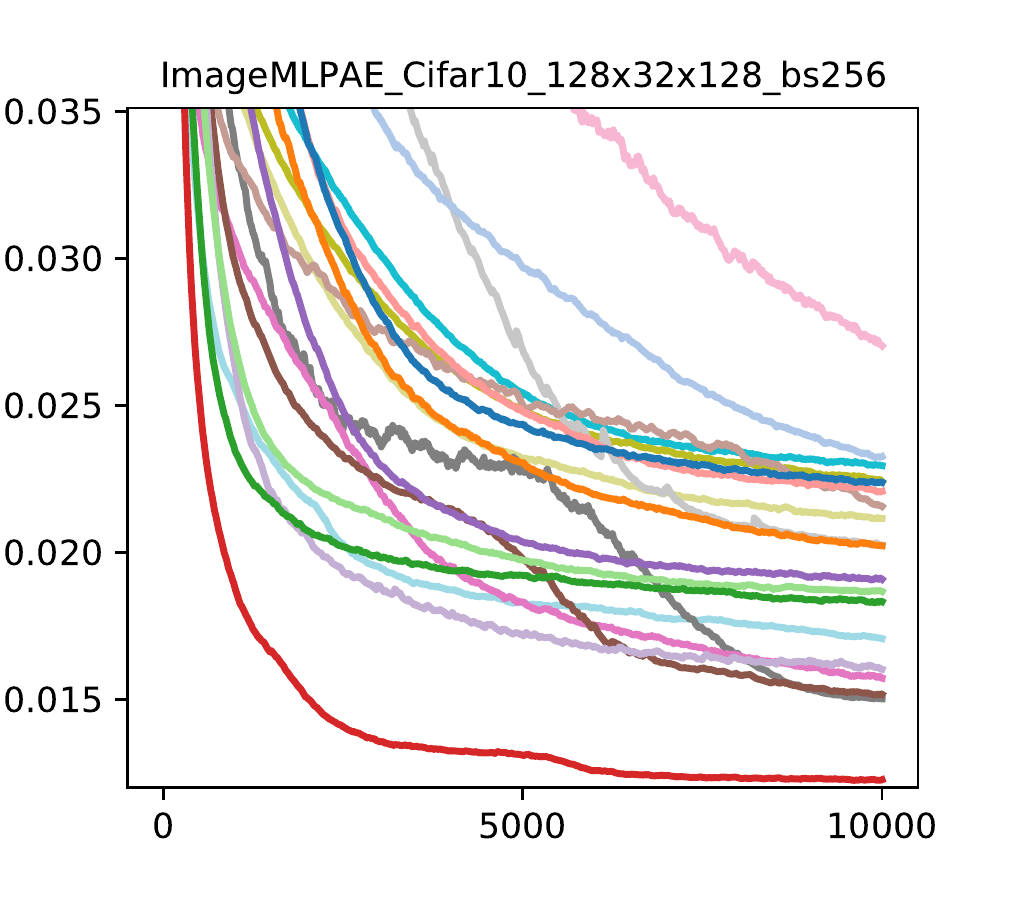}
    \end{overpic}
    \begin{overpic}[width=0.23\textwidth]{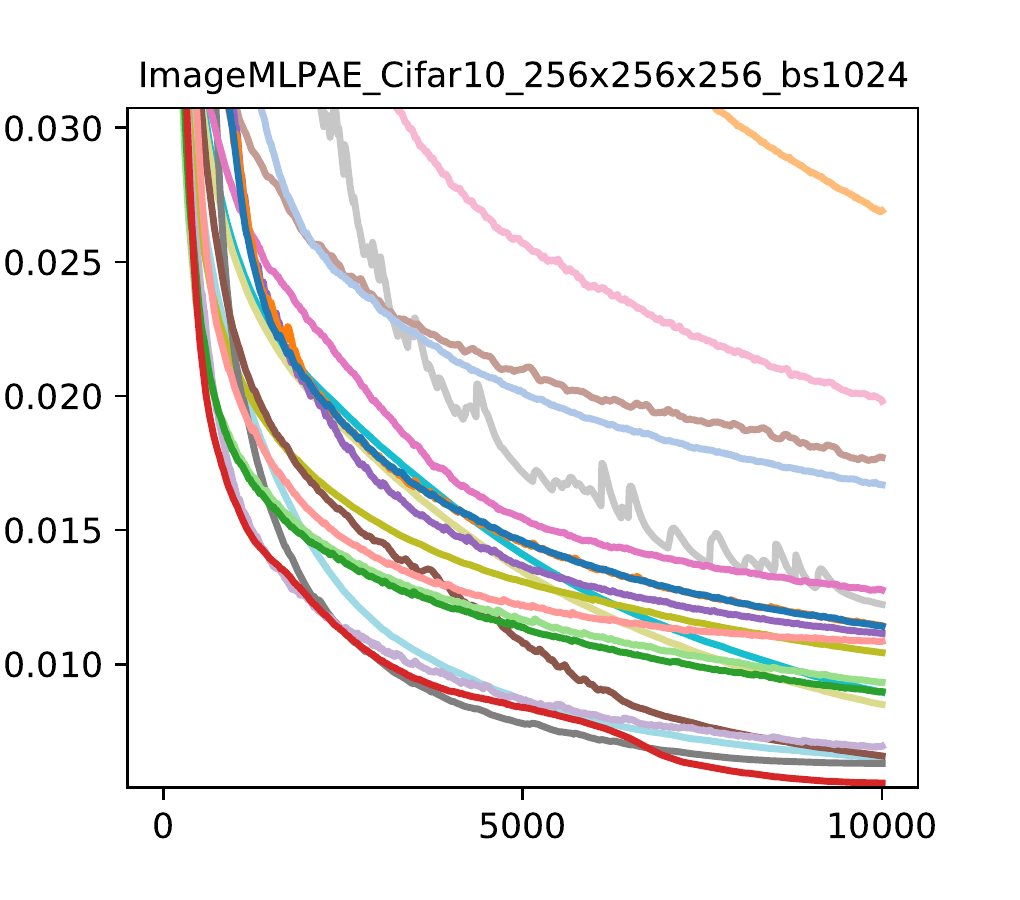}
    \end{overpic}
    \begin{overpic}[width=0.23\textwidth]{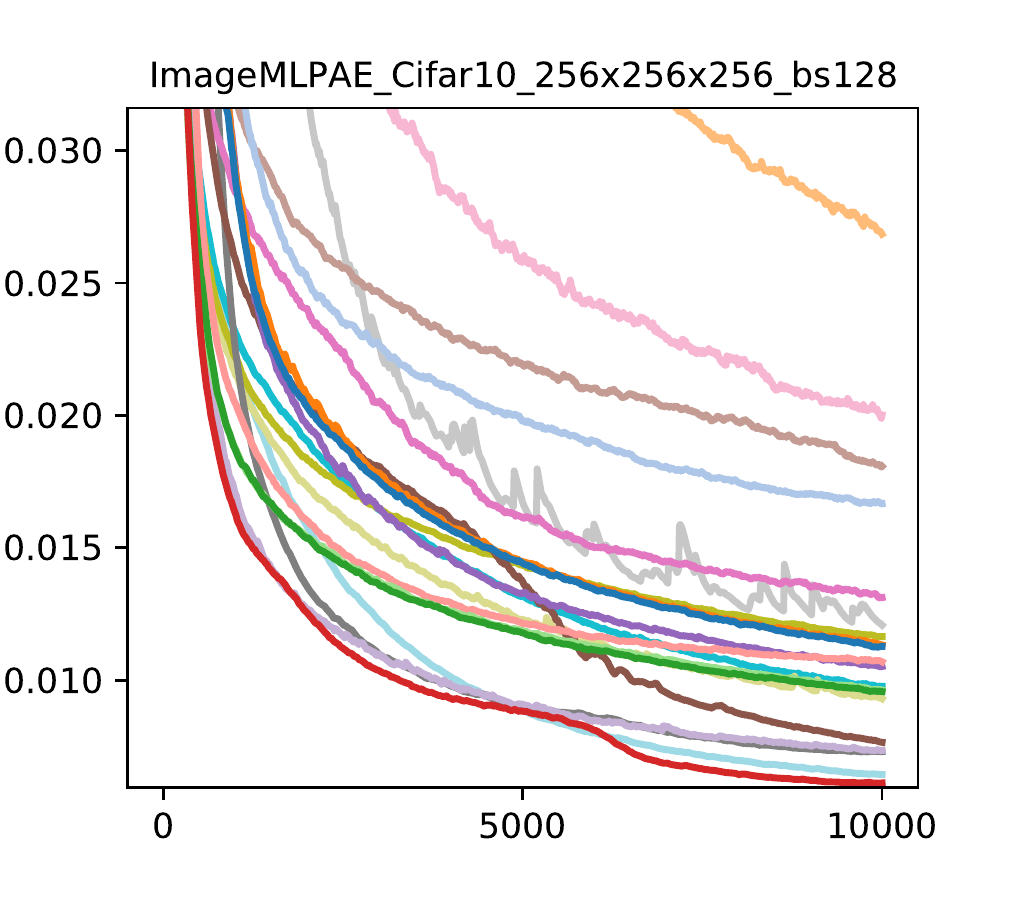}
    \end{overpic}
    \begin{overpic}[width=0.23\textwidth]{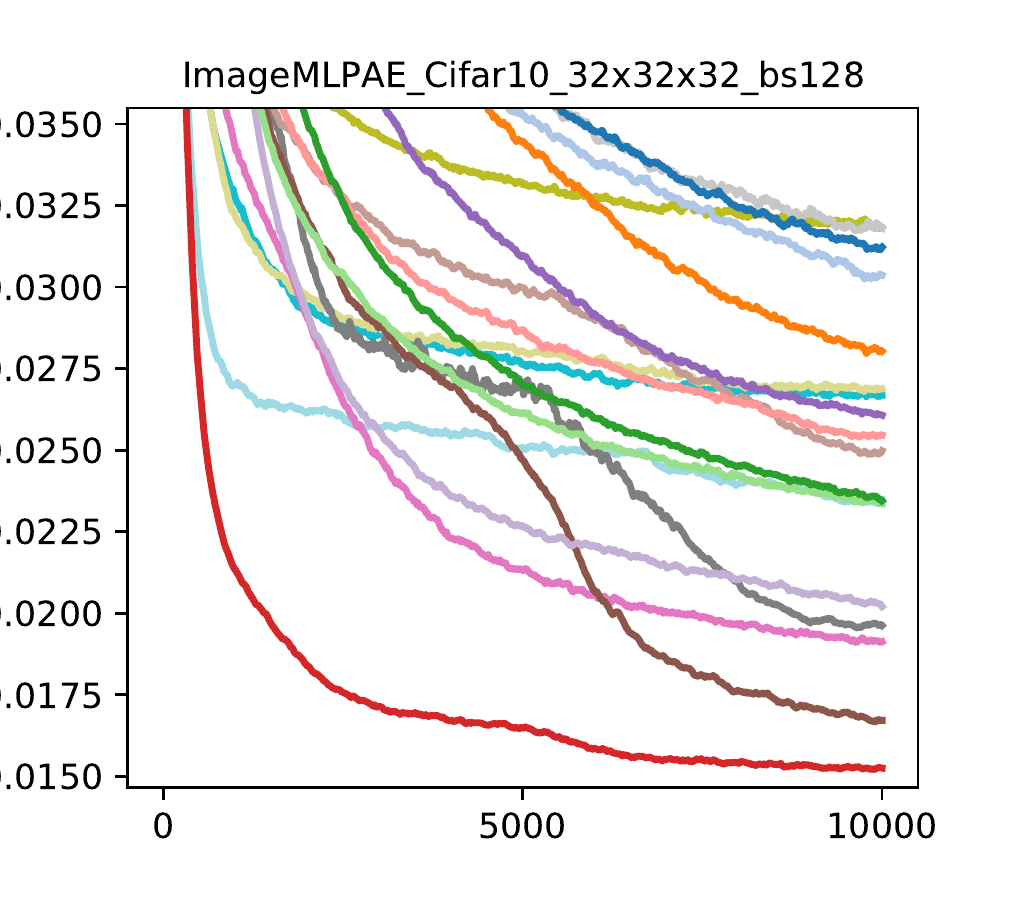}
    \end{overpic}
    \begin{overpic}[width=0.23\textwidth]{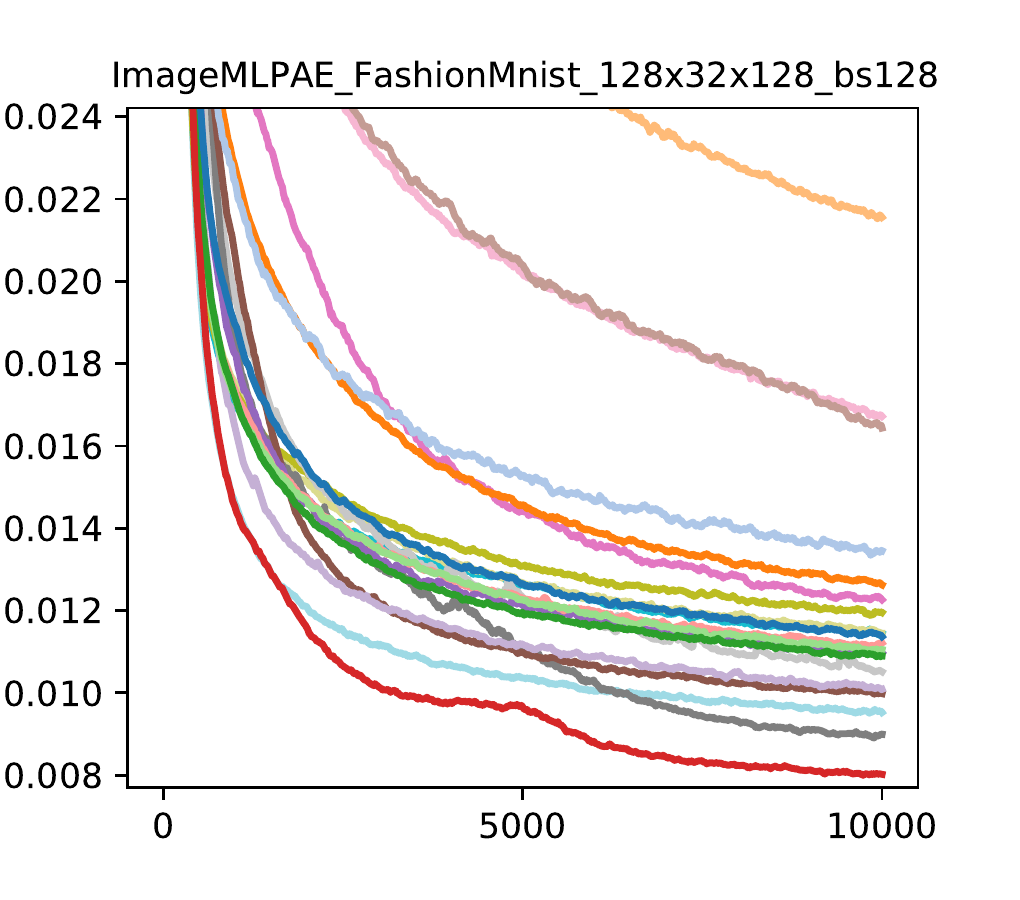}
    \end{overpic}
    }


    \makebox[\textwidth]{%
    \begin{overpic}[width=0.23\textwidth]{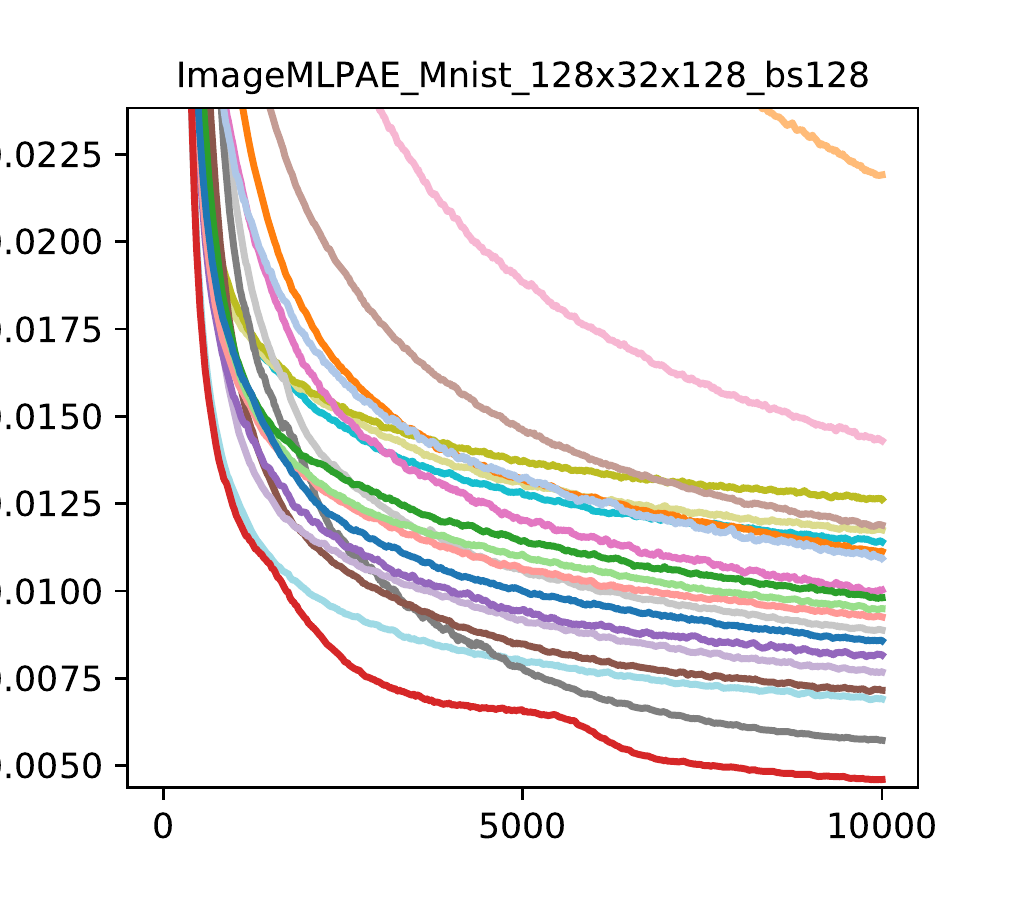}
    \end{overpic}
    \begin{overpic}[width=0.23\textwidth]{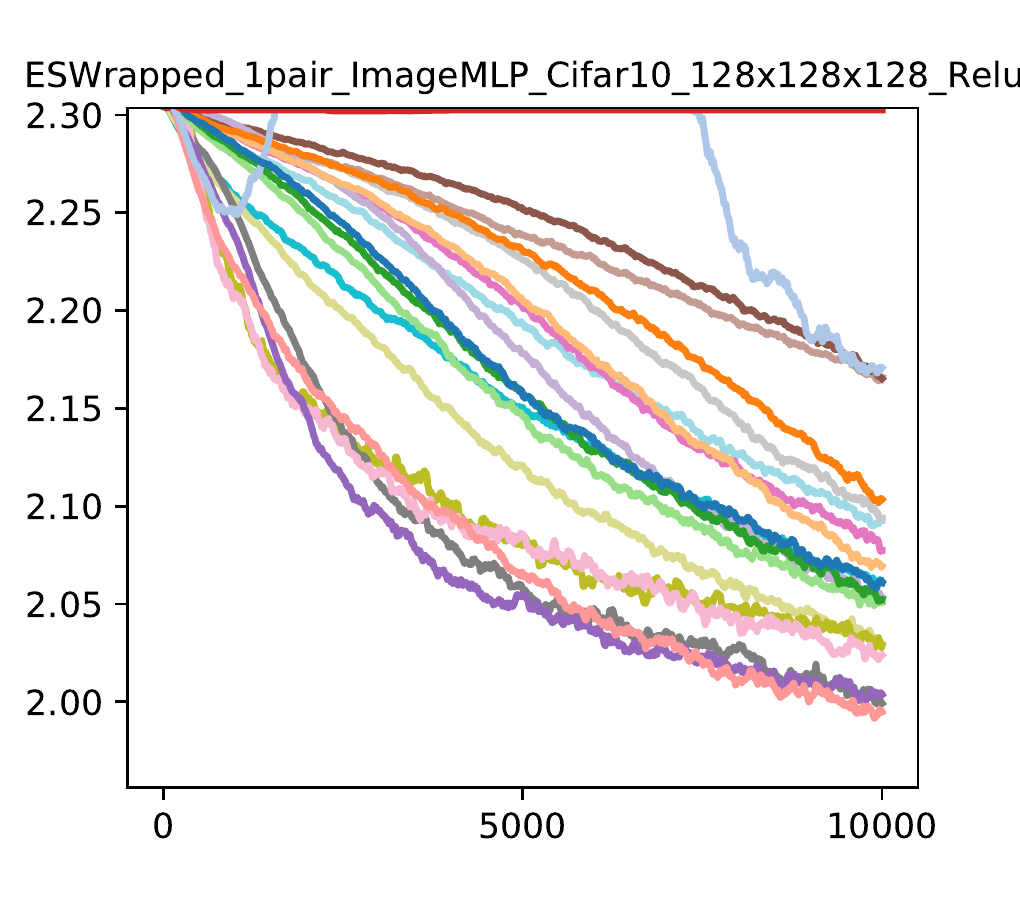}
    \end{overpic}
    \begin{overpic}[width=0.23\textwidth]{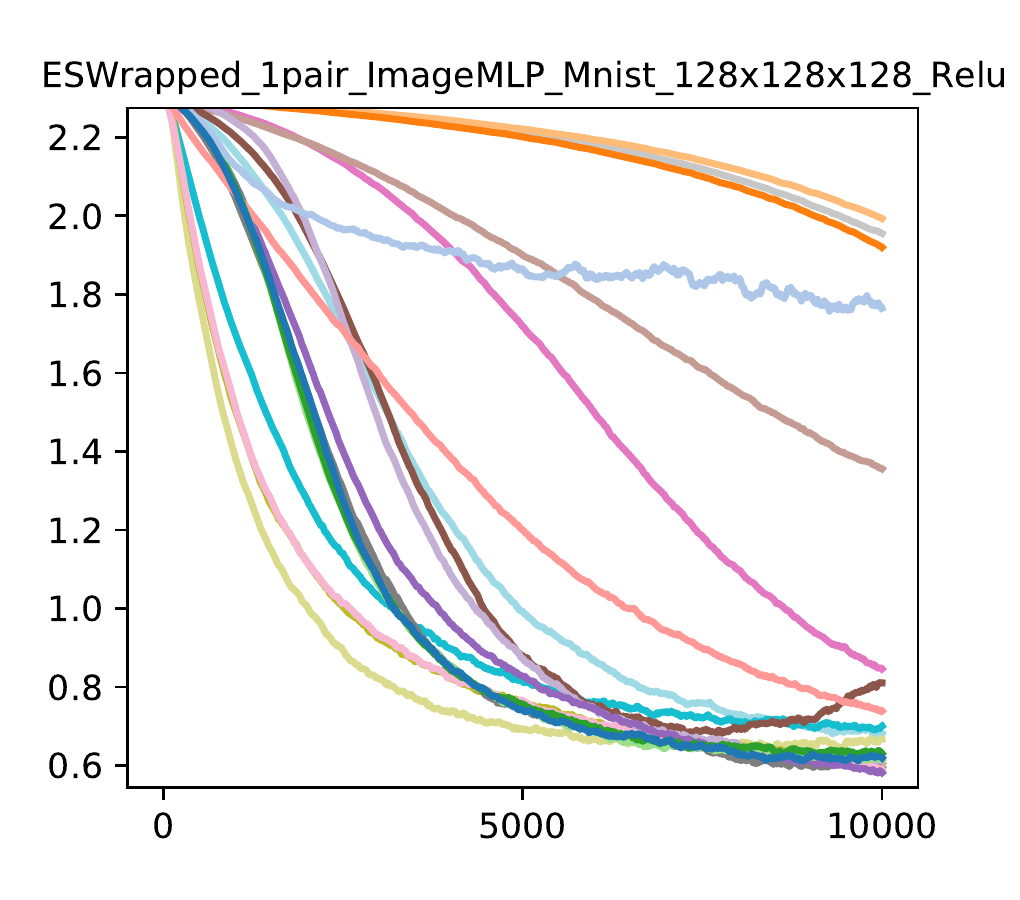}
    \end{overpic}
    \begin{overpic}[width=0.23\textwidth]{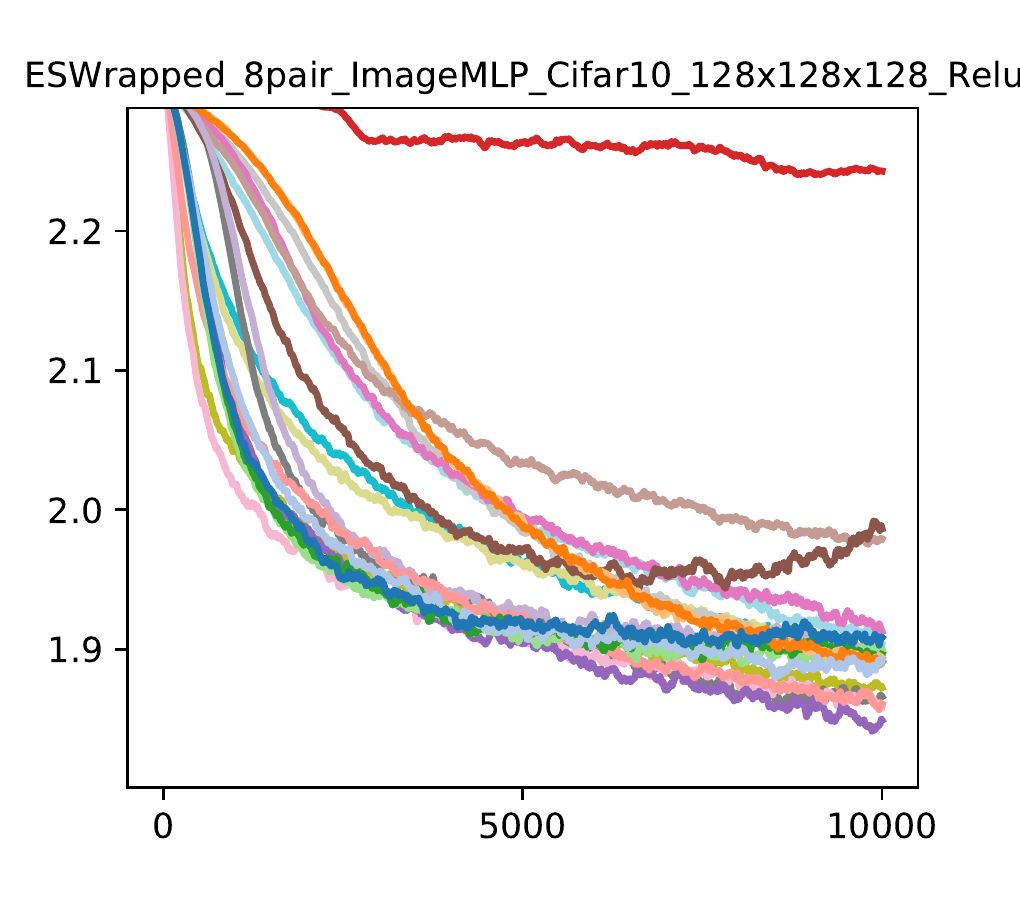}
    \end{overpic}
    \begin{overpic}[width=0.23\textwidth]{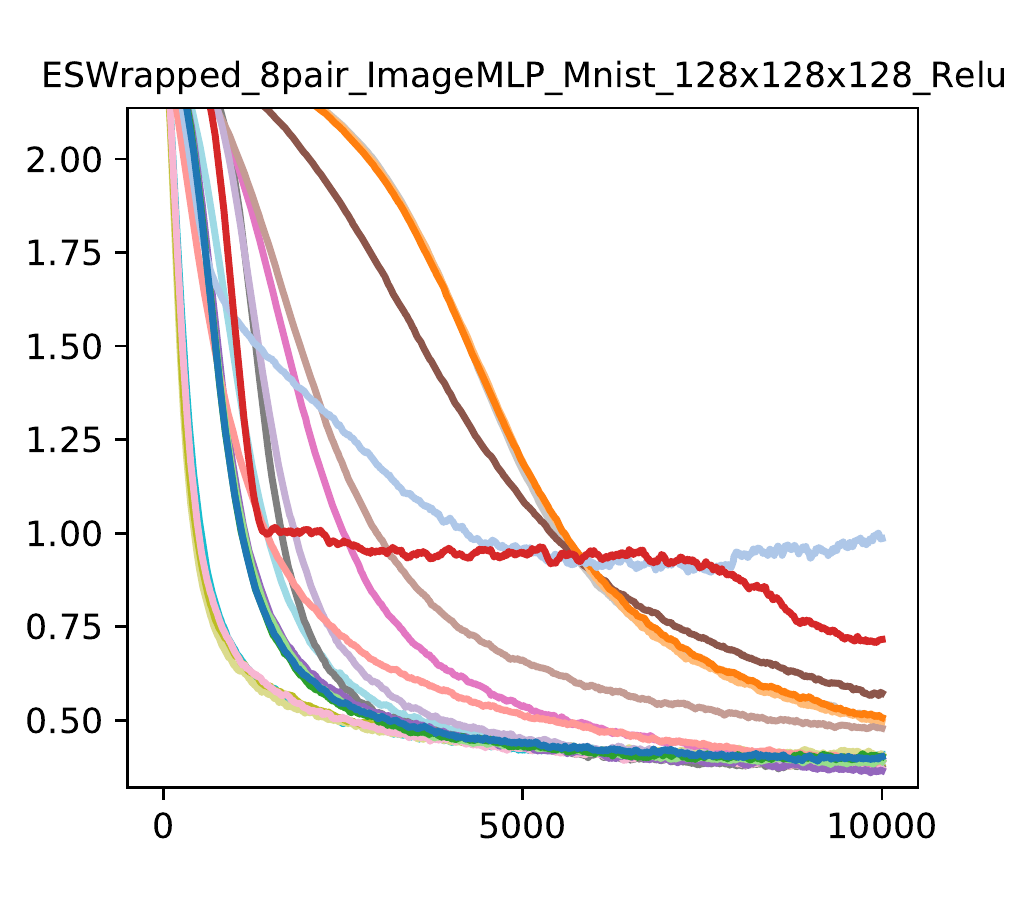}
    \end{overpic}
    }


    \makebox[\textwidth]{%
    \begin{overpic}[width=0.23\textwidth]{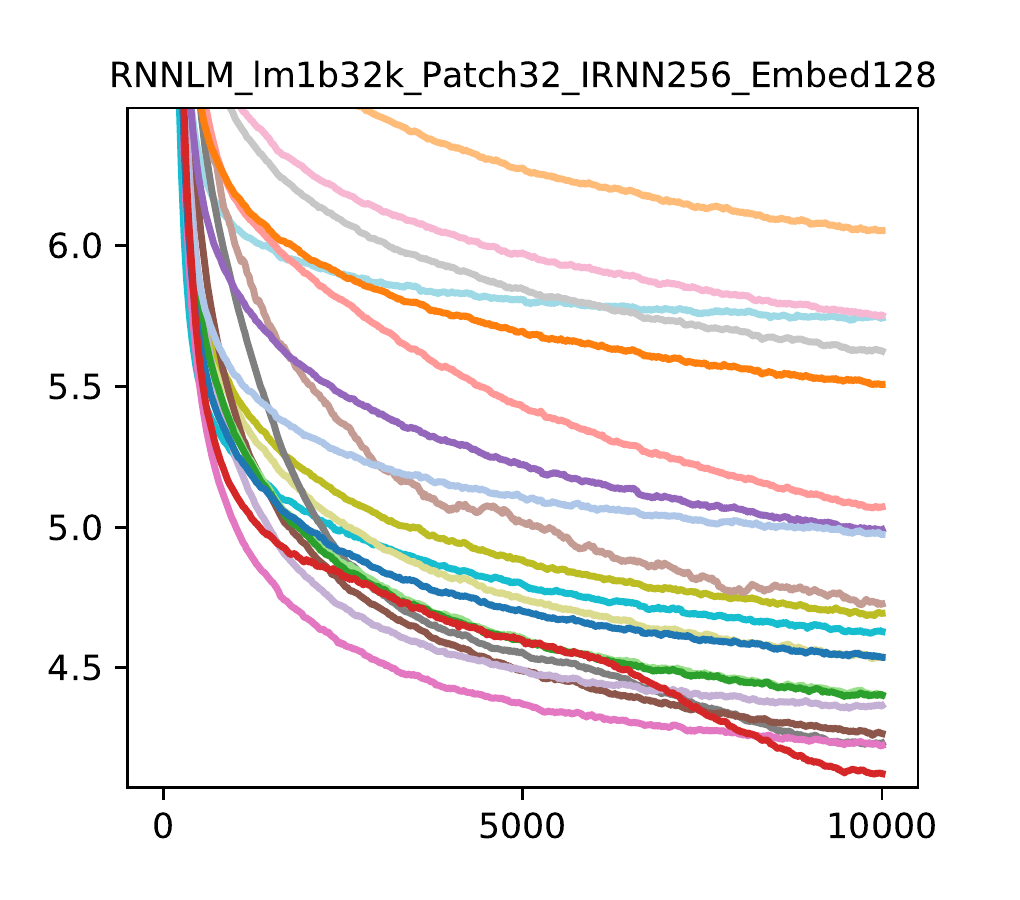}
    \end{overpic}
    \begin{overpic}[width=0.23\textwidth]{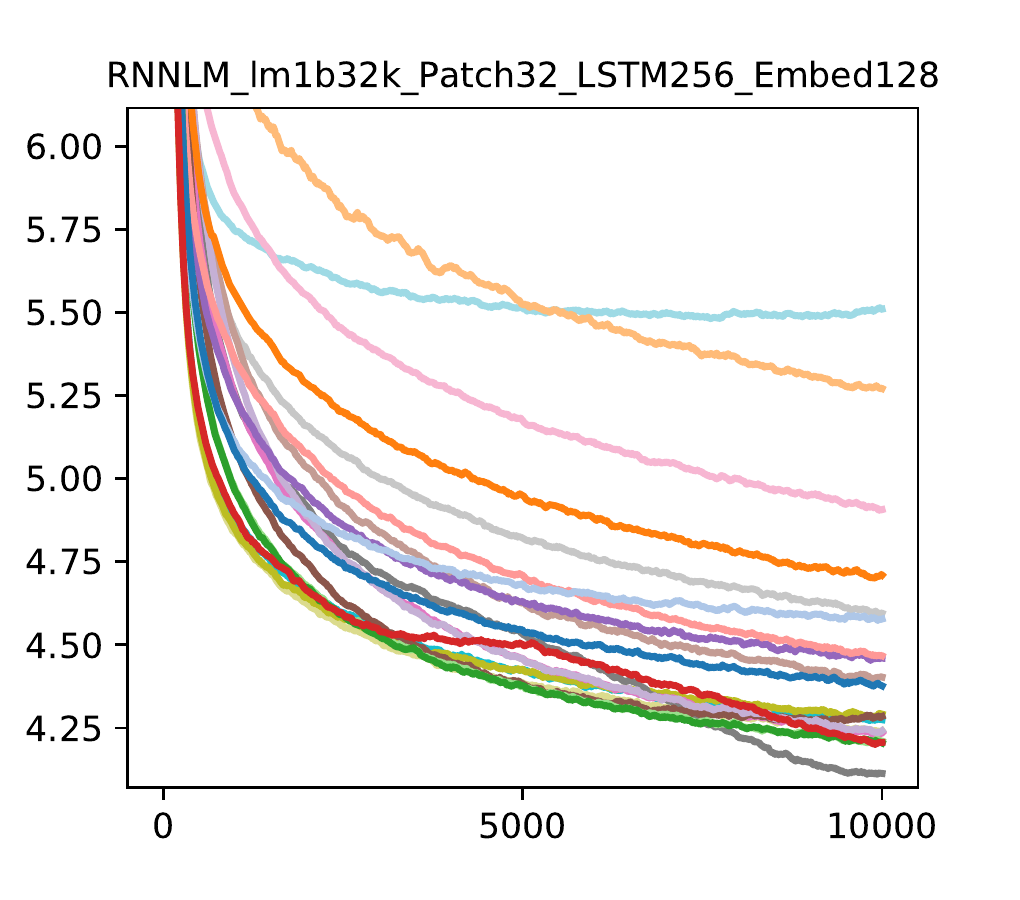}
    \end{overpic}
    \begin{overpic}[width=0.23\textwidth]{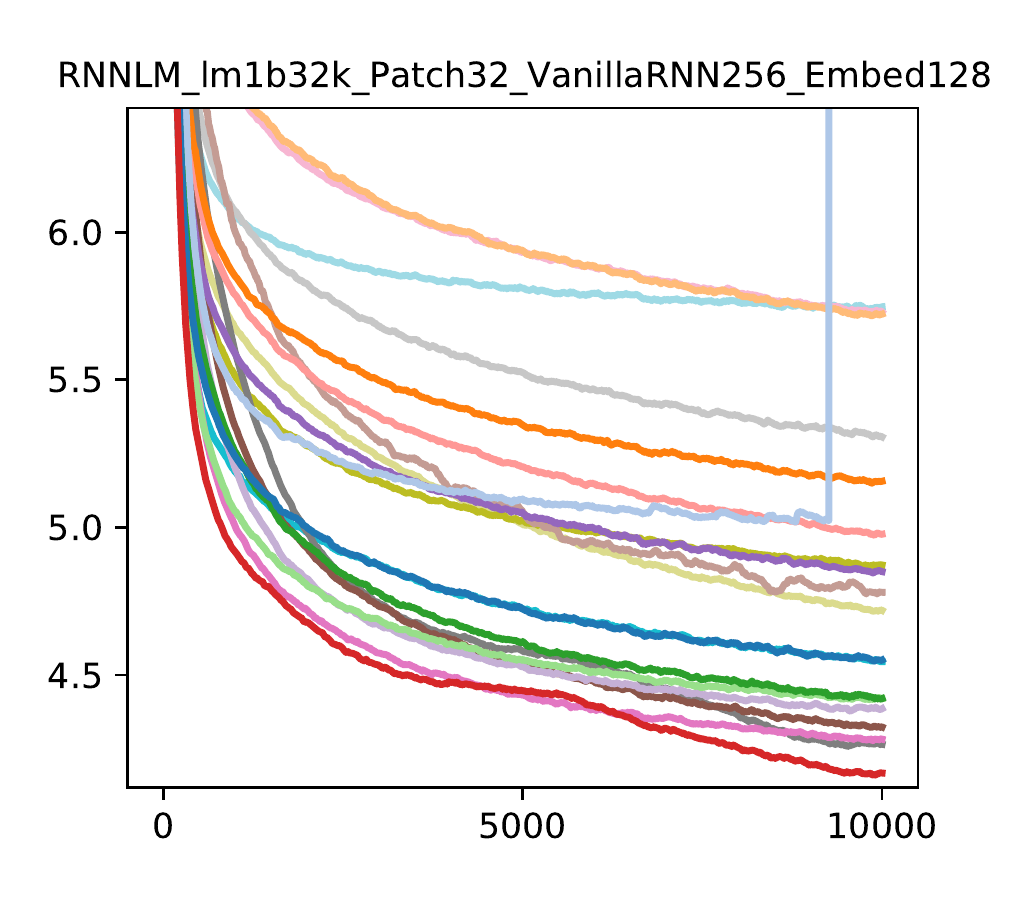}
    \end{overpic}
    \begin{overpic}[width=0.23\textwidth]{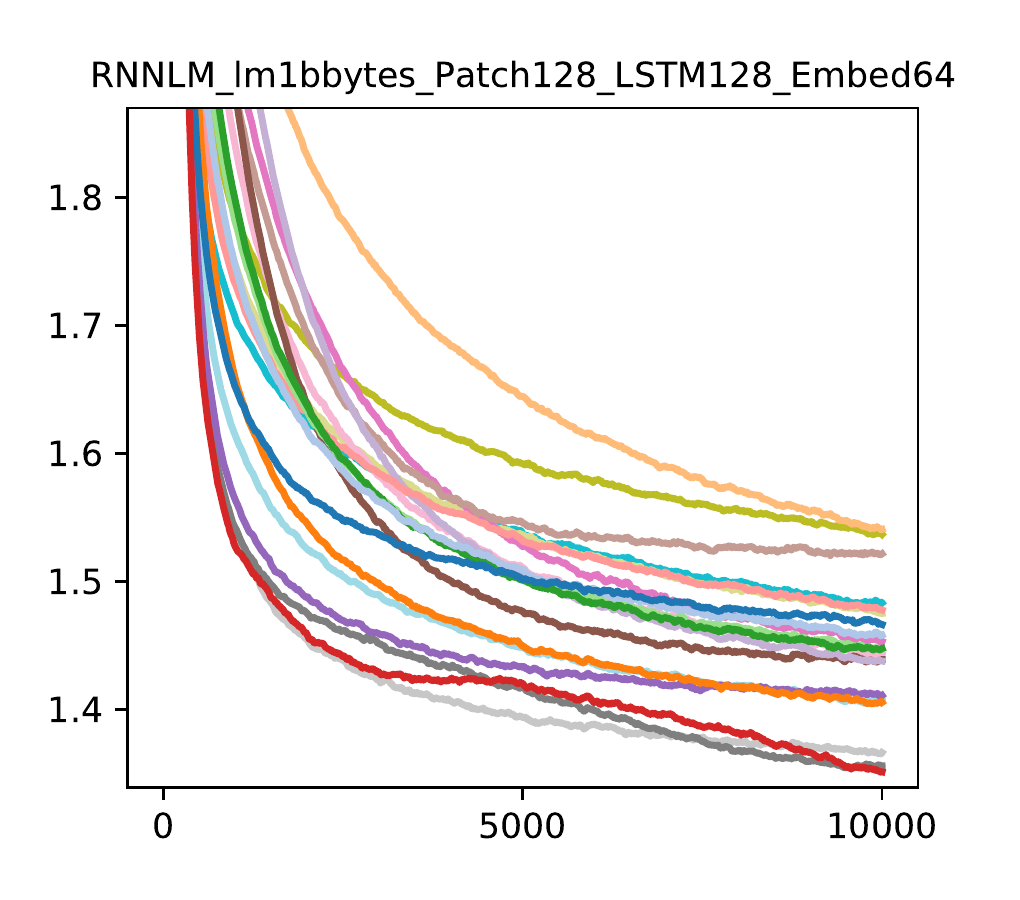}
    \end{overpic}
    \begin{overpic}[width=0.23\textwidth]{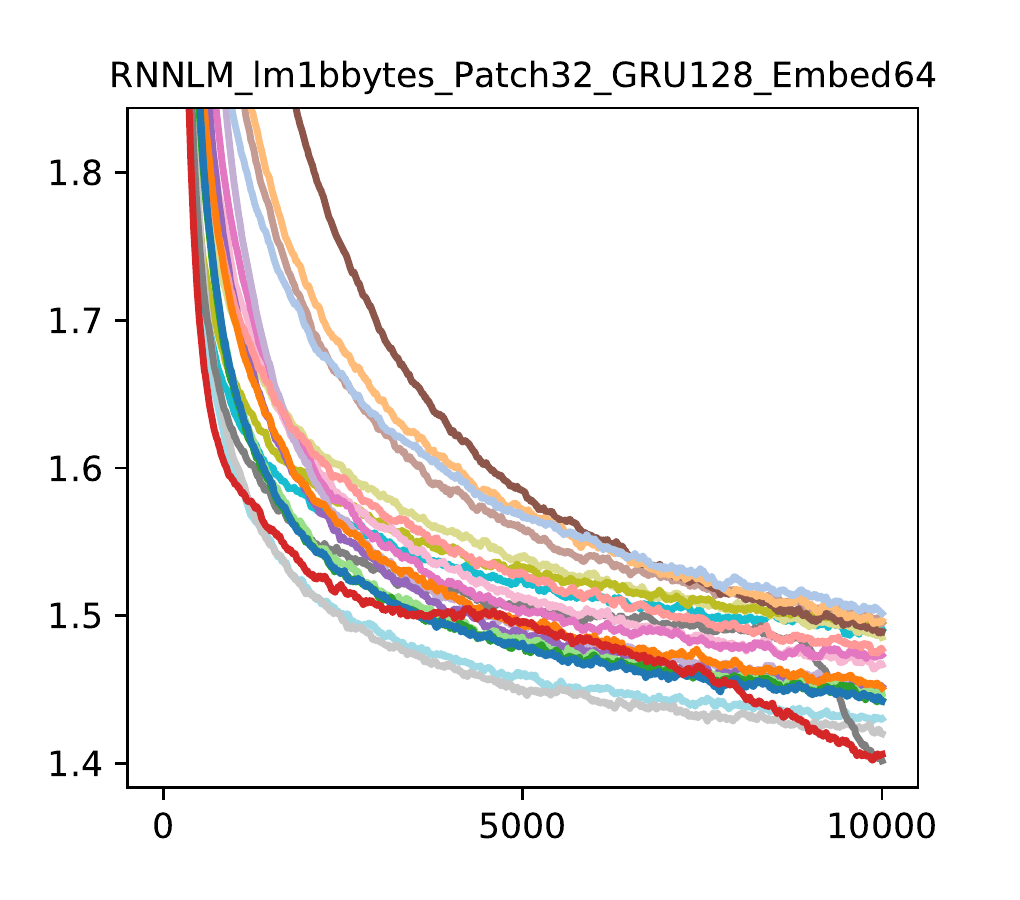}
    \end{overpic}
    }


    \makebox[\textwidth]{%
    \begin{overpic}[width=0.23\textwidth]{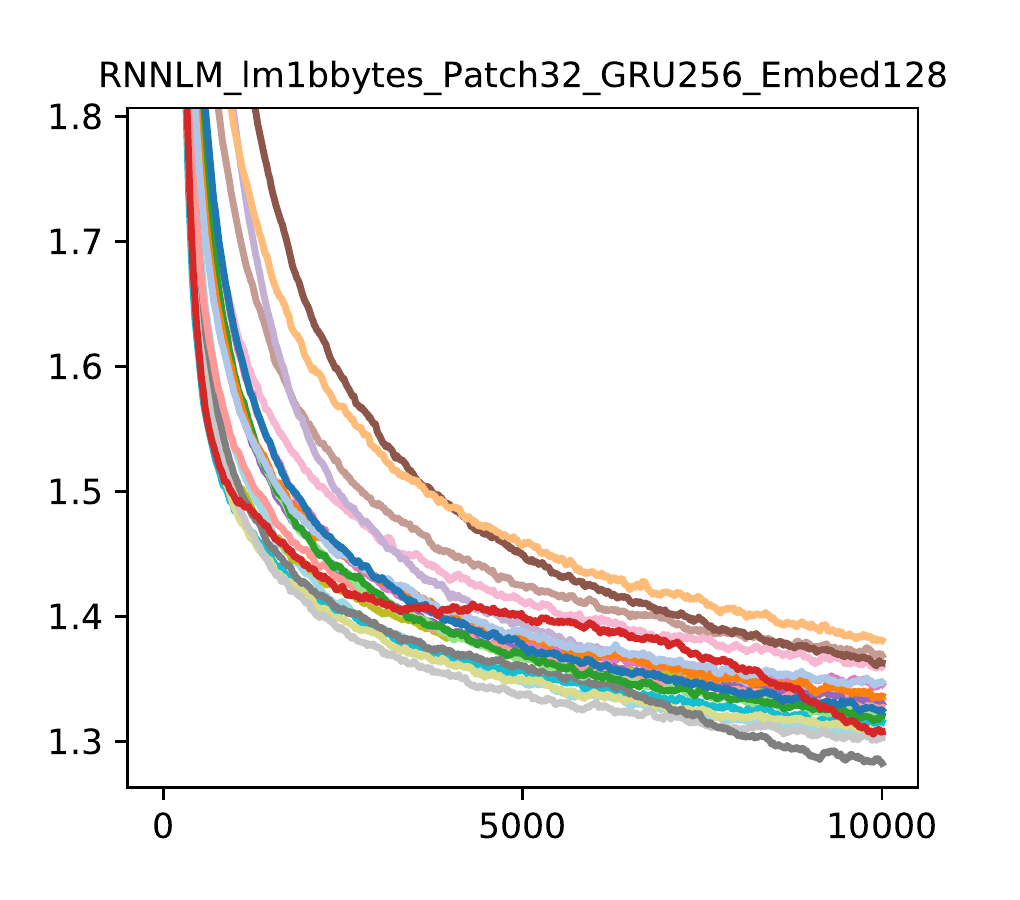}
    \end{overpic}
    \begin{overpic}[width=0.23\textwidth]{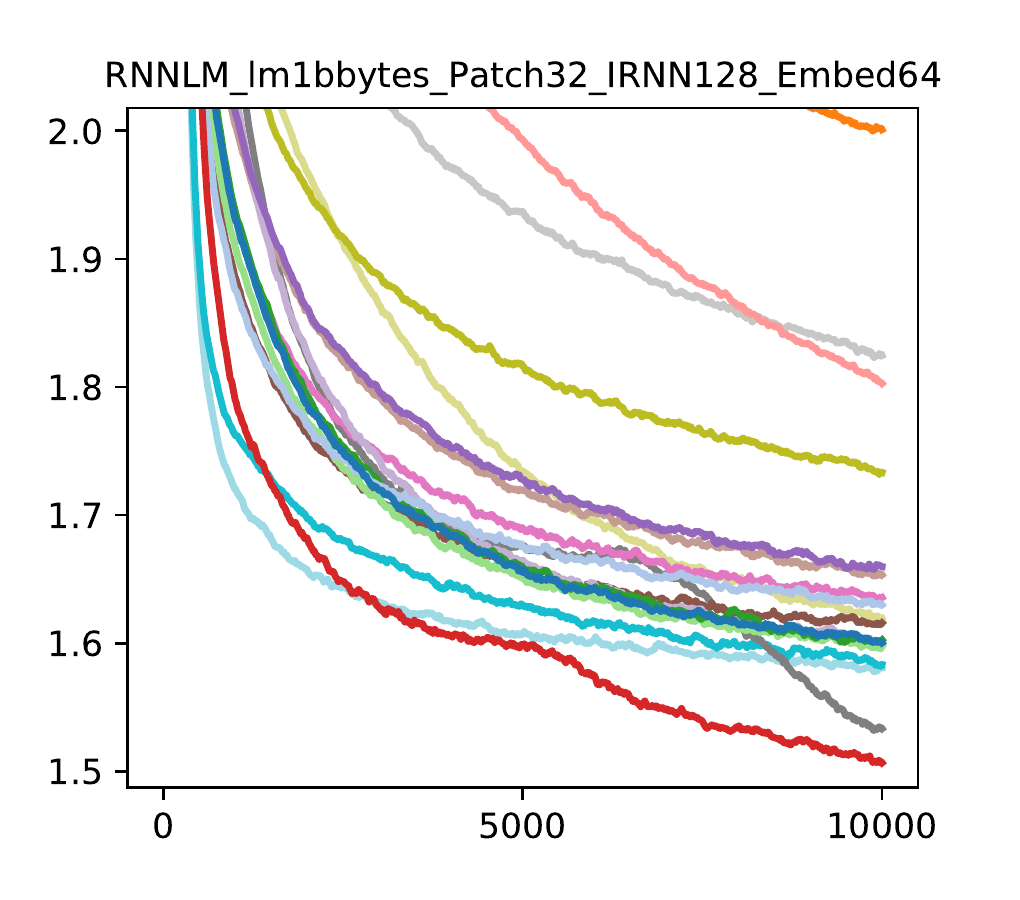}
    \end{overpic}
    \begin{overpic}[width=0.23\textwidth]{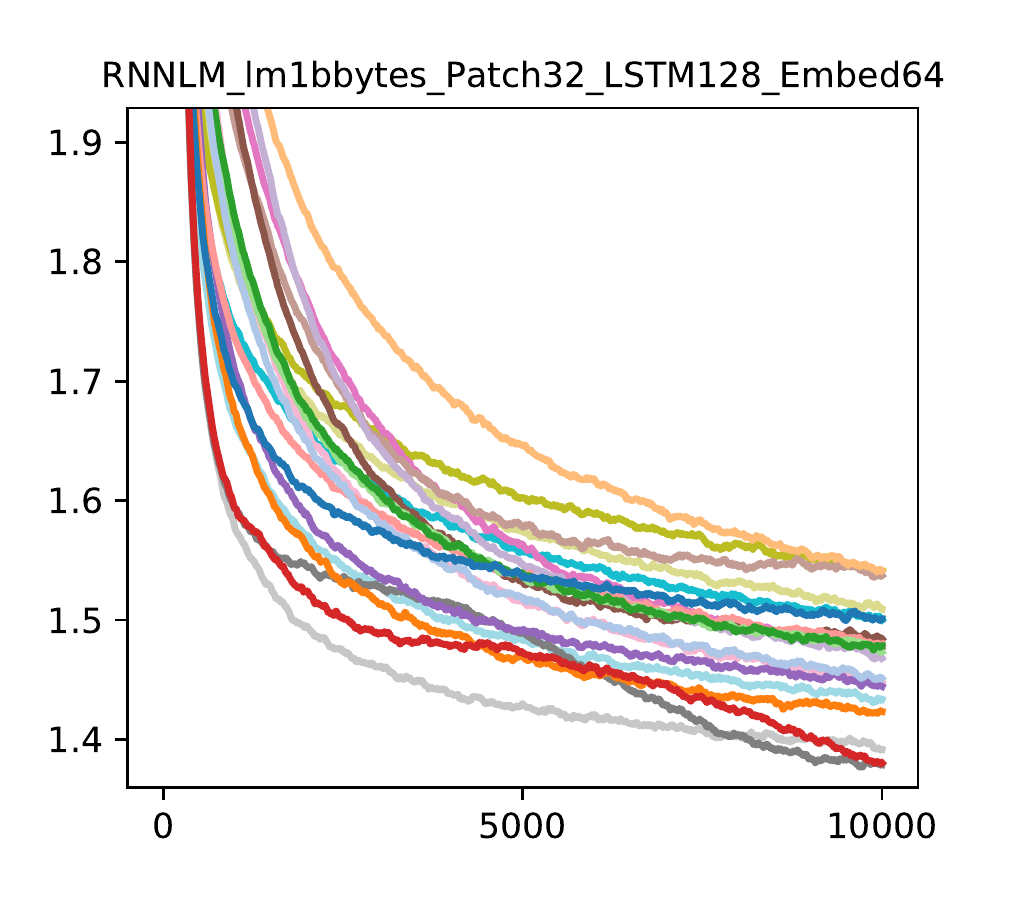}
    \end{overpic}
    \begin{overpic}[width=0.23\textwidth]{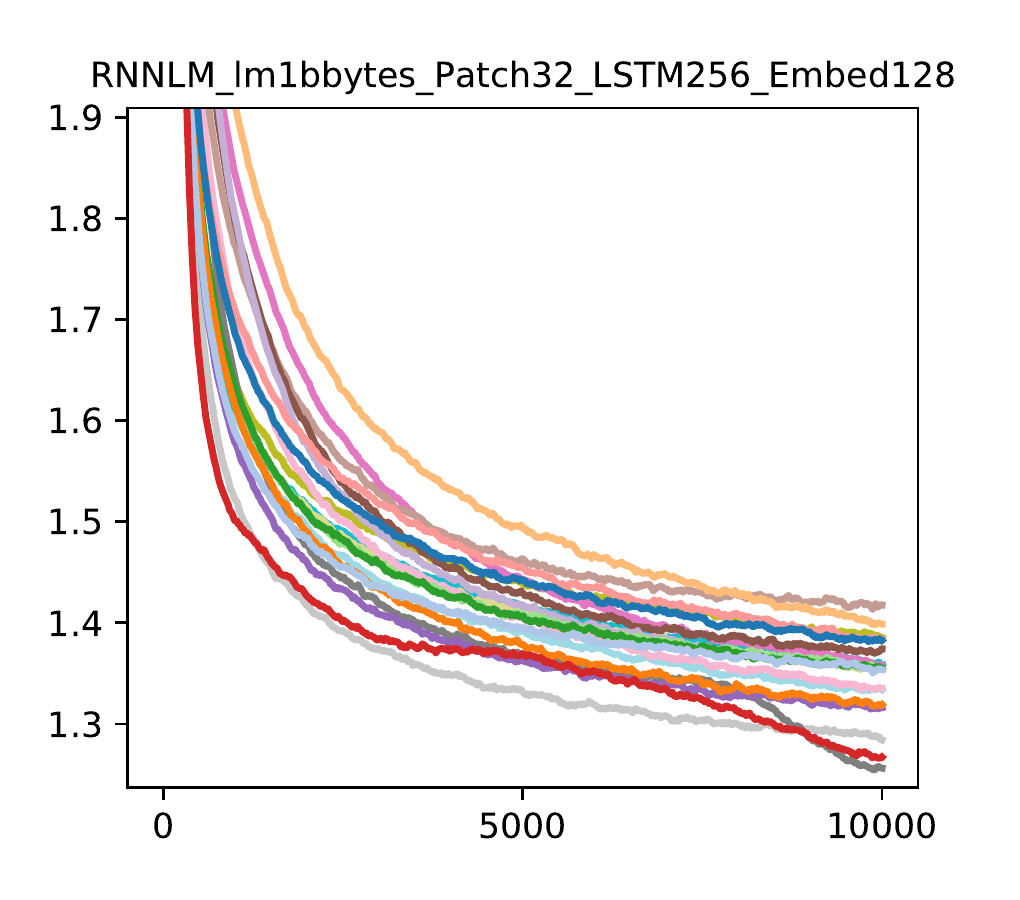}
    \end{overpic}
    \begin{overpic}[width=0.23\textwidth]{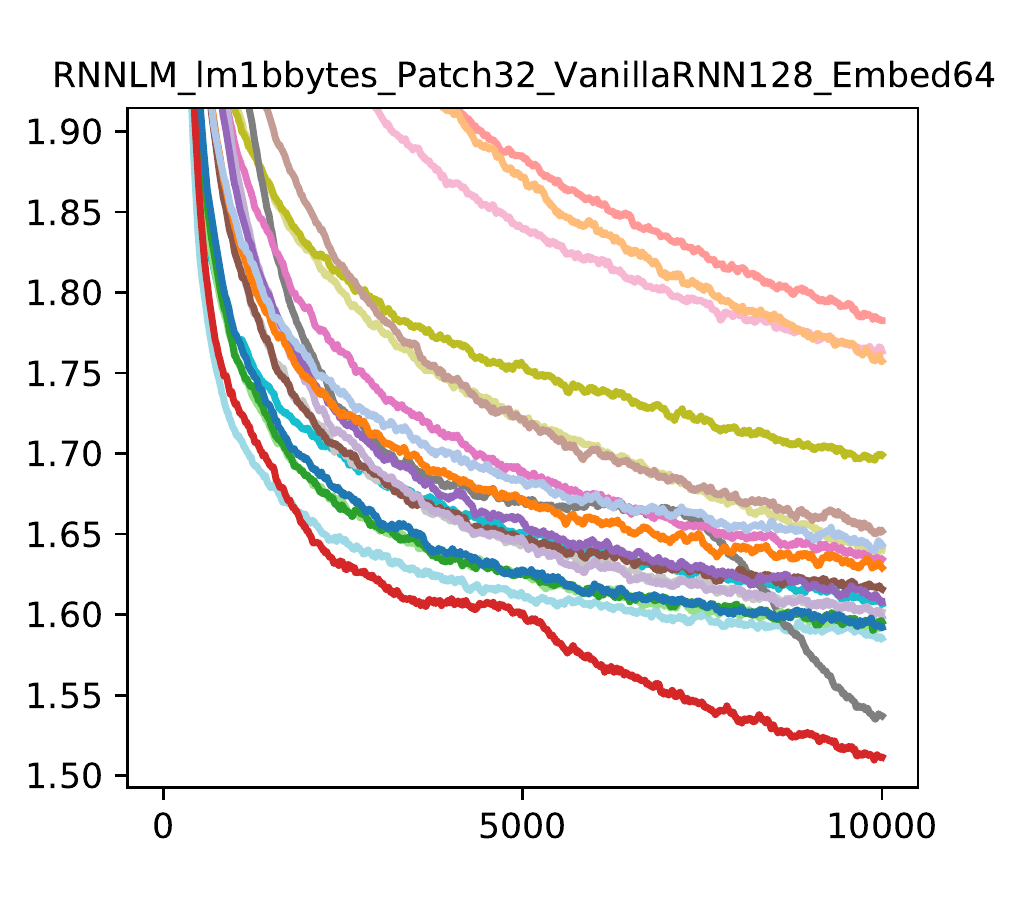}
    \end{overpic}
    }


    \makebox[\textwidth]{%
    \begin{overpic}[width=0.23\textwidth]{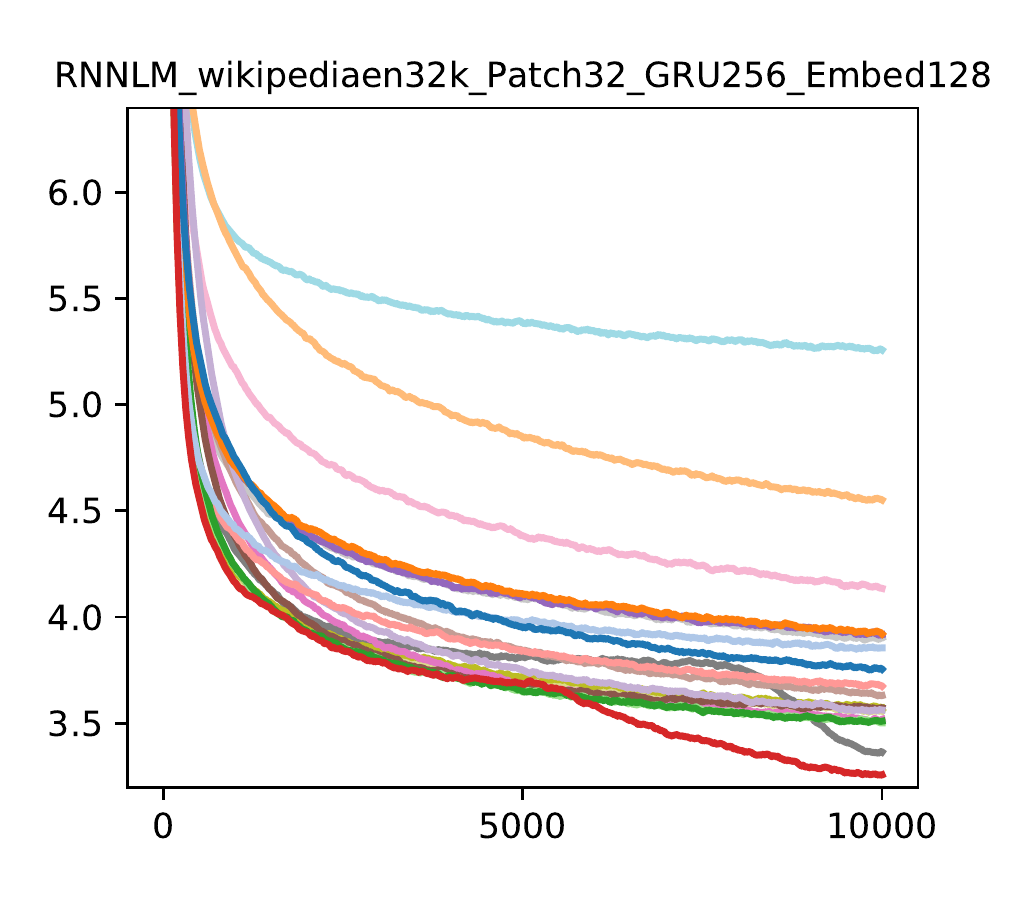}
    \end{overpic}
    \begin{overpic}[width=0.23\textwidth]{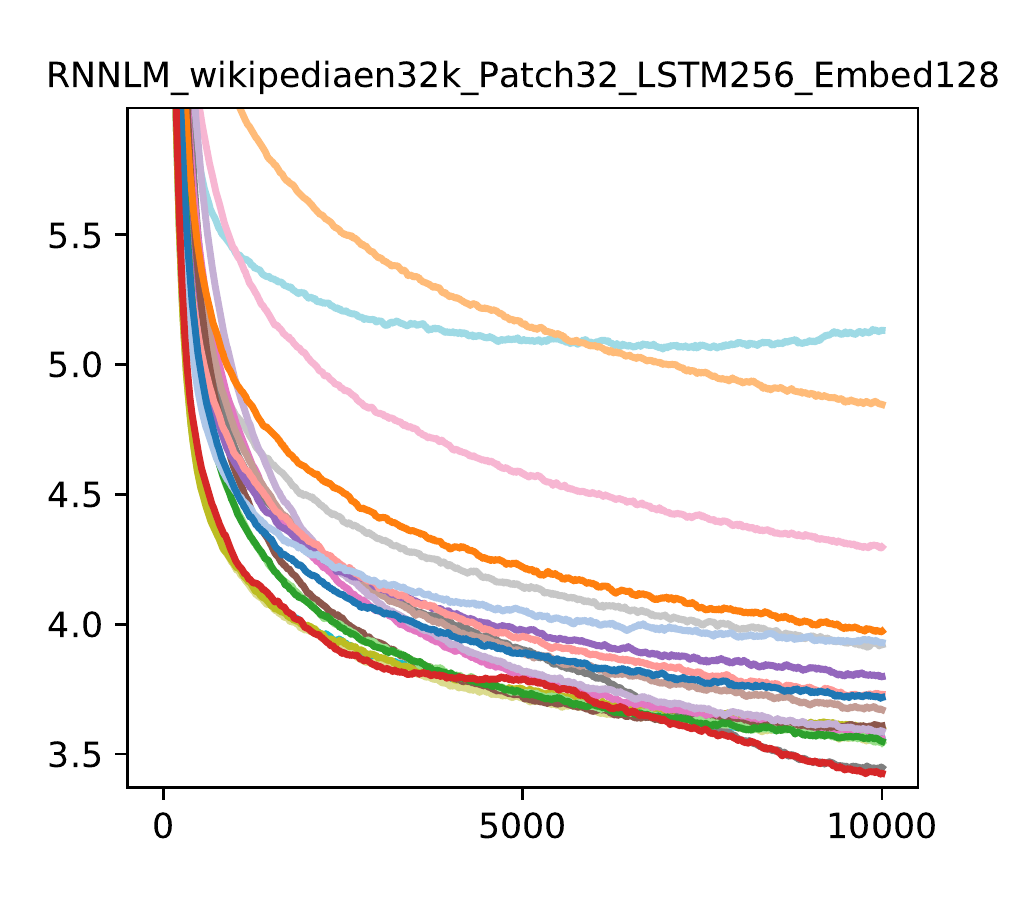}
    \end{overpic}
    \begin{overpic}[width=0.23\textwidth]{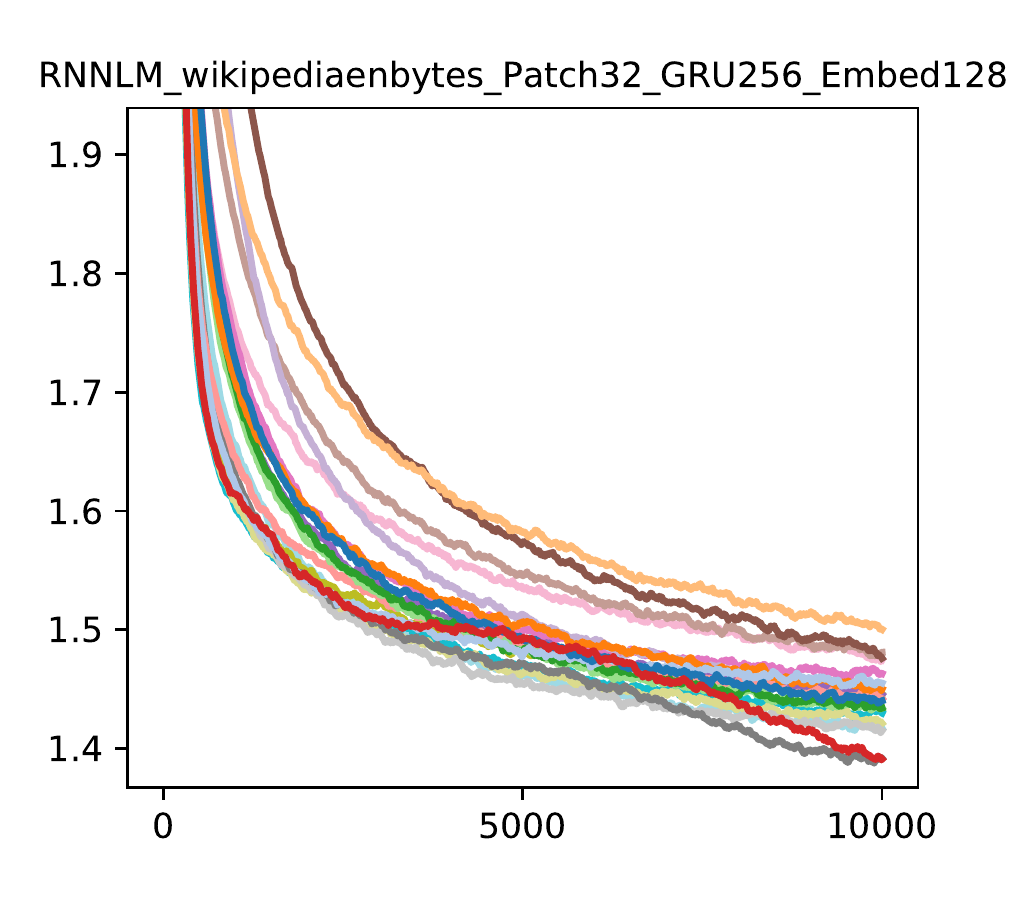}
    \end{overpic}
    \begin{overpic}[width=0.23\textwidth]{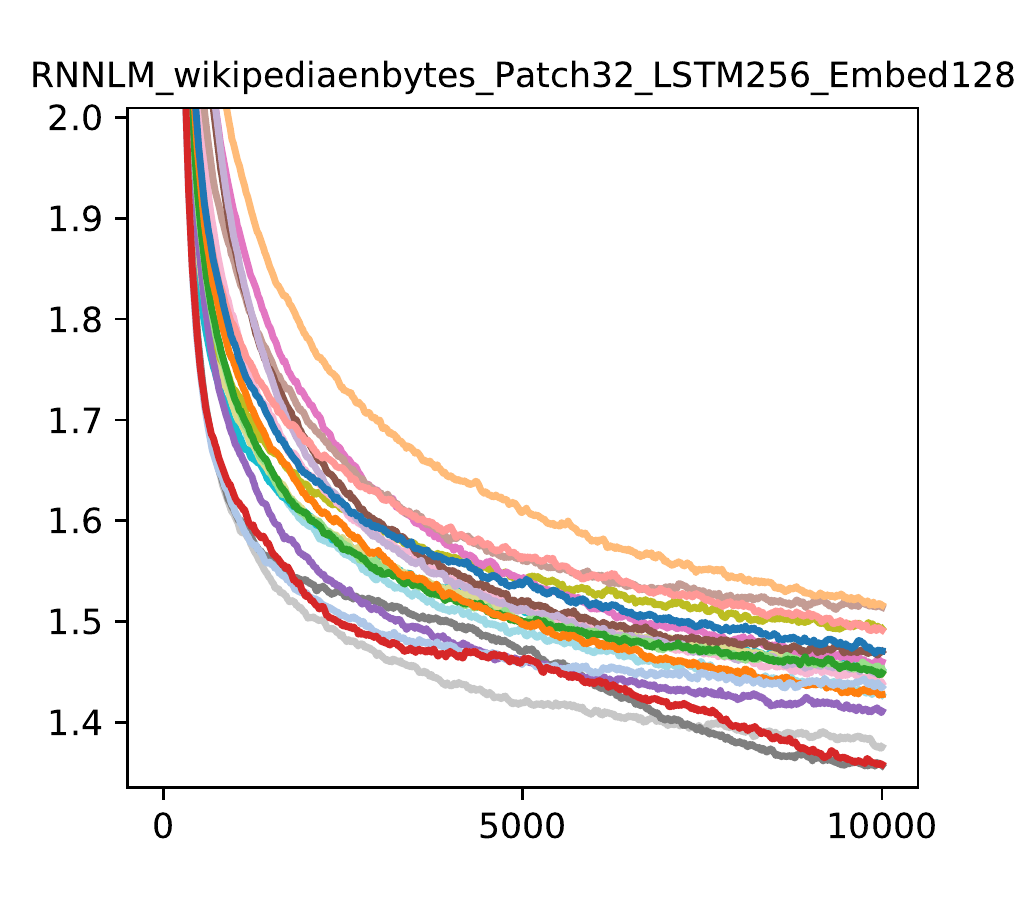}
    \end{overpic}
    \begin{overpic}[width=0.23\textwidth]{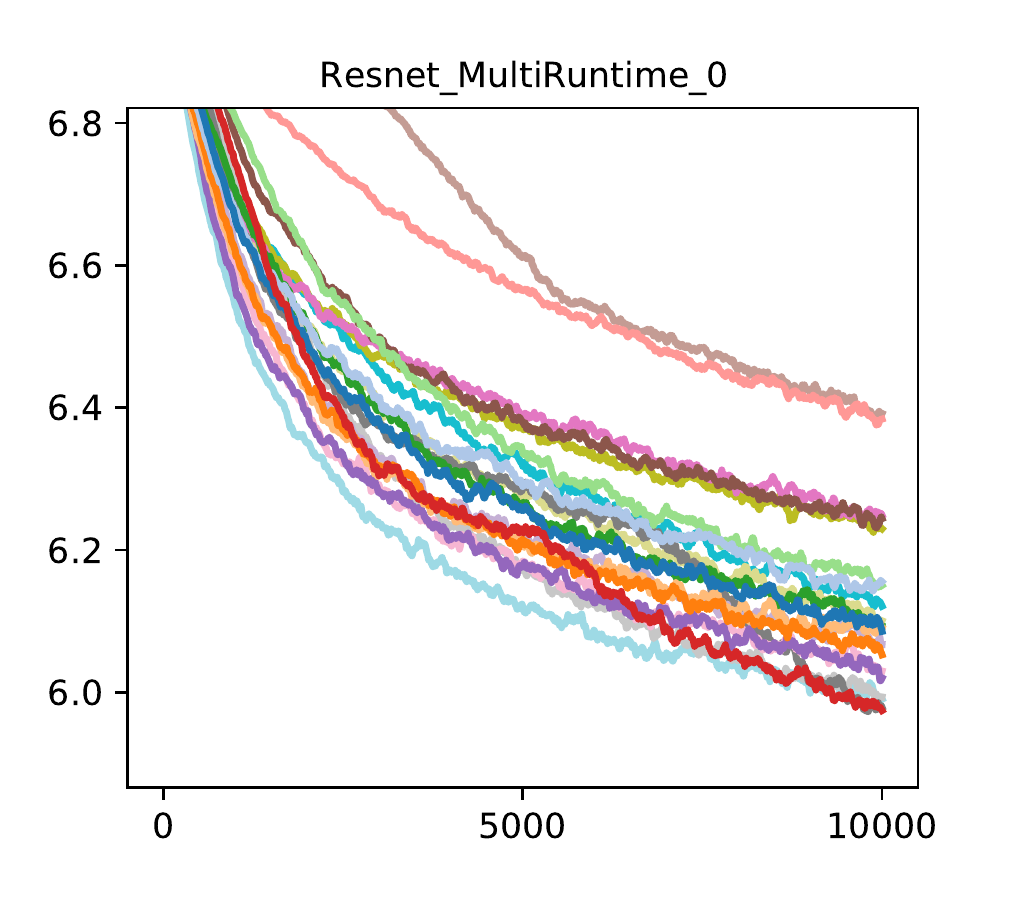}
    \end{overpic}
    }


    \makebox[\textwidth]{%
    \begin{overpic}[width=0.23\textwidth]{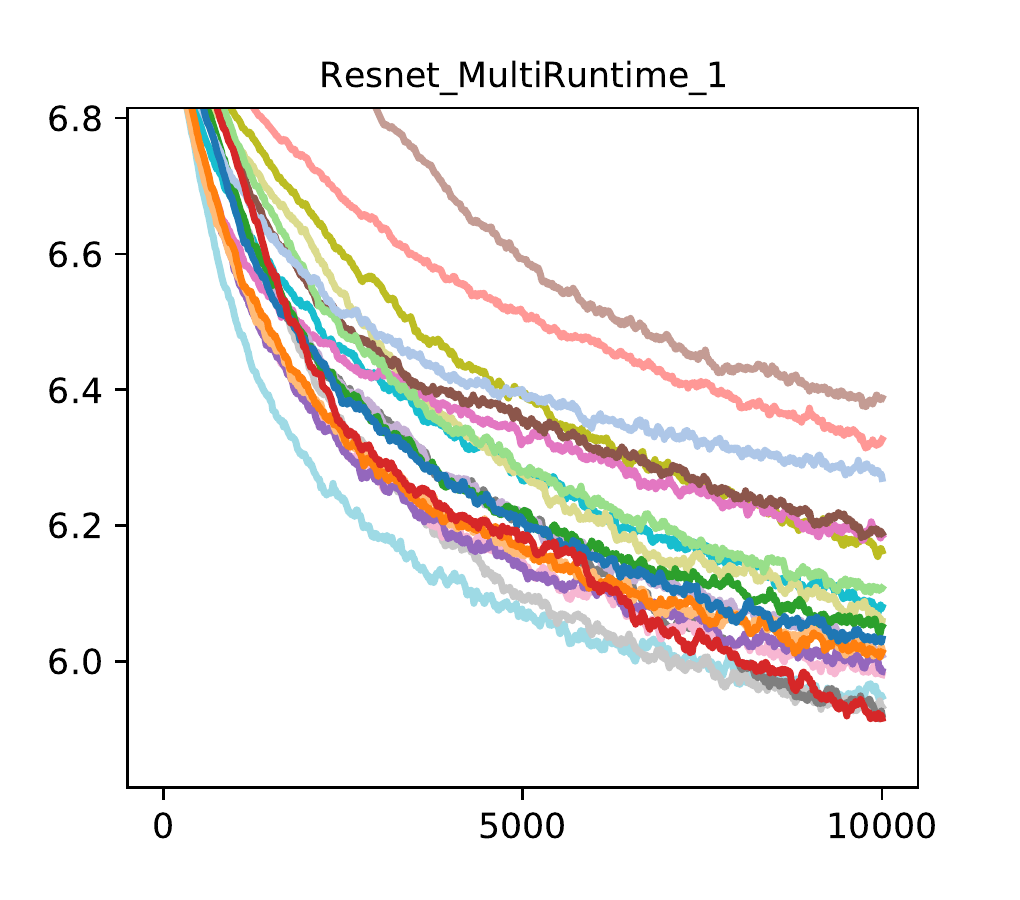}
    \end{overpic}
    \begin{overpic}[width=0.23\textwidth]{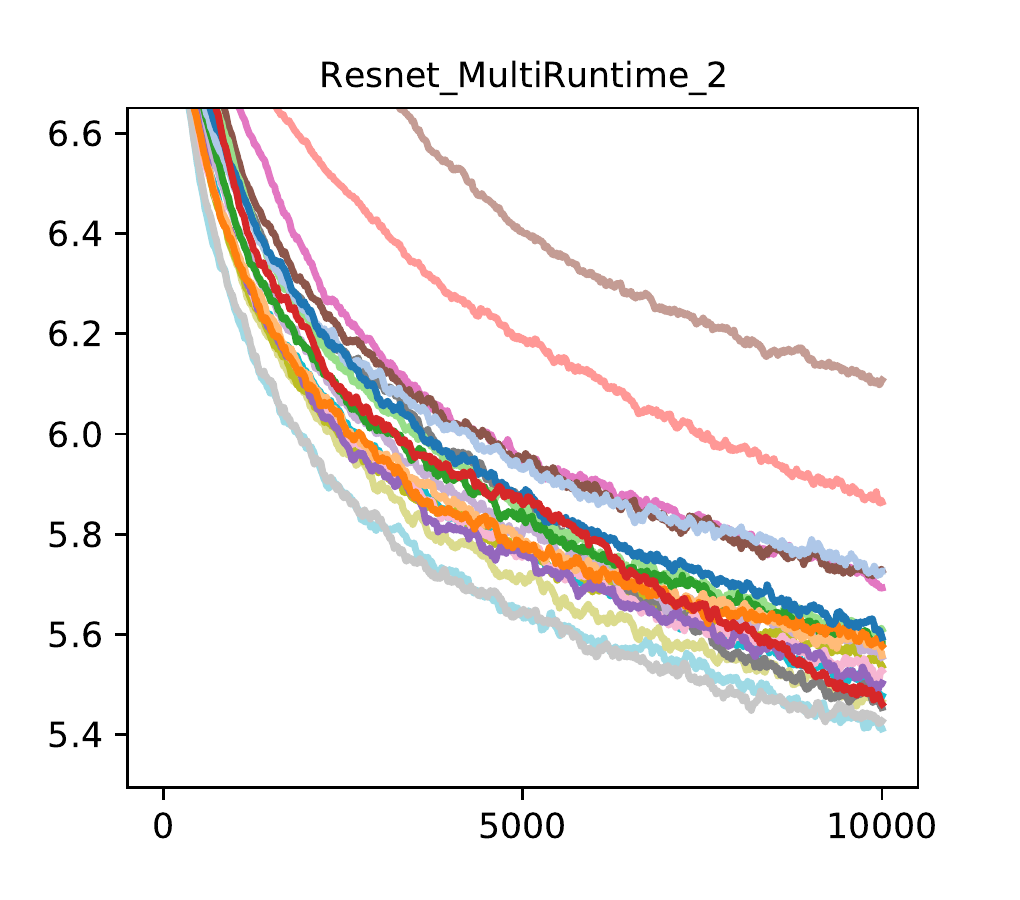}
    \end{overpic}
    \begin{overpic}[width=0.23\textwidth]{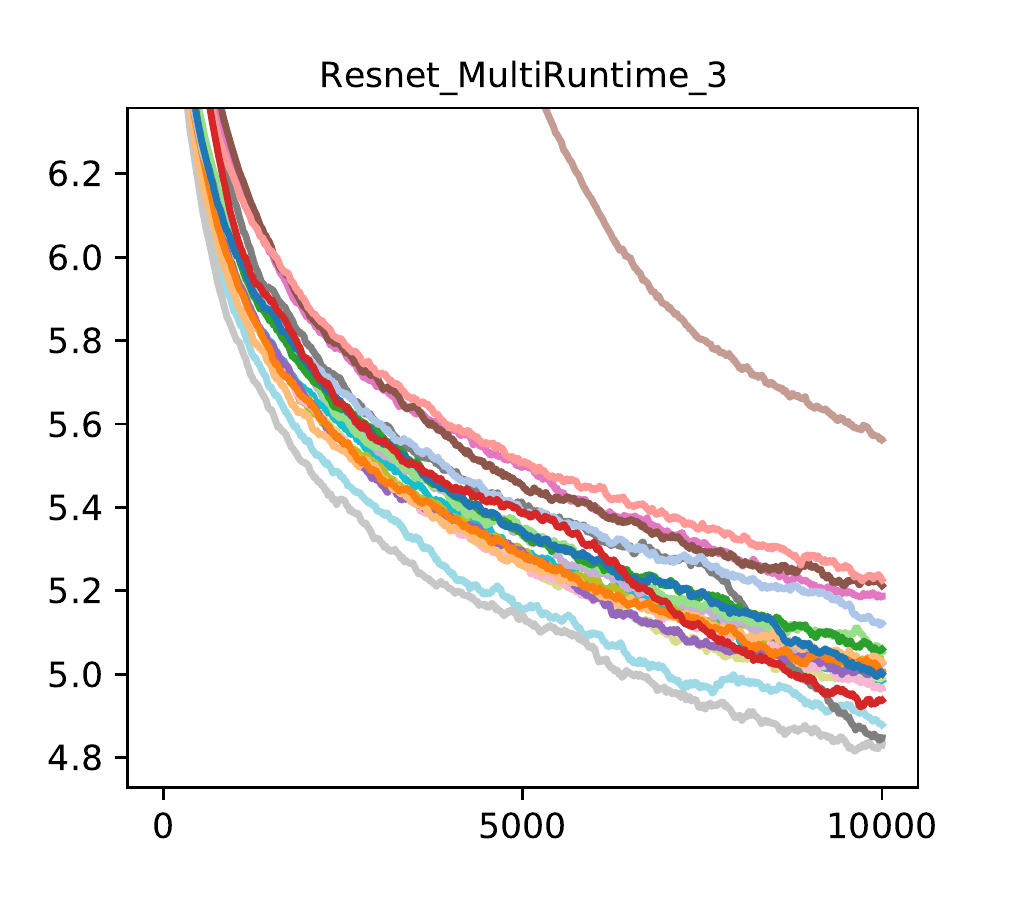}
    \end{overpic}
    \begin{overpic}[width=0.23\textwidth]{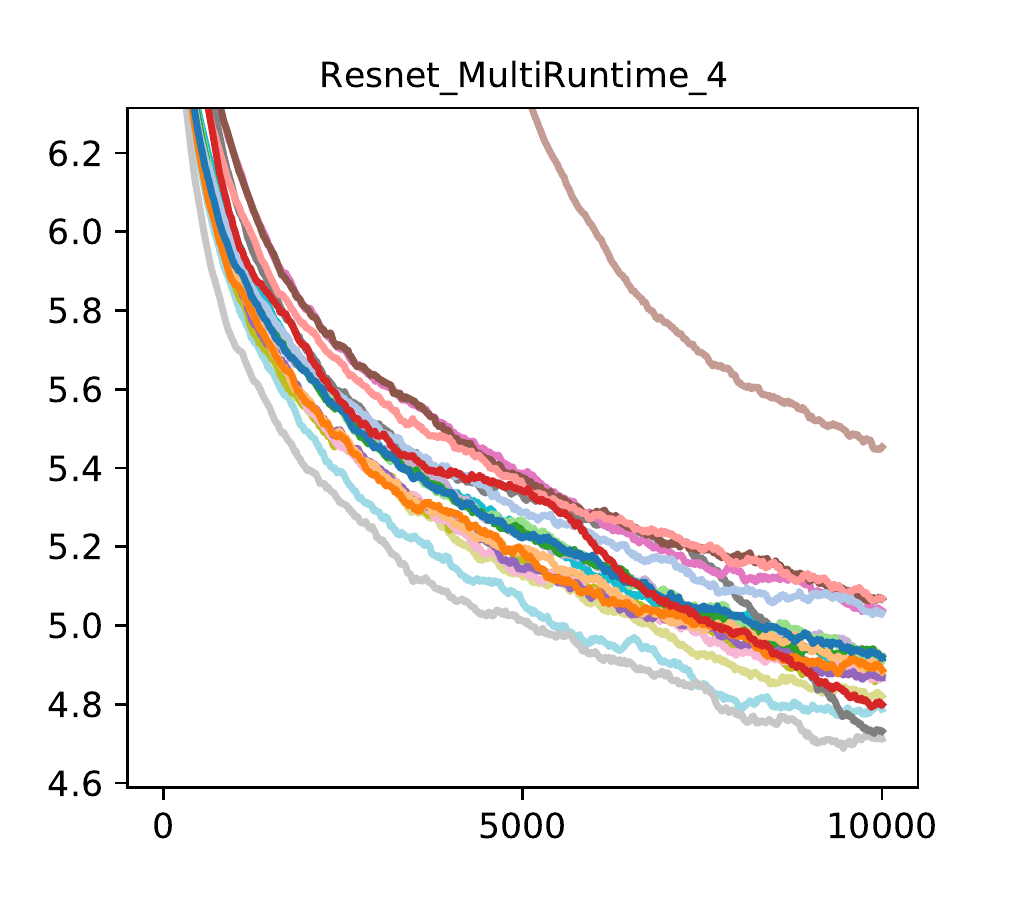}
    \end{overpic}
    \begin{overpic}[width=0.23\textwidth]{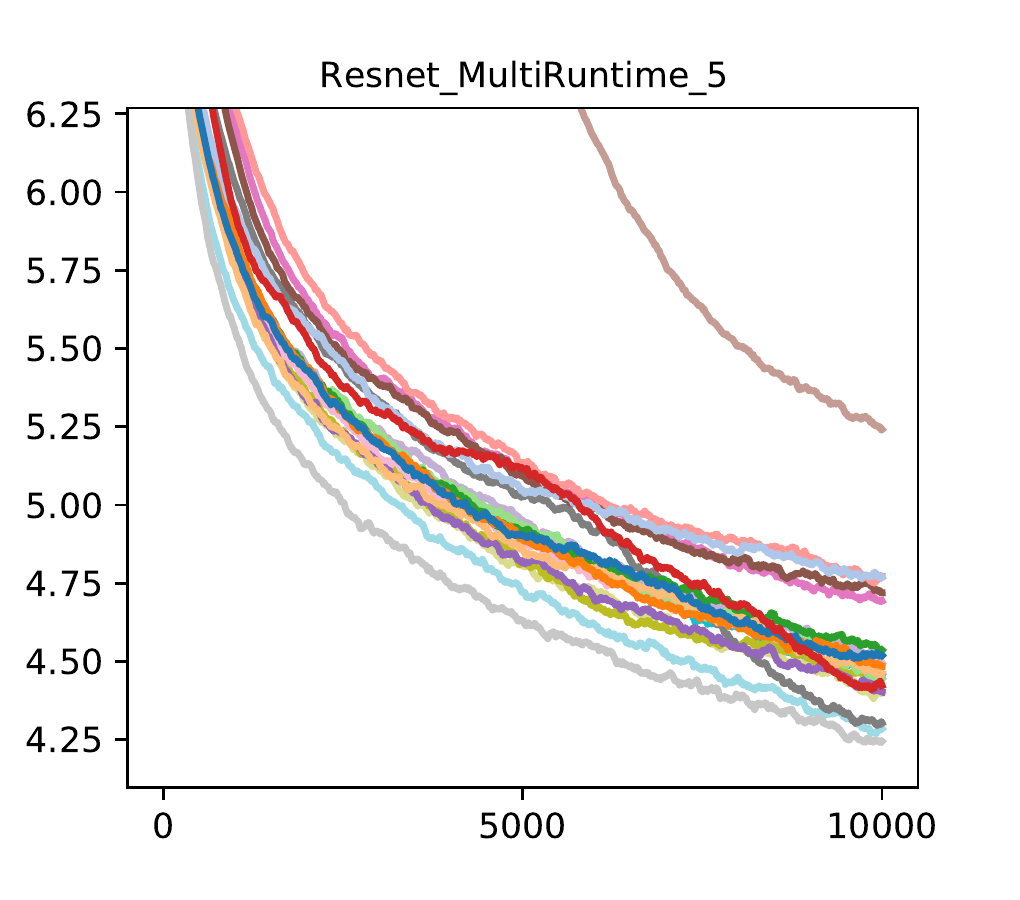}
    \end{overpic}
    }

%% file: app_page3.tex

    \makebox[\textwidth]{%
    \begin{overpic}[width=0.23\textwidth]{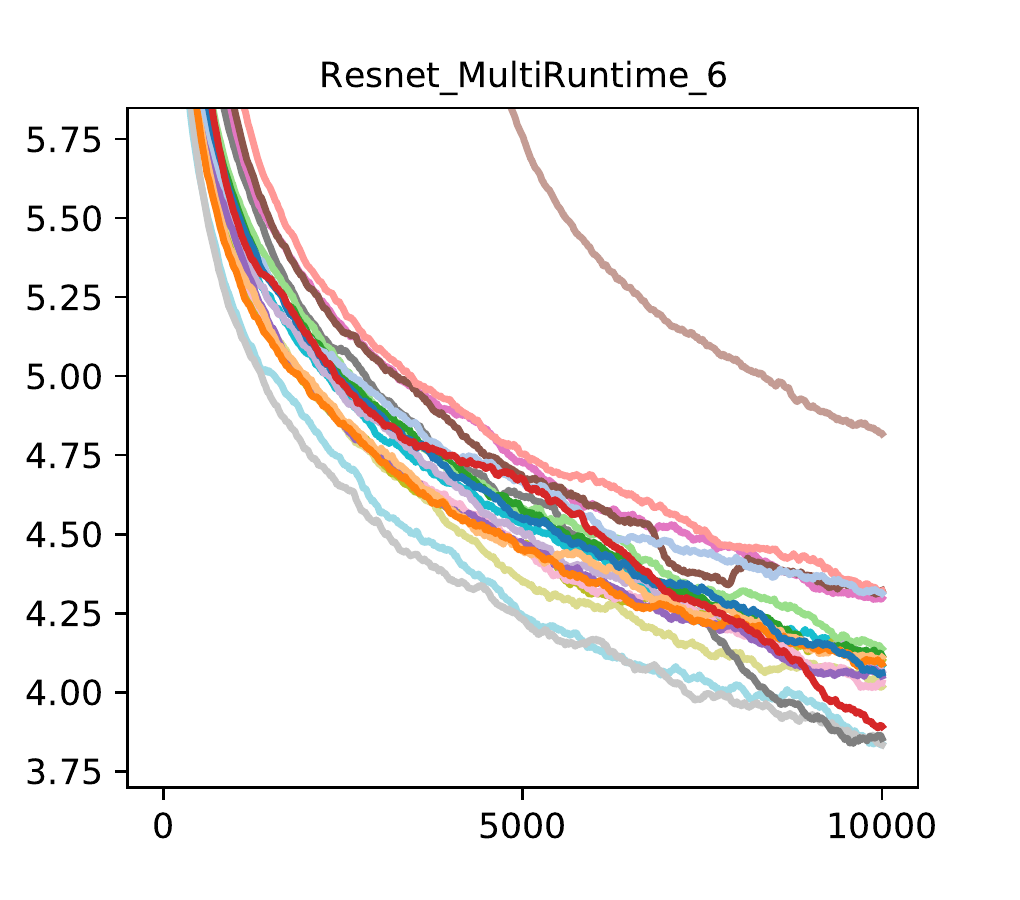}
    \end{overpic}
    \begin{overpic}[width=0.23\textwidth]{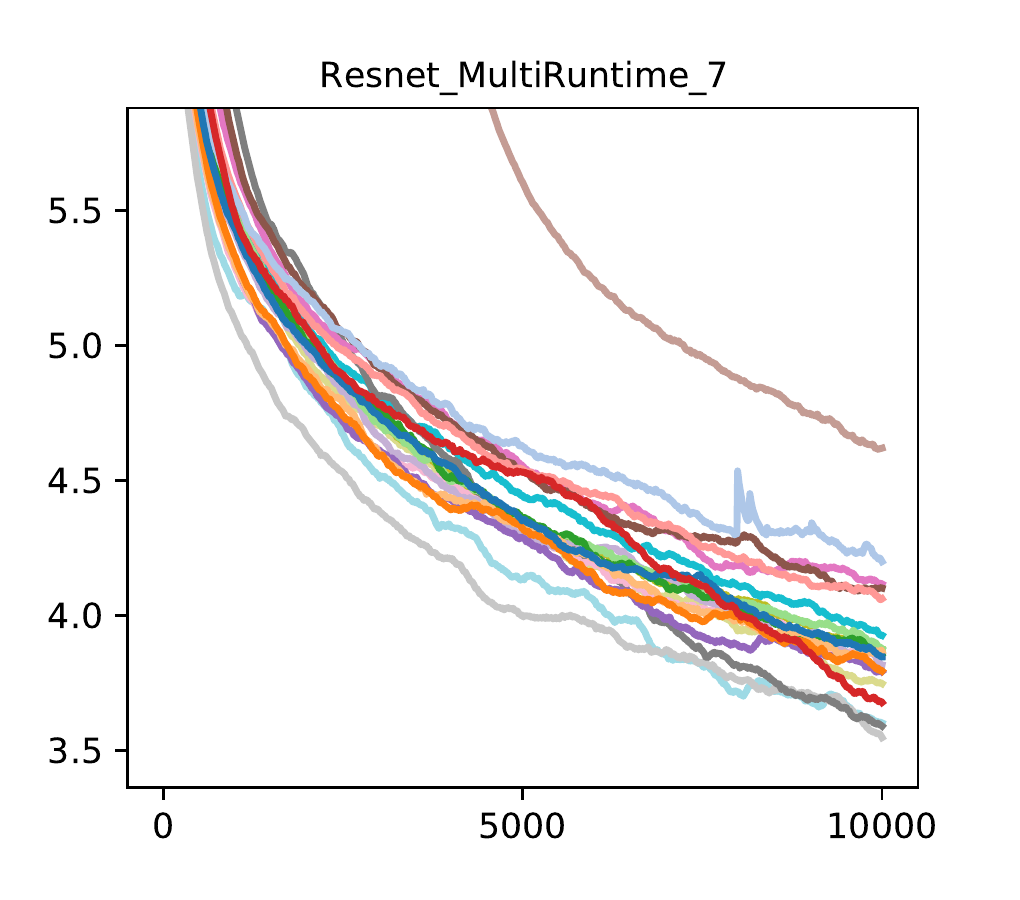}
    \end{overpic}
    \begin{overpic}[width=0.23\textwidth]{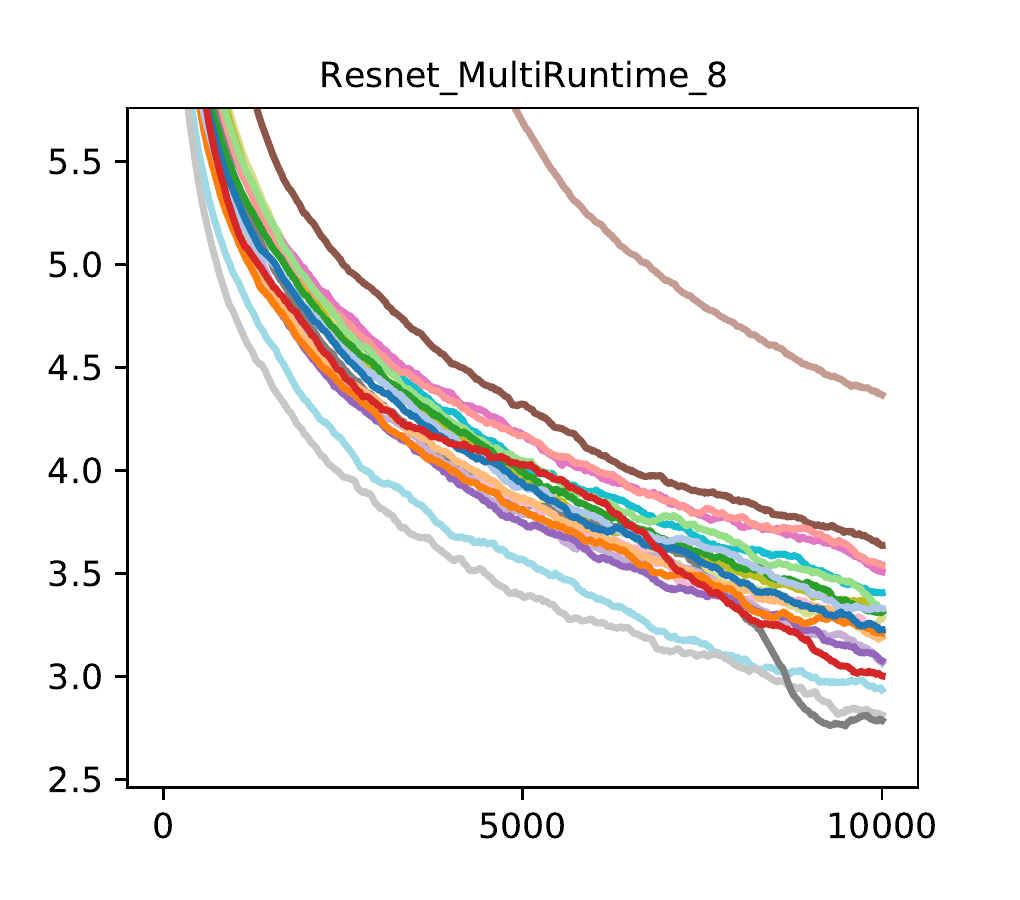}
    \end{overpic}
    \begin{overpic}[width=0.23\textwidth]{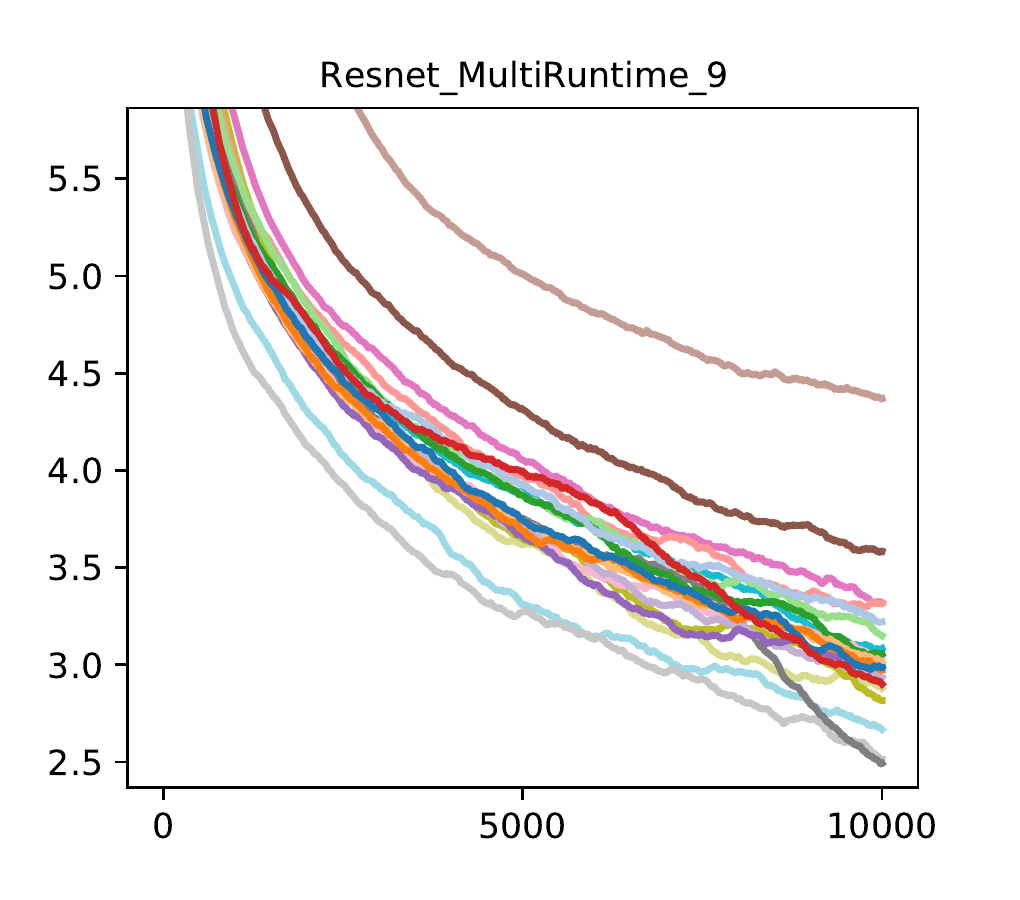}
    \end{overpic}
    \begin{overpic}[width=0.23\textwidth]{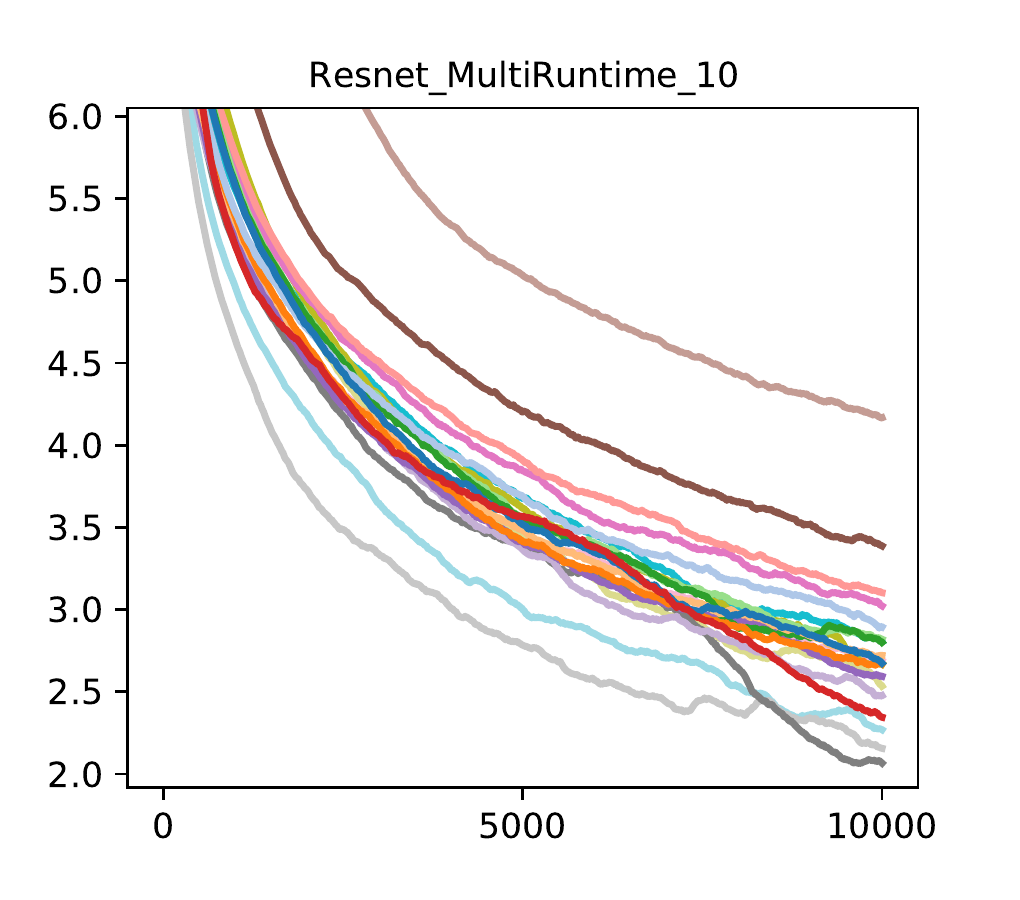}
    \end{overpic}
    }


    \makebox[\textwidth]{%
    \begin{overpic}[width=0.23\textwidth]{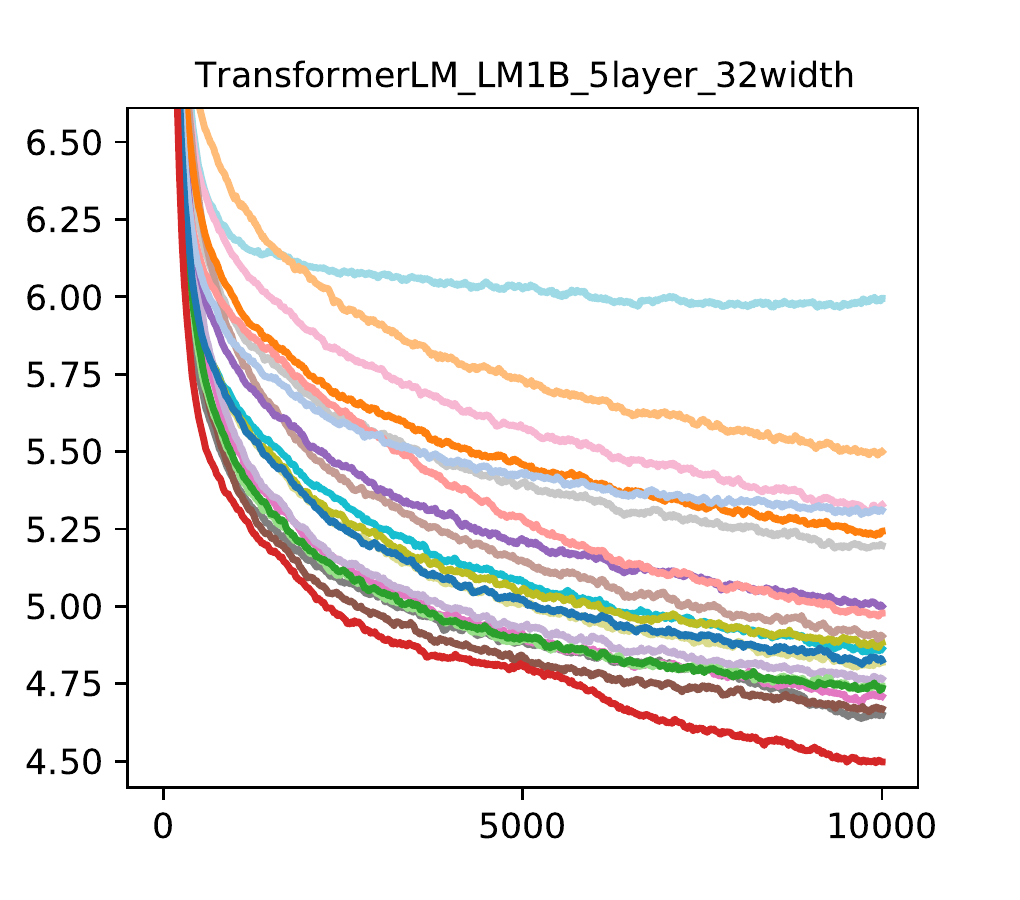}
    \end{overpic}
    \begin{overpic}[width=0.23\textwidth]{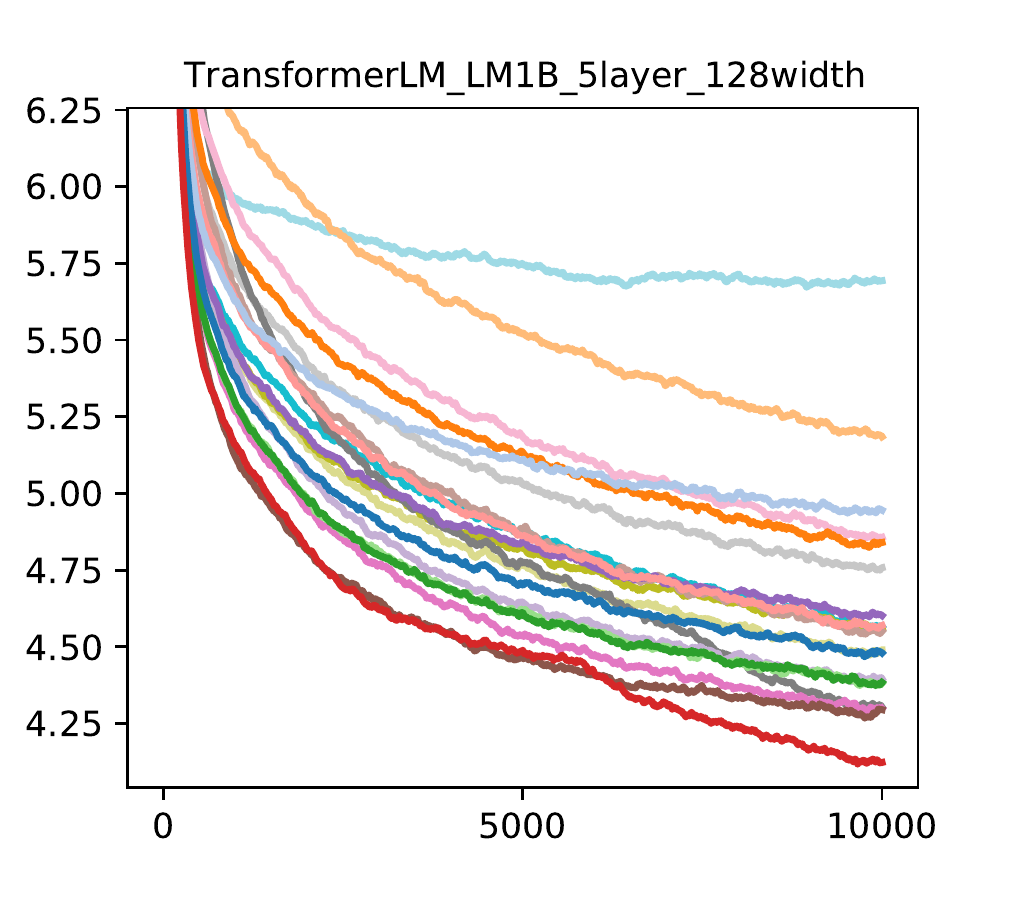}
    \end{overpic}
    \begin{overpic}[width=0.23\textwidth]{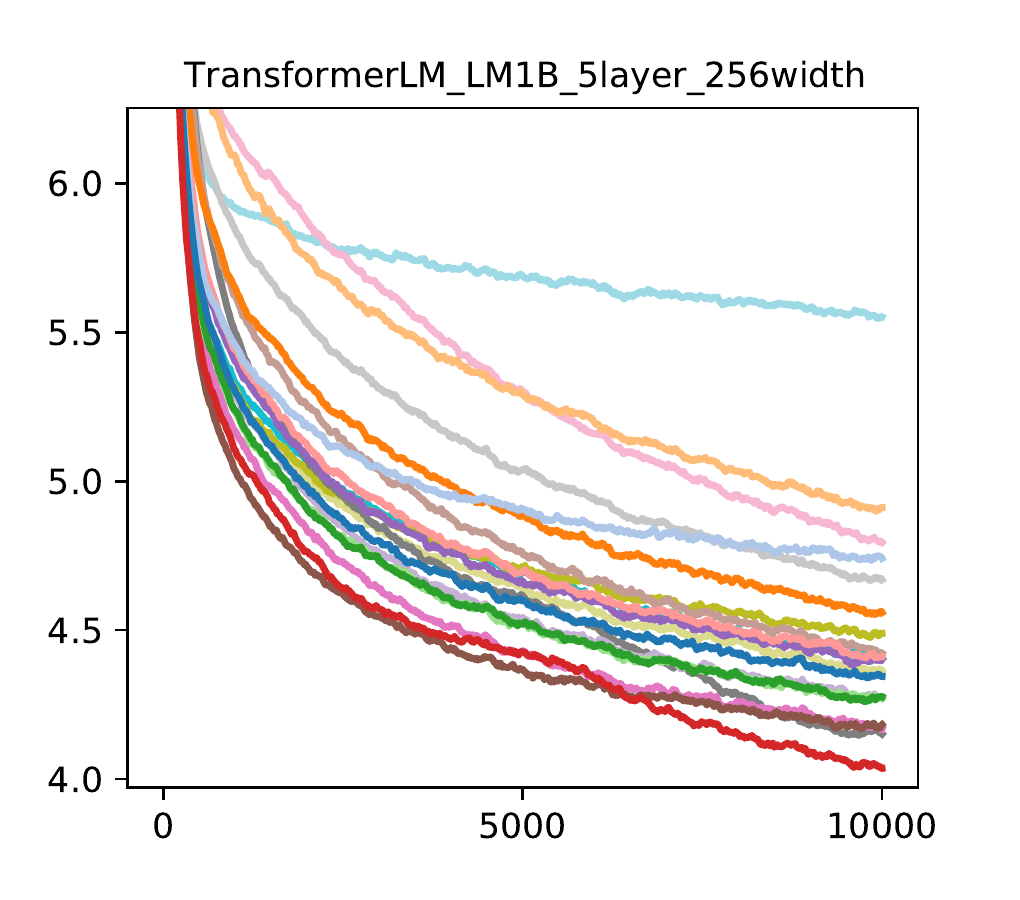}
    \end{overpic}
    \begin{overpic}[width=0.23\textwidth]{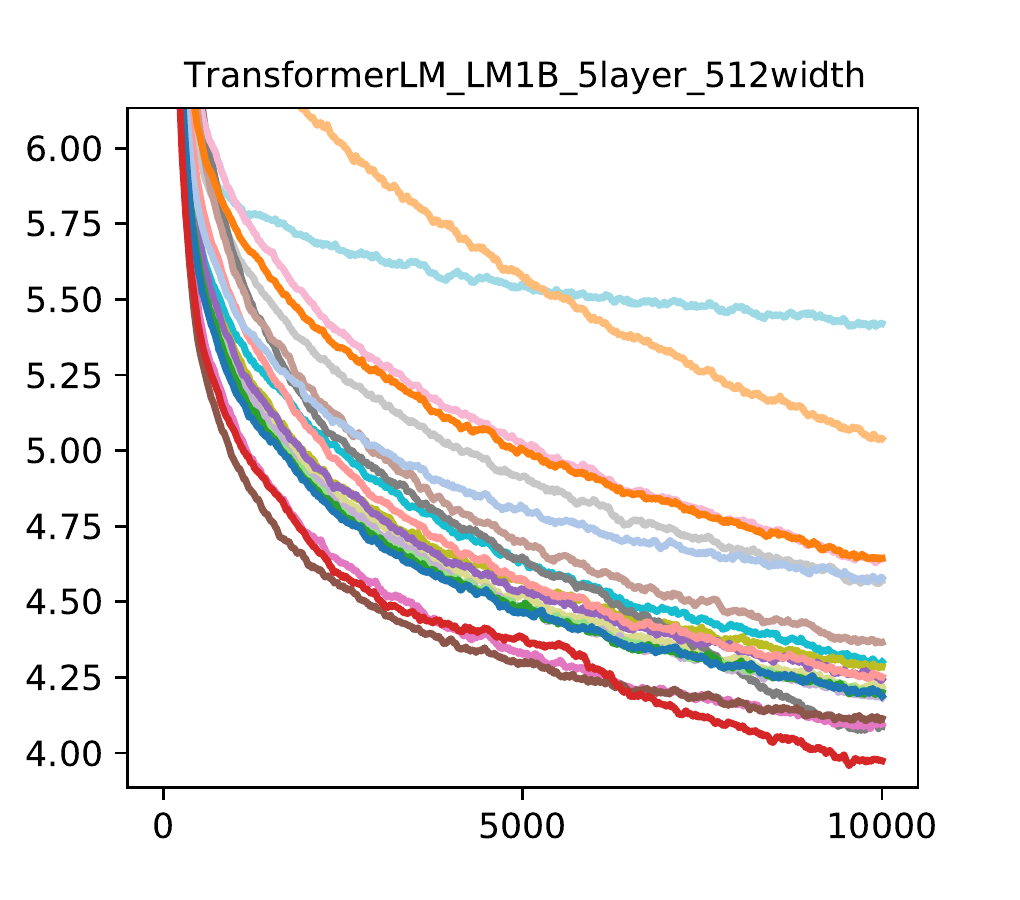}
    \end{overpic}
    \begin{overpic}[width=0.23\textwidth]{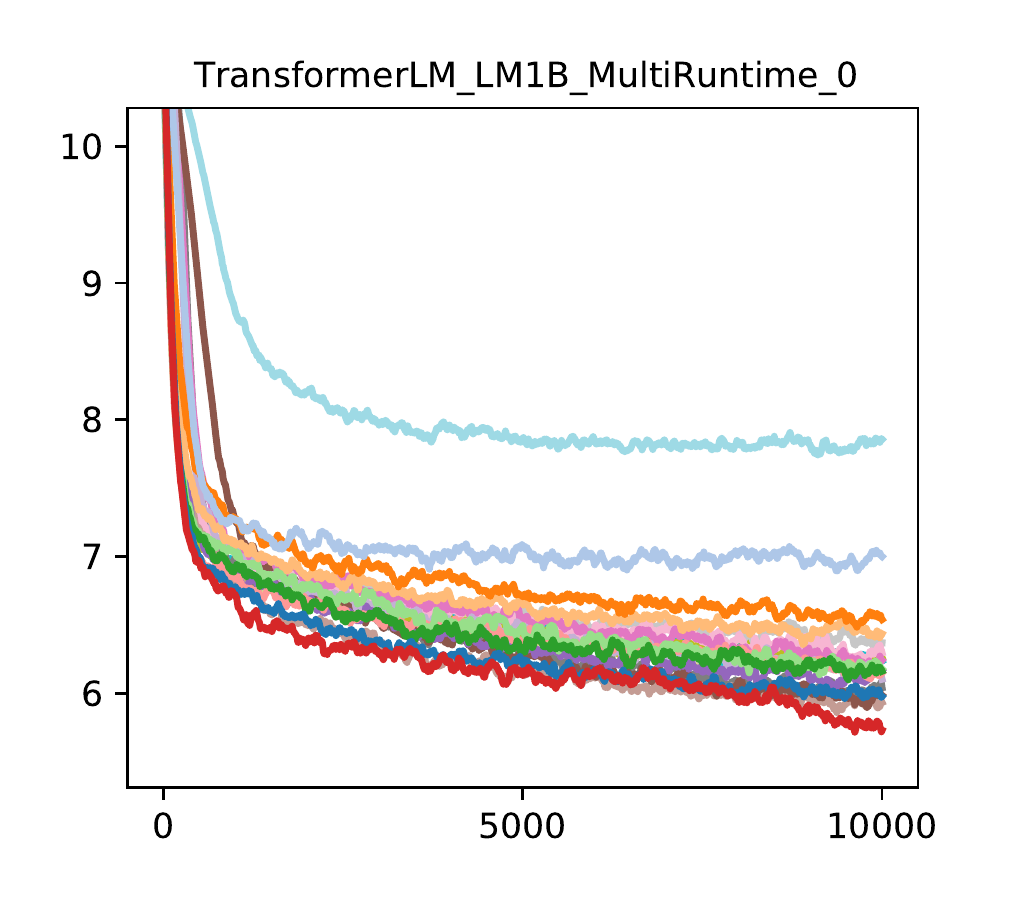}
    \end{overpic}
    }


    \makebox[\textwidth]{%
    \begin{overpic}[width=0.23\textwidth]{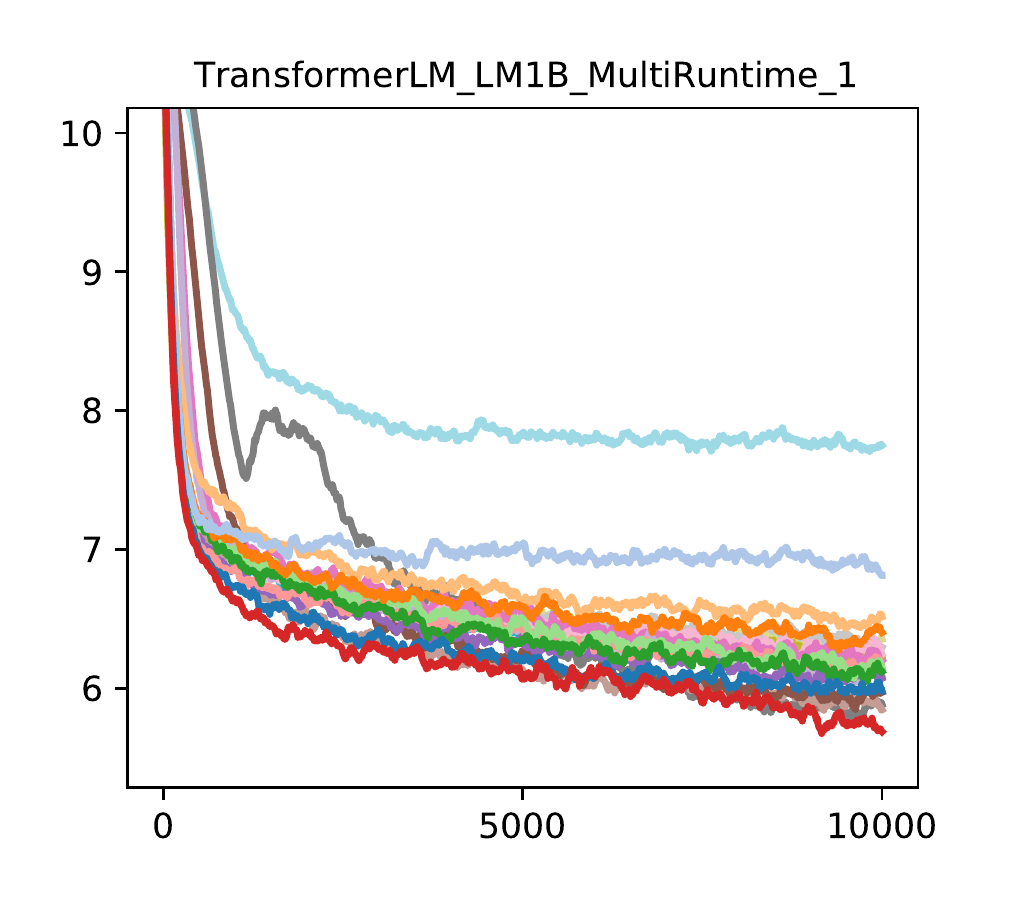}
    \end{overpic}
    \begin{overpic}[width=0.23\textwidth]{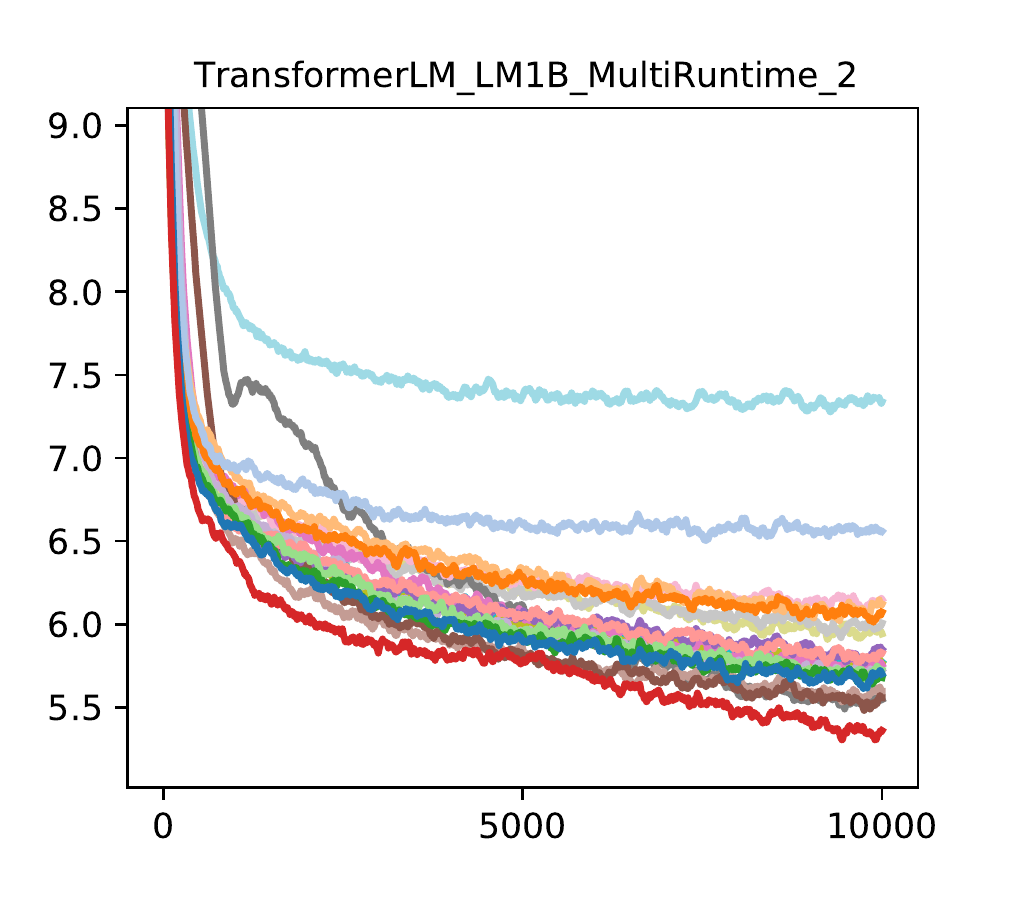}
    \end{overpic}
    \begin{overpic}[width=0.23\textwidth]{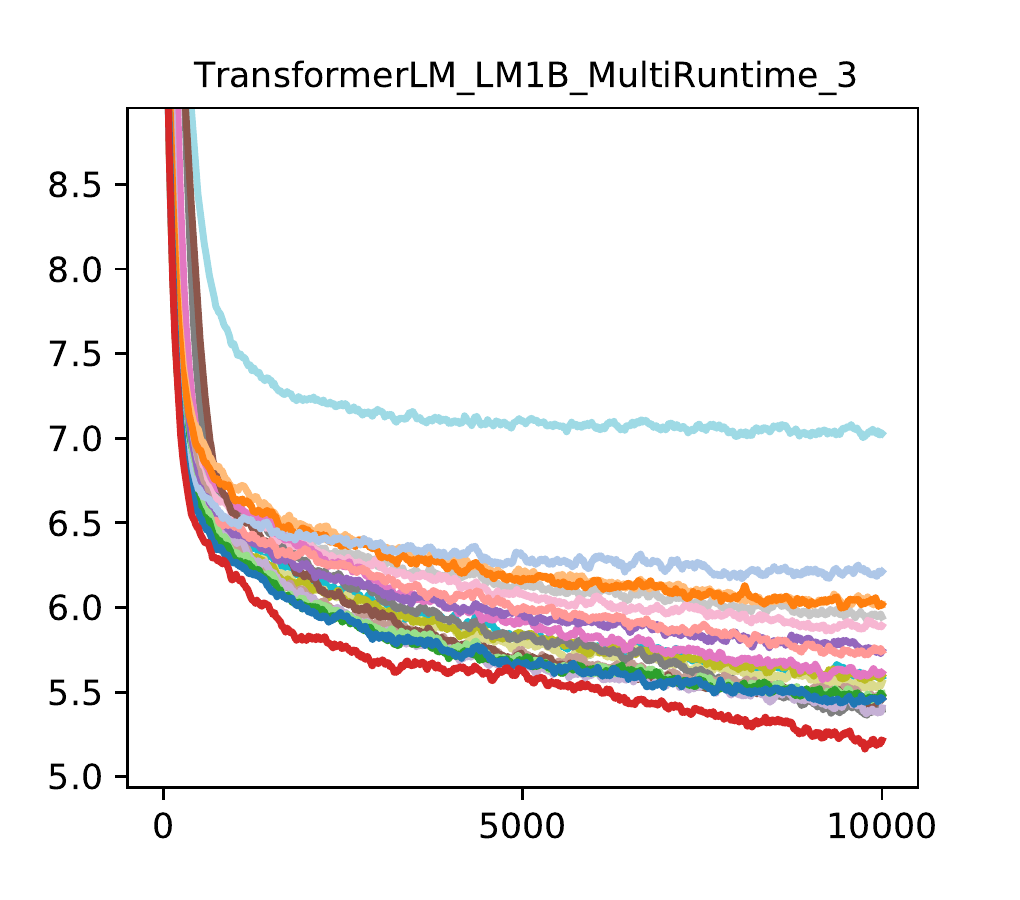}
    \end{overpic}
    \begin{overpic}[width=0.23\textwidth]{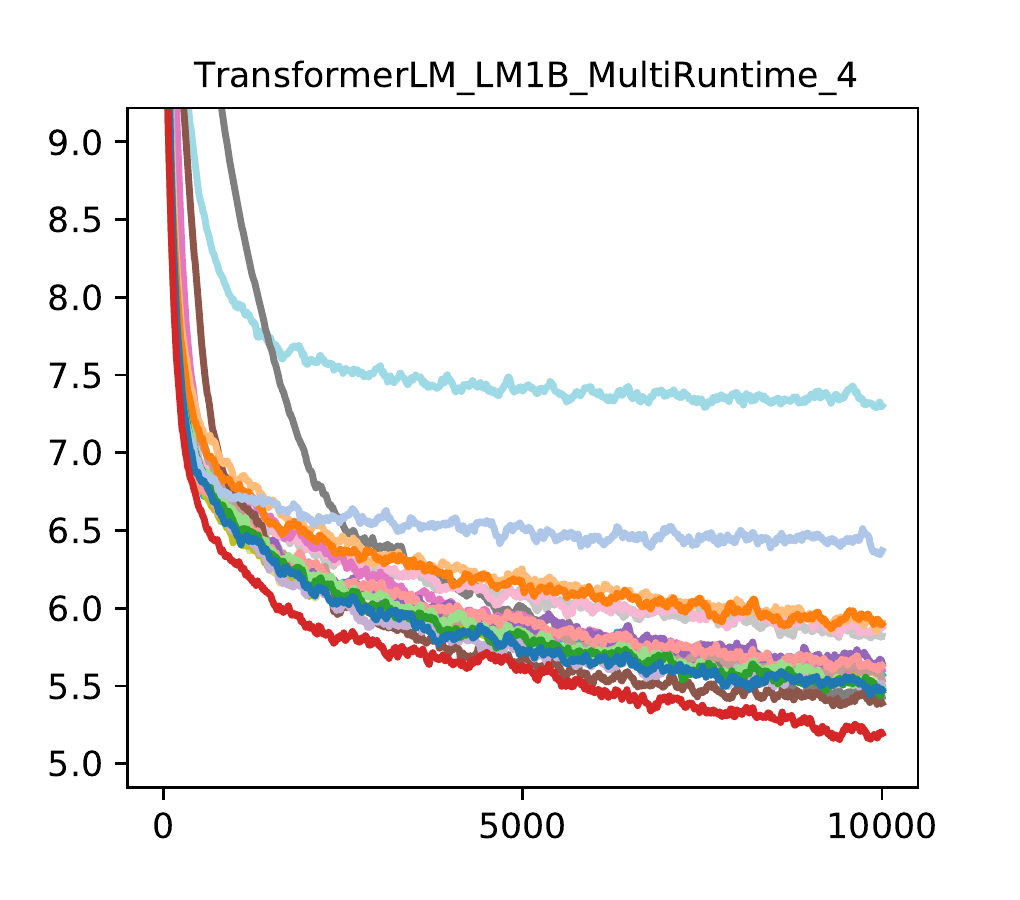}
    \end{overpic}
    \begin{overpic}[width=0.23\textwidth]{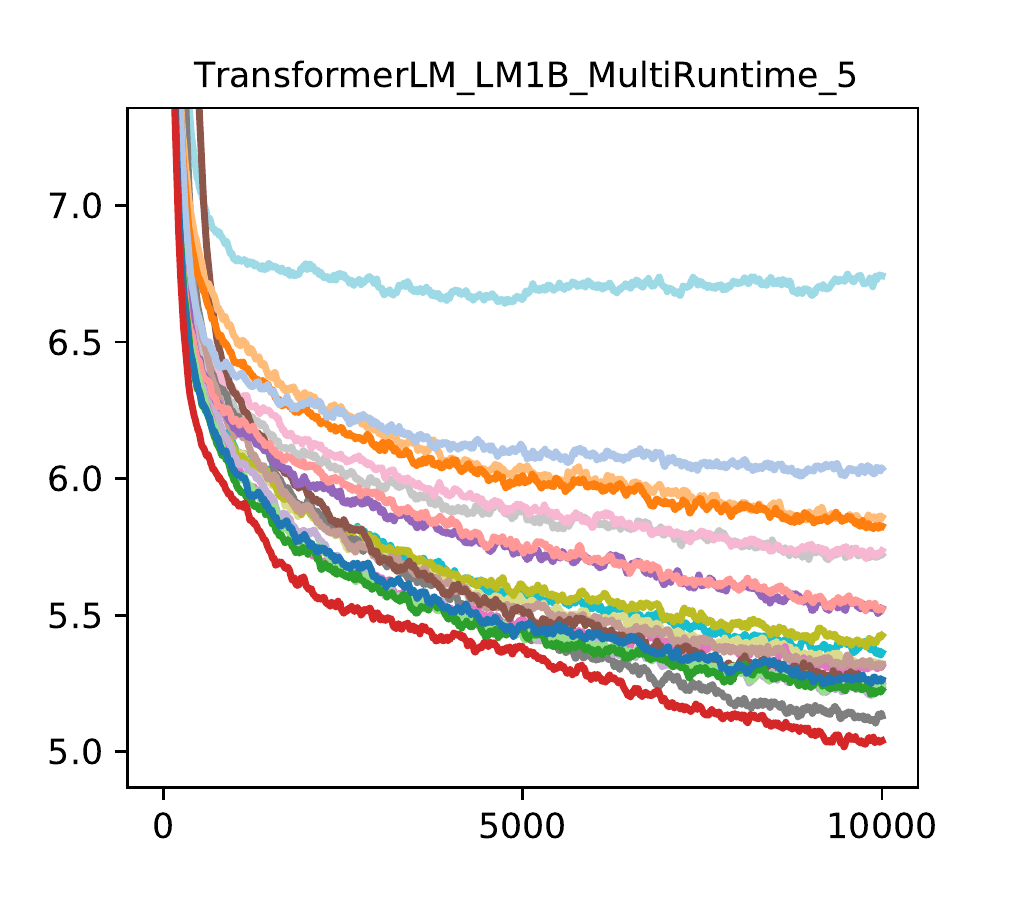}
    \end{overpic}
    }


    \makebox[\textwidth]{%
    \begin{overpic}[width=0.23\textwidth]{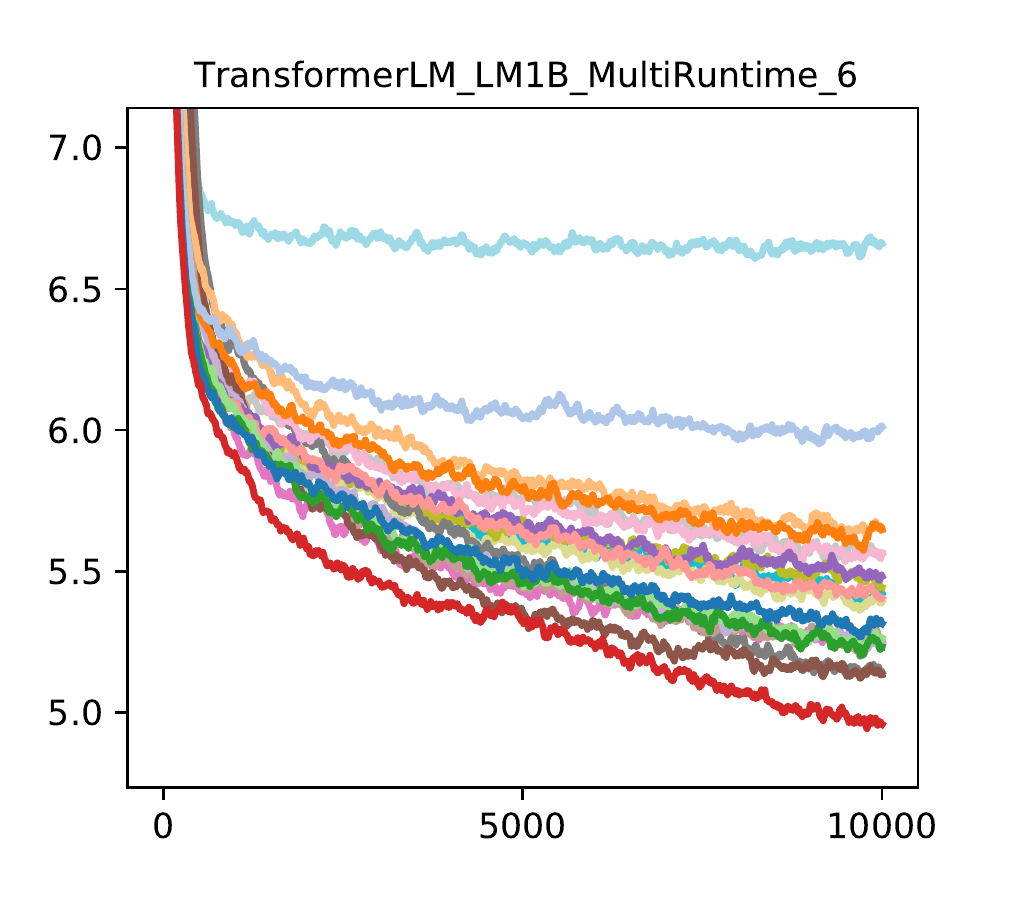}
    \end{overpic}
    \begin{overpic}[width=0.23\textwidth]{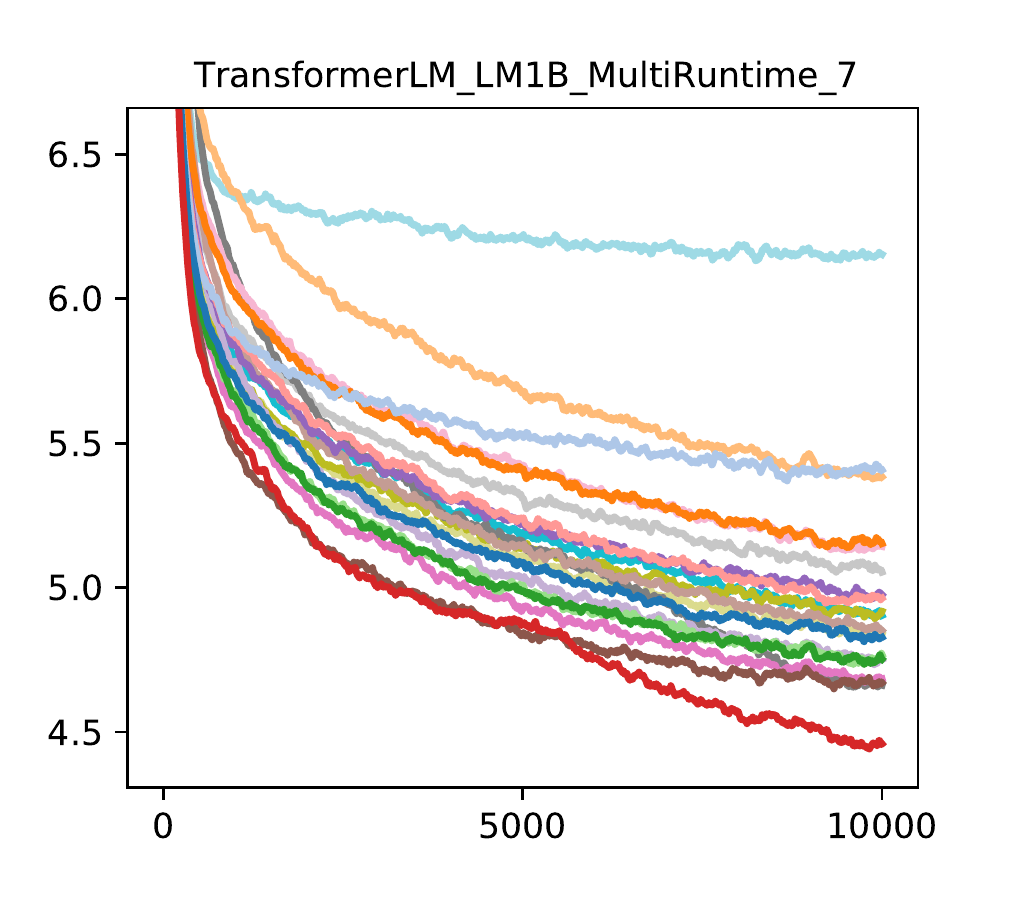}
    \end{overpic}
    \begin{overpic}[width=0.23\textwidth]{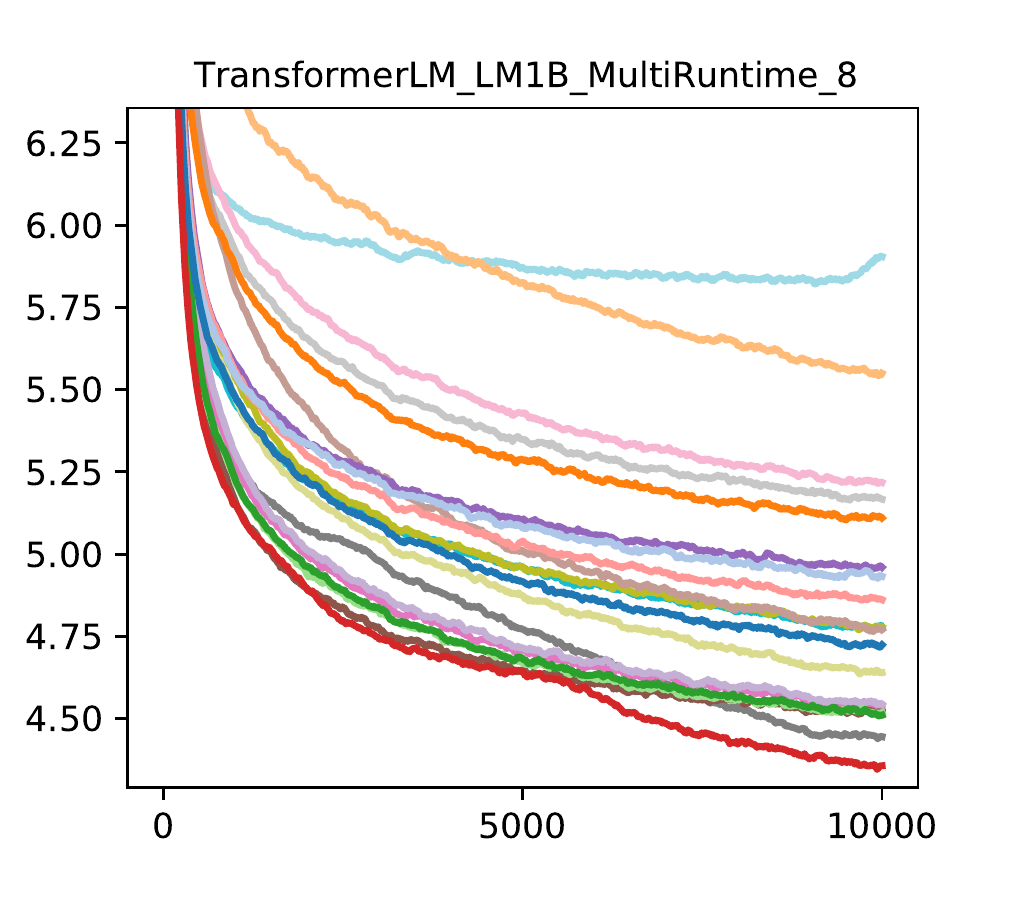}
    \end{overpic}
    \begin{overpic}[width=0.23\textwidth]{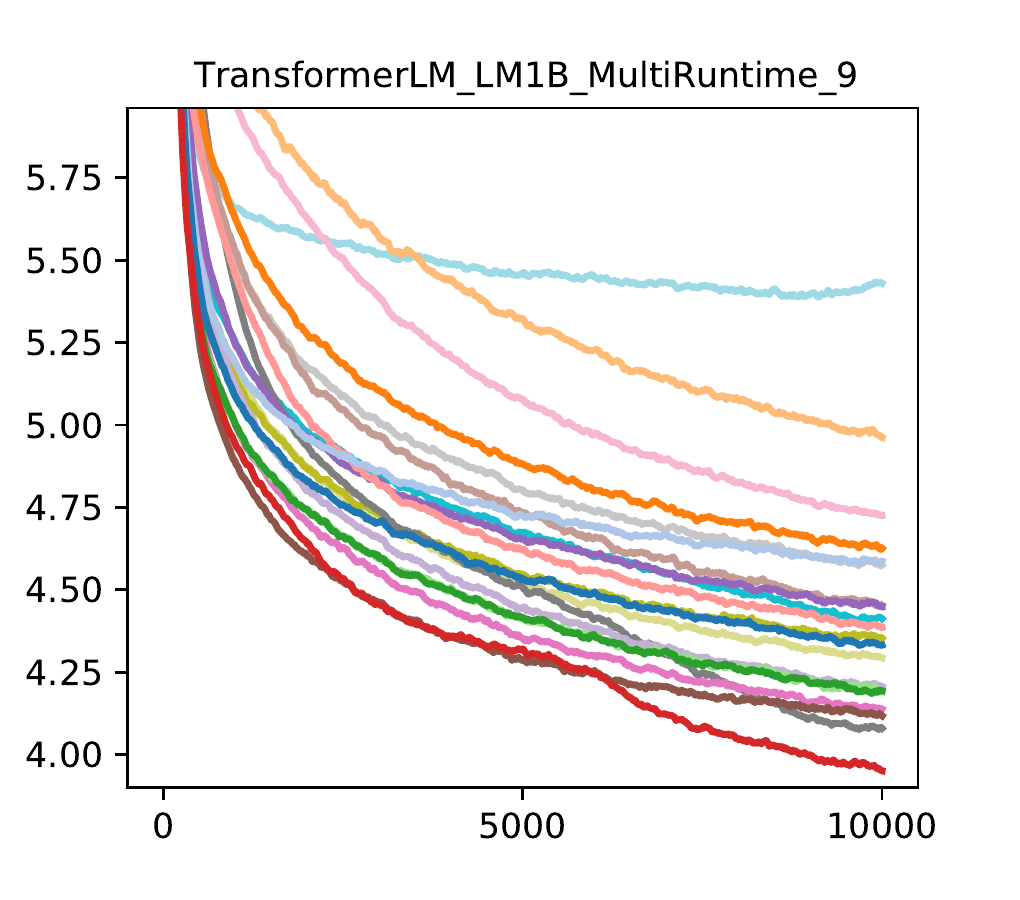}
    \end{overpic}
    \begin{overpic}[width=0.23\textwidth]{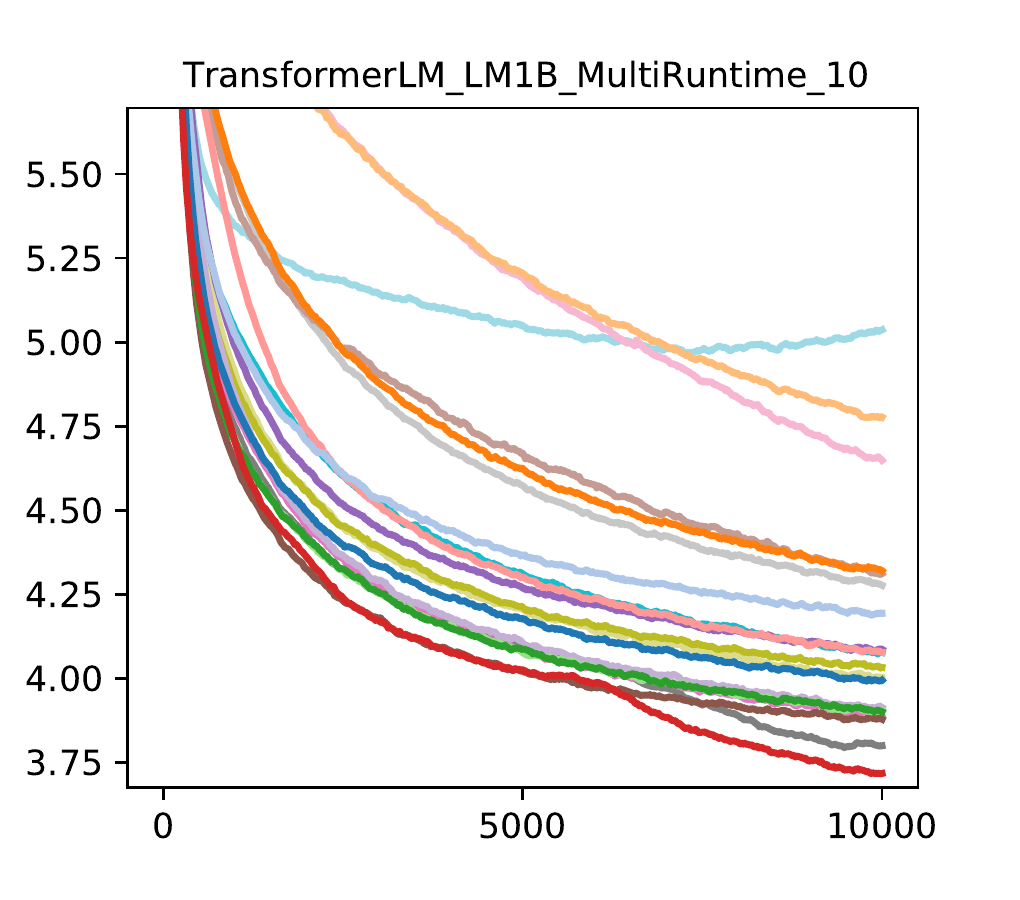}
    \end{overpic}
    }


    \makebox[\textwidth]{%
    \begin{overpic}[width=0.23\textwidth]{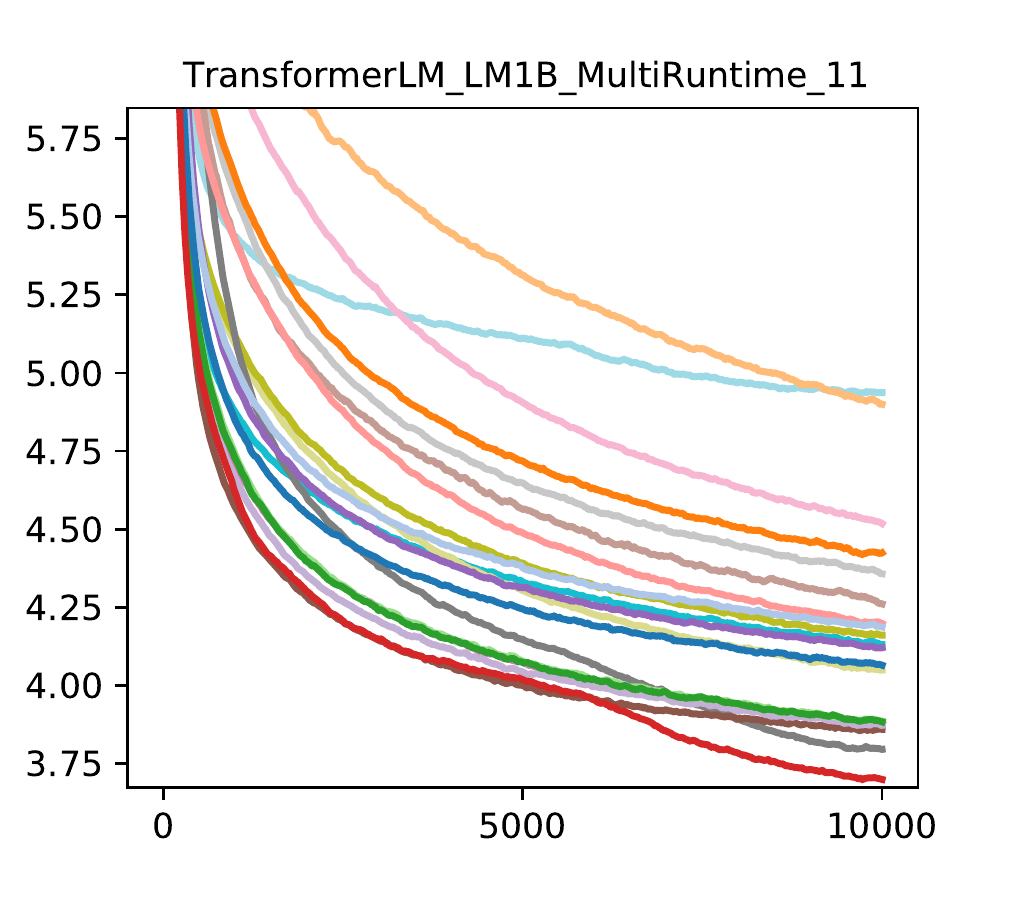}
    \end{overpic}
    \begin{overpic}[width=0.23\textwidth]{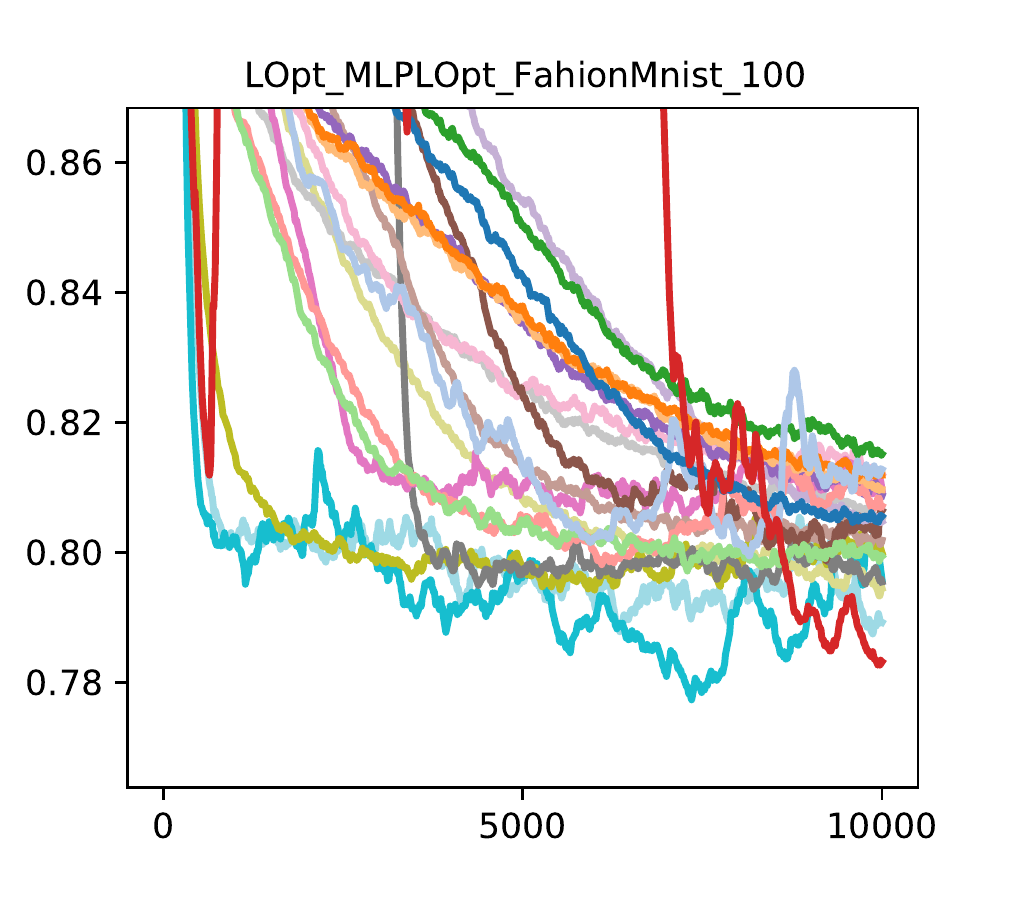}
    \end{overpic}
    \begin{overpic}[width=0.23\textwidth]{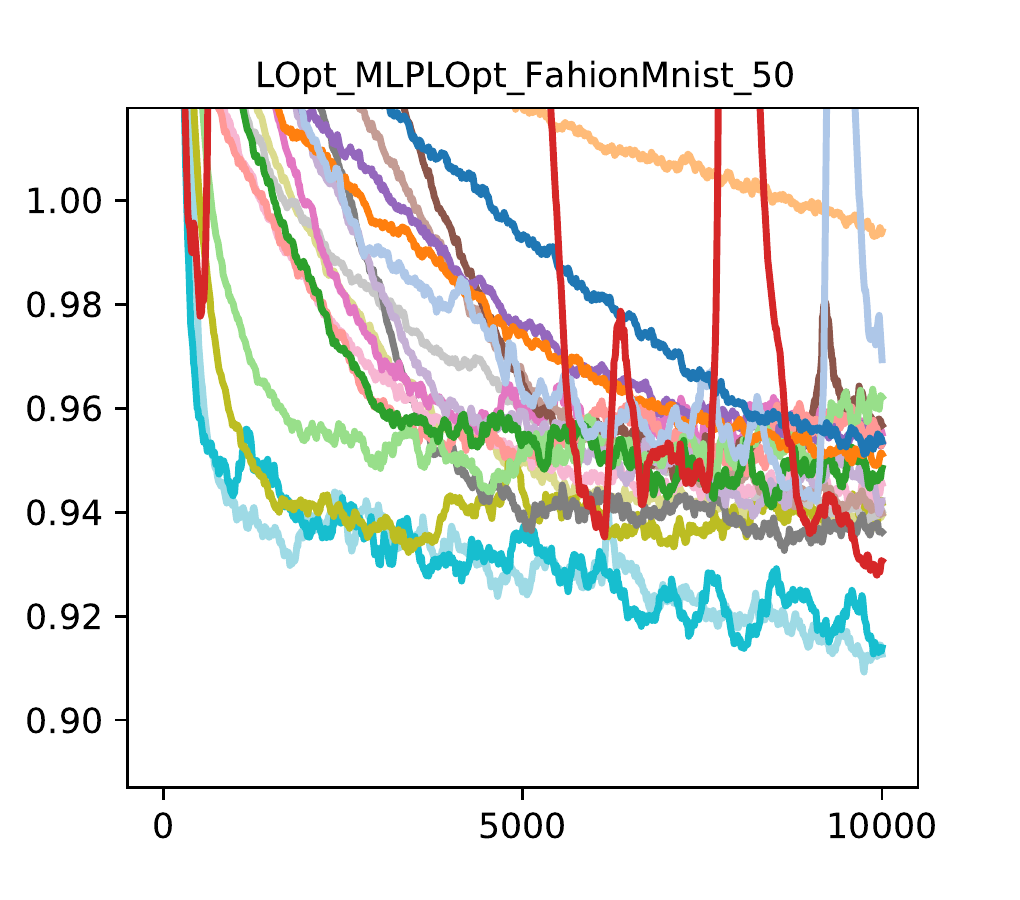}
    \end{overpic}
    \begin{overpic}[width=0.23\textwidth]{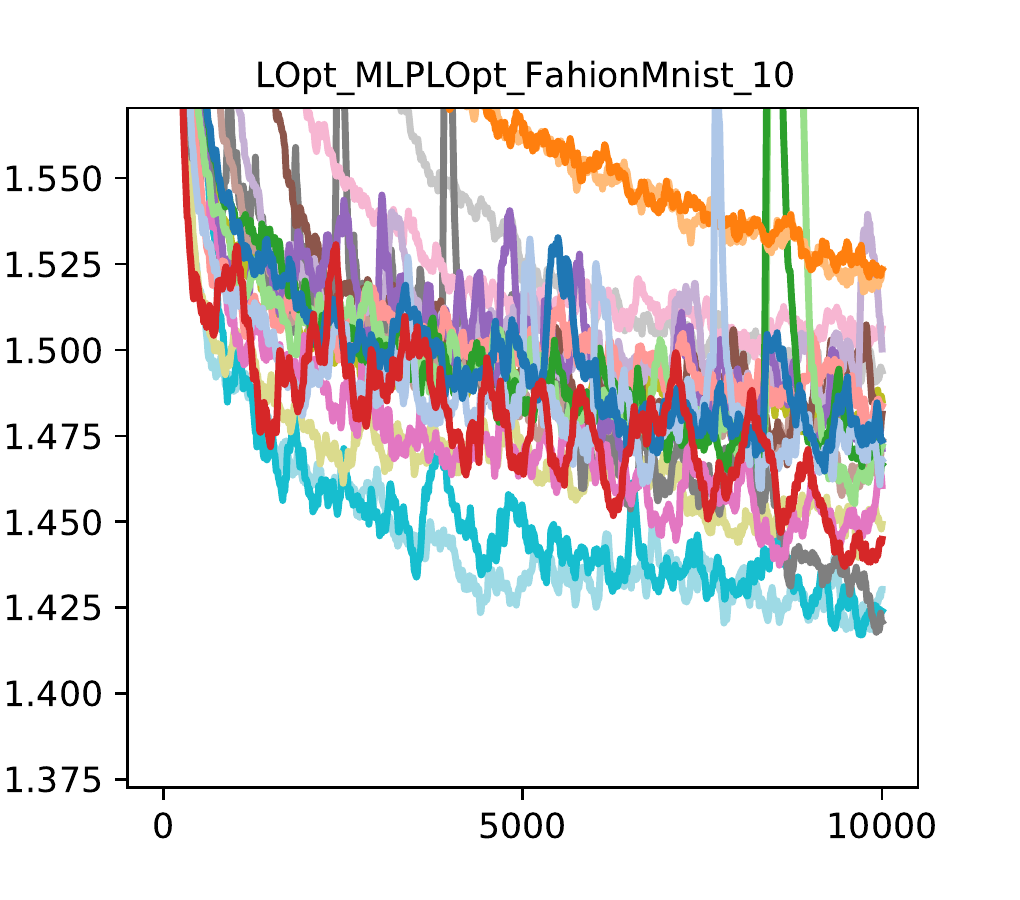}
    \end{overpic}
    \begin{overpic}[width=0.23\textwidth]{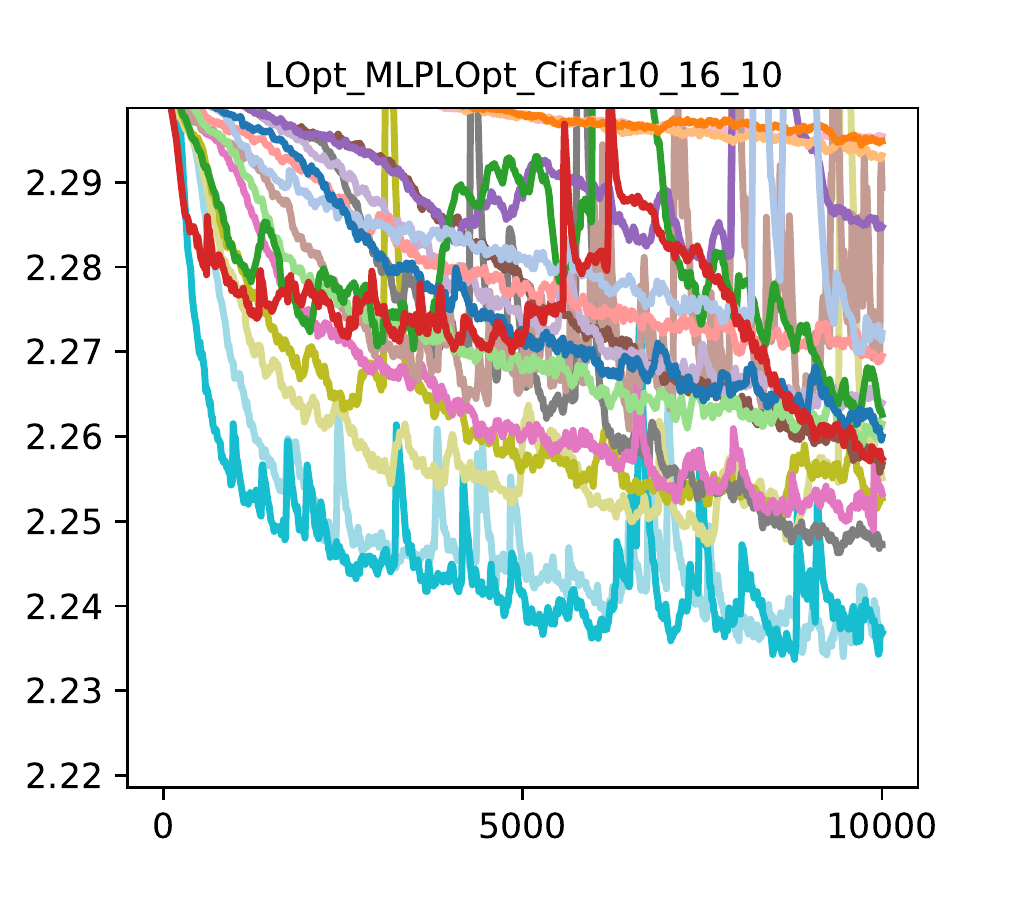}
    \end{overpic}
    }


    \makebox[\textwidth]{%
    \begin{overpic}[width=0.23\textwidth]{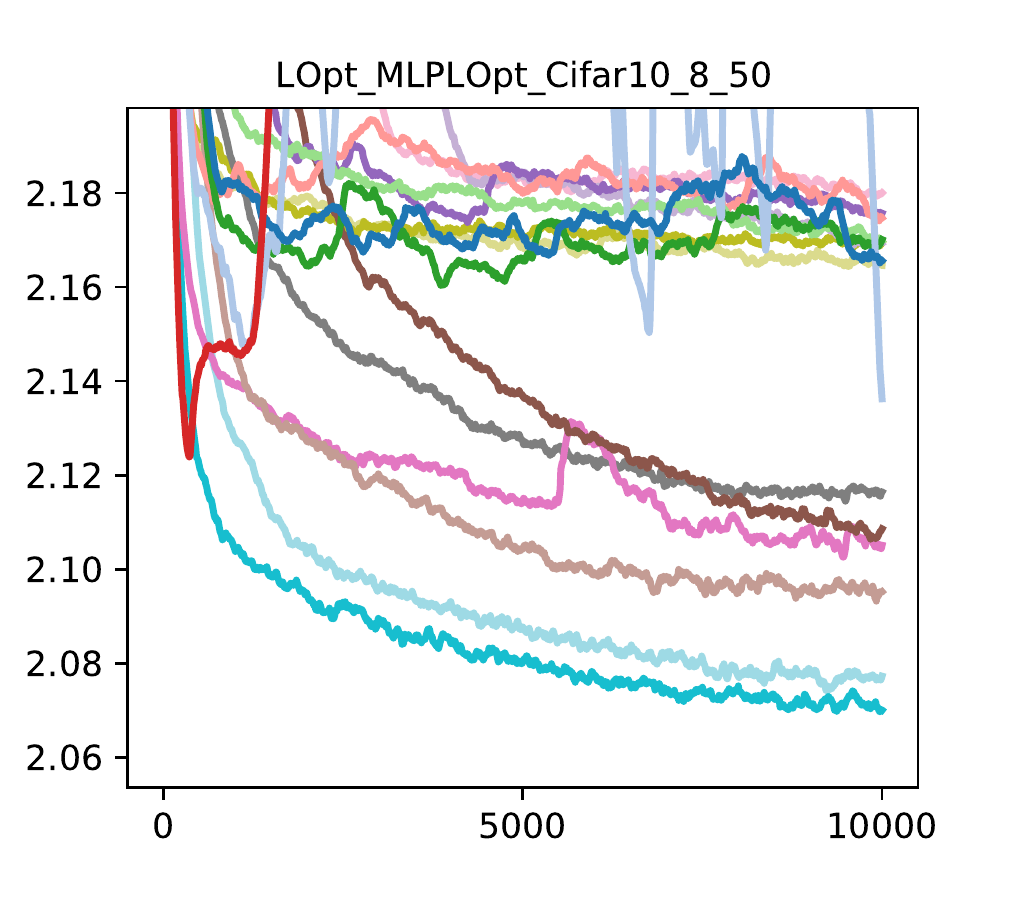}
    \end{overpic}
    \begin{overpic}[width=0.23\textwidth]{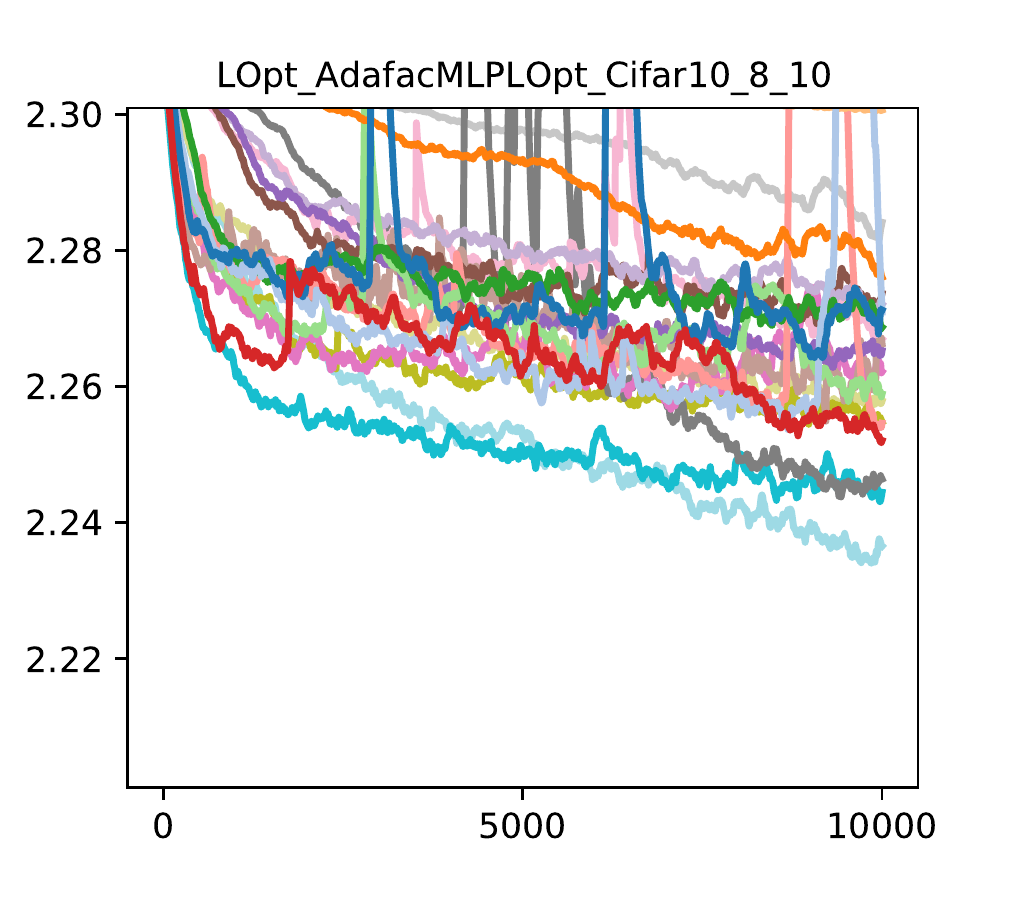}
    \end{overpic}
    \begin{overpic}[width=0.23\textwidth]{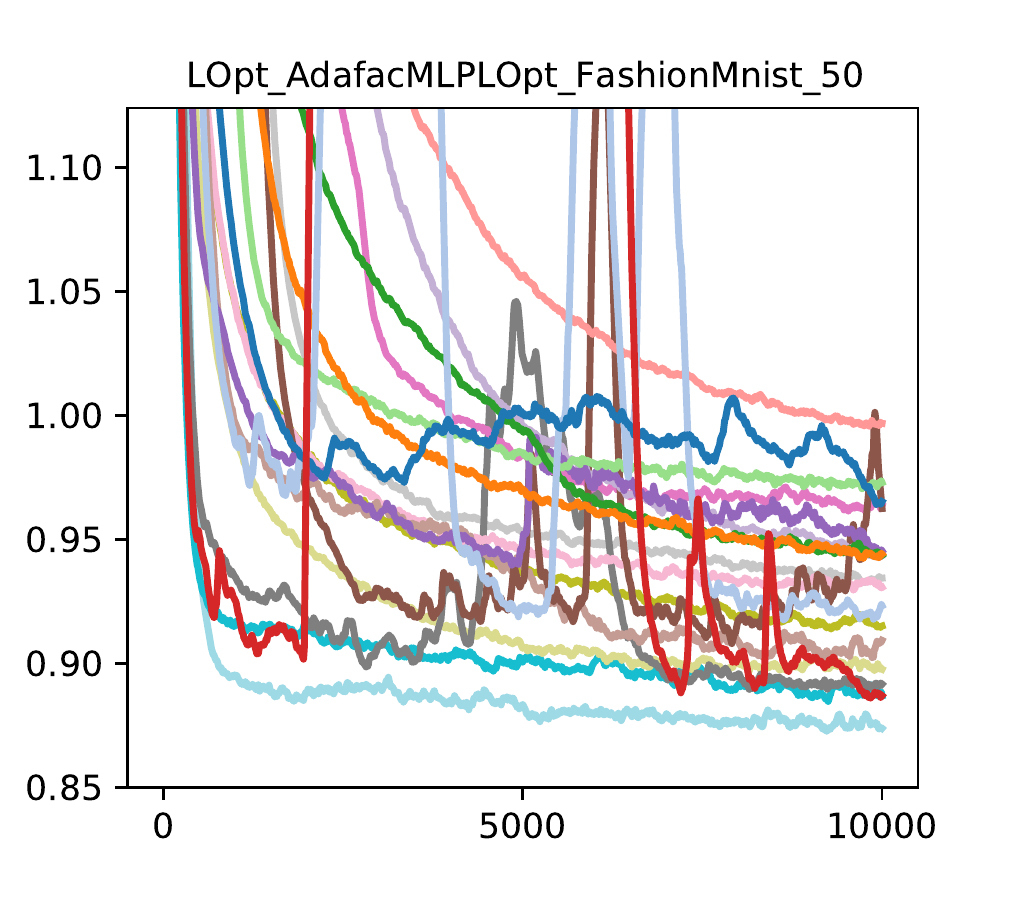}
    \end{overpic}
    \begin{overpic}[width=0.23\textwidth]{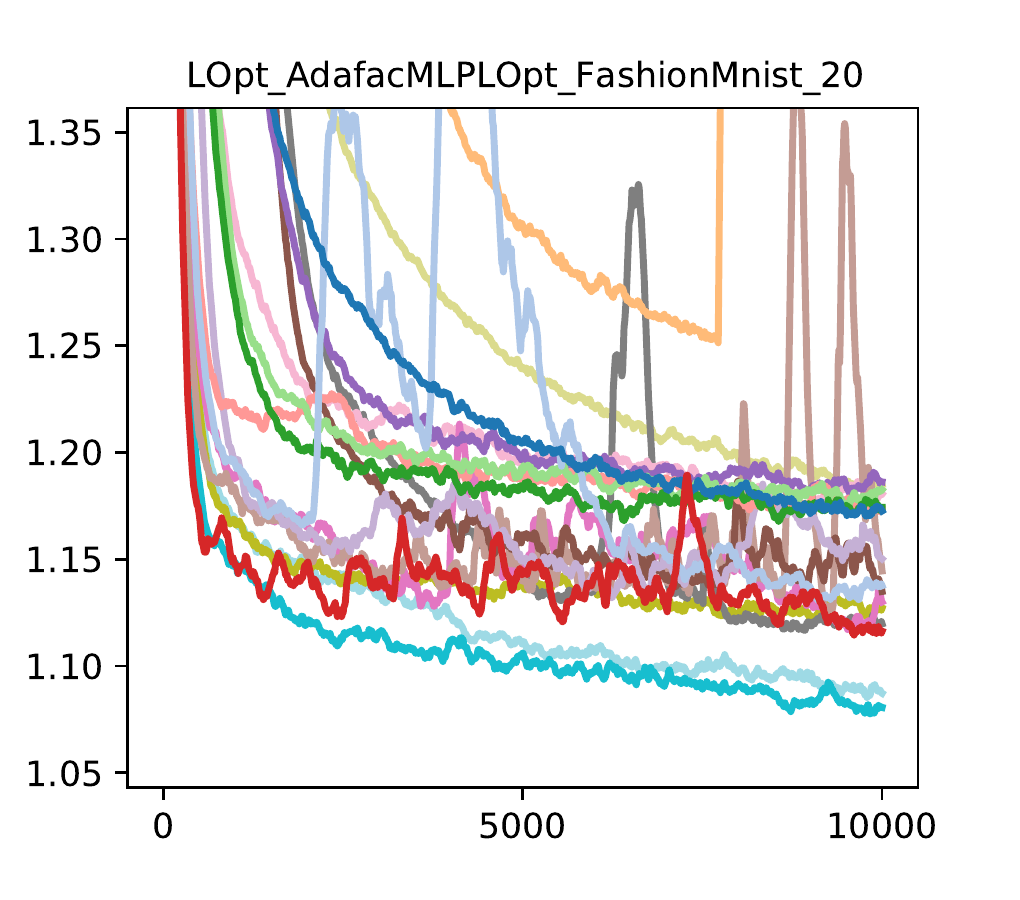}
    \end{overpic}
    \begin{overpic}[width=0.23\textwidth]{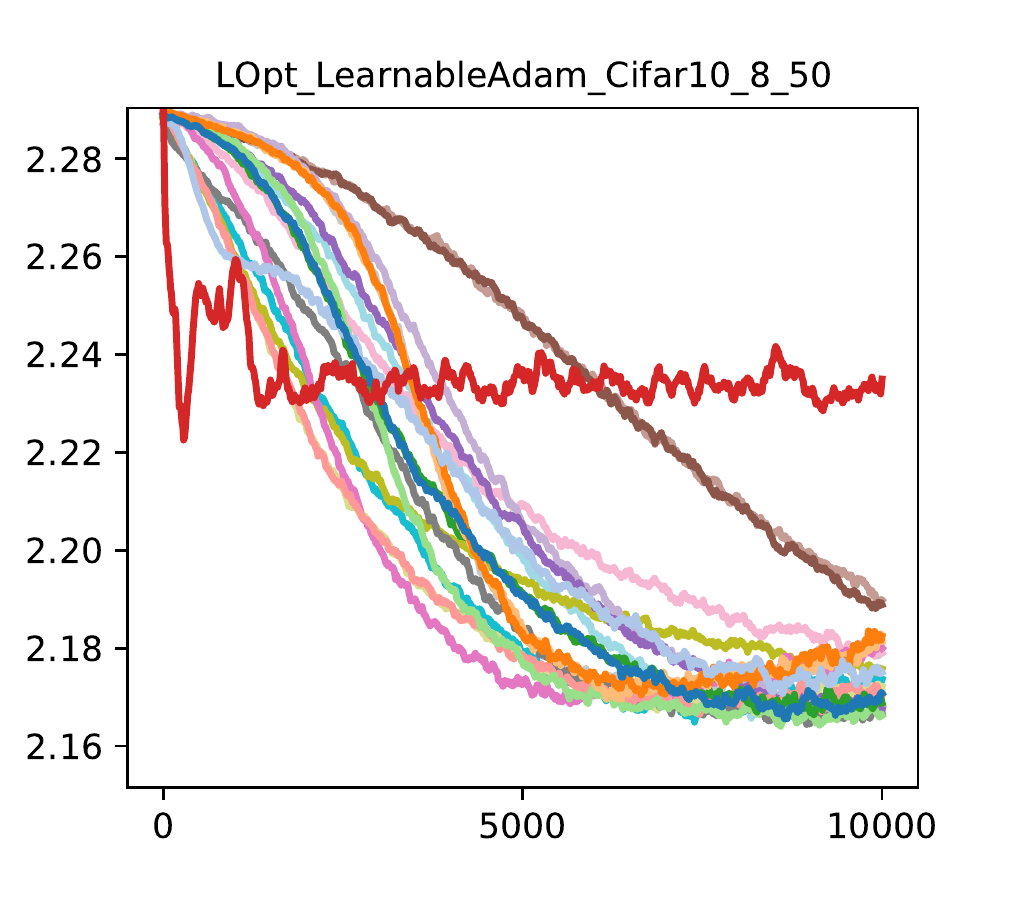}
    \end{overpic}
    }

%% file: app_page4.tex

    \makebox[\textwidth]{%
    \begin{overpic}[width=0.23\textwidth]{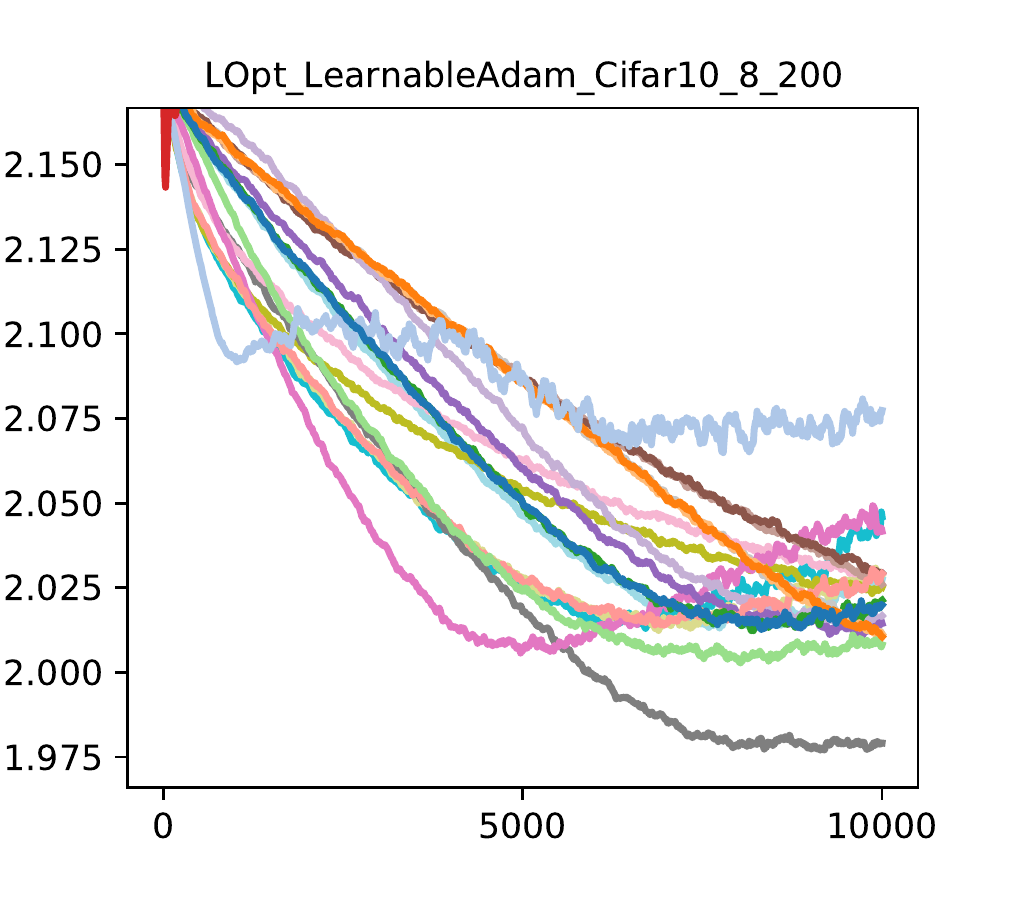}
    \end{overpic}
    \begin{overpic}[width=0.23\textwidth]{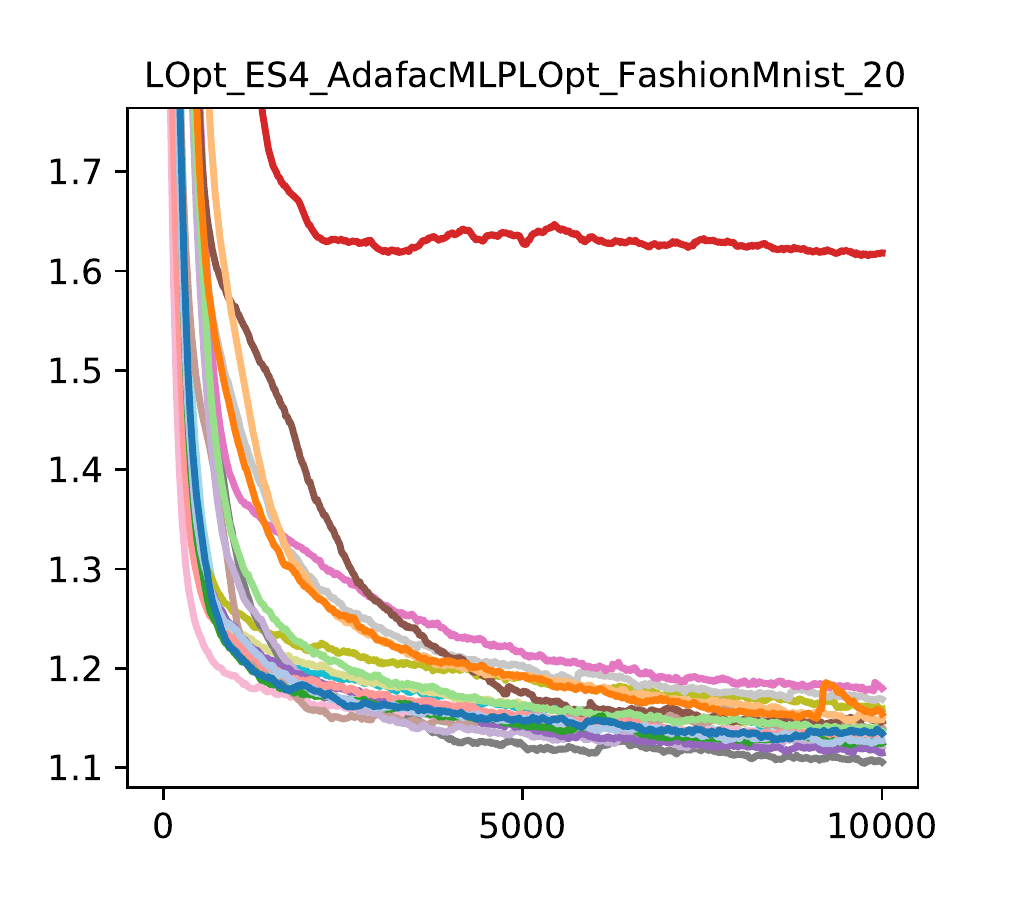}
    \end{overpic}
    \begin{overpic}[width=0.23\textwidth]{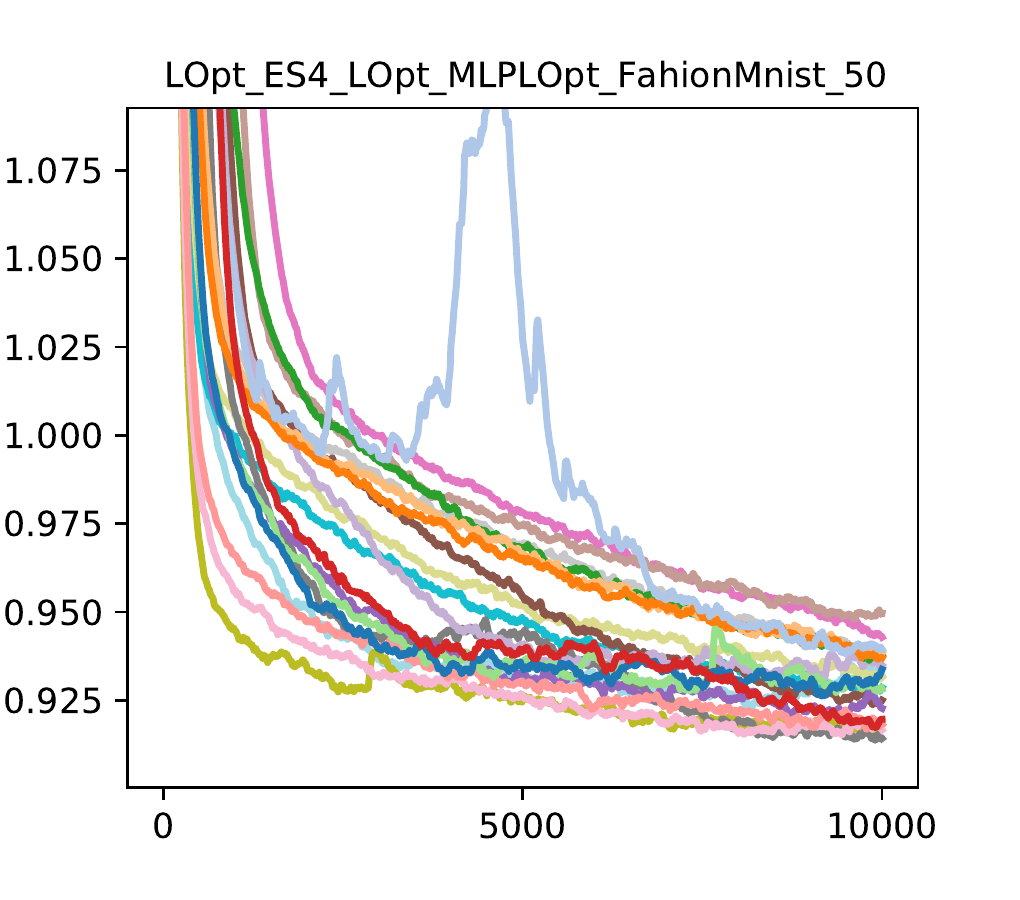}
    \end{overpic}
    \begin{overpic}[width=0.23\textwidth]{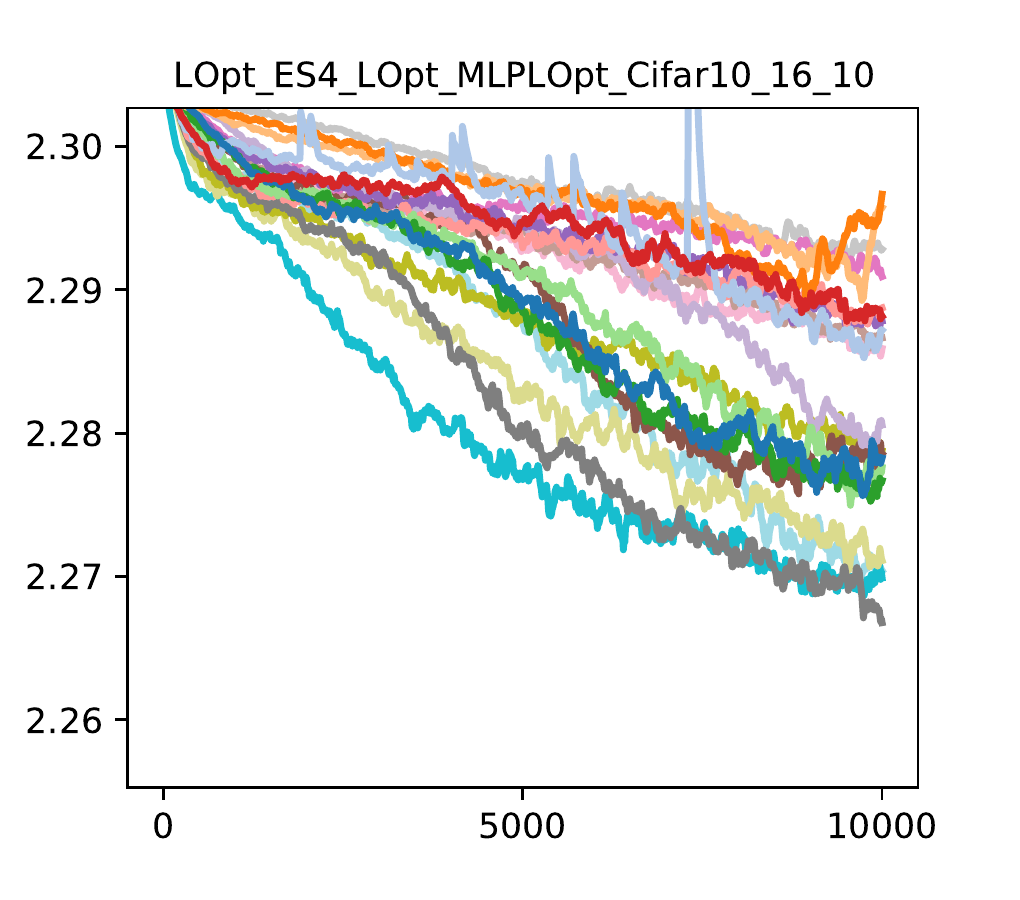}
    \end{overpic}
    \begin{overpic}[width=0.23\textwidth]{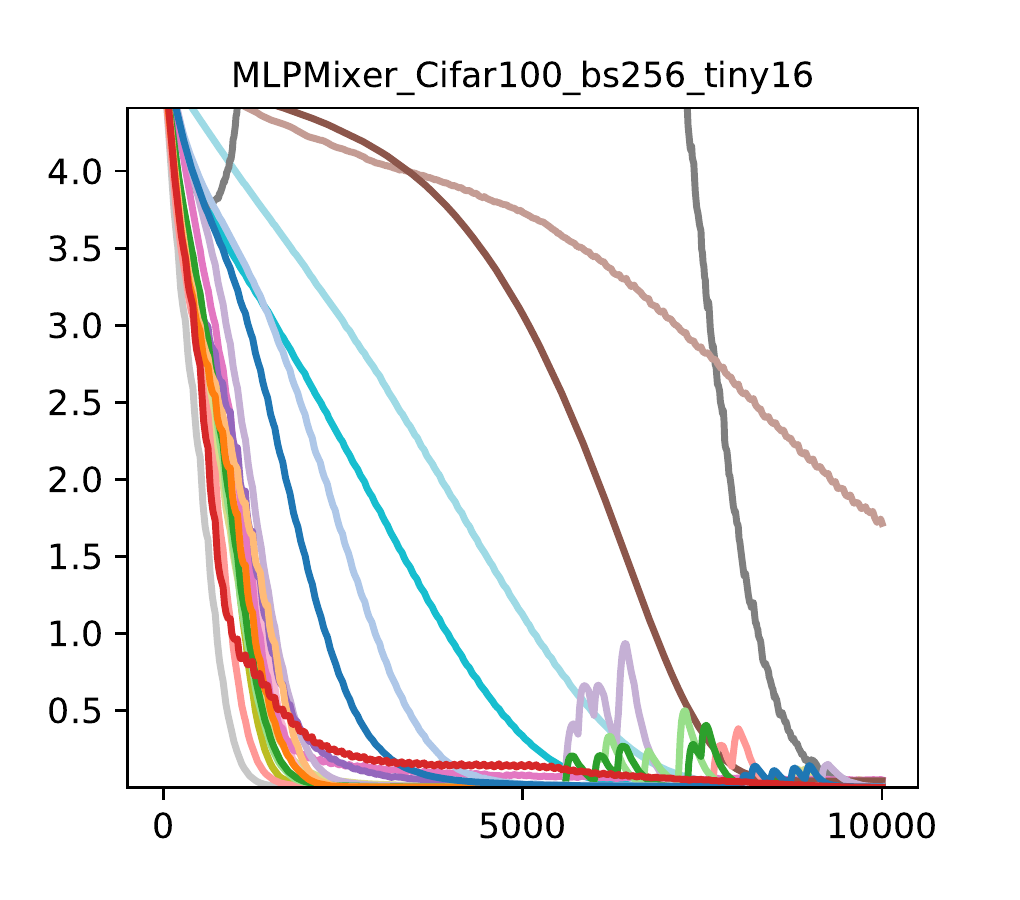}
    \end{overpic}
    }


    \makebox[\textwidth]{%
    \begin{overpic}[width=0.23\textwidth]{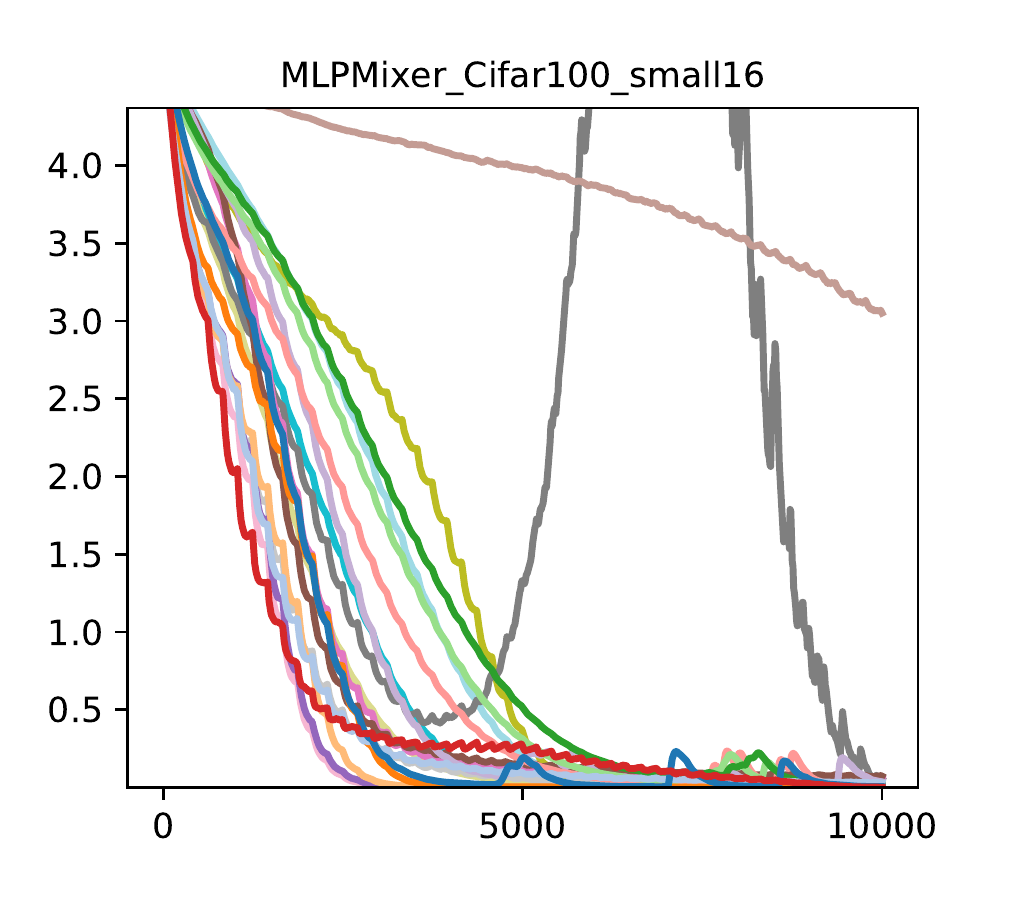}
    \end{overpic}
    \begin{overpic}[width=0.23\textwidth]{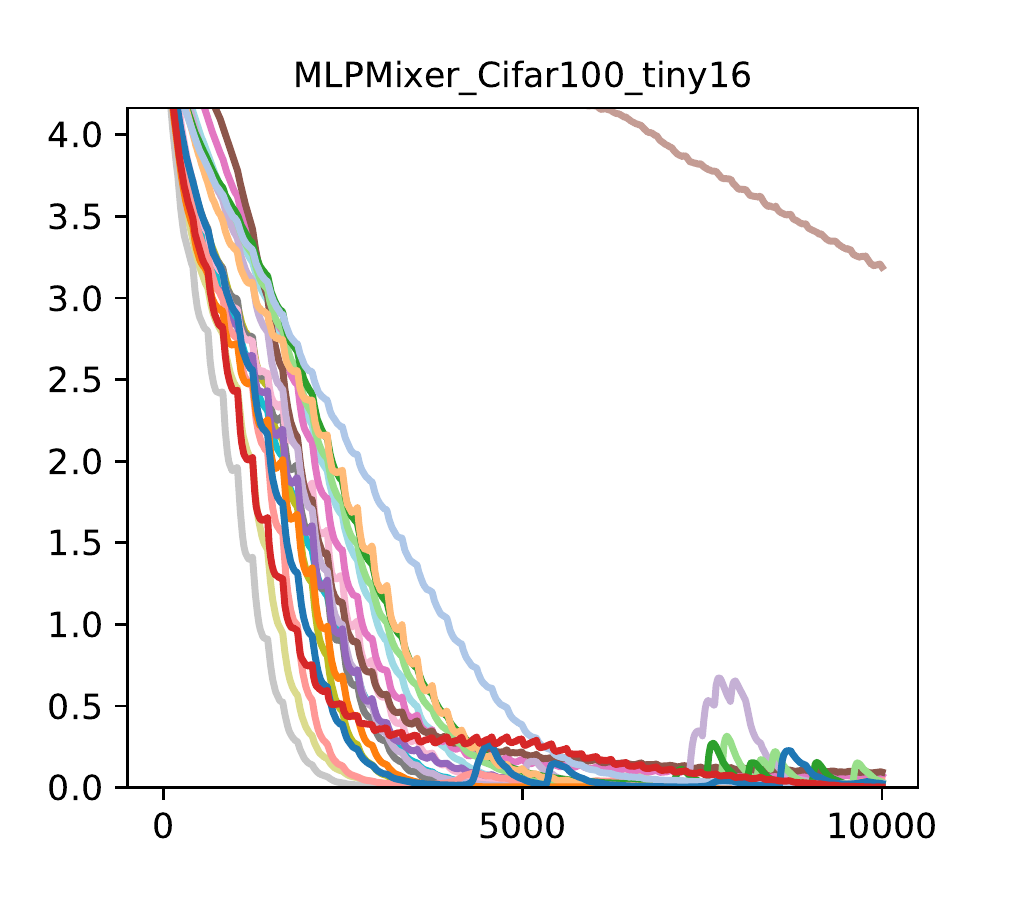}
    \end{overpic}
    \begin{overpic}[width=0.23\textwidth]{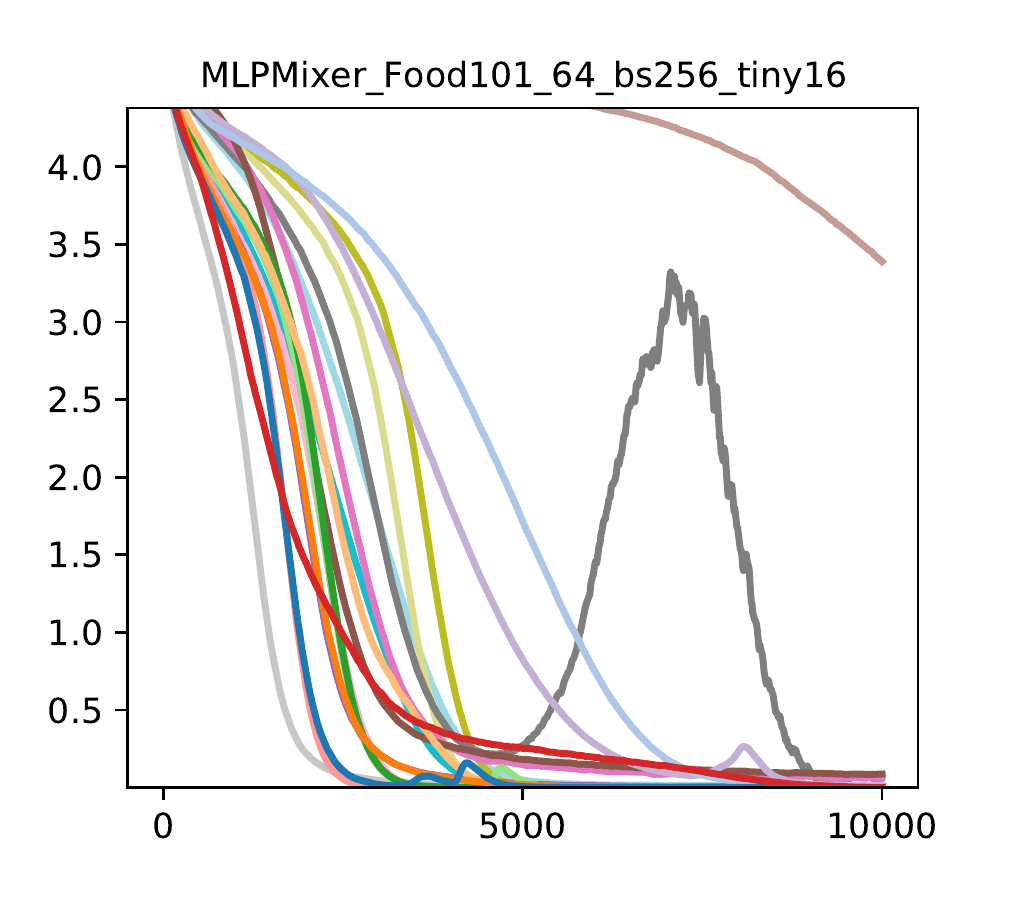}
    \end{overpic}
    \begin{overpic}[width=0.23\textwidth]{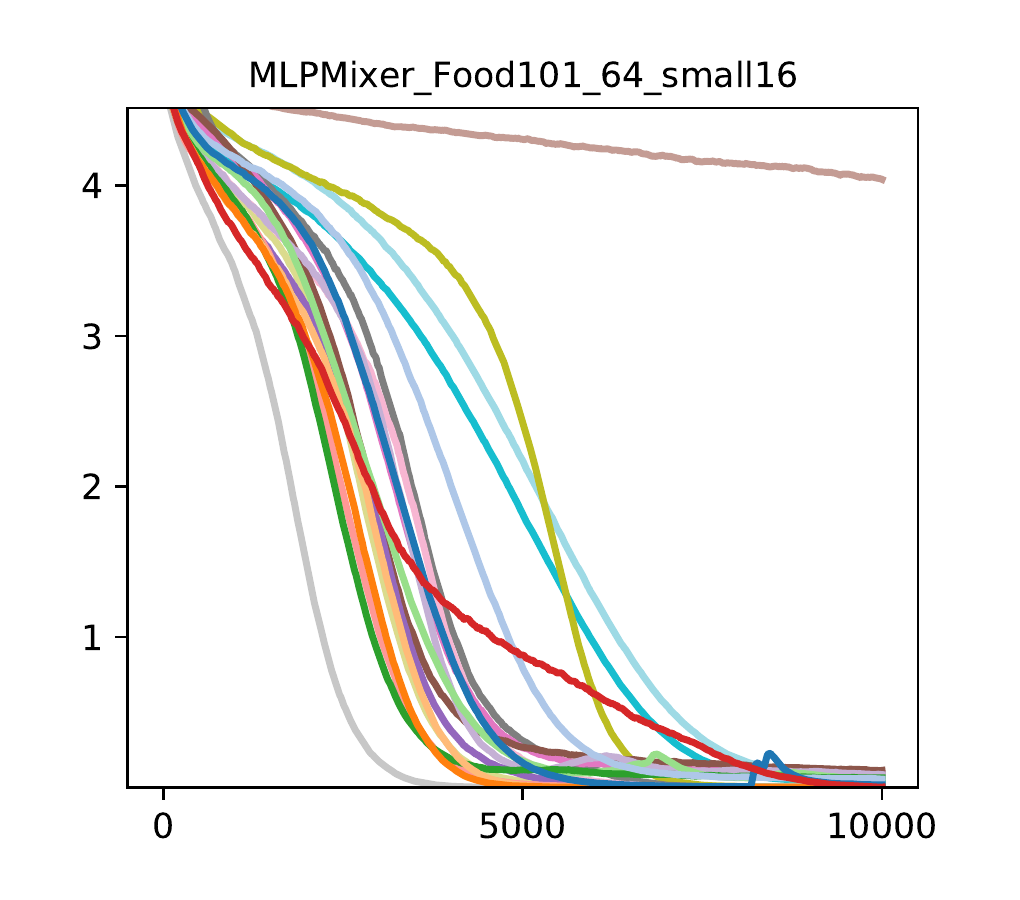}
    \end{overpic}
    \begin{overpic}[width=0.23\textwidth]{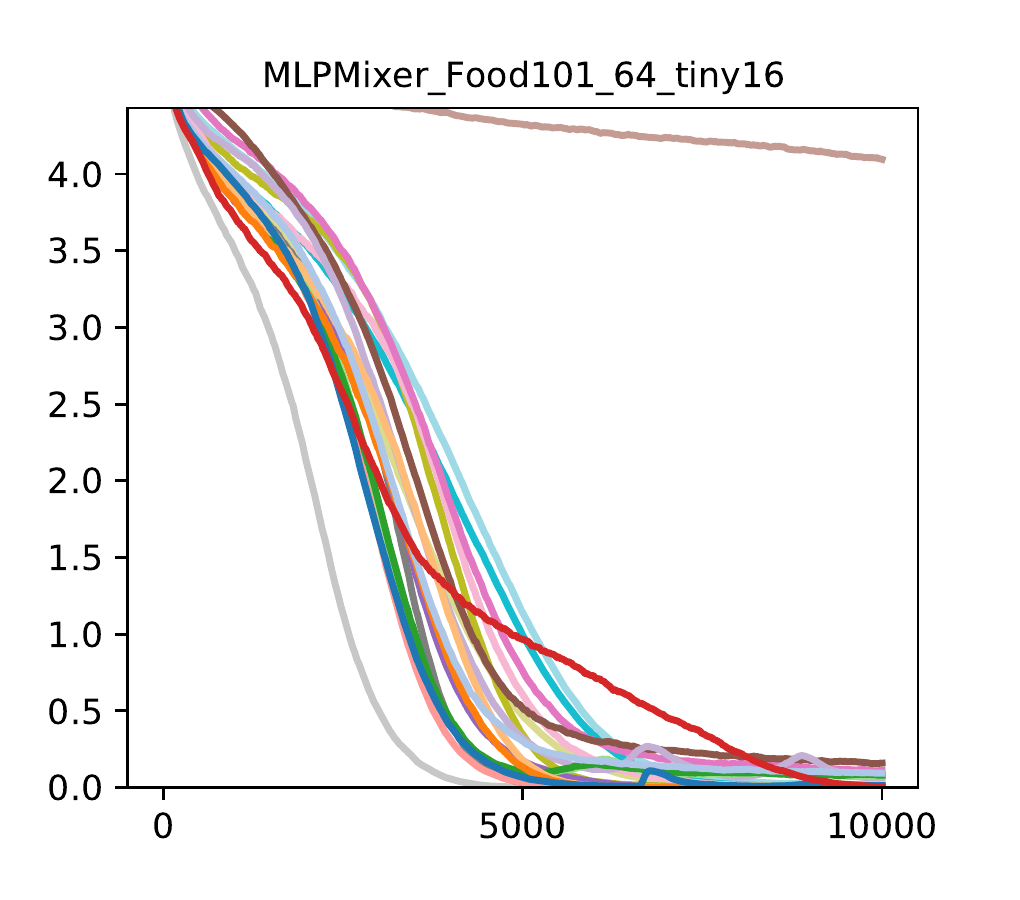}
    \end{overpic}
    }


    \makebox[\textwidth]{%
    \begin{overpic}[width=0.23\textwidth]{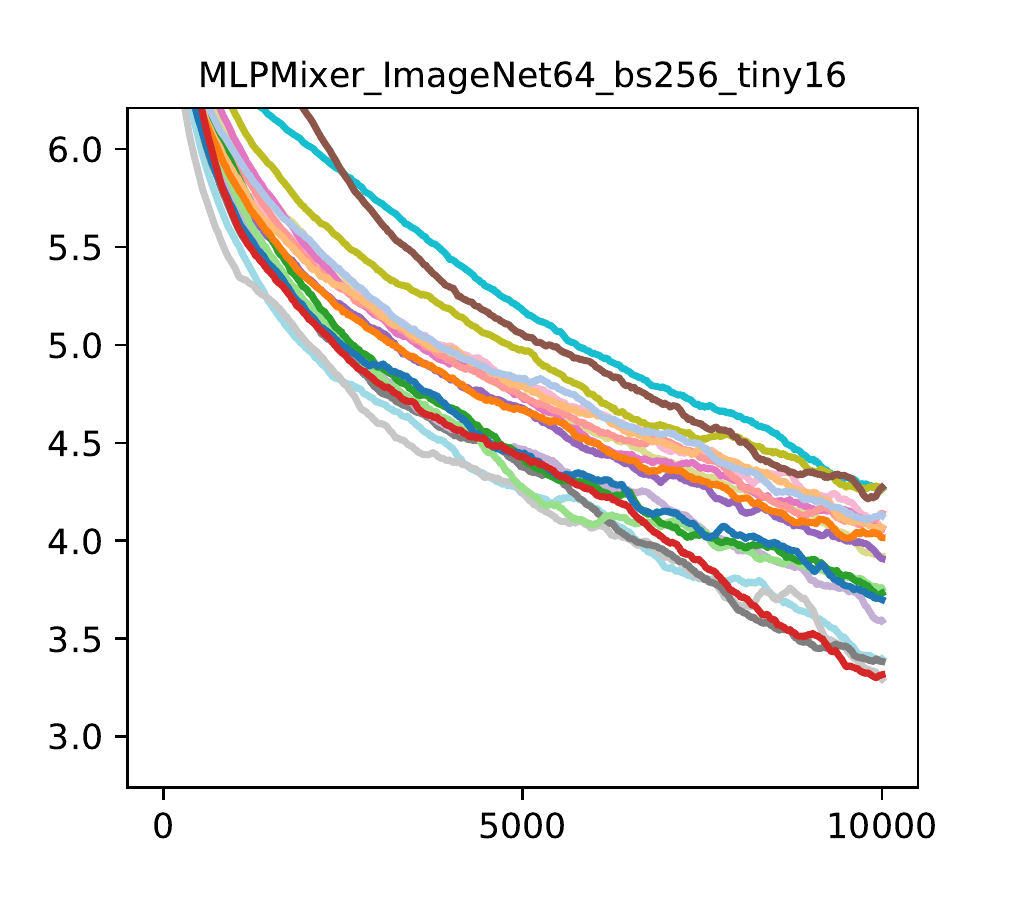}
    \end{overpic}
    \begin{overpic}[width=0.23\textwidth]{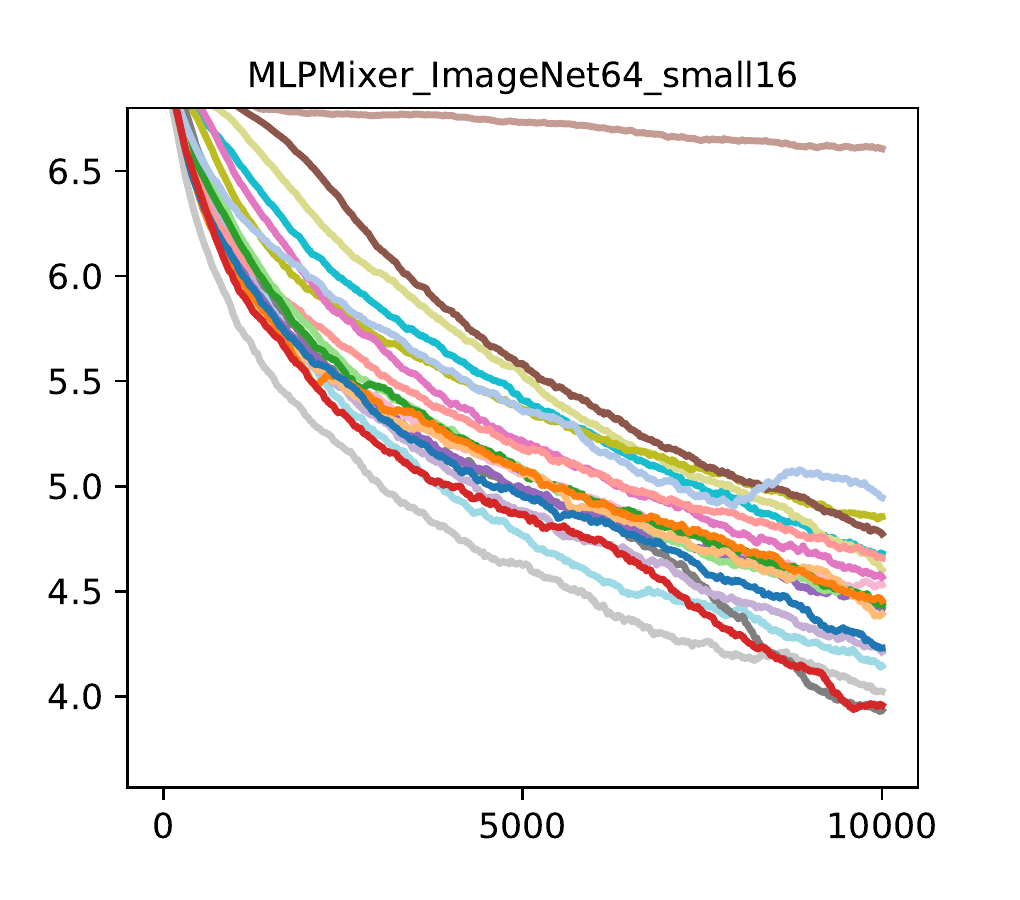}
    \end{overpic}
    \begin{overpic}[width=0.23\textwidth]{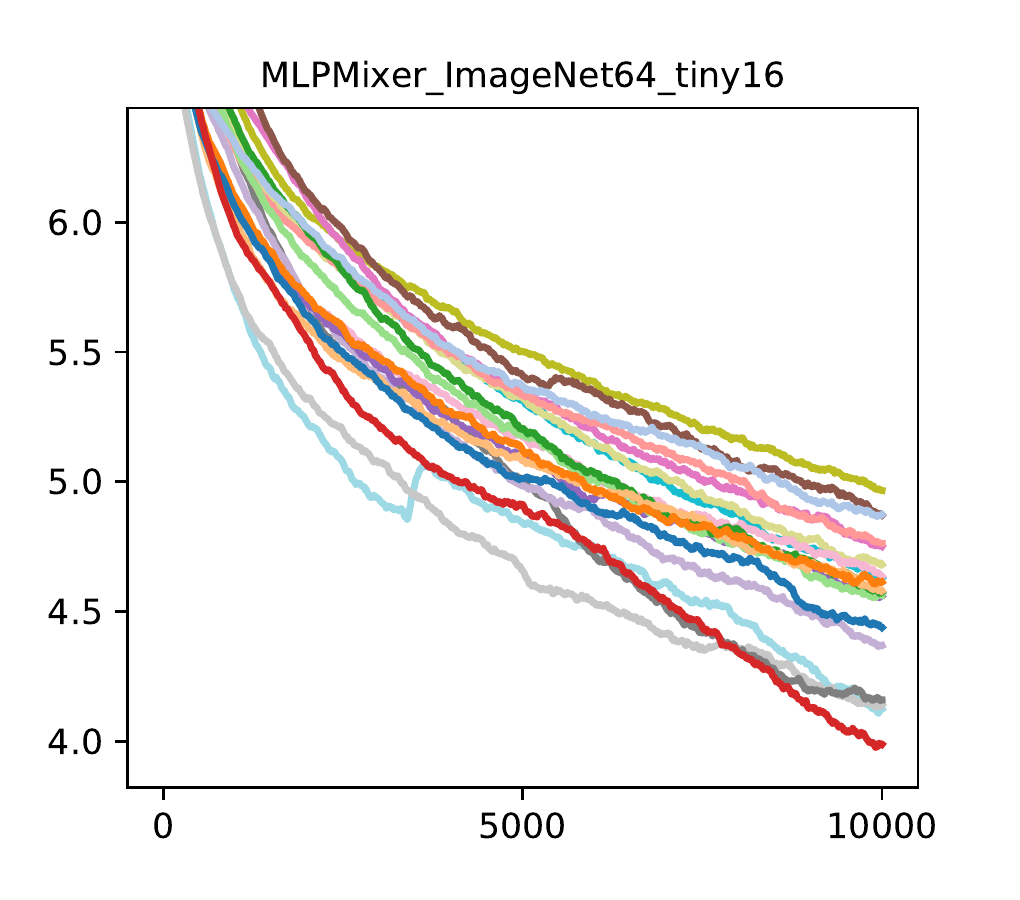}
    \end{overpic}
    \begin{overpic}[width=0.23\textwidth]{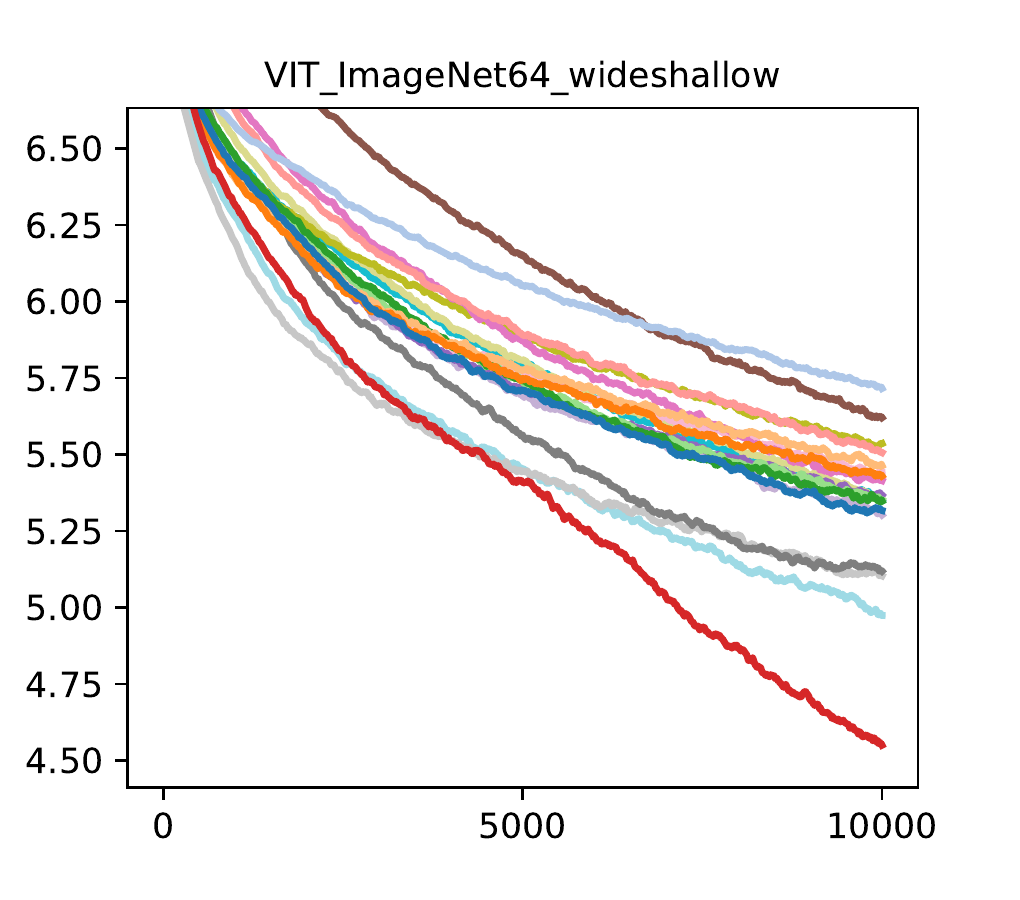}
    \end{overpic}
    \begin{overpic}[width=0.23\textwidth]{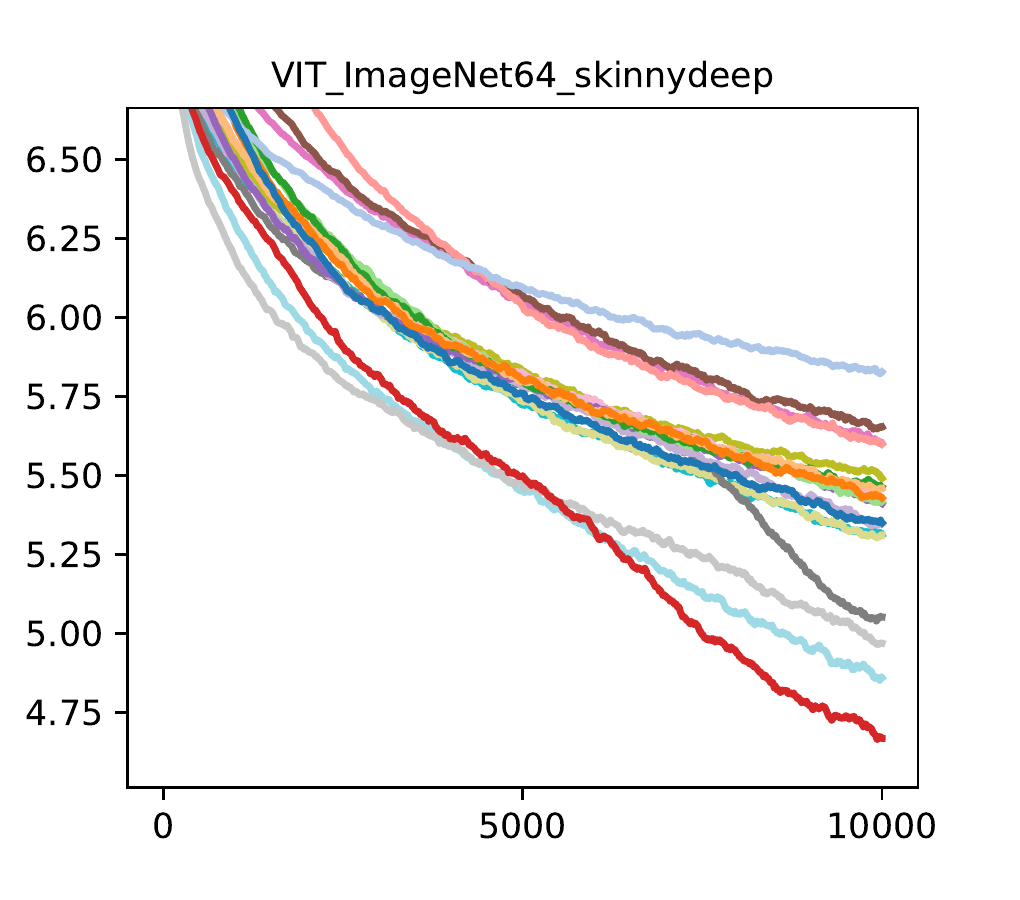}
    \end{overpic}
    }


    \makebox[\textwidth]{%
    \begin{overpic}[width=0.23\textwidth]{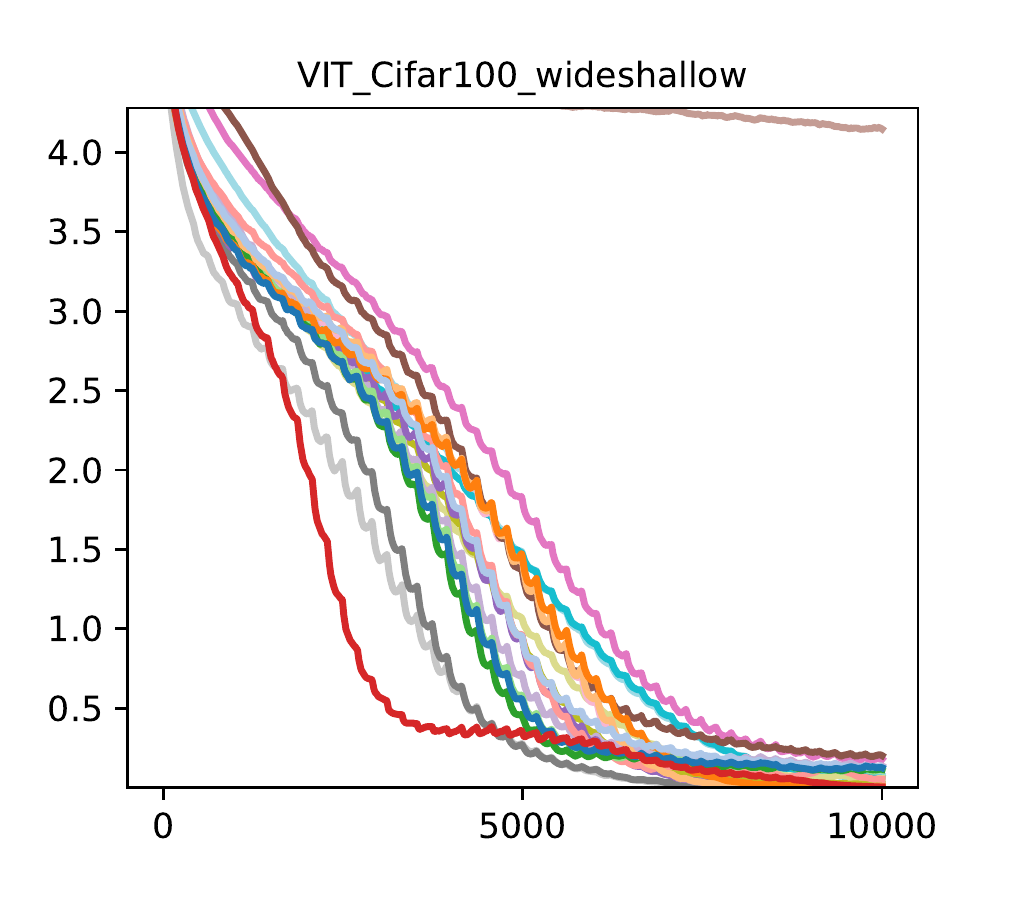}
    \end{overpic}
    \begin{overpic}[width=0.23\textwidth]{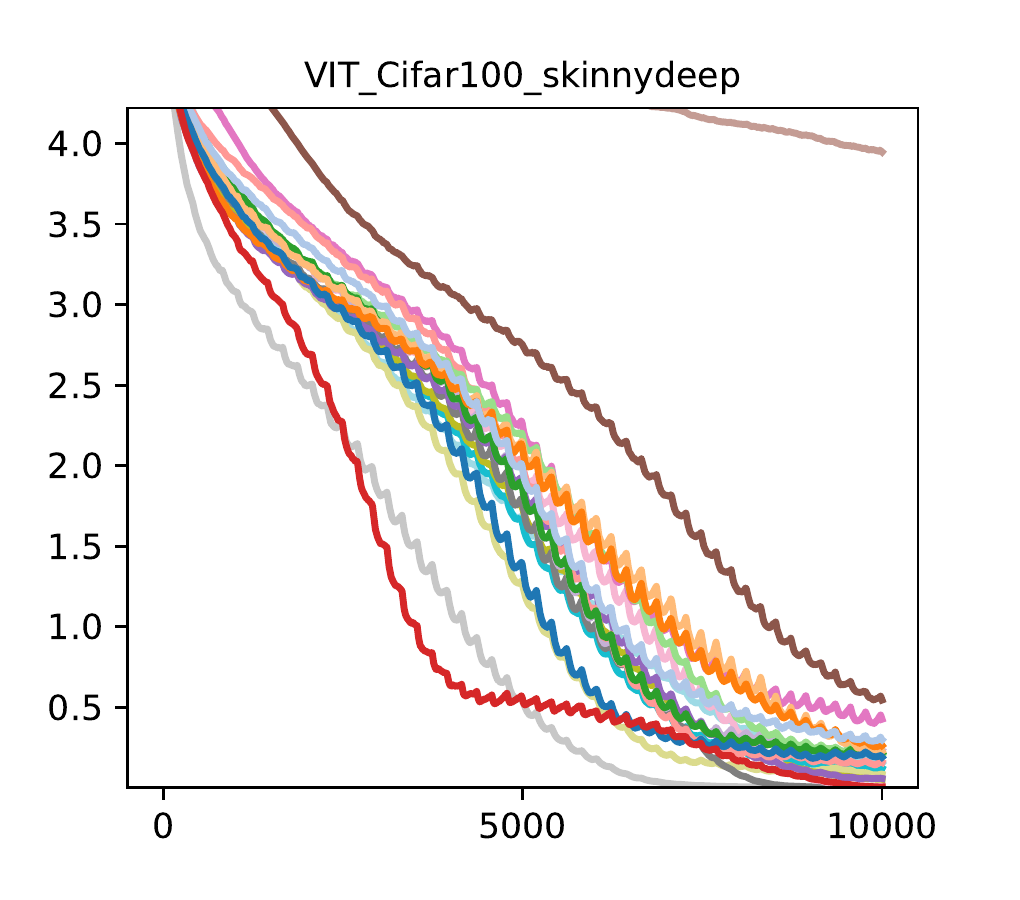}
    \end{overpic}
    \begin{overpic}[width=0.23\textwidth]{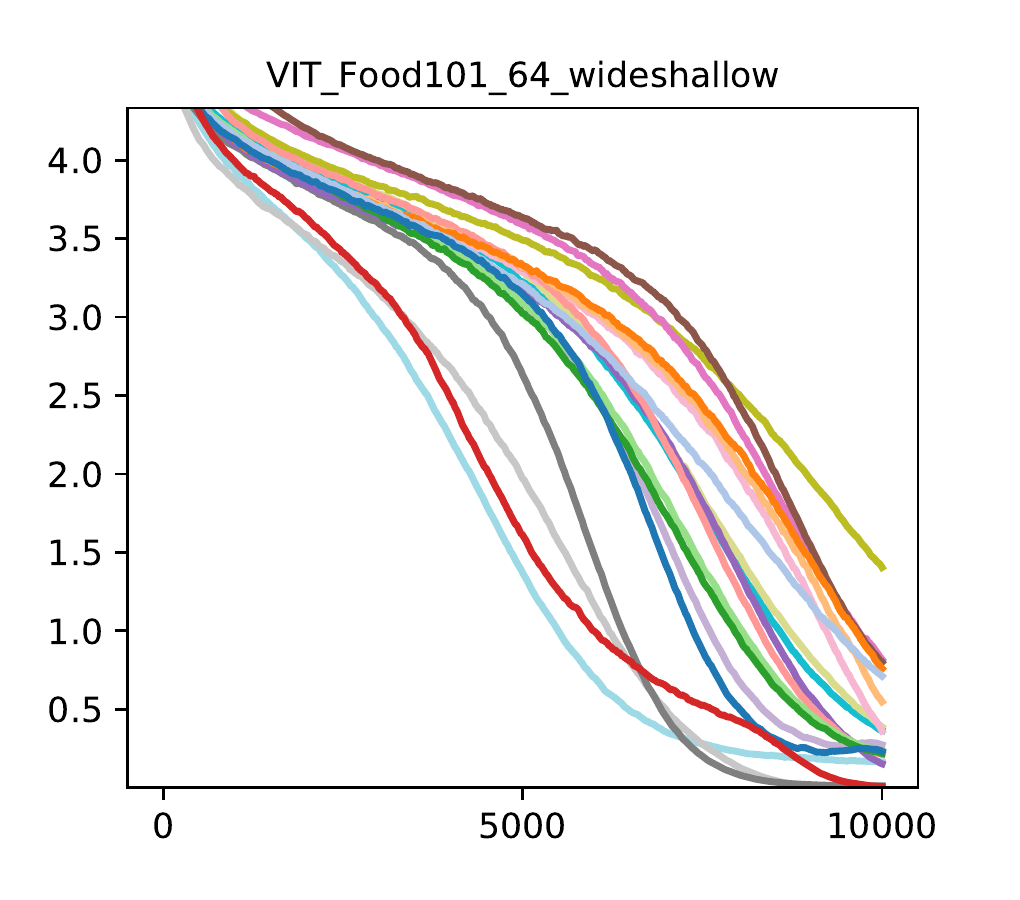}
    \end{overpic}
    \begin{overpic}[width=0.23\textwidth]{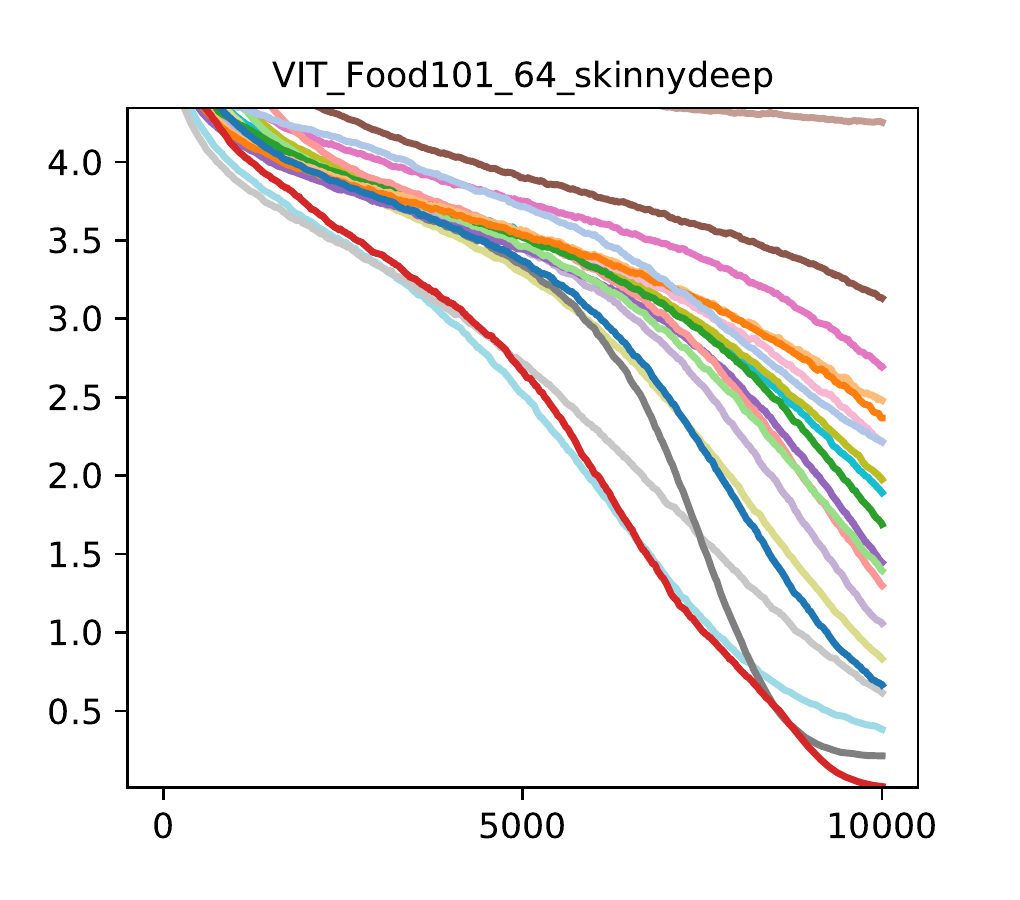}
    \end{overpic}
    }